\documentclass[twoside]{article}
\usepackage[accepted]{aistats2026}

\usepackage{pifont,natbib}
\usepackage[utf8]{inputenc} 
\usepackage[T1]{fontenc}    
\usepackage{hyperref}       
\usepackage{url}            
\usepackage{booktabs}       
\usepackage{amsfonts}       
\usepackage{nicefrac}       
\usepackage{microtype}      
\usepackage{xcolor}         

\usepackage{times,tabularx,makecell}
\usepackage{amssymb,amsmath,graphicx,cite,caption,subcaption,wrapfig}
\usepackage{enumerate}
\usepackage[shortlabels]{enumitem}
\usepackage{multicol}
\usepackage{empheq}
\usepackage{color}
\usepackage{algorithm,algorithmic}
\usepackage{cases}
\usepackage[normalem]{ulem}
\usepackage{bbm}
\usepackage{adjustbox}

\usepackage[capitalise]{cleveref}

\usepackage{rotating}
\usepackage{multirow}

\allowdisplaybreaks
\let\underbrace\LaTeXunderbrace

\usepackage{tikz}

\newcommand{\Acal}{{\cal A}}

\newcommand{\Dcal}{{\cal D}}
\newcommand{\Ecal}{{\cal E}}
\newcommand{\Fcal}{{\cal F}}

\newcommand{\Hcal}{{\cal H}}

\newcommand{\Lcal}{{\cal L}}
\newcommand{\Mcal}{{\cal M}}

\newcommand{\Ocal}{{\cal O}}
\newcommand{\Pcal}{{\cal P}}

\newcommand{\Scal}{{\cal S}}

\newcommand{\1}{{\mathbf{1}}}

\newcommand{\argmin}{\mathop{\rm argmin}}
\newcommand{\argmax}{\mathop{\rm argmax}}

\newtheorem{prop}{Proposition}
\newtheorem{lem}{Lemma}
\newtheorem{thm}{Theorem}

\newtheorem{assump}{Assumption}
\newtheorem{remark}{Remark}

\newcommand{\bias}{\mathcal{L}}
\newcommand{\reweight}{\operatorname{reweight}}

\newcommand{\qed}{\hfill $\blacksquare$}
\usepackage{array}

\usepackage{tcolorbox}

\begin{document}

\runningtitle{A Hessian-Free Actor-Critic Algorithm for Bi-Level Reinforcement Learning
with Applications to LLM Fine-Tuning}

\twocolumn[

\aistatstitle{
A Hessian-Free Actor-Critic Algorithm for Bi-Level Reinforcement Learning
with Applications to LLM Fine-Tuning
}

\aistatsauthor{
Sihan Zeng$^*$ \And
Sujay Bhatt$^*$ \And
Sumitra Ganesh$^*$ \And
Alec Koppel$^{\dagger}$
}

\aistatsaddress{
$^*$JPMorgan AI Research \And
$^\dagger$ Johns Hopkins University Applied Physics Laboratory
}
]

\begin{abstract}
We study a structured bi-level optimization problem where the upper-level objective is a smooth function and the lower-level problem is policy optimization in a Markov decision process (MDP). The upper-level decision variable parameterizes the reward of the lower-level MDP, and the upper-level objective depends on the optimal induced policy. 
Existing methods for bi-level optimization and RL often require second-order information, impose strong regularization at the lower level, or inefficiently use samples through nested-loop procedures. In this work, we propose a single-loop, first-order actor-critic algorithm that optimizes the bi-level objective via a penalty-based reformulation. We introduce into the lower-level RL objective an attenuating entropy regularization, which enables asymptotically unbiased upper-level hyper-gradient estimation without solving the unregularized RL problem exactly. We establish the finite-time and finite-sample convergence of the proposed algorithm to a stationary point of the original, unregularized bi-level optimization problem through a novel lower-level residual analysis under a special type of Polyak–\L ojasiewicz condition. We validate the performance of our method through experiments on a GridWorld goal position problem and on happy tweet generation through reinforcement learning from human feedback (RLHF).\looseness=-1

\end{abstract}

\begin{table*}[!ht]
\centering
\setlength{\tabcolsep}{7pt}
\small
\begin{tabular}{cccccc}
    \toprule
    & \makecell{Lower-Level \\ Structure} & \makecell{Single \\ Loop} & \makecell{Only Using First-\\ Order Information} & \makecell{Sample Complexity}  & \makecell{Anytime\\ Valid}\\
    \midrule
    \citet{kwon2023fully} & Strong convexity & \ding{51} & \ding{51} & $\widetilde{\Ocal}(\epsilon^{-3.5})^\dagger$ & \ding{51}\\
    \citet{shen2023penalty} & PL condition & \ding{55} & \ding{51} & N/A (Iteration complexity derived) & -\\
    \citet{xiao2024unlocking} & PL condition & \ding{51} & \ding{51} & N/A (Deterministic setting studied)& -\\
    \citet{shen2024principled} & Regularized RL & \ding{55} & \ding{51} & N/A (Iteration complexity derived) & -\\
    \citet{chakraborty2024parl} & Regularized RL & \ding{55} & \ding{55} & N/A (Iteration complexity derived) & -\\
    \citet{thoma2024contextual} & Regularized RL & \ding{55} & \ding{51} & $\widetilde{\Ocal}(\epsilon^{-4})$ & \ding{55} \\
    \citet{xiao2025first} & Strong convexity & \ding{55} & \ding{51} & N/A & \ding{55} \\
    \citet{chen2024finding} & PL condition & \ding{55} & \ding{51} & $\widetilde{\Ocal}(\epsilon^{-3})$ & \ding{55} \\
    \citet{yang2025bilevel} & Regularized RL & \ding{55} & \ding{51} & $\widetilde{\Ocal}(\epsilon^{-3.5})$ & \ding{55} \\
    \citet{gaur2025sample} & Regularized RL & \ding{55} & \ding{51} & $\widetilde{\Ocal}(\epsilon^{-3})$ & \ding{55}\\
    \textbf{This work} & \textbf{Regularized RL} & \textbf{\ding{51}} & \textbf{\ding{51}} & \textbf{$\widetilde{\Ocal}(\epsilon^{-3})$} & \textbf{\ding{51}}\\
    \textbf{This work} & \textbf{Original RL} & \textbf{\ding{51}} & \textbf{\ding{51}} & \textbf{$\widetilde{\Ocal}(\epsilon^{-10})$} & \textbf{\ding{51}}\\
    \bottomrule
    \bottomrule
    \end{tabular}
\caption{Assumption, structure, and sample complexity (measured by squared gradient norm) of existing algorithms for bi-level optimization and RL. $^\dagger$This is the complexity of the standard \text{F$^2$SA} (Fully First-order Stochastic Approximation) algorithm. \citet{kwon2023fully} also proposes a momentum-based algorithm (\text{F$^3$SA}) that achieves a better rate. For a fair comparison with other non-momentum algorithms in the table, we report the complexity of \text{F$^2$SA}. Notably, \citet{yang2023achieving} further improve the design and analysis of momentum-based algorithms, achieving a complexity of $\Ocal(\epsilon^{-1.5})$, where the key innovation is estimating the lower-level Hessian via finite differences.}
\vspace{-10pt}
\label{table:literature}
\end{table*}

\section{INTRODUCTION}

We study bi-level reinforcement learning (RL), a structured bi-level optimization program in which the upper-level decision variable determines the reward function of a lower-level RL problem, and the upper-level objective is evaluated under the lower-level optimal policy. This framework abstracts a wide range of applications where the reward must be tuned to achieve high-level goals while the underlying policy adapts to the reward. Examples include reward shaping \citep{ng1999policy,hu2020learning}, inverse RL \citep{zhang2025understanding}, multi-agent incentive design \citep{ma2025adaptive}, contract design \citep{zhu2023sample}, synthetic data generation \citep{wu2024using}, and, notably, reinforcement learning from human feedback (RLHF), one of the central paradigms for training and fine-tuning large language models \citep{chakraborty2024parl,ye2025reward}.

Despite a recent surge of interest in bi-level optimization, bi-level RL remains challenging both in theory and practice, due to the non-convexity at the lower level and the difficulty of estimating the upper-level hyper-gradient. Existing gradient-based approaches to bi-level optimization largely fall into two categories. The first leverages the implicit function theorem to derive the upper-level hyper-gradient \citep{ghadimi2018approximation} and then applies iterative gradient descent along this direction. However, since the hyper-gradient depends on the Jacobian and Hessian of the lower-level objective, these methods are difficult to apply in bi-level RL, where second-order information either requires oracle access to the transition model or is prohibitively expensive to estimate from trajectory samples.

Another approach replaces the lower-level optimality condition with an explicit penalty term \citep{kwon2023fully,shen2023penalty}, enabling an alternative expression of the hyper-gradient that depends only on first-order information. While this approach bypasses Hessian and Jacobian estimation, existing convergence analyses require strong structural assumptions on the lower-level problem, most commonly strong convexity or the Polyak–\L ojasiewicz (PL) condition. In RL, however, these assumptions generally fail: the policy optimization objective may satisfy a weaker “gradient domination” property \citep{agarwal2021theory,mei2020global}, but strong convexity and the PL condition do not hold under common policy parameterizations. 
As a result, existing guarantees in bi-level optimization apply only to regularized or restricted settings of RL, leaving open the question of whether we can provably solve original, unregularized bi-level RL.
Moreover, existing penalty-based methods are often implemented in a nested-loop fashion, repeatedly solving the lower-level problem to high precision before each upper-level update to ensure stability. The price of this stability is inefficient use of samples in practice and limited scalability.\looseness=-1

Our work addresses this research gap. We advance penalty-based approaches in the context of RL and propose a single-loop actor-critic algorithm that provably converges to a stationary point of the bi-level RL objective. The core idea is to enforce the PL condition at the lower level through entropy regularization, and then gradually adjust its weight so that the regularized problem asymptotically recovers the original one. 
Two important technical innovations enable the finite-time and finite-sample analysis for the algorithm. First, we tightly characterize the iteration-wise convergence of the lower-level optimality error for the regularized RL objective, under a time-varying upper-level decision variable which itself is updated in the same loop on a slower time scale. Our innovation is a novel lower-level residual decomposition scheme under the PL condition, which generalizes the technique used in \citet{kwon2023fully} under lower-level strong convexity.
Combined with the penalty reformulation, this innovation allows us to establish the convergence of a fully single-loop penalty-based algorithm in the lower-level PL/regularized RL setting, achieving a convergence rate that surpasses the best known rate derived in \citet{kwon2023fully} under lower-level strong convexity.

Our goal is to optimize the original, unregularized bi-level RL objective. While a large regularization accelerates the solution of the regularized problem, it also enlarges the discrepancy between the regularized and unregularized optima. Our second innovation addresses this trade-off by dynamically decaying the regularization weight, allowing the algorithm to track the regularized optima as the regularized problem gradually approaches the original one. A key challenge arises here from the time-varying lower-level landscape, which we overcome via a careful multi-time-scale stochastic approximation analysis that balances the change rate of the landscape with the algorithm step sizes.
Below we detail and re-iterate our innovation and main contributions.\looseness=-1

\textbf{Main Contributions}\vspace{0.2cm}\\
$\bullet$ We establish several fundamental structural properties of bi-level RL, linking the problem to its entropy-regularized counterpart. In particular, we show that as the regularization weight decays, the optimizer of the entropy-regularized RL objective converges to the unique entropy-maximizing policy within the set of optimizers of the unregularized objective, and we provide a bound on this rate of convergence. 
We also provide bounds on the difference between the bi-level objective (and its gradient) with and without lower-level regularization.
Absent in the prior work, this type of structural analysis lays the foundation for using the regularized objective as a faithful surrogate for the original formulation, and offers insights of potential independent interest to the broader fields of RL and bi-level optimization.\looseness=-1

$\bullet$ We present a sample-based, single-loop bi-level RL algorithm and characterize its finite-sample complexity.
The algorithm optimizes a regularized bi-level RL objective, while the regularization weight dynamically decays over time. We prove that this algorithm converges to a stationary point of the original bi-level objective with a sample complexity of $\Ocal(\epsilon^{-10})$ through a novel five-time-scale analysis, carefully balancing the regularization weight with the update rates of the dual variable, upper-level decision variable, policy iterates, and value function estimates. To our knowledge, this is the first algorithm that provably solves the unregularized bi-level RL problem, and the first that enjoys finite-time and finite-sample guarantees. 

$\bullet$ We show that our algorithm can be instantiated with a constant regularization weight to solve the corresponding regularized bi-level RL problem with sample complexity $\Ocal(\epsilon^{-3})$. This matches the state-of-the-art complexity of a comparable nested-loop method under lower-level PL condition \citep{chen2024finding}, and improves over the $\Ocal(\epsilon^{-3.5})$ complexity of a single-loop algorithm under strong convexity \citet{kwon2023fully}. 

$\bullet$ We verify convergence of the proposed algorithm through experiments in a synthetic RL environment and in finetuning a (small) language model for happy tweet generation, demonstrating that it converges to a substantially better solution than comparable baselines. We also evaluate against a variant of the proposed algorithm with a fixed regularization. The results show that in general solving the regularized problem leads to sub-optimal solutions of the original one.

\subsection{Related Work}
Our work contributes to the increasing volume of literature on bi-level optimization. Here we discuss the most relevant papers on first-order methods to give context to our contributions, and provide a complete literature comparison with other first- and second-order methods in Table~\ref{table:literature}.

\citet{kwon2023fully,shen2023penalty} first introduced the penalty reformulation in bi-level optimization to bypass the estimation of second-order terms when evaluating the bi-level hyper-gradient. Leveraging the reformulation, \citet{kwon2023fully} developed a fully first-order algorithm and showed that, under lower-level strong convexity, it converges to a stationary point of the bi-level objective with sample complexity $\widetilde\Ocal(\epsilon^{-3.5})$. \citet{shen2023penalty} considered lower-level objectives satisfying the PL condition, relaxing the strong convexity assumption. The authors proposed a nested-loop algorithm and established its convergence rate as a function of outer-loop iterations, but the complexity with respect to the total number of iterations or samples remains unknown.

The penalty reformulation has inspired several works in bi-level RL \citep{shen2024principled,yang2025bilevel,gaur2025sample}, which derived first-order expressions for the hyper-gradient and proposed ways of estimating them directly from environment samples.
For technical tractability, these works focused on regularized bi-level RL, where the (fixed) regularization induces the PL condition for the policy optimization objective. Their analyses largely mirror those of generic bi-level optimization under a lower-level PL condition and do not solve the original unregularized problem. The quality of a regularized solution may be highly undesirable when evaluated under the original objective, which we illustrate with an example in Section~\ref{sec:simulations}. In addition, due to their nested-loop structure, the algorithms in these prior works require a target precision to be specified in advance to determine the number of inner-loop iterations, making the convergence guarantees not \textit{anytime valid}: there is no guarantee that the optimality gap decreases monotonically after every iteration, and running the algorithm beyond the prescribed number of iterations does not further reduce the gap. Our work exactly addresses these limitations by proposing a single-loop actor–critic algorithm that converges to a stationary point of the unregularized bi-level RL problem through appropriate attenuation of the regularization, and we establish anytime valid finite-sample guarantees without requiring a pre-specified target precision.

\vspace{-2pt}
\section{FORMULATION}\label{sec:formulation}
\vspace{-2pt}

Consider an infinite-horizon discounted-reward MDP defined as $\Mcal_x = (\mathcal{S}, \mathcal{A}, \mathcal{P}, r_x, \gamma)$, where $x \in \mathbb{R}^d$ is an exogenous control parameter. Under a fixed $x$, $\Mcal_x$ is a standard MDP. The state space $\mathcal{S}$ and action space $\mathcal{A}$ are assumed to be finite. 
The transition kernel is denoted by $\mathcal{P}: \mathcal{S} \times \mathcal{A} \rightarrow \Delta(\mathcal{S})$, and we use $\Pcal(s'\mid s,a)$ to represent the probability that the next state is $s'$ when an agent takes action $a$ in state $s$.
The reward function $r_x: \mathcal{S} \times \mathcal{A} \rightarrow [0, 1]$ is a function of $x$. The discount factor is denoted by $\gamma \in (0,1)$. Our paper considers the setting where the transition kernel is independent of $x$, motivated by applications such as RLHF and reward shaping, where the exogenous variable modulates the reward function but not the system dynamics.

An agent learning in this MDP may not directly observe $x$, and takes actions according to a policy $\pi:\Scal\rightarrow\Delta_{\Acal}$, which we can represent as a table $\Delta_{\Acal}^{\Scal}\in\mathbb{R}^{|\Scal|\times|\Acal|}$.
Given a control-policy pair $(x,\pi)$, we measure its performance in state $s$ by the value function 
\begin{align*}
V^{x,\pi}(s)
&\triangleq \mathbb{E}\Big[\sum_{k=0}^{\infty}\gamma^k r_x(s_k, a_k) \mid s_0=s\Big],
\end{align*}
where the expectation in the first equation is taken over the trajectory $a_k\sim\pi(\cdot\mid s_k),s_{k+1}\sim\Pcal(\cdot\mid s_k,a_k)$ (we similarly omit $a_k,s_{k+1}$ in the expectations below).

Let $d_s^{\pi}\in\Delta_{\Scal}$ and $d_\rho^\pi\triangleq \mathbb{E}_{s\sim\rho}[d_s^\pi]\in\Delta_{\Scal}$ denote the discounted visitation distributions under initial state $s$ and the initial state distribution $\rho\in\Delta_{\Scal}$, respectively
\begin{align*}
d_s^{\pi}(s')\triangleq (1-\gamma)\mathbb{E}\Big[\sum_{k=0}^\infty \gamma^k\1(s_k=s')\mid s_0=s\Big].
\end{align*}
We define the expected cumulative reward under $(x,\pi)$
\begin{align*}
J(x,\pi)\hspace{-2pt}\triangleq\hspace{-2pt}\mathbb{E}_{s\sim\rho}[V^{x,\pi}(s)]\hspace{-2pt}=\hspace{-2pt}\frac{1}{1-\gamma}\mathbb{E}_{s\sim d_\rho^{\pi},a\sim\pi(\cdot\mid s)}[r_x(s,a)].
\end{align*}

If $x$ were fixed, our goal would be to find a policy that maximizes $J(x,\pi)$. In the meantime, the exogenous controller has its own objective to optimize, anticipating the best response from the policy optimization agent. We denote the controller's objective by $f:\mathbb{R}^d\times\Delta_{\Acal}^{\Scal}\rightarrow\mathbb{R}$.
Let $\Pi^\star(x)$ be the set of optimal policies under control $x$, which we note may not be singleton, and $g$ be a function that maps $\Pi^\star(x)$ to a unique optimal policy within the set (we will shortly introduce $g$). The controller’s optimization problem can then be formulated as the following bi-level program
\begin{align}
\begin{aligned}
\min_{x\in\mathbb{R}^{d}} &\; f(x,g(\Pi^\star(x)))&&\textbf{Upper-Level}&\\
\text{s.t.}&\; \Pi^\star(x)\triangleq\argmax_{\pi} J(x,\pi)&&\textbf{Lower-Level RL}\hspace{-2pt}&
\end{aligned}\label{eq:obj}
\end{align}
Our goal in this paper is to solve \eqref{eq:obj}. This is a challenging problem, as the lower-level objective lacks strong structural properties and may not admit a unique solution. To introduce additional structure and enhance the solvability, we add entropy regularization into the lower-level objective, which leads to solution uniqueness and a strong form of ``gradient domination''. We stress that the entropy-regularized formulation serves only as an intermediate tool -- our ultimate aim remains to solve the original, unregularized problem in \eqref{eq:obj}.

\begin{remark}
We discuss how the bi-level RL framework models RLHF, along with experimental results, in Section~\ref{sec:simulations}. More applications of bi-level RL with concrete problem formulations are presented in Appendix~\ref{sec:bilevel_applications}.
\end{remark}

\subsection{Entropy Regularization}

We discuss the regularized objective and its structural properties. 
Given $(x,\pi)$ and regularization weight $\tau$, we define the regularized value function $V_\tau^{x,\pi}\in\mathbb{R}^{|\Scal|}$ and expected cumulative reward $J_\tau$
\begin{align}
V_\tau^{x,\pi}\hspace{-1pt}(s)&\hspace{-2pt}\triangleq \hspace{-2pt}\mathbb{E}\Big[\sum_{k=0}^{\infty}\gamma^k(r_x(s_k, a_k)\hspace{-2pt}-\hspace{-2pt}\tau \log\pi(a_k\mid s_k)) \hspace{-2pt}\mid\hspace{-2pt} s_0\hspace{-2pt}=\hspace{-2pt}s\Big]\notag\\
&\hspace{-15pt}=\hspace{-2pt}\frac{1}{1-\gamma}\mathbb{E}_{s'\sim d_s^{\pi},\,a'\sim\pi(\cdot\mid s')}[r_x(s', a')\hspace{-2pt}+\hspace{-2pt}\tau E(\pi,s')],\hspace{-5pt}\label{eq:def_V_tau}\\
&\hspace{25pt} J_\tau(x,\pi) \triangleq  \mathbb{E}_{s\sim\rho}[V_\tau^{x,\pi}(s)],\label{eq:def_J_regularized}
\end{align}
where $E(\pi,s)=-\sum_{a}\pi(a\mid s)\log \pi(a\mid s)$ is the entropy function. Under regularization weight $\tau\leq1$, we have $|V_\tau^{x,\pi}(s)|\leq B_V$ for all $x,\pi,s$, where $B_V=\frac{1+\log|\Acal|}{1-\gamma}$.

If the initial state distribution has a full support, an assumption we will shortly introduce and impose throughout the paper, then the optimizer of $J_\tau(x,\cdot)$ is unique for any $\tau>0$. We define the operator $\pi_\tau^\star:\mathbb{R}^d\rightarrow\Delta_{\Acal}^{\Scal}$, which maps a control variable to the optimal policy that it induces
\begin{align}
\pi_\tau^\star(x)\triangleq\argmax_{\pi}J_\tau(x,\pi),\quad\forall x\in\mathbb{R}^d.\label{eq:def_pi_tau_star}
\end{align}

As we use the regularized RL problem to approximate the original one, it is important to understand how $\pi_\tau^\star(x)$ relates to $\Pi^\star(x)$. We make the connection in Lemma~\ref{lem:pi_star_unique}, under the following assumption on initial state distribution and ergodicity. The assumption is commonly made in the RL literature to guarantee that the Markov chain of states under any policy has a unique, well-defined stationary distribution \citep{mei2020global,wu2020finite,khodadadian2022finite}.
\begin{assump}[Sufficient Exploration]\label{assump:exploration}
The initial state distribution $\rho$ is bounded away from zero, i.e., there exists a constant $\rho_{\min}>0$ such that $\rho(s)\geq\rho_{\min}$ for all $s\in\Scal$.
Additionally, for any $\pi$, the Markov chain $\{s_t\}$ generated by $P^\pi$ following $s_{t+1}\sim P^\pi(\cdot\mid s_t)$ is ergodic.
\end{assump}

\begin{lem}\label{lem:pi_star_unique}
Let $\pi^\star(x)$ denote the optimal policy for the unregularized MDP with the largest (weighted) entropy
\begin{align}
\pi^\star(x)\triangleq\argmax_{\pi\in\Pi^\star(x)}\mathbb{E}_{s\sim d_\rho^\pi}[E(\pi,s)].\label{eq:def_pi_star}
\end{align}
Then, under Assumption~\ref{assump:exploration}, it holds that $\pi^\star(x)$ is unique for all $x$ and is the limit point of $\{\pi_\tau^\star(x)\}_\tau$
\begin{align*}
\pi^\star(x)=\lim_{\tau\rightarrow0}\pi_\tau^\star(x).
\end{align*}
\end{lem}
The uniqueness of $\pi^\star(x)$ is not obvious and does not directly follow from known results in convex optimization, since $\Pi^\star$ is not a convex set and the weighted entropy objective in \eqref{eq:def_pi_star} is non-concave (see Lemma 3.1 of \citet{hazan2019provably} for a proof of non-concavity). Our proof of Lemma~\ref{lem:pi_star_unique}, presented in Appendix~\ref{sec:pi_star_unique}, exploits the strict concavity of the unweighted entropy function $E(\cdot,s)$ in the interior of the simplex, as well as the fact that $\Pi^\star(x)$, though non-convex, is a connected set with special structure \citep{zeng2023connected}.

Our objective in this work is to solve the problem below, which corresponds to \eqref{eq:obj} with $g$ mapping a set to the (unique) element of the set maximizing the weighted entropy\looseness=-1
\begin{align}
\min_{x} \; \Phi(x)\triangleq f(x,\pi^\star(x)).\label{eq:obj_Phi}
\end{align}

Directly solving \eqref{eq:obj_Phi} is challenging, as the lack of strong structure in $J(x,\cdot)$ makes it difficult to find $\pi^\star(x)$. To bypass the challenge, we introduce the following regularized version of the objective, which serves as an important intermediary in our analysis. Conceptually, the algorithm to be introduced optimizes the regularized objective $\Phi_\tau$ as the regularization attenuation drives $\Phi_\tau$ toward $\Phi$.
\begin{align}
\min_x\Phi_{\tau}(x)\triangleq f(x,\pi_\tau^\star(x)).\label{eq:def_Phi_tau}
\end{align}

\section{ALGORITHM DEVELOPMENT}\label{sec:algorithm}

In this section we develop a single-loop first-order algorithm that optimizes $\Phi_{\tau}$ while gradually decaying $\tau$ to zero, thereby recovering the solution to \eqref{eq:obj_Phi}.
The algorithm operates under stochastic gradient samples of the upper-level objective, as well as state transition and reward samples from the lower-level MDP. We design the algorithm based on a penalty reformulation recently developed in \citep{kwon2023fully} for solving bi-level problems with lower-level strong convexity. Our analysis of the algorithm overcomes the unaddressed challenges specific to bi-level RL, namely, the lower-level non-convexity and the biased, non-i.i.d. stochastic gradients estimated from state-action-reward samples from an MDP. 
We begin by presenting an overview of the reformulation in our context.

\subsection{Preliminaries -- Penalty Reformulation}\label{sec:penalty_reformulation}

Our goal is to 
solve \eqref{eq:def_Phi_tau} via (stochastic) gradient descent. 
By the implicit function theorem \citep{ghadimi2018approximation}, $\nabla_x\Phi_\tau(x)$ admits the closed-form expression below when $\nabla_{\pi,\pi}^2 J_\tau(x,\pi^\star(x))$ is invertible.
\begin{align}
\nabla_x\Phi_\tau(x)&\hspace{-2pt}=\hspace{-2pt}\nabla_x f(x,\pi_\tau^\star(x))\hspace{-2pt}+\hspace{-2pt}\nabla_\pi f(x,\pi_\tau^\star(x))\frac{\partial\pi_\tau^\star(x)}{\partial x}\label{eq:hypergrad_chainrule}\\
&\hspace{-2pt}=\hspace{-2pt}\nabla_x f(x,\pi_\tau^\star(x))-\nabla_{x,\pi}^2 J_\tau(x,\pi_\tau^\star(x))\cdot\label{eq:hypergrad_Phi_tau}\\
&\hspace{30pt}\nabla_{\pi,\pi}^2 J_\tau(x,\pi_\tau^\star(x))^{-1}\cdot\nabla_{\pi}f(x,\pi_\tau^\star(x))\notag
\end{align}
Obtaining unbiased samples of $\nabla_x\Phi_\tau(x)$ based on \eqref{eq:hypergrad_Phi_tau}, however, poses significant challenges, as the expression depends on second-order Jacobian and Hessian terms that cannot be efficiently estimated from state–reward samples.
The penalty reformulation is designed to provide an alternative approach of obtaining (asymptotically) unbiased gradient estimates, only requiring first-order information.

\begin{algorithm}[!ht]
\caption{Single-Loop Actor-Critic Algorithm for Bi-Level RL}
\label{alg:main}
\begin{algorithmic}[1]
\STATE{\textbf{Initialize:} control variable $x_0$, policy parameters $\theta_0$ and $\theta_0^{\bias}$, value function estimates $\hat{V}_0,\hat{V}_0^{\bias}\in\mathbb{R}^{|\Scal|}$
}
\FOR{iteration $k=0,1,2,...$}
\STATE{\textbf{Trajectory 1}: With probability $1-\gamma$, restart the trajectory by taking $s_{k+1}\sim\rho$. With probability $\gamma$, continue following the current trajectory. Take action $a_k \sim \pi_{\theta_k}(\cdot\mid s_k)$, receive rewards $r_{x_k}(s_k,a_k)$, and observe the next state $s_{k+1}\sim \Pcal(\cdot\mid s_k,a_k)$.}
\STATE{\textbf{Trajectory 2}: 
With probability $1-\gamma$, restart the trajectory by taking $\bar{s}_{k+1}\sim\rho$. With probability $\gamma$, continue following the current trajectory. Take action $\bar{a}_k \sim \pi_{\theta_k^{\bias}}(\cdot\mid \bar{s}_k)$, receive rewards $r_{x_k}(\bar{s}_k,\bar{a}_k)$, and observe the next state $\bar{s}_{k+1}\sim \Pcal(\cdot\mid \bar{s}_k,\bar{a}_k)$.}
\STATE{Observe/Obtain $\xi_k\sim \mu$.}
\begin{tcolorbox}[colback=yellow!10, colframe=black, boxrule=0.5pt, width=\dimexpr\linewidth+4.5em\relax, left=2pt, right=2pt, top=2pt, bottom=2pt]
\STATE{Control variable update:\quad\quad \textcolor{orange}{\textbf{Upper-Level Update}}
\begin{align*}
x_{k+1}&=x_k-\zeta_k \Big(\widetilde\nabla_{x}f(x_k,\pi_{\theta_k^{\bias}},\xi_k)\notag\\
&\hspace{45pt}+\frac{1}{w_k}\big(\nabla_x r_{x_k}(s_k,a_k)-\nabla_x r_{x_k}(\bar{s}_k,\bar{a}_k)\big)\Big).
\end{align*}
}
\end{tcolorbox}
\begin{tcolorbox}[colback=blue!10, colframe=black, boxrule=0.5pt, width=\dimexpr\linewidth+6em\relax, left=2pt, right=2pt, top=2pt, bottom=2pt]
\STATE{Policy update:\qquad\qquad\quad \quad\textcolor{blue}{\textbf{Lower-Level Update}}
\begin{align}
\theta_{k+1} &= \theta_k \hspace{-2pt}+\hspace{-2pt} \alpha_k \big(r_{x_k}(s_k,a_k)\hspace{-2pt}-\hspace{-2pt}\tau_k \log\pi_{\theta_k}(a_k\hspace{-1pt}\mid\hspace{-1pt} s_k)\hspace{-2pt}+\hspace{-2pt}\gamma\hat{V}_{k}(s_{k+1})\big)\cdot\notag\\
&\hspace{120pt}\nabla_{\theta}\log\pi_{\theta_k}(a_k\mid s_k),\notag\\
\theta_{k+1}^{\bias} &= \theta_k^{\bias} \hspace{-2pt}+\hspace{-2pt} \alpha_k \Big(\big(r_{x_k}(\bar{s}_k,\bar{a}_k)\hspace{-2pt}\hspace{-1pt}-\hspace{-1pt}\tau_k \log\pi_{\theta_k}^{\bias}(\bar{a}_k\hspace{-2pt}\mid\hspace{-2pt}\bar{s}_k)\hspace{-2pt}+\hspace{-2pt}\gamma\hat{V}_{k}^{\bias}(\bar{s}_{k+1})\big)\hspace{-1pt}\cdot\notag\\
&\hspace{20pt}\nabla_{\theta}\log\pi_{\theta_k^{\bias}}(\bar{a}_k\mid \bar{s}_k)-w_k\widetilde\nabla_{\theta}f(x_k,\pi_{\theta_k^{\bias}},\xi_k)\Big),\notag\\
&\pi_{k+1} = \operatorname{softmax}(\theta_{k+1}),\quad \pi_{k+1}^{\bias} = \operatorname{softmax}(\theta_{k+1}^{\bias}).\notag
\end{align}
}
\STATE{Value function update:
\begin{align*}
\begin{aligned}
\hat{V}_{k+1} &= \Pi_{B_V}\Big(\hat{V}_{k} + \beta_k e_{s_k}\big(r_{x_k}(s_k,a_k)\\
&\hspace{30pt}-\tau_k \log\pi_{\theta_k}(a_k\hspace{-2pt}\mid\hspace{-2pt} s_k)+\gamma\hat{V}_{k}(s_{k+1})-\hat{V}_{k}(s_k)\big)\Big),\\
\hat{V}_{k+1}^{\bias} &= \Pi_{B_V}\Big(\hat{V}_{k}^{\bias} + \beta_k e_{\bar{s}_k}\big(r_{x_k}(\bar{s}_k,\bar{a}_k)\\
&\hspace{30pt}-\tau_k \log\pi_{\theta_k}^{\bias}(\bar{a}_k\hspace{-2pt}\mid\hspace{-2pt}\bar{s}_k)+\gamma\hat{V}_{k}^{\bias}(\bar{s}_{k+1})-\hat{V}_{k}^{\bias}(\bar{s}_k)\big)\Big).
\end{aligned}
\end{align*}
}
\end{tcolorbox}
\ENDFOR
\end{algorithmic}
\end{algorithm}

Recall the definition of $\pi_\tau^\star(x)$ in \eqref{eq:def_pi_tau_star}. We can re-write \eqref{eq:def_Phi_tau} as follows by introducing a constraint
\begin{align}
\min_{x,\pi} f(x,\pi) \quad\text{s.t. }J_{\tau}(x,\pi_\tau^\star(x))-J_{\tau}(x,\pi)\leq0.\label{eq:penalty_reformulation}
\end{align}
Given a positive constant $w$, we define 
\begin{align}
\Lcal_{w,\tau}(x,\pi)&\hspace{-2pt}\triangleq\hspace{-2pt} f(x,\pi)\hspace{-2pt}+\hspace{-2pt}\frac{1}{w}\big(J_{\tau}(x,\pi_\tau^\star(x))\hspace{-2pt}-\hspace{-2pt}J_{\tau}(x,\pi)\big),\label{eq:def_L_w_tau}\\
\Phi_{w,\tau}(x) &\hspace{-2pt}\triangleq \min_\pi \Lcal_{w,\tau}(x,\pi)\notag\\
&\hspace{-23pt}=\hspace{-2pt}\min_\pi f(x,\pi)\hspace{-2pt}+\hspace{-2pt}\frac{1}{w}\big(J_{\tau}(x,\pi_\tau^\star(x))\hspace{-2pt}-\hspace{-2pt}J_{\tau}(x,\pi)\big).\label{eq:def_Phi_w_tau}
\end{align}
We can regard $\Lcal_{w,\tau}$ as the Lagrangian associated with \eqref{eq:penalty_reformulation}, in which $1/w$ plays the role of the dual variable.
To solve \eqref{eq:penalty_reformulation}, it may be tempting to find a minimax saddle point of the Lagrangian using gradient descent ascent.
However, as pointed out in \citet{kwon2023fully}, the solution of \eqref{eq:penalty_reformulation} is only attained in the limit as the dual variable becomes infinitely large (i.e., $w=0$). This motivates us to treat $w$ as a parameter governed by a prescribed decay schedule towards zero, rather than as a dual variable updated via gradient ascent. It is known from \citet{kwon2023fully}[Lemma 3.1] that $\nabla_x\Phi_{w,\tau}(x)$ admits the following expression involving only first-order terms\footnote{We follow the convention and use $\nabla_x \Lcal_{w,\tau}(x,\pi_{w,\tau}^\star(x))$ to denote the partial gradient with respect to $x$ evaluated at $(x,\pi_{w,\tau}^\star(x))$, i.e., $\nabla_x \Lcal_{w,\tau}(x,\pi_{w,\tau}^\star(x))=\nabla_x \Lcal_{w,\tau}(x,\pi)\mid_{\pi=\pi_{w,\tau}^\star(x)}$. The same principle will be used for other functions, such as $f$ and $J_\tau$.}
\begin{align}
&\nabla_x \Phi_{w,\tau}(x) = \nabla_x \Lcal_{w,\tau}(x,\pi_{w,\tau}^\star(x))\notag\\
&=\nabla_x f(x,\pi_{w,\tau}^\star(x))\notag\\
&\hspace{20pt}+\frac{1}{w}\big(\nabla_x J_{\tau}(x,\pi_\tau^\star(x))-\nabla_x J_{\tau}(x,\pi_{w,\tau}^\star(x))\big),\label{eq:grad_Phi_sigma_tau}
\end{align}
where we define $\pi_{w,\tau}^\star(x)\triangleq \argmin_\pi \Lcal_{w,\tau}(x,\pi)$ for all $w,\tau> 0$. Importantly, $\nabla_x\Phi_{w,\tau}(x)$ closely tracks $\nabla_x\Phi_{\tau}(x)$ -- the distance between $\nabla_x\Phi_{w,\tau}(x)$ and $\nabla_x\Phi_{\tau}(x)$ scales linearly in $w$, a result which we establish later in Lemma~\ref{eq:grad_gap_sigma_tau}.

\subsection{Single-Loop Actor-Critic Algorithm}

We introduce $x_k$ as an estimate of the solution to the bi-level objective ($k$ is the iteration index) and design an algorithm that iteratively carries out stochastic gradient descent on $x_k$ in an approximate direction of $\nabla_x\Phi_{w_k,\tau_k}(x_k)$, estimated using online samples from the MDP. Here $w_k$ and $\tau_k$ are time-varying penalty and regularization weights. As $w_k,\tau_k$ decay according to carefully designed schedules, the surrogate objective $\Phi_{w_k,\tau_k}(x_k)$ increasingly approximates $\Phi(x_k)$, allowing us to solve the original bi-level problem.

To estimate $\nabla_x\Phi_{w_k,\tau_k}$ based on \eqref{eq:grad_Phi_sigma_tau}, we need estimates of $\nabla_x J_{\tau_k}(x_k,\pi_{\tau_k}^\star(x_k))$ and $\nabla_x J_{\tau_k}(x_k,\pi_{w_k,\tau_k}^\star(x_k))$. 
Note that $\nabla_x J_\tau(x,\pi)$ can be expressed in the simple form below
\begin{align}
\nabla_x J_\tau(x,\pi)
=\mathbb{E}_{s\sim d_\rho^{\pi},\,a\sim\pi(\cdot\mid s)}[\nabla_x r_x(s, a)].\label{eq:grad_J_eq2}
\end{align}
Given \eqref{eq:grad_J_eq2}, if we had access to an oracle that generates $\pi=\pi_{\tau_k}^\star(x)$ for any $x$, we could obtain asymptotically unbiased samples of $\nabla_x J_{\tau_k}(x_k,\pi_{\tau_k}^\star(x_k))$ by simply generating a Markovian chain $\{s_k,a_k\}$ under $\pi=\pi_{\tau_k}^\star(x_k)$ and evaluating $\nabla_x r_{x_k}(s_k,a_k)$ along the trajectory. The same can be done to estimate $\nabla_x J_{\tau_k}(x_k,\pi_{w_k,\tau_k}^\star(x_k))$ if $\pi_{w_k,\tau_k}^\star(x_k)$ were available. However, $\pi=\pi_{\tau_k}^\star(x_k)$ and $\pi_{w_k,\tau_k}^\star(x_k)$ are solutions to (augmented) lower-level RL problems and cannot be directly accessed. To overcome the oracle unavailability, we introduce the iterates $\pi_k,\pi_k^{\bias}$ as approximations of $\pi_{\tau_k}^\star(x_k),\pi_{w_k,\tau_k}^\star(x_k)$, and update them via another layer of stochastic gradient ascent.

Existing bi-level RL methods \citep{yang2025bilevel,gaur2025sample} typically introduce a nested-loop algorithmic structure when estimating these optimal policies, ensuring that $\pi_k,\pi_k^{\bias}$ from the inner loop fully converges to $\pi_{\tau_k}^\star(x_k),\pi_{w_k,\tau_k}^\star(x_k)$ up to a desired precision. However, such nested-loop algorithms are usually inconvenient to implement in practice and require setting the target precision in advance. While the convergence of nested-loop optimization algorithms may generally be (near) optimal in theory under carefully selected step sizes and inner-loop iterations, how to choose such parameters is often unclear in practical applications, making nested-loop algorithms substantially less sample efficient than their single-loop counterparts.

Our algorithm is completely single-loop and updates $\pi_k,\pi_k^{\bias}$ at the same time as $x_k$, with a larger step size (i.e., on 
a faster time scale) that approximates a nested-loop procedure. Specifically, we maintain policy parameters $\theta_k,\theta_k^{\bias}$ that encode $\pi_k,\pi_k^{\bias}$ (with notation $\pi_k=\pi_{\theta_k},\pi_k^{\bias}=\pi_{\theta_k^{\bias}}$) and iteratively refine them according to
\begin{gather*}
\theta_{k+1} = \theta_k + \alpha_k \widetilde\nabla_\theta J_{\tau_k}(x_k,\pi_{\theta_k}), \\
\theta_{k+1}^{\bias} = \theta_k^{\bias} + \alpha_k \Big(\widetilde\nabla_\theta J_{\tau_k}(x_k,\pi_{\theta_k^{\bias}}) - w_k \widetilde\nabla_\theta f(x_k,\pi_{\theta_k^{\bias}})\Big).
\end{gather*}
Here $\alpha_k$ is a step size properly balanced with the decay rates of $w_k$ and $\tau_k$, and $\widetilde\nabla_\theta f,\widetilde\nabla_\theta J_{\tau_k}$ denote stochastic samples of the true gradients with exact forms presented in line 7 of Algorithm~\ref{alg:main}. Note that $\nabla_\theta J_{\tau_k}$ admits the following closed-form expression, and can be estimated in an asymptotically unbiased way via an actor-critic approach.
\begin{align}
&\nabla_\theta J_\tau(x,\pi_\theta)= \frac{1}{1-\gamma}\mathbb{E}_{s\sim d_{\rho}^{\pi_\theta}, a\sim\pi_\theta(\cdot\mid s), s'\sim\Pcal(\cdot\mid s,a)}\label{eq:grad_J_eq1}\\
&\hspace{5pt}\Big[\big(r_x(s, a)\hspace{-2pt}-\hspace{-2pt}\tau \log \pi_\theta(a \hspace{-2pt}\mid \hspace{-2pt}s)\hspace{-2pt}+\hspace{-2pt}\gamma V_\tau^{x,\pi_\theta}(s')\big) \nabla_\theta \log \pi_\theta(a \hspace{-2pt}\mid \hspace{-2pt}s)\Big]\notag
\end{align}

Actor-critic methods sample stochastic gradients according to \eqref{eq:grad_J_eq1}, replacing the unobservable value function with an estimate updated on an even faster time scale via temporal difference learning. 
Specifically, we introduce two variables $\hat{V}_k,\hat{V}_k^{\bias}$ to track $V_{\tau_k}^{x_k,\pi_k},V_{\tau_k}^{x_k,\pi_k^{\bias}}$ and present their update rules in line 8 of Algorithm~\ref{alg:main}, where $\Pi_{B_V}:\mathbb{R}^{|\Scal|}\rightarrow\mathbb{R}^{|\Scal|}$ denotes the element-wise projection of a vector to the interval $[0,B_V]$. The projection guarantees the stability of the value function estimates, and the interval contains the true (regularization) value function under weight $\tau_k\leq1$.

The algorithm can be described at an abstract level as follows. We perform stochastic gradient descent on $x_k$ along the hyper-gradient direction. The hyper-gradient estimation relies on the solutions of lower-level RL problem and the penalty-augmented RL objective, which we obtain via a single-loop actor-critic method. Highlighted in blue in Algorithm~\ref{alg:main}, our updates of the lower-level variables $\theta_k,\theta_k^{\bias},\hat{V}_k,\hat{V}_k^{\bias}$ follow the standard actor–critic procedure, with the key distinction that we incorporate entropy regularization and gradually attenuate its weight over time. Note that despite the resemblance of our actor–critic updates to existing algorithms, the analysis is significantly more challenging in the bi-level setting. In particular, the learning targets for the lower-level policies are non-stationary, evolving both with the penalty and regularization schedules and with the updates of the upper-level variable. We overcome this challenge with a novel error decomposition scheme that tightly links the sub-optimality gap of the lower-level RL problem to that of the bi-level objective, under the shifting landscape which becomes less structured over time as the penalty and regularization weights decay. 

We present our method in Algorithm~\ref{alg:main}, where we represent the policies through tabular softmax parameterization\footnote{We consider the softmax parameterization for the purpose of mathematical analysis. The algorithm is compatible with any function approximation in practical implementations.}, i.e., the parameter $\theta\in\mathbb{R}^{|\Scal||\Acal|}$ encodes the policy $\pi_\theta$ as below
\[\pi_{\theta}(a\mid s)=\frac{\exp(\theta(s,a))}{\sum_{a'}\exp(\theta(s,a'))}.\]
Algorithm~\ref{alg:main} employs three step size parameters and two penalty/regularization weights, which are all time-decaying sequences: step size $\zeta_k$ for upper-level variable update, step size $\alpha_k$ for policy update, step size $\beta_k$ for value function update, penalty weight $w_k$, and regularization $\tau_k$. The step sizes are associated with the primal variable $(x_k)$ update, and we need to choose $\zeta_k\ll\alpha_k\ll \beta_k$ to approximate the nested-loop dynamics, where we run a large number of value function updates per policy update and a large number of policy updates per upper-level variable update. 

\begin{remark}
As introduced in Section~\ref{sec:formulation}, our work assumes that the upper-level variable influences the lower-level objective $J$ only through the reward function, while the transition kernel is fixed and independent of $x$. We make this assumption for the ease of expressing and estimating the gradient of the lower-level objective with respect to $x$ as in \eqref{eq:grad_J_eq2}. If the transition kernel $\Pcal$ were a function of $x$, the penalty-based method discussed prior to \eqref{eq:grad_J_eq2} would still remain valid, but the gradient $\nabla_x J_\tau(x,\pi)$ would become more challenging to derive and estimate using samples. 
Extending the algorithm and analysis to the setting of $x$-dependent transition kernel can be an important future work.
\end{remark}

\section{CONVERGENCE ANALYSIS}

We start the section by introducing the technical assumptions.

\begin{assump}
\label{assump:invertible}
The Hessian $\nabla_{\theta,\theta}^2 J_\tau(x,\pi_\theta)$ is invertible for all $x,\theta$ and $\tau\geq0$.
\end{assump}

Recall that the Hessian inverse is a component of the hyper-gradient \eqref{eq:hypergrad_Phi_tau}. Although our algorithm does not directly involve the Hessian, the assumption importantly guarantees the differentiability of $\Phi_\tau$. 
Let $\sigma_{\min}(\cdot)$ denote the smallest \textit{singular value} of a matrix.
Assumption~\ref{assump:invertible} implies that there exists a constant $\underline{\sigma}>0$ such that
\begin{align}
\sigma_{\min}\Big(\nabla_{\theta,\theta}^2 J_\tau(x,\pi_\theta)\Big)\geq\underline{\sigma},\quad\forall x,\theta,\tau.\label{assump:invertible:eq1}
\end{align} 
Without loss of generality, we let $\underline{\sigma}\leq 1$ for the convenience of combining terms in the analysis.
Note that \eqref{assump:invertible:eq1} should not be confused with the strong convexity of $J_\tau(x,\pi_\theta)$ with respect to $\theta$, which requires $\nabla_{\theta,\theta}^2 J_\tau(x,\pi_\theta)$ to be positive definite, i.e. its smallest \textit{eigenvalue} is positive.

\begin{assump}
\label{assump:f}
The function $f$ is differentiable, and we have access to unbiased stochastic gradient operators $\widetilde\nabla_x f(x,\pi,\xi),\widetilde\nabla_\pi f(x,\pi,\xi)$ and i.i.d. samples $\xi$ from a distribution $\mu$ such that
\begin{gather*}
\mathbb{E}_{\xi\sim\mu}[\widetilde\nabla_x f(x,\pi,\xi)]=\nabla_x f(x,\pi),\\
\mathbb{E}_{\xi\sim\mu}[\widetilde\nabla_\pi f(x,\pi,\xi)]=\nabla_\pi f(x,\pi).
\end{gather*}
Also, there exists a constant $L_f\hspace{-1pt}<\hspace{-1pt}\infty$ such that $\forall x,\hspace{-1pt}x',\hspace{-1pt}\pi,\hspace{-1pt}\pi',\hspace{-1pt}\xi$
\begin{gather*}
\|\widetilde\nabla_x f(x,\pi,\xi)\|\leq L_f,\quad \|\widetilde\nabla_\pi f(x,\pi,\xi)\|\leq L_f,\\
\|\widetilde\nabla_x f(x,\pi,\xi)\hspace{-2pt}-\hspace{-2pt}\widetilde\nabla_x f(x',\pi',\xi)\|\hspace{-2pt}\leq\hspace{-2pt} L_f(\|x\hspace{-2pt}-\hspace{-2pt}x'\|\hspace{-2pt}+\hspace{-2pt}\|\pi\hspace{-2pt}-\hspace{-2pt}\pi'\|), \\
\|\widetilde\nabla_\pi f(x,\pi,\xi)\hspace{-2pt}-\hspace{-2pt}\widetilde\nabla_\pi f(x',\pi',\xi)\|\hspace{-2pt}\leq\hspace{-2pt} L_f(\|x\hspace{-2pt}-\hspace{-2pt}x'\|\hspace{-2pt}+\hspace{-2pt}\|\pi\hspace{-2pt}-\hspace{-2pt}\pi'\|).
\end{gather*}
\end{assump}
We also assume that the minimizer of $f(\cdot,\pi)$ exists, i.e., $f(x,\pi)$ never blows up to $-\infty$. Without loss of generality, we can shift the function such that $f(x,\pi)\geq0,\, \forall x,\pi$. \looseness=-1

\begin{assump}
\label{assump:reward}
There is a constant $L_r$ such that $\forall s,\hspace{-1pt}a,\hspace{-1pt}x_1,\hspace{-1pt}x_2$
\vspace{-10pt}
\begin{gather*}
|r_{x_1}(s,a)-r_{x_2}(s,a)|\leq L_r\|x_1-x_2\|,\\ \|\nabla_{x}r_{x_1}(s,a)-\nabla_{x}r_{x_2}(s,a)\| \leq L_r\|x_1-x_2\|,\\
\|\nabla_{x,x}^2 r_{x_1}(s,a)-\nabla_{x,x}^2 r_{x_2}(s,a)\| \leq L_r\|x_1-x_2\|.
\end{gather*}
\end{assump}
\vspace{-2pt}

Assumptions~\ref{assump:f} and \ref{assump:reward} are standard regularity assumptions in the bi-level RL literature \citep{chakraborty2024parl,gaur2025sample}. 
Assumption~\ref{assump:f} imposes upper-level Lipschitz continuity, smoothness, and the ability to obtain reliable gradient samples, and Assumption~\ref{assump:reward} requires the reward function to be Lipschitz along with its gradient and Hessian.
Comparable conditions on upper- and lower-level objectives are also commonly imposed by works on generic bi-level optimization \citep{kwon2023fully,shen2023penalty}.


\begin{assump}[Regularization-Dependent PL Condition]\label{assump:PL}
Recall the definition of $\pi_{w,\tau}^\star$ after \eqref{eq:grad_Phi_sigma_tau}. We assume that the minimizer $\pi_{w,\tau}^\star(x)$ is unique for all $x$ and $w,\tau>0$. In addition, there exist constants $C_L,\bar{w}>0$ such that for all $\tau>0$, $w\leq \bar{w}$, and $x,\theta$
\begin{align}
&\|\nabla_\theta \Lcal_{w,\tau}(x,\pi_{\theta})\|^2\geq\notag\\
&\hspace{40pt}\frac{C_L\tau}{w}\Big(\Lcal_{w,\tau}(x,\pi_\theta)-\Lcal_{w,\tau}(x,\pi_{w,\tau}^\star(x))\Big).\label{assump:PL:eq1}
\end{align}
\end{assump}
This structural condition plays an important role and states that the Lagrangian defined in \eqref{eq:def_Phi_w_tau} satisfies the PL condition with respect to the policy parameter $\theta$, while the PL constant attenuates as the regularization weight becomes smaller. While we directly impose \eqref{assump:PL:eq1} for convenience and for potential broader applicability of our analysis beyond bi-level RL, we note that the condition can be derived in the tabular setting for a proper range of $w$ under assumptions of initial state distribution coverage (Assumption~\ref{assump:exploration}) and exploratory policy (i.e., there exists a constant $\underline{\pi}\in(0,1)$ such that $\pi_{\theta}(a\mid s)\geq \underline{\pi}$ for all $\theta$ that we will consider in this work). To see this, note that as $w\rightarrow0$, $L_{w,\tau}(x,\pi)$ approaches a $1/w$-scaled (and shifted) version of $J_\tau(x,\pi)$, which is known to satisfy the PL condition with $C_L=\Ocal\big(\rho_{\min}^2 (\min_{s,a}\pi_\theta(a\mid s))^2\big)$ (see \citet{mei2020global}[Lemma 15]). 
The scaling explains the dependence of the right-hand side of \eqref{assump:PL:eq1} on $1/w$. For sufficiently small $w$, the contribution of the $f$ term in $L_{w,\tau}$ remains negligible, so the PL condition on $J_\tau$ continues to dominate and allows \eqref{assump:PL:eq1} to hold.\looseness=-1
\vspace{-2pt}

\begin{thm}\label{thm:main}
Suppose Assumptions~\ref{assump:exploration}-\ref{assump:PL} hold. Consider the iterates of Algorithm~\ref{alg:main} under the step sizes and weights
\begin{gather*}
\zeta_k=\frac{\zeta_0}{(k+1)^{c_\zeta}}, \quad\alpha_k=\frac{\alpha_0}{(k+1)^{c_\alpha}},\\
\beta_k=\frac{\beta_0}{(k+1)^{c_\beta}},\quad w_k=\frac{w_0}{(k+1)^{c_w}},\quad\tau_k=\frac{\tau_0}{(k+1)^{c_\tau}},
\end{gather*}
with $c_\zeta=\frac{9}{10}, c_\alpha=\frac{1}{2},c_\beta=\frac{1}{2},c_w=\frac{3}{20},c_\tau=\frac{1}{20}$ and properly chosen $\zeta_0,\alpha_0,\beta_0,w_0,\tau_0$.
We have for all $k\geq 0$
\begin{align}
\min_{t<k}\mathbb{E}[\|\nabla_x \Phi(x_t)\|^2] 
&\leq \widetilde\Ocal\left(\frac{1}{(k+1)^{1/10}}\right).
\end{align}\label{thm:main:eq1}
\vspace{-15pt}
\end{thm}

Theorem~\ref{thm:main} establishes the convergence of the best iterate of Algorithm~\ref{alg:main} to a stationary point of the bi-level RL objective, with rate $\widetilde{\Ocal}(k^{-1/10})$. The algorithm draws two samples in each iteration, so \eqref{thm:main:eq1} translates to a sample complexity of the same order. This is the first time an algorithm has been shown to provably solve the original, unregularized bi-level RL problem. The key technical insight and novelty enabling our analysis is 1) that we recognize the RL objective as one observing a regularization-dependent PL condition with the PL constant diminishing as regularization approaches zero, 2) a multi-time-scale stochastic approximation analysis that balances the decay of step sizes and $w_k$ with that of $\tau_k$, allowing the algorithm's convergence to be established under the challenge of a time-varying optimization landscape.

We may also instantiate Algorithm~\ref{alg:main} with a constant regularization weight, thereby optimizing the regularized objective instead of the original one. In this regime, as the lower-level problem satisfies the PL condition with a fixed PL constant, we are able to establish a faster convergence rate.
\begin{thm}\label{thm:main_fixedtau}
Given any fixed regularization weight $\tau_0$, i.e., $\tau_k=\tau_0$ for all $k\geq0$, consider the iterates of Algorithm~\ref{alg:main} under the step sizes and penalty weight
\begin{gather*}
\zeta_k=\frac{\zeta_0}{(k+1)^{c_\zeta}}, \quad\alpha_k=\frac{\alpha_0}{(k+1)^{c_\alpha}},\\
\beta_k=\frac{\beta_0}{(k+1)^{c_\beta}},\quad w_k=\frac{w_0}{(k+1)^{c_w}},
\end{gather*}
with $c_\zeta=\frac{2}{3}, c_\alpha=\frac{1}{2},c_\beta=\frac{1}{2},c_w=\frac{1}{6}$ and properly chosen $\zeta_0,\alpha_0,\beta_0,w_0$. Under Assumptions~\ref{assump:exploration}-\ref{assump:PL}, we have for all $k$
\begin{align}
\min_{t<k}\mathbb{E}[\|\nabla_x \Phi_{\tau_0}(x_t)\|^2] &\leq \widetilde\Ocal\left(\frac{1}{(k+1)^{1/3}}\right).\label{thm:main_fixedtau:eq1}
\end{align}
\vspace{-15pt}
\end{thm}

Theorem~\ref{thm:main_fixedtau} shows that Algorithm~\ref{alg:main} under a constant $\tau_0$ converges to  a stationary point of the regularized bi-level objective, and again \eqref{thm:main_fixedtau:eq1} implies a sample complexity of the same order. Importantly, this rate surpasses that of the \text{F$^2$SA} algorithm in \citet{kwon2023fully}, which is $\widetilde\Ocal(k^{-1/3.5})$ derived under the stronger assumption of lower-level strong convexity. We achieve the rate improvement under weaker lower-level structure by designing a novel error decomposition scheme, which allows us to tightly bound the residuals in the policy iterates based on the PL condition.
Also note that the complexity of Algorithm~\ref{alg:main} matches the best-known complexity of a nested-loop algorithm developed in \citet{gaur2025sample} for solving the regularized bi-level RL problem.

\section{EXPERIMENTAL RESULTS}\label{sec:simulations}

We numerically verify the convergence of Algorithm~\ref{alg:main} in two environments. The first is a synthetic bi-level RL problem of goal placement in a $10\times10$ grid, which is small-scale with fully known reward and transition kernel, allowing us to precisely evaluate $\Phi(x_k)$ along the upper-level variable trajectory $\{x_k\}$ for various algorithm. The second problem is to finetune a language model to generate tweets with positive sentiment, which is large-scale and involves neural networks as the function approximation. 

\textbf{GridWorld Goal Placement.} The reward for the MDP is the negated distance between the current state and a goal position (which encourages the lower-level RL agent to reach the goal in fewest possible steps), whereas the goal is placed by the upper level decision variable. 
We design the upper-level objective, as a function of upper-level decision variable $x$ and the optimal policy $\pi^\star(x)$, to penalize deviations of the goal from the center of the grid, while encouraging $\pi^\star(x)$ to have direction biases towards moving down and right. The environment is small enough that we can exactly compute the bi-level objective $\Phi(x)$ for any $x$, and we report $\Phi(x_k)$ produced by the algorithms.
Details of the experimental setup are deferred to Appendix~\ref{sec:simulation_details}.

In this first environment, we exactly implement the proposed method according to Algorithm~\ref{alg:main} with carefully selected diminishing weights $w_k,\tau_k$, with the purpose of verifying that the proposed algorithm indeed exhibits faster convergence than comparable first-order baselines in the tabular softmax setting where our theoretical guarantees are known to hold (given that the assumptions are satisfied).

\textbf{Language Model Finetuning.} 
The goal in this experiment is to train a language model to generate tweets of positive sentiment given pairwise human preference feedback.
The problem can be formulated as a bi-level RL objective, where the upper-level variable $x$ parameterizes a reward model and the lower-level variable $\theta$ corresponds to the parameters of a language model \citep{chakraborty2024parl}. 
\begin{align*}
\min_x \quad&\mathbb{E}_{(p_0,p_1,y)\sim \psi(\pi^\star(x))}[y\log P_x(p_0>p_1)\notag\\
&\hspace{80pt}+(1-y)\log P_x(p_0<p_1)]\\
\text{s.t.}\quad& \pi^\star(x)=\arg\max_\pi J(x,\pi)
\end{align*}
Here $(p_0,p_1,y)\sim\psi(\pi^\star(x))$ denotes the two sample trajectories $p_0,p_1$ (completed texts) generated independently using the policy $\pi^\star(x)$, on which a human determines a preference $y\in\{0,1\}$, with $1$ indicating $p_0$ is preferred over $p_1$ and $0$ indicating the opposite. 
We model the preference probability $P_x$ using the Bradley-Terry model \citep{bradley1952rank}, calculated from the reward function as follows
\[P_x(p_0>p_1)=\frac{\exp (r_x(p_0))}{\exp (r_x(p_0))+\exp (r_x(p_1))}.\]
Conceptually, this objective means that the language model is optimized to generate paragraphs of positive sentiment that achieve high rewards under the reward model, while the upper-level variable updates the reward model to align with pairwise human preferences. 

We simulate human preference by a rule-based sentiment analyzer \citep{hutto2014vader}, which assigns scores in the range $[-1,1]$ to a completed body of text, with a larger value corresponding to more positive sentiment. We determine that $y=1$ if the score of the first response is higher than the second, and $y=0$ otherwise.
To scale up the proposed algorithm we employ function approximation (small 4-layer transformer-based neural networks) to parameterize both the reward and language models. We set the regularization weight to $0$ in this set of experiments for simplicity.\looseness=-1

The bi-level objective $\Phi$ is not easily computable in this environment, and we report as a proxy the negated sentiment score (given by the rule-based analyzer) of paragraphs generated under policy $\pi_{\theta_k}$.

\begin{figure}
\begin{center}
\includegraphics[width=.48\textwidth]{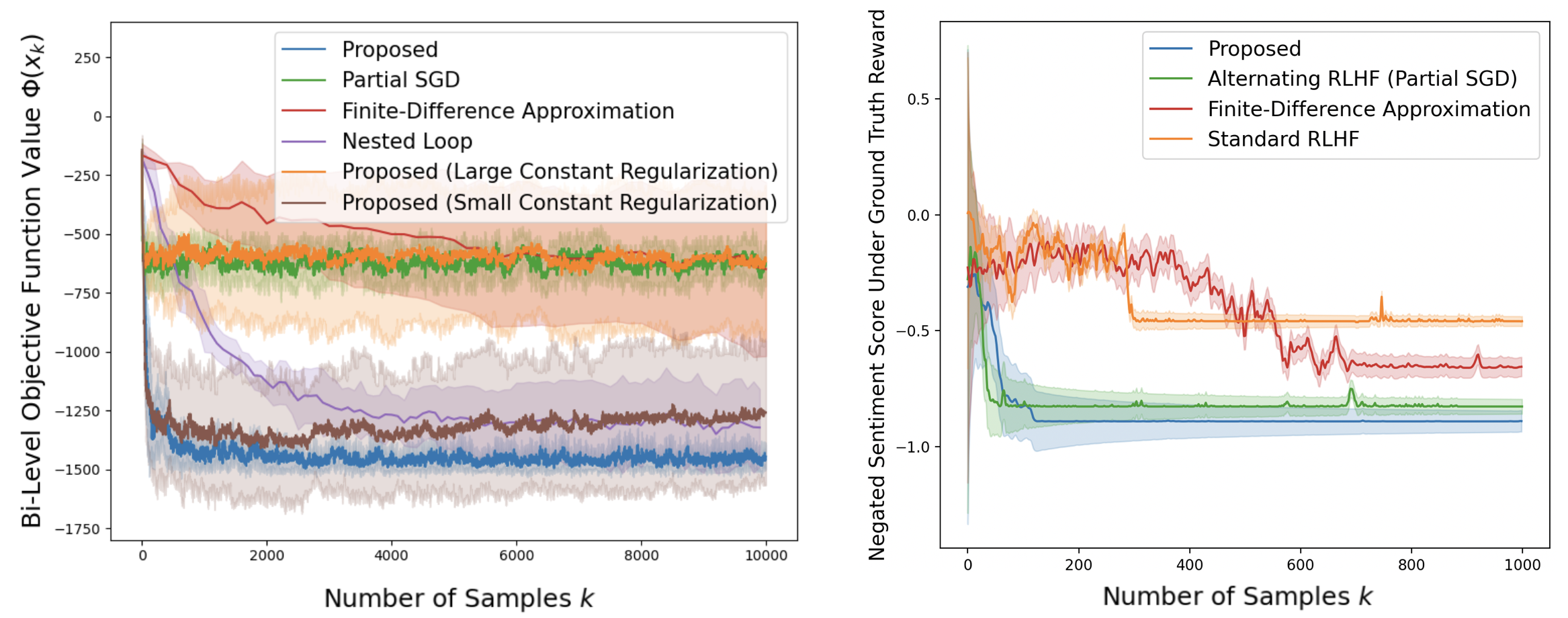}
\end{center}
\vspace{-10pt}
\caption{Algorithm Performance on GridWorld Goal Placement (Left) and Language Model Finetuning (Right)}
\vspace{-10pt}
\label{fig:GridWorld_HappyTweet}
\end{figure}

In terms of baselines, a first natural choice is an alternating (partial) gradient descent–ascent algorithm, which is commonly adopted by empirical works in bi-level RL \citep{zhang2020bi,hu2024bi}: we maintain iterates $(x_k,\theta_k)$ and update them in the direction of $\nabla_x f(x_k,\pi_{\theta_k})$ and $\nabla_\theta J(x_k,\pi_{\theta_k})$, respectively, in an alternating fashion. We refer to this approach as ``Partial SGD'', as the update direction for $x_k$ is a partial component of the full hyper-gradient $\nabla_x \Phi(x_k)$. 
In the context of the second experiment, Partial SGD resembles the standard RLHF procedure: according to standard RLHF, we draw a large number of text samples from a pre-finetuned language model to train the reward model and then train the language model with the reward model frozen; in comparison, Partial SGD alternates between reward model and language model updates with every paired text sample draw.
Note that Partial SGD may be stuck at sub-optimal solutions when the partial gradient is misaligned with the true hyper-gradient. This phenomenon is evident in Figure~\ref{fig:GridWorld_HappyTweet}, where Partial SGD fails to achieve the minimum objective value. Standard RLHF also exhibits a substantial performance gap compared with the proposed algorithm, as it fails to properly account for the coupling between reward model training and policy optimization, an observation also made in \citet{chakraborty2024parl}.\looseness=-1

We also compare against an algorithm that performs stochastic gradient descent with  gradients estimated from the chain-rule expression \eqref{eq:hypergrad_chainrule}, where $\frac{\partial\pi^\star(x)}{\partial x}$ is approximated via finite differencing. Specifically, for a scalar $x$, we approximate $\frac{\partial \pi^\star(x)}{\partial x} \approx \frac{ \pi^\star(x + \epsilon) - \pi^\star(x-\epsilon)}{2\epsilon}$,
where $\pi^\star(x+\epsilon)$ and $\pi^\star(x-\epsilon)$ are approximated by running an actor-critic algorithm within a large number of inner-loop iterations. This can be extended to vector or tensor $x$ with simultaneous perturbation stochastic approximation \citep{spall2002multivariate}. We refer to this method as ``Finite-Difference Approximation'' in Figure~\ref{fig:GridWorld_HappyTweet} and note that it is sample-inefficient, owing to both the computational overhead and the inaccuracy inherent in finite-difference-based gradient estimation.

In the first experiment, we additionally compare with a nested-loop variant of Algorithm~\ref{alg:main}, where we run a large number of inner-loop iterations to approximately compute $\pi_{w_k,\tau_k}^\star(x_k),\pi_{\tau_k}^\star(x_k)$ as $x_k$ is updated much less frequently in the outer loop. In spirit this nested-loop procedure resembles the algorithm proposed in \citet{gaur2025sample}. It has been commonly observed in the literature that single-loop algorithms often outperform their nested-loop alternatives due to more efficient use of inner-loop samples, and we see a similar relative performance gap in our setting as well.

Finally, in the first experiment, we implement the proposed algorithm under fixed regularization weights, with the purpose of showing that in general solving the regularized problem leads to a sub-optimal solution of the original problem. Specifically, we test two variants: one with a relatively large regularization weight, which overly restricts the solution and prevents progress toward the true optimum, and another with a smaller regularization weight, which still falls short of the performance attained with a decaying regularization weight. These two variants are labeled ``Proposed (Large Constant Regularization)'' and ``Proposed (Small Constant Regularization)'' in Figure~\ref{fig:GridWorld_HappyTweet} (left).



\vspace{-5pt}
\section{CONCLUSION \& FUTURE WORK}
\vspace{-5pt}

We present a novel single-loop, first-order actor-critic algorithm for bi-level RL. By introducing the entropy regularization, we enforce a special type of PL condition (with diminishing strength) at the lower level, enabling the algorithm to converge to a stationary point of the original, unregularized objective with a sample complexity of $\widetilde{\Ocal}(\epsilon^{-1/10})$. A future direction is to investigate whether the convergence rate can be improved by incorporating variance reduction and/or momentum, under which the decision variables are updated along an averaged stochastic gradient direction rather than the instantaneous one. Momentum has been shown to improve the convergence rate of penalty-based bi-level optimization algorithms under lower-level strong convexity, giving promise that it may also lead to convergence acceleration under our weaker version of PL condition.

\section*{Disclaimer}
This paper was prepared for informational purposes by the Artificial Intelligence Research group of JPMorgan Chase \& Co. and its affiliates (``JP Morgan'') and is not a product of the Research Department of JP Morgan. JP Morgan makes no representation and warranty whatsoever and disclaims all liability, for the completeness, accuracy or reliability of the information contained herein. This document is not intended as investment research or investment advice, or a recommendation, offer or solicitation for the purchase or sale of any security, financial instrument, financial product or service, or to be used in any way for evaluating the merits of participating in any transaction, and shall not constitute a solicitation under any jurisdiction or to any person, if such solicitation under such jurisdiction or to such person would be unlawful.\looseness=-1

\bibliographystyle{plainnat} 
\bibliography{references}

@inproceedings{ng1999policy,
  title={Policy invariance under reward transformations: Theory and application to reward shaping},
  author={Ng, Andrew Y and Harada, Daishi and Russell, Stuart},
  booktitle={Icml},
  volume={99},
  pages={278--287},
  year={1999},
  organization={Citeseer}
}

@article{ghadimi2018approximation,
  title={Approximation methods for bilevel programming},
  author={Ghadimi, Saeed and Wang, Mengdi},
  journal={arXiv preprint arXiv:1802.02246},
  year={2018}
}

@inproceedings{zhu2023sample,
  title={The Sample Complexity of Online Contract Design},
  author={Zhu, Banghua and Bates, Stephen and Yang, Zhuoran and Wang, Yixin and Jiao, Jiantao and Jordan, Michael I},
  booktitle={Proceedings of the 24th ACM Conference on Economics and Computation},
  pages={1188--1188},
  year={2023}
}

@article{thoma2024contextual,
  title={Contextual bilevel reinforcement learning for incentive alignment},
  author={Thoma, Vinzenz and P{\'a}sztor, Barna and Krause, Andreas and Ramponi, Giorgia and Hu, Yifan},
  journal={Advances in Neural Information Processing Systems},
  volume={37},
  pages={127369--127435},
  year={2024}
}

@inproceedings{mei2020global,
  title={On the global convergence rates of softmax policy gradient methods},
  author={Mei, Jincheng and Xiao, Chenjun and Szepesvari, Csaba and Schuurmans, Dale},
  booktitle={International conference on machine learning},
  pages={6820--6829},
  year={2020},
  organization={PMLR}
}

@article{zeng2022regularized,
  title={Regularized gradient descent ascent for two-player zero-sum Markov games},
  author={Zeng, Sihan and Doan, Thinh and Romberg, Justin},
  journal={Advances in Neural Information Processing Systems},
  volume={35},
  pages={34546--34558},
  year={2022}
}

@article{zou2019finite,
  title={Finite-sample analysis for sarsa with linear function approximation},
  author={Zou, Shaofeng and Xu, Tengyu and Liang, Yingbin},
  journal={Advances in neural information processing systems},
  volume={32},
  year={2019}
}

@article{agarwal2021theory,
  title={On the theory of policy gradient methods: Optimality, approximation, and distribution shift},
  author={Agarwal, Alekh and Kakade, Sham M and Lee, Jason D and Mahajan, Gaurav},
  journal={Journal of Machine Learning Research},
  volume={22},
  number={98},
  pages={1--76},
  year={2021}
}

@inproceedings{zeng2021decentralized,
  title={A decentralized policy gradient approach to multi-task reinforcement learning},
  author={Zeng, Sihan and Anwar, Malik Aqeel and Doan, Thinh T and Raychowdhury, Arijit and Romberg, Justin},
  booktitle={Uncertainty in Artificial Intelligence},
  pages={1002--1012},
  year={2021},
  organization={PMLR}
}

@inproceedings{shen2024principled,
  title={Principled Penalty-based Methods for Bilevel Reinforcement Learning and RLHF},
  author={Shen, Han and Yang, Zhuoran and Chen, Tianyi},
  booktitle={International Conference on Machine Learning},
  pages={44774--44799},
  year={2024},
  organization={PMLR}
}

@inproceedings{kwon2023fully,
  title={A fully first-order method for stochastic bilevel optimization},
  author={Kwon, Jeongyeol and Kwon, Dohyun and Wright, Stephen and Nowak, Robert D},
  booktitle={International Conference on Machine Learning},
  pages={18083--18113},
  year={2023},
  organization={PMLR}
}

@article{gaur2025sample,
  title={On The Sample Complexity Bounds In Bilevel Reinforcement Learning},
  author={Gaur, Mudit and Bedi, Amrit Singh and Pasupathu, Raghu and Aggarwal, Vaneet},
  journal={arXiv preprint arXiv:2503.17644},
  year={2025}
}

@inproceedings{shen2023penalty,
  title={On penalty-based bilevel gradient descent method},
  author={Shen, Han and Chen, Tianyi},
  booktitle={International Conference on Machine Learning},
  pages={30992--31015},
  year={2023},
  organization={PMLR}
}

@article{xiao2024unlocking,
  title={Unlocking Global Optimality in Bilevel Optimization: A Pilot Study},
  author={Xiao, Quan and Chen, Tianyi},
  journal={arXiv preprint arXiv:2408.16087},
  year={2024}
}

@article{xiao2025first,
  title={A First-order Generative Bilevel Optimization Framework for Diffusion Models},
  author={Xiao, Quan and Yuan, Hui and Saif, AFM and Liu, Gaowen and Kompella, Ramana and Wang, Mengdi and Chen, Tianyi},
  journal={arXiv preprint arXiv:2502.08808},
  year={2025}
}

@inproceedings{hazan2019provably,
  title={Provably efficient maximum entropy exploration},
  author={Hazan, Elad and Kakade, Sham and Singh, Karan and Van Soest, Abby},
  booktitle={International Conference on Machine Learning},
  pages={2681--2691},
  year={2019},
  organization={PMLR}
}

@article{zeng2023connected,
  title={Connected superlevel set in (deep) reinforcement learning and its application to minimax theorems},
  author={Zeng, Sihan and Doan, Thinh and Romberg, Justin},
  journal={Advances in Neural Information Processing Systems},
  volume={36},
  pages={20146--20163},
  year={2023}
}

@inproceedings{shen2019hessian,
  title={Hessian aided policy gradient},
  author={Shen, Zebang and Ribeiro, Alejandro and Hassani, Hamed and Qian, Hui and Mi, Chao},
  booktitle={International conference on machine learning},
  pages={5729--5738},
  year={2019},
  organization={PMLR}
}

@inproceedings{chakraborty2024parl,
  title={PARL: A Unified Framework for Policy Alignment in Reinforcement Learning from Human Feedback},
  author={Chakraborty, Souradip and Bedi, Amrit S and Koppel, Alec and Wang, Huazheng and Manocha, Dinesh and Wang, Mengdi and Huang, Furong},
  booktitle={ICLR},
  year={2024}
}

@inproceedings{yang2025bilevel,
  title={Bilevel Reinforcement Learning via the Development of Hyper-gradient without Lower-Level Convexity},
  author={Yang, Yan and Gao, Bin and Yuan, Ya-xiang},
  booktitle={The 28th International Conference on Artificial Intelligence and Statistics},
  year={2025}
}

@inproceedings{chen2024finding,
  title={On finding small hyper-gradients in bilevel optimization: Hardness results and improved analysis},
  author={Chen, Lesi and Xu, Jing and Zhang, Jingzhao},
  booktitle={The Thirty Seventh Annual Conference on Learning Theory},
  pages={947--980},
  year={2024},
  organization={PMLR}
}

@article{khodadadian2022finite,
  title={Finite-sample analysis of two-time-scale natural actor--critic algorithm},
  author={Khodadadian, Sajad and Doan, Thinh T and Romberg, Justin and Maguluri, Siva Theja},
  journal={IEEE Transactions on Automatic Control},
  volume={68},
  number={6},
  pages={3273--3284},
  year={2022},
  publisher={IEEE}
}

@article{wu2020finite,
  title={A finite-time analysis of two time-scale actor-critic methods},
  author={Wu, Yue Frank and Zhang, Weitong and Xu, Pan and Gu, Quanquan},
  journal={Advances in Neural Information Processing Systems},
  volume={33},
  pages={17617--17628},
  year={2020}
}

@article{yang2023achieving,
  title={Achieving $\mathcal{O}(\epsilon^{-1.5})$ Complexity in Hessian/Jacobian-free Stochastic Bilevel Optimization},
  author={Yang, Yifan and Xiao, Peiyao and Ji, Kaiyi},
  journal={Advances in Neural Information Processing Systems},
  volume={36},
  pages={39491--39503},
  year={2023}
}

@inproceedings{ye2025reward,
  title={Reward-Guided Prompt Evolving in Reinforcement Learning for LLMs},
  author={Ye, Ziyu and Agarwal, Rishabh and Liu, Tianqi and Joshi, Rishabh and Velury, Sarmishta and Le, Quoc V and Tan, Qijun and Liu, Yuan},
  booktitle={Forty-second International Conference on Machine Learning},
  year={2025}
}

@article{hu2020learning,
  title={Learning to utilize shaping rewards: A new approach of reward shaping},
  author={Hu, Yujing and Wang, Weixun and Jia, Hangtian and Wang, Yixiang and Chen, Yingfeng and Hao, Jianye and Wu, Feng and Fan, Changjie},
  journal={Advances in Neural Information Processing Systems},
  volume={33},
  pages={15931--15941},
  year={2020}
}

@article{ma2025adaptive,
  title={Adaptive Incentive Design for Markov Decision Processes with Unknown Rewards},
  author={Ma, Haoxiang and Han, Shuo and Hemida, Ahmed and Fu, Jie and others},
  journal={Transactions on Machine Learning Research},
  year={2025}
}

@inproceedings{hu2024bi,
  title={Bi-cl: A reinforcement learning framework for robots coordination through bi-level optimization},
  author={Hu, Zechen and Shishika, Daigo and Xiao, Xuesu and Wang, Xuan},
  booktitle={2024 IEEE/RSJ International Conference on Intelligent Robots and Systems (IROS)},
  pages={581--586},
  year={2024},
  organization={IEEE}
}

@inproceedings{zhang2020bi,
  title={Bi-level actor-critic for multi-agent coordination},
  author={Zhang, Haifeng and Chen, Weizhe and Huang, Zeren and Li, Minne and Yang, Yaodong and Zhang, Weinan and Wang, Jun},
  booktitle={Proceedings of the AAAI conference on artificial intelligence},
  volume={34},
  pages={7325--7332},
  year={2020}
}

@inproceedings{zhang2025understanding,
  title={Understanding Inverse Reinforcement Learning under Overparameterization: Non-Asymptotic Analysis and Global Optimality},
  author={Zhang, Ruijia and Zeng, Siliang and Li, Chenliang and Garcia, Alfredo and Hong, Mingyi},
  booktitle={International Conference on Artificial Intelligence and Statistics},
  pages={2944--2952},
  year={2025},
  organization={PMLR}
}

@inproceedings{hutto2014vader,
  title={Vader: A parsimonious rule-based model for sentiment analysis of social media text},
  author={Hutto, Clayton and Gilbert, Eric},
  booktitle={Proceedings of the international AAAI conference on web and social media},
  volume={8},
  pages={216--225},
  year={2014}
}

@article{wu2024using,
  title={Using Synthetic Data to Mitigate Unfairness and Preserve Privacy in Collaborative Machine Learning},
  author={Wu, Chia-Yuan and Curtis, Frank E and Robinson, Daniel P},
  journal={arXiv preprint arXiv:2409.09532},
  year={2024}
}

@article{bradley1952rank,
  title={Rank analysis of incomplete block designs: I. the method of paired comparisons},
  author={Bradley, Ralph Allan and Terry, Milton E},
  journal={Biometrika},
  volume={39},
  number={3/4},
  pages={324--345},
  year={1952},
  publisher={JSTOR}
}

@article{spall2002multivariate,
  title={Multivariate stochastic approximation using a simultaneous perturbation gradient approximation},
  author={Spall, James C},
  journal={IEEE transactions on automatic control},
  volume={37},
  number={3},
  pages={332--341},
  year={2002},
  publisher={IEEE}
}

\clearpage
\onecolumn
\tableofcontents

\appendix

\section{Frequently Used Notations, Equations, and Inequalities}

\begin{itemize}
    \item Besides the value function defined in \eqref{eq:def_V_tau}, we also define the regularized Q function and advantage function
    \begin{align}
    \begin{gathered}
    Q_\tau^{x,\pi}(s,a)\triangleq r_x(s,a)+\gamma\sum_{s'\in\Scal}\Pcal(s'\mid s,a)V_\tau^{x,\pi}(s),\\
    A_\tau^{x,\pi}(s,a)\triangleq Q_\tau^{x,\pi}(s,a)-\tau\log\pi(a\mid s)-V_\tau^{x,\pi}(s).
    \end{gathered}
    \label{eq:def_Q_A}
    \end{align}

    \item We define the filtration $\Fcal_{k}\triangleq\{\xi_0,\cdots,\xi_k,s_0,\cdots,s_k,a_0,\cdots,a_k,\bar{s}_0,\cdots,\bar{s}_k,\bar{a}_0,\cdots,\bar{a}_k\}$.

    \item The PL condition in Assumption~\ref{assump:PL} implies quadratic growth, i.e. for any $x$
    \begin{align}
    \Lcal_{w,\tau}(x,\pi_\theta)-\Lcal_{w,\tau}(x,\pi_{w,\tau}^\star(x))\geq\frac{C_L\tau}{4w}\|\pi_\theta-\pi_{w,\tau}^\star(x)\|^2.\label{eq:qudratic_growth}
    \end{align}
    In combination with \eqref{assump:PL:eq1}, the inequality implies
    \begin{align}
    \|\nabla_\theta \Lcal_{w,\tau}(x,\pi_\theta)\|\geq\frac{C_L\tau}{2w}\|\pi_\theta-\pi_{w,\tau}^\star(x)\|.\label{eq:PL_QG_combined}
    \end{align}

    \item We introduce the following shorthand notations that abstract the update operators in Algorithm~\ref{alg:main}. For any $x,\pi,\pi^{\bias},\theta,V,s,a,s',\bar{s},\bar{a},\xi$, we define
    \begin{gather}
    D_{w}(x,\pi,\pi^{\bias},s,a,\bar{s},\bar{a},\xi)=\widetilde\nabla_{x}f(x,\pi^{\bias},\xi)+\frac{1}{w}\Big(\nabla_x r_{x}(s,a)-\nabla_x r_{x}(\bar{s},\bar{a})\Big),\label{eq:def_D}\\
    F_{w, \tau}(x,\theta,V,s,a,s',\xi)=\big(r_{x}(s,a)-\tau \log \pi_\theta(a \mid s)+\gamma V(s')-V(s)\big)\nabla_{\theta}\log\pi_{\theta}(a\mid s)-w\widetilde\nabla_\theta f(x,\pi_\theta,\xi),\label{eq:def_F}\\
    G_{\tau}(x,\theta,V,s,a,s')=e_s\big(r_{x}(s,a)+\tau \log\pi_\theta(a\mid s)+\gamma V(s')-V(s)\big),\label{eq:def_G}
    \end{gather}
    where $e_s$ is the indicator function, i.e., the entry $s$ has a value of one and all other entries are zero.
    
    With \eqref{eq:def_D}-\eqref{eq:def_G}, we can rewrite the updates of Algorithm~\ref{alg:main}
    \begin{gather*}
    x_{k+1}=x_k-\zeta_k D_{w_k}(x_k,\pi_k,\pi_k^{\bias},s_k,a_k,\bar{s}_k,\bar{a}_k,\xi_k),\\
    \theta_{k+1}=\theta_k+\alpha_k F_{0,\tau_k}(x_k,\theta_k,\hat{V}_{k},s_k,a_k,s_k',\xi_k),\\
    \theta_{k+1}^{\bias}=\theta_k^{\bias}+\alpha_k F_{w_k, \tau_k}(x_k,\theta_k^{\bias},\hat{V}_{k}^{\bias},\bar{s}_k,\bar{a}_k,\bar{s}_k',\xi_k),\\
    \hat{V}_{k+1} = \hat{V}_{k} + \beta_k G_{\tau_k}(x_k,\theta_k,\hat{V}_{k},s_k,a_k,s_k'),\\
    \hat{V}_{k+1}^{\bias} = \hat{V}_{k}^{\bias} + \beta_k G_{\tau_k}(x_k,\theta_k^{\bias},\hat{V}_{k}^{\bias},\bar{s}_k,\bar{a}_k,\bar{s}_k').
    \end{gather*}
    
    We also define the expected versions $\bar{D},\bar{F},\bar{F}$ of the (semi-)gradient operators above. The expectation is over stochastic samples from a proper stationary distribution.
    \begin{gather}
    \bar{D}_w(x,\pi,\pi^{\bias}) \triangleq \mathbb{E}_{s \sim d_\rho^{\pi}, a \sim \pi(\cdot \mid s), \bar{s} \sim d_\rho^{\pi^{\bias}}, \bar{a} \sim \pi^{\bias}(\cdot \mid \bar{s})),\xi\sim\mu}[D_w(x,\pi,\pi^{\bias}, s, a, \bar{s},\bar{a},\xi)],\label{eq:def_bar_D}\\
    \bar{F}_{w,\tau}(x,\theta,V) \triangleq \mathbb{E}_{s \sim d_\rho^{\pi_\theta}, a \sim \pi_\theta(\cdot \mid s), s'\sim \Pcal(\cdot \mid s, a),\xi\sim\mu}[F_{w,\tau}(x,\theta, V, s, a, s',\xi)],\label{eq:def_bar_F}\\
    \bar{G}_\tau(x,\theta,V)\triangleq \mathbb{E}_{s \sim d_\rho^{\pi_\theta}, a \sim \pi_\theta(\cdot \mid s), s'\sim \Pcal(\cdot \mid s, a)}[G_\tau(x,\theta,V,s,a,s')].\label{eq:def_bar_G}
    \end{gather}

    \item It can be seen from \eqref{eq:grad_Phi_sigma_tau} that the following relationship holds for any $x$
    \begin{align}
    \nabla_{x} \Phi_{w,\tau}(x)=\bar{D}_{w}(x,\pi_\tau^\star(x),\pi_{w,\tau}^\star(x)).\label{eq:grad_Phi_barD_equality}
    \end{align}
    In addition, we have for any $x,\theta,\tau$
    \begin{gather}
    \bar{F}_{0,\tau}(x,\theta,V_\tau^{x,\pi_\theta})=\nabla_\theta J_\tau(x,\pi_\theta),\label{eq:grad_J_barF_equality}\\
    \bar{F}_{w,\tau}(x,\theta,V_\tau^{x,\pi_\theta})=w\nabla_\theta L_{w,\tau}(x,\pi_\theta).\label{eq:grad_L_barF_equality}
    \end{gather} 

    \item We define for $\tau>0$
    \begin{align}
    \ell_\tau(x)=J_\tau(x,\pi_\tau^\star(x)).\label{eq:def_ell}
    \end{align}
    Note that $\ell_\tau$ is not to be confused with $\Phi_{\tau}$ defined in \eqref{eq:def_Phi_tau}, which is the regularized upper-level objective.

    \item  While our goal is to quantify the convergence to a stationary point of the bi-level objective by the metric $\|\nabla_x \Phi(x_t)\|^2$ (or $\|\nabla_x \Phi_{\tau_0}(x_t)\|^2$ under a fixed regularization weight $\tau_0$), the analysis relies on jointly bounding the convergence of all variables of Algorithm~\ref{alg:main} through a coupled Lyapunov function, which combines the residuals shown below. To measure the convergence of $\theta_k$ and $\theta_k^{\bias}$, we consider their distance to the optimal policy in the function space. The value function estimates $\hat{V}_k,\hat{V}_k^{\bias}$ are measured by their $\ell_2$ distance to the value functions under the latest upper-level decision variable and policy. 
    \begin{gather}
    \varepsilon_k^{\theta,\bias}=w_k\Big(\Lcal_{w_k,\tau_k}(x_k,\pi_{\theta_k^{\bias}})-\Lcal_{w_k,\tau_k}(x_k,\pi_{w_k,\tau_k}^\star(x_k))\Big),\notag\\
    \varepsilon_k^{\theta}=J_{\tau_k}(x_k,\pi_{\tau_k}^\star(x_k))-J_{\tau_k}(x_k,\pi_{\theta_k}),\label{eq:def_metrics}\\
    \varepsilon_k^V=\|\hat{V}_{k}-V_{\tau_k}^{x_k,\pi_{\theta_k}}\|^2,\quad
    \varepsilon_k^{V,\bias}=\|\hat{V}_{k}^{\bias}-V_{\tau_k}^{x_k,\pi_{\theta_k^{\bias}}}\|^2.\notag
    \end{gather}

\end{itemize}

We also introduce a number of technical lemmas, which will be used in the proofs of the propositions and theorems. We defer the proofs of the lemmas to Appendix~\ref{sec:proof_lemma}.

\begin{lem}\label{lem:tau_diff}
For any $k\geq0$, we have
\[\tau_k-\tau_{k+1}\leq\frac{8\tau_k}{3(k+1)}.\]
\end{lem}

Lemma~\ref{lem:tau_diff} derives a simple bound on the rate of change of the regularization weight $\tau_k$.

\begin{lem}\label{lem:Lipschitz_V}
Define $L_V=\max\{\frac{2L_r|\Scal||\Acal|}{1-\gamma},\frac{(12+8\log|\Acal|)\sqrt{|\Scal|}}{(1-\gamma)^3}\}$. We have for all $w,\tau\leq 1$ and $x,x',\theta,\theta'$
\begin{gather}
|J_{\tau}(x,\pi_\theta)-J_{\tau}(x',\pi_{\theta'})|\leq L_V(\|x-x'\|+\|\theta-\theta'\|),\label{lem:Lipschitz_V:eq1}\\
\|\nabla_\theta J_\tau(x,\pi_{\theta})-\nabla_\theta J_\tau(x',\pi_{\theta'})\|\leq L_V(\|x-x'\|+\|\theta-\theta'\|),\label{lem:Lipschitz_V:eq2}\\
\|\nabla_x J_\tau(x,\pi)-\nabla_x J_\tau(x',\pi')\|\leq L_V(\|x-x'\|+\|\pi-\pi'\|),\label{lem:Lipschitz_V:eq2.5}\\
\|V_\tau^{x,\pi_{\theta}}-V_\tau^{x',\pi_{\theta'}}\|\leq L_V(\|x-x'\|+\|\theta-\theta'\|),\label{lem:Lipschitz_V:eq3}\\
\|\nabla_\theta V_\tau^{x,\pi_{\theta}}-\nabla_\theta V_\tau^{x',\pi_{\theta'}}\|\leq L_V(\|x-x'\|+\|\theta-\theta'\|),\label{lem:Lipschitz_V:eq4}\\
\|\nabla_{\theta,\theta}\mathbb{E}_{s\sim d_\rho^{\pi_{\theta}}}[E(\pi_{\theta},s)]\|\leq L_V.\label{lem:Lipschitz_V:eq5}
\end{gather}

In addition, there exists a bounded constant $L_{V,2}$\footnote{We skip showing the exact constant here, but note that it depends polynomially on the structural parameters of the problem.} such that for all $\tau\leq 1$ and $x,x',\theta,\theta'$
\begin{gather*}
\|\nabla_{x,\theta}^2 J_\tau(x,\pi_\theta)-\nabla_{x,\theta}^2 J_\tau(x',\pi_{\theta'})\|\leq L_{V,2}(\|x-x'\|+\|\theta-\theta'\|),\\
\|\nabla_{\theta,\theta}^2 J_\tau(x,\pi_\theta)-\nabla_{\theta,\theta}^2 J_\tau(x',\pi_{\theta'})\|\leq L_{V,2}(\|x-x'\|+\|\theta-\theta'\|).
\end{gather*}
\end{lem}

Lemma~\ref{lem:Lipschitz_V} shows that the value functions/cumulative returns are Lipschitz, and have Lipschitz continuous gradients and Hessians.

\begin{lem}\label{lem:Lipschitz_discountedvisitation}
For any $\pi,\pi'$, we have
\begin{align*}
\|d_\rho^\pi-d_\rho^{\pi'}\|\leq\frac{\gamma}{1-\gamma}\|\pi-\pi'\|.
\end{align*}
\end{lem}
Lemma~\ref{lem:Lipschitz_discountedvisitation} shows that the occupancy measure is a Lipschitz function of the policy, a well-known result in the literature. We include the proof in Section~\ref{sec:proof_lem:Lipschitz_discountedvisitation} for completeness.

\begin{lem}\label{lem:bounded_DFG}
Recall that $B_V=\frac{1+\log|\Acal|}{1-\gamma}$ is the entry-wise upper bound on the magnitude of the value function introduced in the paragraph after \eqref{eq:def_J_regularized}. We define the constants
\begin{align}
B_D=3L_r, \quad B_F=2(1+\gamma)B_V+2|\log\underline{\pi}|+2+L_r,\quad B_G=(1+\gamma)B_V+\log|\log\underline{\pi}|+1.\label{lem:bounded_DFG:eq1}
\end{align}
Suppose that the regularization parameters satisfy $w\leq\frac{L_r}{L_f},\tau\leq1$. For all $x,\theta,\theta^{\bias},s,a,s',\bar{s},\bar{a},\xi$ and $V\in\mathbb{R}^{|\Scal|}$ satisfying $V(s)\leq B_V$, we have
\begin{gather*}
\|D_w(x,\pi_{\theta},\pi_{\theta^{\bias}}, s, a, \bar{s},\bar{a},\xi)\|\leq\frac{B_D}{w},\\
\|F_{w,\tau}(x,\theta,V,s,a,s',\xi)\|\leq B_F,\\
\|G_{\tau}(x,\theta,V,s,a,s')\|\leq B_G.
\end{gather*}
\end{lem}

\begin{lem}\label{lem:Lipschitz_DFG}
We define the constants
\begin{align}
L_D=3L_r+\frac{2B_D}{1-\gamma},\quad L_F=3L_r+\frac{2}{\underline{\pi}}+|\log\underline{\pi}|+\frac{2+B_F}{1-\gamma}+4,\quad L_G=L_r+\frac{B_G}{1-\gamma}+\frac{1}{\underline{\pi}}+2.\label{lem:Lipschitz_DFG:eq1}
\end{align}
Suppose that the regularization parameters satisfy $w\leq\min\{\frac{L_r}{L_f},\frac{B_D}{(1-\gamma)L_f}\},\tau\leq1$. 
We have for all $x_1,x_2,\pi_1,\pi_1^{\bias},\pi_2,\pi_2^{\bias}$
\begin{gather*}
\|\bar{D}_w(x_1,\pi_1,\pi_1^{\bias})-\bar{D}_w(x_2,\pi_2,\pi_2^{\bias})\|\leq\frac{L_D}{w}\Big(\|x_1-x_2\|+\|\pi_1-\pi_2\|+\|\pi_1^{\bias}-\pi_2^{\bias}\|\Big)\\
\|\bar{F}_{w,\tau}(x_1,\theta_1,V_1) - \bar{F}_{w,\tau}(x_2,\theta_2,V_2)\|\leq L_F(\|x_1-x_2\|+\|\theta_1-\theta_2\|+\|V_1-V_2\|),\\
\|\bar{G}_{\tau}(x_1,\theta_1,V_1) - \bar{G}_{\tau}(x_2,\theta_2,V_2)\|\leq L_G(\|x_1-x_2\|+\|\theta_1-\theta_2\|+\|V_1-V_2\|).
\end{gather*}
\end{lem}

\begin{lem}\label{lem:Lipschitz_BR}
Recall the definition of $\pi_{w,\tau}^\star$ in Section~\ref{sec:penalty_reformulation}. For any $w_1,w_2,\tau_1,\tau_2,x_1,x_2$, we have
\begin{align*}
\|\pi_{w_1,\tau_1}^\star(x_1)-\pi_{w_2,\tau_2}^\star(x_2)\|&\leq (\frac{2L_f w_2}{C_L\tau_1} + \frac{2L_V}{C_L\tau_1})\|x_1-x_2\|+\frac{2L_f|w_1-w_2|}{C_L\tau_1}+\frac{6|\tau_1-\tau_2||\Scal|\log|\Acal|}{(1-\gamma)C_L\tau_1}.
\end{align*}
In addition, for any $w,\tau>0$, we have
\begin{align*}
\|\pi_{\tau}^\star(x)-\pi_{w,\tau}^\star(x)\|\leq\frac{2L_f w}{C_L\tau}.
\end{align*}
\end{lem}

Lemmas~\ref{lem:Lipschitz_DFG} and \ref{lem:Lipschitz_BR} show that the update operators introduced in \eqref{eq:def_D}-\eqref{eq:def_bar_G} are (approximately) bounded and Lipschitz.

\begin{lem}\label{lem:distance_pistar}
Define $L_\star=\frac{4+8\log|\Acal|}{\underline{\sigma}(1-\gamma)^4}$. For any $x$ and $\tau\geq0$, we have
\begin{align*}
\|\pi_{\tau}^\star(x)-\pi^\star(x)\|\leq L_\star\tau.
\end{align*}
\end{lem}
Lemma~\ref{lem:distance_pistar} bounds the distance between the regularized best response $\pi_\tau^\star(x)$ to $\pi^\star(x)$ defined in \eqref{eq:def_pi_star} by a linear function of $\tau$.

\begin{lem}\label{lem:Lipschitz_Lcal}
Define $L_L=L_V+L_f+\frac{L_V(C_L+2L_V)}{C_L}$. We have for all $w,\tau\leq 1$ and $x,x',\theta,\theta'$
\begin{gather}
\|\nabla_\theta\Lcal_{w,\tau}(x,\pi_\theta)-\nabla_\theta\Lcal_{w,\tau}(x',\pi_{\theta'})\|\leq \frac{L_L}{w}\|x-x'\|+\frac{L_L}{w}\|\theta-\theta'\|,\label{lem:Lipschitz_Lcal:eq1}\\
\|\nabla_x\Lcal_{w,\tau}(x,\pi_\theta)-\nabla_x\Lcal_{w,\tau}(x',\pi_{\theta'})\|\leq \frac{L_L}{w\tau}\|x-x'\|+\frac{L_L}{w}\|\theta-\theta'\|.\label{lem:Lipschitz_Lcal:eq2}
\end{gather}
\end{lem}

Lemma~\ref{lem:Lipschitz_Lcal} establishes the Lipschitz continuity of the gradients of $\Lcal_{w,\tau}$.

\begin{lem}\label{lem:Lipschitz_Phi}
Recall the definition of $\ell_\tau$ in \eqref{eq:def_ell}. We have for all $x_1,x_2$
\begin{gather*}
\|\nabla \ell_\tau(x_1)-\nabla \ell_\tau(x_2)\|\leq\Big(L_V+\frac{2L_V^2}{C_L\tau}\Big)\|x_1-x_2\|,\\
\|\nabla_x \Phi_{w,\tau}(x_1)-\nabla_x \Phi_{w,\tau}(x_2)\| \leq \Big(L_f+\frac{4L_f L_V}{C_L\tau}+\frac{4L_V^2}{C_L w\tau}\Big)\|x_1-x_2\|,\\
\|\nabla_x\Phi_\tau(x_1)-\nabla_x\Phi_\tau(x_2)\| \leq (1+\frac{2L_V}{C_L\tau})\Big(\frac{2L_fL_V}{C_L\tau}+\frac{2L_fL_VL_{V,2}}{\underline{\sigma}C_L\tau}+\frac{2L_fL_V^2L_{V,2}}{\underline{\sigma}^2C_L\tau}+\frac{2L_fL_V^2}{\underline{\sigma}C_L\tau}\Big)\|x_1-x_2\|.
\end{gather*}
Particularly, if $w,\tau\leq1$, we have
\begin{gather*}
\|\nabla \ell_\tau(x_1)-\nabla \ell_\tau(x_2)\|\leq \frac{L_\Phi}{\tau}\|x_1-x_2\|,\\
\|\nabla_x \Phi_{w,\tau}(x_1)-\nabla_x \Phi_{w,\tau}(x_2)\| \leq \frac{L_\Phi}{w\tau}\|x_1-x_2\|,
\end{gather*}
where $L_\Phi=\max\{L_V+\frac{2L_V^2}{C_L},L_f+\frac{4L_f L_V}{C_L}+\frac{4L_V^2}{C_L},\frac{4L_fL_V}{C_L\tau}+\frac{4L_fL_VL_{V,2}}{\underline{\sigma}C_L\tau}+\frac{4L_fL_V^2L_{V,2}}{\underline{\sigma}^2C_L\tau}+\frac{4L_fL_V^2}{\underline{\sigma}C_L\tau}\}$.

In addition, if $\tau\leq\frac{2L_V}{C_L}$, we have
\begin{align*}
\|\nabla_x\Phi_\tau(x_1)-\nabla_x\Phi_\tau(x_2)\|
&\leq \frac{L_\Phi}{\tau}\|x_1-x_2\|.
\end{align*}
\end{lem}

Lemma~\ref{lem:Lipschitz_Phi} establishes the Lipschitz continuity of the gradients of $\ell_\tau$, $\Phi_{w,\tau}$, and $\Phi_\tau$.

\begin{lem}\label{eq:grad_gap_sigma_tau}
We have for any $w,\tau>0$
\begin{align*}
\|\nabla_x \Phi_\tau(x)-\nabla_x \Phi_{w,\tau}(x)\|&\leq\frac{4L_f L_V w}{C_L \underline{\sigma} \tau}(L_f+\frac{2L_f L_{V,2}}{C_L\tau}),\\
\|\nabla_x\Phi(x)-\nabla_x\Phi_{w,\tau}(x)\|
&\leq\frac{4L_f L_V w}{C_L \underline{\sigma} \tau}(L_f+\frac{2L_f L_{V,2}}{C_L\tau})\notag\\
&\hspace{20pt}+\frac{L_\star L_f (L_V+1)+L_\star L_f L_{V,2}+L_\star L_V L_{V,2}+L_f L_{V,2}(4+8\log|\Acal|)}{(1-\gamma)^4\underline{\sigma}^2}\tau.
\end{align*}
\end{lem}

The lemma demonstrate how the magnitude of the difference between $\nabla_x\Phi_\tau(x)$ and $\nabla_x\Phi_{w,\tau}(x)$ and that between $\nabla_x\Phi(x)$ and $\nabla_x\Phi_{w,\tau}(x)$ scale with $w$ and $\tau$.

\begin{lem}\label{lem:Phi_tau}
For any $\tau>0$, we have   
\begin{align*}
\|\nabla_x\Phi_\tau(x)-\nabla_x\Phi(x)\|\leq \frac{L_\star L_f L_V^2 L_{V,2}\tau}{\underline{\sigma}^2}.
\end{align*}
\end{lem}

This lemma shows that distance between $\nabla_x\Phi_\tau(x)$ and $\nabla_x\Phi(x)$ is bounded by a function linear in $\tau$.

\section{Applications of Bi-Level Reinforcement Learning}\label{sec:bilevel_applications}

Bi-level RL provides a unifying framework for problems in which a high-level objective depends on the optimal solution of a lower-level policy optimization problem. In this section, we describe several representative applications that naturally fit into the bi-level RL formulation studied in this paper, besides RLHF discussed in Section~\ref{sec:simulations}.

\subsection{Reward Shaping}

In many reinforcement learning environments, the reward signal is sparse or delayed, making policy optimization difficult in practice. A common approach to address this challenge is reward shaping, where an auxiliary reward is added to the original environment reward to provide denser learning signals.

Let $r:\Scal\times\Acal\rightarrow[0,1]$ denote the original reward function (which we have no control over), and $\hat{r}_x$ an auxiliary reward parameterized by a variable $x$. The lower-level agent optimizes the shaped reward $r+\hat{r}_x$, while the upper-level optimization adjusts the shaping reward so that the induced optimal policy performs well with respect to the true task objective. This leads to the following formulation
\begin{align}
\max_x J_r\left(\pi_{r+\hat{r}_x}\right) \quad \text { s.t. } \quad \pi_{r+\hat{r}_x} \in \argmax _\pi J_{r+\hat{r}_x}(\pi),\label{eq:obj_rewardshaping}
\end{align}
where $J_r(\pi)=\mathbb{E}_\pi\left[\sum_{k=0}^{\infty} \gamma^k r\left(s_k, a_k\right)\right]$ for any policy $\pi$ and reward $r$. For \eqref{eq:obj_rewardshaping} to be well-defined, we need either $\pi^\star(x)$ to be unique or all lower-level optimal policies to evaluate to the same upper-level objective. When the problem structure does not guarantee that this is always true, we can slightly revise the formulation to include a selection mapping on the set of lower-level optimal policies as in \eqref{eq:obj} and \eqref{eq:obj_Phi}.

\subsection{Inverse Reinforcement Learning}

Inverse reinforcement learning (IRL) aims to recover a reward function that explains observed expert behavior. Given a collection of expert demonstrations $\Dcal$, assumed to be generated by an optimal policy under some unknown reward, the goal is to infer a reward such that the induced optimal policy matches the demonstrations.

Let $L(\pi,\Dcal)$ be a loss measuring the discrepancy between a policy $\pi$ and the expert data, for example via maximum likelihood or a divergence between trajectory distributions. IRL can be expressed as the bi-level problem
\begin{align*}
\min_r L\left(\pi_r, \Dcal\right) \quad \text { s.t. } \quad \pi_r \in \argmax _\pi J_r(\pi) .
\end{align*}
Here, the reward function plays the role of the upper-level decision variable, while the lower-level problem corresponds to policy optimization under that reward.
Again, we may need to include a selection mapping on the set of lower-level optimal policies as in \eqref{eq:obj} and \eqref{eq:obj_Phi}, to ensure that the problem is well-defined.

\section{Proof of Theorems}

We study a slightly simplified variant of Algorithm~\ref{alg:main}, which is presented as Algorithm~\ref{alg:analysis}. The sole distinction between the two is that Algorithm~\ref{alg:analysis} uses i.i.d. samples drawn from the stationary distribution, instead of continuously generated Markovian samples. Stochastic approximation and RL algorithms have been extensively analyzed under Markovian sampling \citep{zou2019finite,wu2020finite}, and it is well-established that Markovian samples affect convergence rates only by a logarithmic factor. This simplification enables us to concentrate on the novel aspects we introduce to bi-level RL, without being distracted by standard technical considerations related to Markovian samples.

\begin{algorithm}[!ht]
\caption{Actor Critic Algorithm for Bi-Level RL}
\label{alg:analysis}
\begin{algorithmic}[1]
\STATE{\textbf{Initialize:} control variable $x_0$, policy parameters $\theta_0$ and $\theta_0^{\bias}$, value function estimates $\hat{V}_0,\hat{V}_0^{\bias}\in\mathbb{R}^{|\Scal|}$
}
\FOR{iteration $k=0,1,2,...$}
\STATE{Trajectory 1:\\
Get samples $s_k\sim d_{\rho}^{\pi_{\theta_k}}, a_k\sim\pi_{\theta_k}(\cdot\mid s_k),s_k'\sim\Pcal(\cdot\mid s_k,a_k)$. Receive reward $r_{x_k}(s_k,a_k)$.}
\STATE{Trajectory 2:\\
Get samples $\bar{s}_k\sim d_{\rho}^{\pi_{\theta_k^{\bias}}}, \bar{a}_k\sim\pi_{\theta_k^{\bias}}(\cdot\mid s_k),\bar{s}_k'\sim\Pcal(\cdot\mid \bar{s}_k,a_k)$. Receive reward $r_{x_k}(\bar{s}_k,\bar{a}_k)$.}
\STATE{Observe/Obtain $\xi_k\sim \mu$}
\STATE{Control variable update:
\begin{align}
x_{k+1}=x_k-\zeta_k \left(\widetilde\nabla_{x}f(x_k,\pi_{\theta_k},\xi_k)+\frac{1}{w_k}\Big(\nabla_x r_{x_k}(s_k,a_k)-\nabla_x r_{x_k}(\bar{s}_k,\bar{a}_k)\Big)\right).\label{alg:analysis:leader}
\end{align}
}
\STATE{Policy update:
\begin{align}
\theta_{k+1} &= \theta_k + \alpha_k \Big(r_{x_k}(s_k,a_k)+\tau_k E(\pi_{\theta_k},s_k)+\gamma\hat{V}_{k}(s_{k+1})\Big)\nabla_{\theta}\log\pi_{\theta_k}(a_k\mid s_k),\label{alg:analysis:policy}\\
\theta_{k+1}^{\bias} &= \theta_k^{\bias} + \alpha_k \Big(\big(r_{x_k}(\bar{s}_k,\bar{a}_k)+\tau_k E(\pi_{\theta_k^{\bias}},\bar{s}_k)+\gamma\hat{V}_{k}^{\bias}(\bar{s}_{k+1})\big)\nabla_{\theta}\log\pi_{\theta_k^{\bias}}(\bar{a}_k\mid \bar{s}_k)\notag\\
&\hspace{200pt}-w_k\widetilde\nabla_{\theta}f(x_k,\pi_{\theta_k^{\bias}},\xi_k)\Big),\label{alg:analysis:policy_bias}\\
&\hspace{70pt}\pi_k = \operatorname{softmax}(\theta_k),\quad \pi_k^{\bias} = \operatorname{softmax}(\theta_k^{\bias}).
\end{align}
}
\STATE{Value function update:
\begin{align}
\begin{aligned}
\hat{V}_{k+1} &= \Pi_{B_V}\left(\hat{V}_{k} + \beta_k e_{s_k}\Big(r_{x_k}(s_k,a_k)+\tau_k E(\pi_{\theta_k},s_k)+\gamma\hat{V}_{k}(s_{k+1})-\hat{V}_{k}(s_k)\Big)\right),\\
\hat{V}_{k+1}^{\bias} &= \Pi_{B_V}\left(\hat{V}_{k}^{\bias} + \beta_k e_{\bar{s}_k}\Big(r_{x_k}(\bar{s}_k,\bar{a}_k)+\tau_k E(\pi_{\theta_k}^{\bias},\bar{s}_k)+\gamma\hat{V}_{k}^{\bias}(\bar{s}_{k+1})-\hat{V}_{k}^{\bias}(\bar{s}_k)\Big)\right).
\end{aligned}\label{alg:analysis:critic}
\end{align}
}
\ENDFOR
\end{algorithmic}
\end{algorithm}

\begin{thm}[replication of Theorem~\ref{thm:main} under i.i.d. samples]\label{thm:analysis}
Consider the iterates of Algorithm~\ref{alg:analysis} under the step sizes and weights
\[\zeta_k=\frac{\zeta_0}{(k+1)^{c_\zeta}}, \quad\alpha_k=\frac{\alpha_0}{(k+1)^{c_\alpha}},\quad\beta_k=\frac{\beta_0}{(k+1)^{c_\beta}},\quad w_k=\frac{w_0}{(k+1)^{c_w}},\quad\tau_k=\frac{\tau_0}{(k+1)^{c_\tau}},\]
with $c_\zeta=\frac{9}{10}, c_\alpha=\frac{1}{2},c_\beta=\frac{1}{2},c_w=\frac{3}{20},c_\tau=\frac{1}{20}$ and $\zeta_0,\alpha_0,\beta_0,w_0,\tau_0$ selected such that\footnote{Note that the step sizes satisfying the conditions always exist and can be found in the order of $\tau_0,w_0,\beta_0,\alpha_0,\zeta_0$ -- we first select $w_0,\tau_0$ small enough to satisfy their upper bounds; then we select $\beta_0$; then we select $\alpha_0$ with respect to $\beta_0,w_0,\tau_0$; and finally we select $\zeta_0$ with respect to $\alpha_0,\beta_0,w_0,\tau_0$.}
\begin{align}
\begin{gathered}
\zeta_0\leq\alpha_0\leq\beta_0\leq w_0\leq\tau_0\leq 1,\\
\alpha_0\leq\min\{\frac{4L_\Phi}{3L_V^2},\frac{4L_\Phi}{3L_L^2},\frac{3B_F}{8}\},\quad\beta_0\leq\min\{\frac{1-\gamma}{L_G^2},\frac{(1-\gamma)L_V^2}{8|\Scal|\log^2|\Acal|}\},\\
w_0\leq\min\{\bar{w},\frac{L_r}{L_f},\frac{B_D}{(1-\gamma)L_f}\},\quad\tau_0\leq\min\{\frac{L_f|\Scal|}{C_L},\frac{2L_V}{C_L},\frac{2L_{V,2}}{C_L}\},\\
\frac{\zeta_0}{\alpha_0}\leq\min\left\{\frac{C_L^2\tau_0^2}{1024(L_L^2+L_V^2)},\frac{C_L^2w_0^2\tau_0^2}{512L_D^2},\frac{C_L^2 w_0\tau_0^2}{128L_D L_L},\frac{(1-\gamma)C_L^2\tau_0^2}{6144L_V^2 L_D^2},\frac{B_F}{B_D}\right\},\\
\frac{\alpha_0}{\beta_0}\leq\left\{\frac{1-\gamma}{2\sqrt{6}L_V L_F},\frac{1-\gamma}{48L_V^2},\frac{1-\gamma}{8L_F^2},\frac{8(1-\gamma)(L_L^2+L_V^2)}{3L_V^2 C_L^2},\sqrt{\frac{32B_G}{B_F^2(L_L+L_V)}}, \frac{1-\gamma}{36B_F^2L_V\tau_0}\right\},\\
\frac{\alpha_0}{\beta_0^2}\leq\frac{4B_G}{11B_F^2 L_V},\quad\frac{\alpha_0}{\tau_0}\leq\frac{1}{L_V}.
\end{gathered}
\label{thm:analysis:eq1}
\end{align}
Then, under Assumptions~\ref{assump:exploration}-\ref{assump:PL}, we have for all $k\geq 0$,
\begin{align*}
\min_{t<k}\mathbb{E}[\|\nabla_x \Phi(x_t)\|^2] &\leq \frac{40}{3\zeta_0(k+1)^{1/10}}\left(\Phi_{\tau_0}(x_0)+\varepsilon_0^{\theta}+\varepsilon_0^{\theta,\bias}+\varepsilon_0^V+\varepsilon_0^{V,\bias}\right)+\Ocal\left(\frac{\log(k+1)}{(k+1)^{1/10}}\right).
\end{align*}
\end{thm}

\begin{thm}[replication of Theorem~\ref{thm:main_fixedtau} under i.i.d. samples]\label{thm:analysis_fixedtau}
Given any fixed regularization weight $\tau_0\leq\min\{1,\frac{L_f|\Scal|}{C_L},\frac{2L_V}{C_L},\frac{2L_{V,2}}{C_L}\}$\footnote{Note that Propositions~\ref{prop:policy_conv}-\ref{prop:value_conv} use the Lipschitz continuity conditions established on operator/functions such as $L_{w,\tau}$ and $\ell_\tau$ under $\tau\leq\min\{1,\frac{L_f|\Scal|}{C_L},\frac{2L_V}{C_L},\frac{2L_{V,2}}{C_L}\}$, a condition imposed so that we can present the associated Lipschitz constants in a more concise form. As the proof of Theorem 4 is based on Propositions~\ref{prop:policy_conv}-\ref{prop:value_conv}, we state that the result holds for this range of $\tau_0$. However, the same proof technique applies verbatim for any arbitrary $\tau_0>0$; only the values of the Lipschitz constants would change accordingly.}, i.e., $\tau_k=\tau_0$ for all $k\geq0$, consider the iterates of Algorithm~\ref{alg:analysis} under the step sizes and penalty weight
\[\zeta_k=\frac{\zeta_0}{(k+1)^{c_\zeta}}, \quad\alpha_k=\frac{\alpha_0}{(k+1)^{c_\alpha}},\quad\beta_k=\frac{\beta_0}{(k+1)^{c_\beta}},\quad w_k=\frac{w_0}{(k+1)^{c_w}}\]
with $c_\zeta=\frac{2}{3}, c_\alpha=\frac{1}{2},c_\beta=\frac{1}{2},c_w=\frac{1}{6}$ and $\zeta_0,\alpha_0,\beta_0,w_0$ selected such that \eqref{thm:analysis:eq1} holds.
Then, under Assumptions~\ref{assump:exploration}-\ref{assump:PL}, we have for all $k\geq 0$,
\begin{align*}
\min_{t<k}\mathbb{E}[\|\nabla_x \Phi_{\tau_0}(x_t)\|^2] &\leq \frac{20}{3\zeta_0(k+1)^{1/3}}\left(\Phi_{\tau_0}(x_0)+\varepsilon_0^{\theta}+\varepsilon_0^{\theta,\bias}+\varepsilon_0^V+\varepsilon_0^{V,\bias}\right)+\Ocal\left(\frac{\log(k+1)}{(k+1)^{1/3}}\right).
\end{align*}
\end{thm}

We break down the proofs of the theorems into two parts. First, in the following propositions, we individually establish the iteration-wise convergence of the upper-level decision variable (in Proposition~\ref{prop:control_conv} under decaying regularization and Proposition~\ref{prop:control_conv_fixedtau} under fixed regularization), policy iterates (in Propositions~\ref{prop:policy_conv}-\ref{prop:policybias_conv}), and value function estimates (in Proposition~\ref{prop:value_conv}). Then, in Sections~\ref{sec:thm:analysis}-\ref{sec:thm:analysis_fixedtau}, we combine the convergence of these variables and bound their joint convergence through a coupled Lyapunov function. The proofs of the propositions are deferred to Section~\ref{sec:proof:prop}.

\begin{prop}\label{prop:control_conv}
Under Assumptions~\ref{assump:exploration}-\ref{assump:PL} and step sizes satisfying \eqref{thm:analysis:eq1}, the iterates of Algorithm~\ref{alg:analysis} satisfy for all $k\geq0$
\begin{align*}
\frac{\zeta_k}{4}\mathbb{E}[\|\nabla_x \Phi(x_{k})\|^2]&\leq\mathbb{E}[\Phi_{\tau_k}(x_{k})-\Phi_{\tau_{k+1}}(x_{k+1})]+ \frac{2L_D^2\zeta_k}{w_k^2}\mathbb{E}[\|\pi_k-\pi_{\tau_k}^\star(x_k)\|^2] \\
&\hspace{20pt}+ \frac{2L_D^2\zeta_k}{w_k^2}\mathbb{E}[\|\pi_k^{\bias}-\pi_{w_k,\tau_k}^\star(x_k)\|^2]+\frac{L_\star^2 L_f^2 L_V^4L_{V,2}^2\zeta_k\tau_k^2}{2\underline{\sigma}^4}\\
&\hspace{20pt}+\frac{256L_f^4L_V^2L_{V,2}^2 \zeta_k w_k^2}{C_L^4\underline{\sigma}^2\tau_k^4} + \frac{B_D^2 L_\Phi \zeta_k^2}{2\tau_k w_k^2} + \frac{16L_f|\Scal|\log|\Acal|}{(1-\gamma)C_L(k+1)}.
\end{align*}
\end{prop}

\begin{prop}\label{prop:control_conv_fixedtau}
Under Assumptions~\ref{assump:exploration}-\ref{assump:PL} and step sizes satisfying \eqref{thm:analysis:eq1}, the iterates of Algorithm~\ref{alg:analysis} satisfy for all $k\geq0$
\begin{align*}
\frac{\zeta_k}{2}\mathbb{E}[\|\nabla_x \Phi_{\tau_0}(x_{k})\|^2]&\leq\mathbb{E}[\Phi_{\tau_0}(x_{k})-\Phi_{\tau_0}(x_{k+1})]+ \frac{2L_D^2\zeta_k}{w_k^2}\mathbb{E}[\|\pi_k-\pi_{\tau_0}^\star(x_k)\|^2]\\
&\hspace{20pt} + \frac{2L_D^2\zeta_k}{w_k^2}\mathbb{E}[\|\pi_k^{\bias}-\pi_{w_k,\tau_0}^\star(x_k)\|^2]+C_{2,\tau_0}\zeta_k w_k^2 + \frac{B_D^2 L_{\Phi,\tau_0} \zeta_k^2 }{2w_k^2},
\end{align*}
where $C_{2,\tau_0} = \Big(\frac{4L_f L_V}{C_L \underline{\sigma} \tau_0}(L_f+\frac{2L_f L_{V,2}}{C_L\tau_0})\Big)^2$ and $L_{\Phi,\tau_0}=(1+\frac{2L_V}{C_L\tau_0})\Big(\frac{2L_fL_V}{C_L\tau_0}+\frac{2L_fL_VL_{V,2}}{\underline{\sigma}C_L\tau_0}+\frac{2L_fL_V^2L_{V,2}}{\underline{\sigma}^2C_L\tau_0}+\frac{2L_fL_V^2}{\underline{\sigma}C_L\tau}\Big)$.
\end{prop}

\begin{prop}\label{prop:policy_conv}
Under Assumptions~\ref{assump:exploration}-\ref{assump:PL} and step sizes satisfying \eqref{thm:analysis:eq1}, the iterates of Algorithm~\ref{alg:analysis} satisfy for all $k\geq0$
\begin{align*}
\mathbb{E}[\varepsilon_{k+1}^{\theta}-\varepsilon_k^{\theta}]&\leq-\frac{\alpha_k}{8} \mathbb{E}[\|\nabla_{\theta} J_{\tau_k}(x_{k},\pi_{\theta_k})\|^2] + 2L_F^2\alpha_k\mathbb{E}[\varepsilon_k^V] +\frac{32L_V^2 \zeta_k^2}{ C_L^2\alpha_k\tau_k^2}\mathbb{E}[\|\nabla_{x} \Phi_{w_k,\tau_k}(x_k)\|^2]\notag\\
&\hspace{20pt}-\frac{ C_L^2\alpha_k\tau_k^2}{64}\mathbb{E}[\|\pi_k-\pi_{\tau_k}^\star(x_k)\|^2]+\frac{64 L_D^2 L_V^2 \zeta_k^2}{ C_L^2\alpha_k w_k^2 \tau_k^2} \mathbb{E}[\|\pi_k-\pi_{\tau_k}^\star(x_k)\|^2+\|\pi_k^{\bias}-\pi_{w_k,\tau_k}^\star(x_k)\|^2]\notag\\
&\hspace{20pt}+ \frac{2B_D^2 L_\Phi\zeta_k^2}{w_k^2\tau_k} + \frac{2B_D B_F L_V\zeta_k\alpha_k}{w_k} + \frac{B_F^2 L_V \alpha_k^2}{2}+\frac{16\log|\Acal|\tau_k}{3(1-\gamma)(k+1)}.
\end{align*}
\end{prop}

\begin{prop}\label{prop:policybias_conv}
Under Assumptions~\ref{assump:exploration}-\ref{assump:PL} and step sizes satisfying \eqref{thm:analysis:eq1}, the iterates of Algorithm~\ref{alg:analysis} satisfy for all $k\geq0$
\begin{align*}
&\mathbb{E}[\varepsilon_{k+1}^{\theta,\bias}-\varepsilon_k^{\theta,\bias}]\notag\\
&\leq -\frac{w_k^2 \alpha_k}{8} \mathbb{E}[\|\nabla_{\theta} \Lcal_{w_k,\tau_k}(x_{k},\pi_{\theta_k^{\bias}})\|^2] + 2L_F^2\alpha_k\mathbb{E}[\varepsilon_k^{V,\bias}] +\frac{32L_L^2 \zeta_k^2}{ C_L^2\alpha_k\tau_k^2}\mathbb{E}[\|\nabla_{x} \Phi_{w_k,\tau_k}(x_k)\|^2]\notag\\
&\hspace{20pt}-\frac{ C_L^2\alpha_k\tau_k^2}{64}\mathbb{E}[\|\pi_k^{\bias}-\pi_{w_k,\tau_k}^\star(x_k)\|^2]+\frac{64 L_D^2 L_L^2 \zeta_k^2}{ C_L^2\alpha_k w_k^2\tau_k^2} \mathbb{E}[\|\pi_k-\pi_{\tau_k}^\star(x_k)\|^2+\|\pi_k^{\bias}-\pi_{w_k,\tau_k}^\star(x_k)\|^2]\notag\\
&\hspace{20pt}+ \frac{2B_D^2 L_\Phi\zeta_k^2}{w_k^2\tau_k} + \frac{2B_D B_F L_L\zeta_k\alpha_k}{w_k}+\frac{B_F^2 L_L\alpha_k^2}{2}+\frac{32\log|\Acal|\tau_k}{3(1-\gamma)(k+1)}.
\end{align*}
\end{prop}

\begin{prop}[Value Function Convergence]\label{prop:value_conv}
Under Assumptions~\ref{assump:exploration}-\ref{assump:PL} and step sizes satisfying \eqref{thm:analysis:eq1}, the iterates of Algorithm~\ref{alg:analysis} satisfy for all $k\geq0$
\begin{align*}
\mathbb{E}[\varepsilon_{k+1}^V] 
&\leq \Big(1-\frac{(1-\gamma)\beta_k}{4}\Big)\mathbb{E}[\varepsilon_k^V] + \frac{12L_V^2 L_D^2\zeta_k^2}{(1-\gamma)\beta_k}\mathbb{E}[\|\pi_k-\pi_{\tau_k}^\star(x_k)\|^2+\|\pi_k^{\bias}-\pi_{w_k,\tau_k}^\star(x_k)\|^2]\notag\\
&\hspace{20pt}+\frac{6L_V^2\alpha_k^2}{(1-\gamma)\beta_k}\mathbb{E}[\|\nabla_{\theta}J_{\tau_k}(x_k,\pi_{\theta_k})\|^2]+\frac{6L_V^2\zeta_k^2}{(1-\gamma)\beta_k}\mathbb{E}[\|\nabla_{x} \Phi_{w_k,\tau_k}(x_k)\|^2]\notag\\
&\hspace{20pt}+\frac{22B_F^2 L_V\tau_0\alpha_k^2}{\alpha_0}+\frac{64L_V^2\tau_k^2}{3(1-\gamma)\beta_k(k+1)^2} + 8B_G\beta_k^2,\\
\mathbb{E}[\varepsilon_{k+1}^{V,\bias}] 
&\leq\Big(1-\frac{(1-\gamma)\beta_k}{4}\Big)\mathbb{E}[\varepsilon_k^{V,\bias}] + \frac{12L_V^2 L_D^2\zeta_k^2}{(1-\gamma)\beta_k}\mathbb{E}[\|\pi_k-\pi_{\tau_k}^\star(x_k)\|^2+\|\pi_k^{\bias}-\pi_{w_k,\tau_k}^\star(x_k)\|^2]\notag\\
&\hspace{20pt}+\frac{6L_V^2 w_k^2 \alpha_k^2}{(1-\gamma)\beta_k}\mathbb{E}[\|\nabla_{\theta} \Lcal_{w_k,\tau_k}(x_{k},\pi_{\theta_k^{\bias}})\|^2]+\frac{6L_V^2\zeta_k^2}{(1-\gamma)\beta_k}\mathbb{E}[\|\nabla_{x} \Phi_{w_k,\tau_k}(x_k)\|^2]\notag\\
&\hspace{20pt}+\frac{22B_F^2 L_V\tau_0\alpha_k^2}{\alpha_0}+\frac{64L_V^2\tau_k^2}{3(1-\gamma)\beta_k(k+1)^2} + 8B_G\beta_k^2.
\end{align*}
\end{prop}

\subsection{Proof of Theorem~\ref{thm:analysis} (replication of Theorem~\ref{thm:main} under i.i.d. samples)}\label{sec:thm:analysis}

Combining the bounds in Propositions~\ref{prop:control_conv}-\ref{prop:value_conv}, we have for any $k\geq 0$
\begin{align*}
&\frac{\zeta_k}{4}\mathbb{E}[\|\nabla_x \Phi(x_{k})\|^2]\notag\\
&\leq\mathbb{E}[\Phi_{\tau_k}(x_{k})-\Phi_{\tau_{k+1}}(x_{k+1})]+ \frac{2L_D^2\zeta_k}{w_k^2}\mathbb{E}[\|\pi_k-\pi_{\tau_k}^\star(x_k)\|^2] + \frac{2L_D^2\zeta_k}{w_k^2}\mathbb{E}[\|\pi_k^{\bias}-\pi_{w_k,\tau_k}^\star(x_k)\|^2]\\
&\hspace{20pt}+\frac{L_\star^2 L_f^2L_V^4L_{V,2}^2\zeta_k\tau_k^2}{2\underline{\sigma}^4}+\frac{256L_f^4L_V^2L_{V,2}^2 \zeta_k w_k^2}{C_L^4\underline{\sigma}^2\tau_k^4} + \frac{B_D^2 L_\Phi \zeta_k^2}{2\tau_k w_k^2} + \frac{16L_f|\Scal|\log|\Acal|}{(1-\gamma)C_L(k+1)}\notag\\
&\hspace{20pt}+\mathbb{E}[\varepsilon_k^{\theta,\bias}-\varepsilon_{k+1}^{\theta,\bias}]-\frac{w_k^2 \alpha_k}{8} \mathbb{E}[\|\nabla_{\theta} \Lcal_{w_k,\tau_k}(x_{k},\pi_{\theta_k^{\bias}})\|^2] + 2L_F^2\alpha_k\mathbb{E}[\varepsilon_k^{V,\bias}] +\frac{32L_L^2 \zeta_k^2}{ C_L^2\alpha_k\tau_k^2}\mathbb{E}[\|\nabla_{x} \Phi_{w_k,\tau_k}(x_k)\|^2]\notag\\
&\hspace{20pt}-\frac{ C_L^2\alpha_k\tau_k^2}{64}\mathbb{E}[\|\pi_k^{\bias}-\pi_{w_k,\tau_k}^\star(x_k)\|^2]+\frac{64 L_D^2 L_L^2 \zeta_k^2}{ C_L^2\alpha_k w_k^2\tau_k^2} \mathbb{E}[\|\pi_k-\pi_{\tau_k}^\star(x_k)\|^2+\|\pi_k^{\bias}-\pi_{w_k,\tau_k}^\star(x_k)\|^2]\notag\\
&\hspace{20pt}+ \frac{2B_D^2 L_\Phi\zeta_k^2}{w_k^2\tau_k} + \frac{2B_D B_F L_L\zeta_k\alpha_k}{w_k}+\frac{B_F^2 L_L\alpha_k^2}{2}+\frac{32\log|\Acal|\tau_k}{3(1-\gamma)(k+1)}\notag\\
&\hspace{20pt}+\mathbb{E}[\varepsilon_k^{\theta}-\varepsilon_{k+1}^{\theta}]-\frac{\alpha_k}{8} \mathbb{E}[\|\nabla_{\theta} J_{\tau_k}(x_{k},\pi_{\theta_k})\|^2] + 2L_F^2\alpha_k\mathbb{E}[\varepsilon_k^V] +\frac{32L_V^2 \zeta_k^2}{ C_L^2\alpha_k\tau_k^2}\mathbb{E}[\|\nabla_{x} \Phi_{w_k,\tau_k}(x_k)\|^2]\notag\\
&\hspace{20pt}-\frac{ C_L^2\alpha_k\tau_k^2}{64}\mathbb{E}[\|\pi_k-\pi_{\tau_k}^\star(x_k)\|^2]+\frac{64 L_D^2 L_V^2 \zeta_k^2}{ C_L^2\alpha_k w_k^2 \tau_k^2} \mathbb{E}[\|\pi_k-\pi_{\tau_k}^\star(x_k)\|^2+\|\pi_k^{\bias}-\pi_{w_k,\tau_k}^\star(x_k)\|^2]\notag\\
&\hspace{20pt}+ \frac{2B_D^2 L_\Phi\zeta_k^2}{w_k^2\tau_k} + \frac{2B_D B_F L_V\zeta_k\alpha_k}{w_k} + \frac{B_F^2 L_V \alpha_k^2}{2}+\frac{16\log|\Acal|\tau_k}{3(1-\gamma)(k+1)}\notag\\
&\hspace{20pt}-\mathbb{E}[\varepsilon_{k+1}^V]+\Big(1-\frac{(1-\gamma)\beta_k}{4}\Big)\mathbb{E}[\varepsilon_k^V] + \frac{12L_V^2 L_D^2\zeta_k^2}{(1-\gamma)\beta_k}\mathbb{E}[\|\pi_k-\pi_{\tau_k}^\star(x_k)\|^2+\|\pi_k^{\bias}-\pi_{w_k,\tau_k}^\star(x_k)\|^2]\notag\\
&\hspace{20pt}+\frac{6L_V^2\alpha_k^2}{(1-\gamma)\beta_k}\mathbb{E}[\|\nabla_{\theta}J_{\tau_k}(x_k,\pi_{\theta_k})\|^2]+\frac{6L_V^2\zeta_k^2}{(1-\gamma)\beta_k}\mathbb{E}[\|\nabla_{x} \Phi_{w_k,\tau_k}(x_k)\|^2]\notag\\
&\hspace{20pt}+\frac{22B_F^2 L_V\tau_0\alpha_k^2}{\alpha_0}+\frac{64L_V^2\tau_k^2}{3(1-\gamma)\beta_k(k+1)^2} + 8B_G\beta_k^2\notag\\
&\hspace{20pt}-\mathbb{E}[\varepsilon_{k+1}^{V,\bias}]+\Big(1-\frac{(1-\gamma)\beta_k}{4}\Big)\mathbb{E}[\varepsilon_k^{V,\bias}] + \frac{12L_V^2 L_D^2\zeta_k^2}{(1-\gamma)\beta_k}\mathbb{E}[\|\pi_k-\pi_{\tau_k}^\star(x_k)\|^2+\|\pi_k^{\bias}-\pi_{w_k,\tau_k}^\star(x_k)\|^2]\notag\\
&\hspace{20pt}+\frac{6L_V^2 w_k^2 \alpha_k^2}{(1-\gamma)\beta_k}\mathbb{E}[\|\nabla_{\theta} \Lcal_{w_k,\tau_k}(x_{k},\pi_{\theta_k^{\bias}})\|^2]+\frac{6L_V^2\zeta_k^2}{(1-\gamma)\beta_k}\mathbb{E}[\|\nabla_{x} \Phi_{w_k,\tau_k}(x_k)\|^2]\notag\\
&\hspace{20pt}+\frac{22B_F^2 L_V\tau_0\alpha_k^2}{\alpha_0}+\frac{64L_V^2\tau_k^2}{3(1-\gamma)\beta_k(k+1)^2} + 8B_G\beta_k^2\notag\\
&\leq \mathbb{E}[\Phi_{\tau_k}(x_{k})-\Phi_{\tau_{k+1}}(x_{k+1})+\varepsilon_k^{\theta,\bias}-\varepsilon_{k+1}^{\theta,\bias}+\varepsilon_k^{\theta}-\varepsilon_{k+1}^{\theta}+\varepsilon_k^V-\varepsilon_{k+1}^{V}+\varepsilon_k^{V,\bias}-\varepsilon_{k+1}^{V,\bias}]\notag\\
&\hspace{20pt}+\textcolor{blue}{\Big(\frac{32 (L_L^2+L_V^2) \zeta_k^2}{ C_L^2\alpha_k\tau_k^2}+\frac{12L_V^2\zeta_k^2}{(1-\gamma)\beta_k}\Big)}\mathbb{E}[\|\nabla_{x} \Phi_{w_k,\tau_k}(x_k)\|^2]+\textcolor{red}{\Big(-\frac{ \alpha_k}{8}+\frac{6L_V^2 \alpha_k^2}{(1-\gamma)\beta_k}\Big)} \mathbb{E}[\|\|\nabla_{\theta} \Lcal_{w_k,\tau_k}(x_{k},\pi_{\theta_k^{\bias}})\|^2]\notag\\
&\hspace{20pt}+\textcolor{red}{\Big(-\frac{ w_k^2\alpha_k}{8}+\frac{6L_V^2 w_k^2 \alpha_k^2}{(1-\gamma)\beta_k}\Big)} \mathbb{E}[\|\|\nabla_{\theta} \Lcal_{w_k,\tau_k}(x_{k},\pi_{\theta_k^{\bias}})\|^2]\notag\\
&\hspace{20pt}+\textcolor{red}{\Big(-\frac{ C_L^2\alpha_k\tau_k^2}{64}+\frac{2L_D^2\zeta_k}{w_k^2}+\frac{128 L_D^2 L_L^2 \zeta_k^2}{ C_L^2\alpha_k w_k^2\tau_k^2}+\frac{24 L_V^2 L_D^2\zeta_k^2}{(1-\gamma)\beta_k}\Big)}\mathbb{E}[\|\pi_k-\pi_{\tau_k}^\star(x_k)\|^2]\notag\\
&\hspace{20pt}+\textcolor{red}{\Big(-\frac{ C_L^2\alpha_k\tau_k^2}{64}+\frac{2L_D^2\zeta_k}{w_k^2}+\frac{128 L_D^2 L_L^2 \zeta_k^2}{ C_L^2\alpha_k w_k^2\tau_k^2}+\frac{24 L_V^2 L_D^2\zeta_k^2}{(1-\gamma)\beta_k}\Big)}\mathbb{E}[\|\pi_k^{\bias}-\pi_{w_k,\tau_k}^\star(x_k)\|^2]\notag\\
&\hspace{20pt}+\textcolor{red}{\Big(-\frac{(1-\gamma)\beta_k}{4}+2L_F^2\alpha_k\Big)}\mathbb{E}[\varepsilon_k^V]+\textcolor{red}{\Big(-\frac{(1-\gamma)\beta_k}{4}+2L_F^2\alpha_k\Big)}\mathbb{E}[\varepsilon_k^{V,\bias}]\notag\\
&\hspace{20pt}+\frac{256L_f^4L_V^2L_{V,2}^2 \zeta_k w_k^2}{C_L^4\underline{\sigma}^2\tau_k^4} +\frac{L_\star^2 L_f^2 L_V^4L_{V,2}^2\zeta_k\tau_k^2}{2\underline{\sigma}^4}+\frac{9B_D^2 L_\Phi\zeta_k^2}{2w_k^2\tau_k}+\frac{4B_D B_F L_L\zeta_k\alpha_k}{w_k}+48B_G\beta_k^2\notag\\
&\hspace{20pt}+\frac{128L_V^2\tau_k^2}{3(1-\gamma)\beta_k(k+1)^2}+\textcolor{blue}{\frac{16\log|\Acal|\tau_k}{(1-\gamma)(k+1)}+ \frac{16L_f|\Scal|\log|\Acal|}{(1-\gamma)C_L(k+1)}},
\end{align*}
where to get the second inequality we combine the terms $\frac{22B_F^2 L_V\tau_0\alpha_k^2}{\alpha_0}$ and $8B_G\beta_k^2$ under the condition $\alpha_0\leq\frac{4B_G\beta_0^2}{11B_F^2L_V}$, and the terms $\frac{B_F^2L_L\alpha_k^2}{2}+\frac{B_F^2L_V\alpha_k^2}{2}$ and $16B_G\beta_k^2$ under the condition $\frac{\alpha_k}{\beta_k}\leq\sqrt{\frac{32B_G}{B_F^2(L_L+L_V)}}$.

Note that the highlighted red coefficients in the inequality above are non-positive and the blue coefficients can be combined under the step size conditions 
$\zeta_0\leq\beta_0$, $\tau_k\leq\frac{L_f|\Scal|}{C_L}$, and
\begin{align*}
\frac{\alpha_0}{\beta_0}\leq\left\{\frac{1-\gamma}{48L_V^2},\frac{1-\gamma}{8L_F^2},\frac{8(1-\gamma)(L_L^2+L_V^2)}{3L_V^2 C_L^2}\right\},\quad\frac{\zeta_0}{\alpha_0}\leq\min\left\{\frac{C_L^2w_0^2\tau_0^2}{512L_D^2},\frac{C_L^2 w_0\tau_0^2}{128L_D L_L},\frac{(1-\gamma)C_L^2\tau_0^2}{6144L_V^2 L_D^2}\right\}. 
\end{align*}
This allows us to simplify the inequality and obtain
\begin{align}
&\frac{\zeta_k}{4}\mathbb{E}[\|\nabla_x \Phi(x_{k})\|^2]\notag\\
&\leq \mathbb{E}[\Phi_{\tau_k}(x_{k})-\Phi_{\tau_{k+1}}(x_{k+1})+\varepsilon_k^{\theta,\bias}-\varepsilon_{k+1}^{\theta,\bias}+\varepsilon_k^{\theta}-\varepsilon_{k+1}^{\theta}+\varepsilon_k^V-\varepsilon_{k+1}^{V}+\varepsilon_k^{V,\bias}-\varepsilon_{k+1}^{V,\bias}]\notag\\
&\hspace{20pt}+\frac{64(L_L^2+L_V^2) \zeta_k^2}{ C_L^2\alpha_k\tau_k^2}\mathbb{E}[\|\nabla_{x} \Phi_{w_k,\tau_k}(x_k)\|^2]+\frac{256L_f^4L_V^2L_{V,2}^2 \zeta_k w_k^2}{C_L^4\underline{\sigma}^2\tau_k^4} +\frac{L_\star^2 L_f^2 L_V^4L_{V,2}^2\zeta_k\tau_k^2}{2\underline{\sigma}^4}+\frac{9B_D^2 L_\Phi\zeta_k^2}{2w_k^2\tau_k}\notag\\
&\hspace{20pt}+\frac{4B_D B_F L_L\zeta_k\alpha_k}{w_k}+48B_G\beta_k^2+\frac{128L_V^2\tau_k^2}{3(1-\gamma)\beta_k(k+1)^2}+\frac{32L_f|\Scal|\log|\Acal|}{(1-\gamma)C_L(k+1)}.\label{thm:analysis:proof_eq1}
\end{align}

We can relate the second term on the right hand side of \eqref{thm:analysis:proof_eq1} to $\|\nabla_x \Phi(x_{k})\|^2$ using Lemma~\ref{eq:grad_gap_sigma_tau}
\begin{align}
&\frac{64(L_L^2+L_V^2) \zeta_k^2}{ C_L^2\alpha_k\tau_k^2}\|\nabla_{x} \Phi_{w_k,\tau_k}(x_k)\|^2\notag\\
&\leq\frac{128(L_L^2+L_V^2) \zeta_k^2}{ C_L^2\alpha_k\tau_k^2}\|\nabla_x \Phi(x_{k})\|^2+\frac{128(L_L^2+L_V^2) \zeta_k^2}{ C_L^2\alpha_k\tau_k^2}\|\nabla_x \Phi(x_{k})-\nabla_{x} \Phi_{w_k,\tau_k}(x_k)\|^2\notag\\
&\leq \frac{128(L_L^2+L_V^2) \zeta_k^2}{ C_L^2\alpha_k\tau_k^2}\|\nabla_x \Phi(x_{k})\|^2+\frac{128(L_L^2+L_V^2) \zeta_k^2}{ C_L^2\alpha_k\tau_k^2}\Big(\frac{4L_f L_V w_k}{C_L \underline{\sigma} \tau_k}(L_f+\frac{2L_f L_{V,2}}{C_L\tau_k})\notag\\
&\hspace{50pt}+\frac{L_\star L_f (L_V+1)+L_\star L_f L_{V,2}+L_\star L_V L_{V,2}+L_f L_{V,2}(4+8\log|\Acal|)}{(1-\gamma)^4\underline{\sigma}^2}\tau_k\Big)^2\notag\\
&\leq \frac{\zeta_k}{ 8}\|\nabla_x \Phi(x_{k})\|^2+\frac{128(L_L^2+L_V^2) \zeta_k^2}{ C_L^2\alpha_k\tau_k^2}\Big(\frac{4L_f L_V w_k}{C_L \underline{\sigma} \tau_k}(L_f+\frac{2L_f L_{V,2}}{C_L\tau_k})\notag\\
&\hspace{50pt}+\frac{L_\star L_f (L_V+1)+L_\star L_f L_{V,2}+L_\star L_V L_{V,2}+L_f L_{V,2}(4+8\log|\Acal|)}{(1-\gamma)^4\underline{\sigma}^2}\tau_k\Big)^2\notag\\
&= \frac{\zeta_k}{8}\|\nabla_x \Phi(x_{k})\|^2+\frac{128(L_L^2+L_V^2)\zeta_k^2}{ C_L^2\alpha_k\tau_k^2}\Big(\frac{4L_f L_V w_k}{C_L \underline{\sigma} \tau_k}(L_f+\frac{2L_f L_{V,2}}{C_L\tau_k})+C_1\tau_k\Big)^2,\label{thm:analysis:proof_eq2}
\end{align}
where we derive the third inequality from the step size condition $\frac{\zeta_0}{\alpha_0}\leq\frac{C_L^2\tau_0^2}{1024(L_L^2+L_V^2)}$, and define in the last equation $C_1=\frac{L_\star L_f (L_V+1)+L_\star L_f L_{V,2}+L_\star L_V L_{V,2}+L_f L_{V,2}(4+8\log|\Acal|)}{(1-\gamma)^4\underline{\sigma}^2}$.

Combining \eqref{thm:analysis:proof_eq1} and \eqref{thm:analysis:proof_eq2} and substituting in the step size decay rates,
\begin{align}
\frac{\zeta_k}{8}\mathbb{E}[\|\nabla_x \Phi(x_{k})\|^2]&\leq \mathbb{E}[\Phi_{\tau_k}(x_{k})-\Phi_{\tau_{k+1}}(x_{k+1})+\varepsilon_k^{\theta,\bias}-\varepsilon_{k+1}^{\theta,\bias}+\varepsilon_k^{\theta}-\varepsilon_{k+1}^{\theta}+\varepsilon_k^V-\varepsilon_{k+1}^{V}+\varepsilon_k^{V,\bias}-\varepsilon_{k+1}^{V,\bias}]\notag\\
&\hspace{20pt}+\frac{128(L_L^2+L_V^2)\zeta_k^2}{ C_L^2\alpha_k\tau_k^2}\Big(\frac{4L_f L_V w_k}{C_L \underline{\sigma} \tau_k}(L_f+\frac{2L_f L_{V,2}}{C_L\tau_k})+C_1\tau_k\Big)^2\notag\\
&\hspace{20pt}+\frac{256L_f^4L_V^2L_{V,2}^2 \zeta_k w_k^2}{C_L^4\underline{\sigma}^2\tau_k^4}+\frac{L_\star^2 L_f^2 L_V^4L_{V,2}^2\zeta_k\tau_k^2}{2\underline{\sigma}^4}+\frac{9B_D^2 L_\Phi\zeta_k^2}{2w_k^2\tau_k}\notag\\
&\hspace{20pt}+\frac{4B_D B_F L_L\zeta_k\alpha_k}{w_k}+48B_G\beta_k^2+\frac{128L_V^2\tau_k^2}{3(1-\gamma)\beta_k(k+1)^2}+\frac{32L_f|\Scal|\log|\Acal|}{(1-\gamma)C_L(k+1)}\label{thm:analysis:proof_eq3}\\
&\leq \mathbb{E}[\Phi_{\tau_k}(x_{k})-\Phi_{\tau_{k+1}}(x_{k+1})+\varepsilon_k^{\theta,\bias}-\varepsilon_{k+1}^{\theta,\bias}+\varepsilon_k^{\theta}-\varepsilon_{k+1}^{\theta}+\varepsilon_k^V-\varepsilon_{k+1}^{V}+\varepsilon_k^{V,\bias}-\varepsilon_{k+1}^{V,\bias}]\notag\\
&\hspace{20pt}+\Ocal\left(\frac{1}{(k+1)}\right).\notag
\end{align}

Re-arranging the terms and summing over iterations,
\begin{align}
\sum_{t=0}^{k-1}\frac{\zeta_0}{8(t+1)^{9/10}} \mathbb{E}[\|\nabla_x \Phi(x_t)\|^2] &\leq \Phi_{\tau_0}(x_0)+\varepsilon_0^{\theta}+\varepsilon_0^{\theta,\bias}+\varepsilon_0^V+\varepsilon_0^{V,\bias}+\Ocal\left(\sum_{t=0}^{k-1}\frac{1}{(t+1)}\right).\label{thm:analysis:proof_eq4}
\end{align}

The following inequalities on the summation of step sizes are standard results in the literature and easy to verify.
\begin{gather*}
\sum_{t=0}^{k-1} \frac{1}{t+1} \leqslant \frac{\log (k+1)}{\log(2)},\\
\sum_{t=0}^k \frac{1}{(t+1)^u}\geq\frac{(1-\frac{1}{2^{1-u}})(k+1)^{1-u}}{1-u} ,\quad\forall u\in(0,1)
\end{gather*}
and with $u=9/10$,
\[\sum_{t=0}^k \frac{1}{(t+1)^{9/10}}\geq\frac{0.06(k+1)^{1/10}}{1/10}=\frac{3(k+1)^{1/10}}{5}.\]

This allows us to further simplify \eqref{thm:analysis:proof_eq4}
\begin{align*}
\min_{t<k}\mathbb{E}[\|\nabla_x \Phi(x_t)\|^2] &\leq \frac{1}{\sum_{t=0}^{k-1}\frac{\zeta_0}{8(t+1)^{9/10}}}\sum_{t=0}^{k-1}\frac{\zeta_0}{8(t+1)^{9/10}} \mathbb{E}[\|\nabla_x \Phi(x_t)\|^2] \notag\\
&\leq \frac{40}{3\zeta_0(k+1)^{1/10}}\left(\Phi_{\tau_0}(x_0)+\varepsilon_0^{\theta}+\varepsilon_0^{\theta,\bias}+\varepsilon_0^V+\varepsilon_0^{V,\bias}+\Ocal(\log(k+1))\right).
\end{align*}

\qed

\subsection{Proof of Theorem~\ref{thm:analysis_fixedtau} (replication of Theorem~\ref{thm:main_fixedtau} under i.i.d. samples)}\label{sec:thm:analysis_fixedtau}

We combine Propositions~\ref{prop:control_conv_fixedtau} and \ref{prop:policy_conv}. Note that $\tau_k=\tau_0$.
We can follow a line of analysis identical to what leads to \eqref{thm:analysis:proof_eq1} in the proof of Theorem~\ref{thm:analysis} and show the following inequality 
\begin{align}
\frac{\zeta_k}{2}\mathbb{E}[\|\nabla_x \Phi_{\tau_0}(x_{k})\|^2]&\leq \mathbb{E}[\Phi_{\tau_0}(x_{k})-\Phi_{\tau_0}(x_{k+1})+\varepsilon_k^{\theta,\bias}-\varepsilon_{k+1}^{\theta,\bias}+\varepsilon_k^{\theta}-\varepsilon_{k+1}^{\theta}+\varepsilon_k^V-\varepsilon_{k+1}^{V}+\varepsilon_k^{V,\bias}-\varepsilon_{k+1}^{V,\bias}]\notag\\
&\hspace{20pt}+\frac{64(L_L^2+L_V^2) \zeta_k^2}{ C_L^2\alpha_k\tau_0^2}\mathbb{E}[\|\nabla_{x} \Phi_{w_k,\tau_k}(x_k)\|^2]+C_{2,\tau_0}\zeta_k w_k^2+\frac{4B_D^2 L_\Phi\zeta_k^2}{w_k^2\tau_0}+\frac{B_D^2 L_{\Phi,\tau_0}\zeta_k^2}{2w_k^2}\notag\\
&\hspace{20pt}+\frac{4B_D B_F L_L\zeta_k\alpha_k}{w_k}+48B_G\beta_k^2+\frac{128L_V^2\tau_0^2}{3(1-\gamma)\beta_k(k+1)^2}+\frac{48\log|\Acal|\tau_0}{3(1-\gamma)(k+1)}.\label{thm:analysis_fixedtau:proof_eq1}
\end{align}
The step size conditions required to show \eqref{thm:analysis_fixedtau:proof_eq1} are
\begin{gather*}
\frac{\alpha_0}{\beta_0}\leq\left\{\frac{1-\gamma}{48L_V^2},\frac{1-\gamma}{8L_F^2},\frac{8(1-\gamma)(L_L^2+L_V^2)}{3L_V^2 C_L^2},\sqrt{\frac{32B_G}{B_F^2(L_L+L_V)}}\right\},\quad\frac{\alpha_0}{\beta_0^2}\leq\frac{4B_G}{11B_F^2 L_V},\\
\zeta_0\leq\beta_0,\quad\frac{\zeta_0}{\alpha_0}\leq\min\left\{\frac{C_L^2w_0^2\tau_0^2}{512L_D^2},\frac{C_L^2 w_0\tau_0^2}{128L_D L_L},\frac{(1-\gamma)C_L^2\tau_0^2}{6144L_V^2 L_D^2}\right\}. 
\end{gather*}

We can relate the second term on the right hand side of \eqref{thm:analysis_fixedtau:proof_eq1} to $\|\nabla_x \Phi_{\tau_0}(x_{k})\|^2$ using Lemma~\ref{eq:grad_gap_sigma_tau}
\begin{align}
&\frac{64(L_L^2+L_V^2) \zeta_k^2}{ C_L^2\alpha_k\tau_0^2}\|\nabla_{x} \Phi_{w_k,\tau_k}(x_k)\|^2\notag\\
&\leq\frac{128(L_L^2+L_V^2) \zeta_k^2}{ C_L^2\alpha_k\tau_0^2}\|\nabla_x \Phi_{\tau_0}(x_{k})\|^2+\frac{128(L_L^2+L_V^2) \zeta_k^2}{ C_L^2\alpha_k\tau_0^2}\|\nabla_x \Phi_{\tau_0}(x_{k})-\nabla_{x} \Phi_{w_k,\tau_0}(x_k)\|^2\notag\\
&\leq \frac{128(L_L^2+L_V^2) \zeta_k^2}{ C_L^2\alpha_k\tau_0^2}\|\nabla_x \Phi_{\tau_0}(x_{k})\|^2+\frac{128(L_L^2+L_V^2) \zeta_k^2}{ C_L^2\alpha_k\tau_0^2}\Big(\frac{4L_f L_V w_k}{C_L \underline{\sigma} \tau_0}(L_f+\frac{2L_f L_{V,2}}{C_L\tau_0})\Big)^2\notag\\
&\leq \frac{\zeta_k}{ 4}\|\nabla_x \Phi_{\tau_0}(x_{k})\|^2+\frac{128(L_L^2+L_V^2) \zeta_k^2 w_k^2}{ C_L^2\alpha_k\tau_0^2}\Big(\frac{4L_f L_V}{C_L \underline{\sigma} \tau_0}(L_f+\frac{2L_f L_{V,2}}{C_L\tau_0})\Big)^2\notag\\
&= \frac{\zeta_k}{ 4}\|\nabla_x \Phi_{\tau_0}(x_{k})\|^2+\frac{128(L_L^2+L_V^2) C_{2,\tau_0}\zeta_k^2 w_k^2}{ C_L^2\alpha_k\tau_0^2},\label{thm:analysis_fixedtau:proof_eq2}
\end{align}
where the third inequality is due to the step size condition $\frac{\zeta_0}{\alpha_0}\leq\frac{C_L^2\tau_0^2}{512(L_L^2+L_V^2)}$.

Combining \eqref{thm:analysis_fixedtau:proof_eq1} and \eqref{thm:analysis_fixedtau:proof_eq2} and plugging in the step size decay rates,
\begin{align*}
\frac{\zeta_k}{4}\mathbb{E}[\|\nabla_x \Phi_{\tau_0}(x_{k})\|^2]&\leq \mathbb{E}[\Phi_{\tau_0}(x_{k})-\Phi_{\tau_0}(x_{k+1})+\varepsilon_k^{\theta,\bias}-\varepsilon_{k+1}^{\theta,\bias}+\varepsilon_k^{\theta}-\varepsilon_{k+1}^{\theta}+\varepsilon_k^V-\varepsilon_{k+1}^{V}+\varepsilon_k^{V,\bias}-\varepsilon_{k+1}^{V,\bias}]\notag\\
&\hspace{20pt}+\frac{128(L_L^2+L_V^2)C_{2,\tau_0}\zeta_k^2 w_k^2}{C_L^2\alpha_k\tau_0^2}+2C_{2,\tau_0}\zeta_k w_k^2+\frac{4B_D^2 L_\Phi\zeta_k^2}{w_k^2\tau_0}+\frac{B_D^2 L_{\Phi,\tau_0}\zeta_k^2}{2w_k^2}\notag\\
&\hspace{20pt}+\frac{4B_D B_F L_L\zeta_k\alpha_k}{w_k}+48B_G\beta_k^2+\frac{128L_V^2\tau_0^2}{3(1-\gamma)\beta_k(k+1)^2}+\frac{48\log|\Acal|\tau_0}{3(1-\gamma)(k+1)}\notag\\
&\leq \mathbb{E}[\Phi_{\tau_0}(x_{k})-\Phi_{\tau_0}(x_{k+1})+\varepsilon_k^{\theta,\bias}-\varepsilon_{k+1}^{\theta,\bias}+\varepsilon_k^{\theta}-\varepsilon_{k+1}^{\theta}+\varepsilon_k^V-\varepsilon_{k+1}^{V}+\varepsilon_k^{V,\bias}-\varepsilon_{k+1}^{V,\bias}]\notag\\
&\hspace{20pt}+\Ocal\left(\frac{\zeta_k^2 w_k^2}{\alpha_k}+\zeta_k w_k^2+\frac{\zeta_k^2}{w_k^2}+\frac{\zeta_k\alpha_k}{w_k}+\beta_k^2+\frac{1}{\beta_k(k+1)^2}+\frac{1}{k+1}\right)\notag\\
&\leq \mathbb{E}[\Phi_{\tau_0}(x_{k})-\Phi_{\tau_0}(x_{k+1})+\varepsilon_k^{\theta,\bias}-\varepsilon_{k+1}^{\theta,\bias}+\varepsilon_k^{\theta}-\varepsilon_{k+1}^{\theta}+\varepsilon_k^V-\varepsilon_{k+1}^{V}+\varepsilon_k^{V,\bias}-\varepsilon_{k+1}^{V,\bias}]\notag\\
&\hspace{20pt}+\Ocal\left(\frac{1}{k+1}\right).
\end{align*}

Re-arranging the terms and summing over iterations,
\begin{align}
\sum_{t=0}^{k-1}\frac{\zeta_0}{4(t+1)^{2/3}} \mathbb{E}[\|\nabla_x \Phi_{\tau_0}(x_t)\|^2] &\leq \Phi_{\tau_0}(x_0)+\varepsilon_0^{\theta}+\varepsilon_0^{\theta,\bias}+\varepsilon_0^V+\varepsilon_0^{V,\bias}+\Ocal\left(\sum_{t=0}^{k-1}\frac{1}{(t+1)}\right).\label{thm:analysis_fixedtau:proof_eq3}
\end{align}

Again, the following inequalities on the summation of step sizes are standard results in the literature.
\begin{gather*}
\sum_{t=0}^{k-1} \frac{1}{t+1} \leqslant \frac{\log (k+1)}{\log(2)},\\
\sum_{t=0}^k \frac{1}{(t+1)^u}\geq\frac{(1-\frac{1}{2^{1-u}})(k+1)^{1-u}}{1-u} ,\quad\forall u\in(0,1)
\end{gather*}
and with $u=2/3$,
\[\sum_{t=0}^k \frac{1}{(t+1)^{2/3}}\geq\frac{0.2(k+1)^{1/3}}{1/3}=\frac{3(k+1)^{1/3}}{5}.\]

This allows us to further simplify \eqref{thm:analysis_fixedtau:proof_eq3}
\begin{align*}
\min_{t<k}\mathbb{E}[\|\nabla_x \Phi_{\tau_0}(x_t)\|^2] &\leq \frac{1}{\sum_{t=0}^{k-1}\frac{\zeta_0}{4(t+1)^{2/3}}}\sum_{t=0}^{k-1}\frac{\zeta_0}{4(t+1)^{2/3}} \mathbb{E}[\|\nabla_x \Phi_{\tau_0}(x_t)\|^2] \notag\\
&\leq \frac{20}{3\zeta_0(k+1)^{1/3}}\left(\Phi_{\tau_0}(x_0)+\varepsilon_0^{\theta}+\varepsilon_0^{\theta,\bias}+\varepsilon_0^V+\varepsilon_0^{V,\bias}+\Ocal(\log(k+1))\right).
\end{align*}

\qed

\section{Proof of Propositions}\label{sec:proof:prop}

\subsection{Proof of Proposition~\ref{prop:control_conv}}

We know from Lemma~\ref{lem:Lipschitz_Phi} that under the step size condition $\tau_k\leq\frac{2L_V}{C_L}$, the objective $\Phi_{\tau_k}$ has $ \frac{L_\Phi}{\tau_k}$-Lipschitz gradients. This implies
\begin{align*}
&\Phi_{\tau_k}(x_{k+1})-\Phi_{\tau_k}(x_{k})\notag\\
&\leq \langle\nabla_x \Phi_{\tau_k}(x_{k}), x_{k+1}-x_k\rangle + \frac{L_\Phi}{2\tau_k}\|x_{k+1}-x_k\|^2\notag\\
&= -\zeta_k \langle\nabla_x \Phi_{\tau_k}(x_{k}),D_{w_k}(x_k,\pi_k,\pi_k^{\bias},s_k,a_k,\bar{s}_k,\bar{a}_k,\xi_k)\rangle+\frac{L_\Phi \zeta_k^2}{2\tau_k}\|D_{w_k}(x_k,\pi_k,\pi_k^{\bias},s_k,a_k,\bar{s}_k,\bar{a}_k,\xi_k)\|^2\notag\\
&=-\zeta_k \|\nabla_x \Phi_{\tau_k}(x_{k})\|^2 + \zeta_k\langle\nabla_x \Phi_{\tau_k}(x_{k}),\nabla_x \Phi_{\tau_k}(x_{k})-D_{w_k}(x_k,\pi_k,\pi_k^{\bias},s_k,a_k,\bar{s}_k,\bar{a}_k,\xi_k)\rangle\notag\\
&\hspace{20pt}+\frac{L_\Phi \zeta_k^2}{2\tau_k}\|D_{w_k}(x_k,\pi_k,\pi_k^{\bias},s_k,a_k,\bar{s}_k,\bar{a}_k,\xi_k)\|^2.
\end{align*}

By the law of total expectation,
\begin{align}
&\mathbb{E}[\Phi_{\tau_k}(x_{k+1})-\Phi_{\tau_k}(x_{k})]\notag\\
&\leq -\zeta_k \mathbb{E}[\|\nabla_x \Phi_{\tau_k}(x_{k})\|^2] +\frac{L_\Phi \zeta_k^2}{2\tau_k}\mathbb{E}[\|D_{w_k}(x_k,\pi_k,\pi_k^{\bias},s_k,a_k,\bar{s}_k,\bar{a}_k,\xi_k)\|^2]\notag\\
&\hspace{20pt}+ \zeta_k\mathbb{E}[\langle\nabla_x \Phi_{\tau_k}(x_{k}),\nabla_x \Phi_{\tau_k}(x_{k})-\mathbb{E}[D_{w_k}(x_k,\pi_k,\pi_k^{\bias},s_k,a_k,\bar{s}_k,\bar{a}_k,\xi_k)\mid\Fcal_{k-1}]\rangle]\notag\\
&=-\zeta_k \mathbb{E}[\|\nabla_x \Phi_{\tau_k}(x_{k})\|^2] +\frac{L_\Phi \zeta_k^2}{2\tau_k}\mathbb{E}[\|D_{w_k}(x_k,\pi_k,\pi_k^{\bias},s_k,a_k,\bar{s}_k,\bar{a}_k,\xi_k)\|^2]\notag\\
&\hspace{20pt}+ \zeta_k\mathbb{E}[\langle\nabla_x \Phi_{\tau_k}(x_{k}),\nabla_x \Phi_{\tau_k}(x_{k})-\bar{D}_{w_k}(x_k,\pi_k,\pi_k^{\bias})\rangle]\notag\\
&\leq-\zeta_k \mathbb{E}[\|\nabla_x \Phi_{\tau_k}(x_{k})\|^2] +\frac{B_D^2 L_\Phi \zeta_k^2}{2\tau_k w_k^2}+\frac{\zeta_k}{2}\mathbb{E}[\|\nabla_x \Phi_{\tau_k}(x_{k})\|^2]+\frac{\zeta_k}{2}\mathbb{E}[\|\nabla_x \Phi_{\tau_k}(x_{k})-\bar{D}_{w_k}(x_k,\pi_k,\pi_k^{\bias})\|^2]\notag\\
&=-\frac{\zeta_k}{2}\mathbb{E}[\|\nabla_x \Phi_{\tau_k}(x_{k})\|^2] +\frac{B_D^2 L_\Phi \zeta_k^2}{2\tau_k w_k^2}\notag\\
&\hspace{20pt}+\frac{\zeta_k}{2}\mathbb{E}\left[\Big\|\Big(\nabla_x \Phi_{\tau_k}(x_{k})-\nabla_x \Phi_{w_k,\tau_k}(x_{k})\Big)+\Big(\bar{D}_{w_k}(x_k,\pi_{\tau_k}^\star(x_k),\pi_{w_k,\tau_k}^\star(x_k))-\bar{D}_{w_k}(x_k,\pi_k,\pi_k^{\bias})\Big)\Big\|^2\right]\notag\\
&\leq -\frac{\zeta_k}{2}\mathbb{E}[\|\nabla_x \Phi_{\tau_k}(x_{k})\|^2]+\zeta_k \mathbb{E}[\|\bar{D}_{w_k}(x_k,\pi_{\tau_k}^\star(x_k),\pi_{w_k,\tau_k}^\star(x_k))-\bar{D}_{w_k}(x_k,\pi_k,\pi_k^{\bias})\|^2]\notag\\
&\hspace{20pt}+\zeta_k \mathbb{E}[\|\nabla_x \Phi_{\tau_k}(x_{k})-\nabla_x \Phi_{w_k,\tau_k}(x_{k})\|^2]+\frac{B_D^2 L_\Phi \zeta_k^2}{2\tau_k w_k^2},\label{prop:control_conv:proof_eq1}
\end{align}
where the second inequality applies Lemma~\ref{lem:bounded_DFG} and the last equation follows from \eqref{eq:grad_Phi_barD_equality}.

To bound the second term on the right hand side of \eqref{prop:control_conv:proof_eq1}, we apply Lemma~\ref{lem:Lipschitz_DFG}
\begin{align}
&\|\bar{D}_{w_k}(x_k,\pi_{\tau_k}^\star(x_k),\pi_{w_k,\tau_k}^\star(x_k))-\bar{D}_{w_k}(x_k,\pi_k,\pi_k^{\bias})\|^2\notag\\
&\leq \frac{L_D^2}{w_k^2}\Big(\|\pi_k-\pi_{\tau_k}^\star(x_k)\|+\|\pi_k^{\bias}-\pi_{w_k,\tau_k}^\star(x_k)\|\Big)^2\notag\\
&\leq \frac{2L_D^2}{w_k^2}\|\pi_k-\pi_{\tau_k}^\star(x_k)\|^2+\frac{2L_D^2}{w_k^2}\|\pi_k^{\bias}-\pi_{w_k,\tau_k}^\star(x_k)\|^2.\label{prop:control_conv:proof_eq2}
\end{align}

For the third term of \eqref{prop:control_conv:proof_eq1}, we have from Lemma~\ref{eq:grad_gap_sigma_tau}
\begin{align}
\|\nabla_x\Phi_{\tau_k}(x_k)-\nabla_x\Phi_{w_k,\tau_k}(x_k)\|^2
&\leq\Big(\frac{4L_f L_V w_k}{C_L \underline{\sigma} \tau_k}(L_f+\frac{2L_f L_{V,2}}{C_L\tau_k})\Big)^2.\label{prop:control_conv:proof_eq3}
\end{align}

Substituting \eqref{prop:control_conv:proof_eq2} and \eqref{prop:control_conv:proof_eq3} into \eqref{prop:control_conv:proof_eq1},
\begin{align}
&\mathbb{E}[\Phi_{\tau_k}(x_{k+1})-\Phi_{\tau_k}(x_k)]\notag\\
&\leq -\frac{\zeta_k}{2}\mathbb{E}[\|\nabla_x \Phi_{\tau_k}(x_{k})\|^2]+ \frac{2L_D^2\zeta_k}{w_k^2}\mathbb{E}[\|\pi_k-\pi_{\tau_k}^\star(x_k)\|^2] + \frac{2L_D^2\zeta_k}{w_k^2}\mathbb{E}[\|\pi_k^{\bias}-\pi_{w_k,\tau_k}^\star(x_k)\|^2]\notag\\
&\hspace{20pt}+\zeta_k (\frac{4L_f L_V w_k}{C_L \underline{\sigma} \tau_k}(L_f+\frac{2L_f L_{V,2}}{C_L\tau_k})\Big)^2+ \frac{B_D^2 L_\Phi \zeta_k^2 }{2\tau_k w_k^2}.\label{prop:control_conv:proof_eq4}
\end{align}

Under the choice of step size $\tau_k\leq \frac{2L_{V,2}}{C_L}$, we can further simplify \eqref{prop:control_conv:proof_eq4}
\begin{align}
&\mathbb{E}[\Phi_{\tau_k}(x_{k+1})-\Phi_{\tau_k}(x_k)]\notag\\
&\leq -\frac{\zeta_k}{2}\mathbb{E}[\|\nabla_x \Phi_{\tau_k}(x_{k})\|^2]+ \frac{2L_D^2\zeta_k}{w_k^2}\mathbb{E}[\|\pi_k-\pi_{\tau_k}^\star(x_k)\|^2] + \frac{2L_D^2\zeta_k}{w_k^2}\mathbb{E}[\|\pi_k^{\bias}-\pi_{w_k,\tau_k}^\star(x_k)\|^2]\notag\\
&\hspace{20pt}+\zeta_k (\frac{4L_f L_V w_k}{C_L \underline{\sigma} \tau_k}\cdot\frac{4L_f L_{V,2}}{C_L\tau_k}\Big)^2+ \frac{B_D^2 L_\Phi \zeta_k^2 }{2\tau_k w_k^2}\notag\\
&=-\frac{\zeta_k}{2}\mathbb{E}[\|\nabla_x \Phi_{\tau_k}(x_{k})\|^2]+ \frac{2L_D^2\zeta_k}{w_k^2}\mathbb{E}[\|\pi_k-\pi_{\tau_k}^\star(x_k)\|^2] + \frac{2L_D^2\zeta_k}{w_k^2}\mathbb{E}[\|\pi_k^{\bias}-\pi_{w_k,\tau_k}^\star(x_k)\|^2]\notag\\
&\hspace{20pt}+\frac{256L_f^4L_V^2L_{V,2}^2 \zeta_k w_k^2}{C_L^4\underline{\sigma}^2\tau_k^4} + \frac{B_D^2 L_\Phi \zeta_k^2 }{2\tau_k w_k^2}.\label{prop:control_conv:proof_eq5}
\end{align}

The next step is to bridge the gap between $\Phi_{\tau_k}(x_{k+1})$ and $\Phi_{\tau_{k+1}}(x_{k+1})$. By the definition of $\Phi_\tau$ in \eqref{eq:def_Phi_tau}
\begin{align}
\Phi_{\tau_{k+1}}(x_{k+1})-\Phi_{\tau_k}(x_{k+1}) &= f(x_{k+1},\pi_{\tau_{k+1}}^\star(x_{k+1}))-f(x_{k+1},\pi_{\tau_k}^\star(x_{k+1}))\notag\\
&\leq L_f\|\pi_{\tau_{k+1}}^\star(x_{k+1}) - \pi_{\tau_k}^\star(x_{k+1})\|\notag\\
&\leq L_f\cdot\frac{6(\tau_k-\tau_{k+1})|\Scal|\log|\Acal|}{(1-\gamma)C_L\tau_k}\notag\\
&\leq \frac{16L_f|\Scal|\log|\Acal|}{(1-\gamma)C_L(k+1)},\label{prop:control_conv:proof_eq6}
\end{align}
where the first inequality is due to Assumption~\ref{assump:f}, and the second inequality is due to Lemma~\ref{lem:Lipschitz_BR}, and the last inequality applies Lemma~\ref{lem:tau_diff}.

Finally, we bridge the gap between $\|\nabla_x \Phi_{\tau_k}(x_{k})\|^2$ and $\|\nabla_x \Phi(x_{k})\|^2$ by invoking Lemma~\ref{lem:Phi_tau}
\begin{align}
\|\nabla_x\Phi_{\tau_k}(x)-\nabla_x\Phi(x)\|\leq \frac{L_\star L_f L_V^2 L_{V,2}\tau_k}{\underline{\sigma}^2}.\label{prop:control_conv:proof_eq7}
\end{align}

Combining \eqref{prop:control_conv:proof_eq5}-\eqref{prop:control_conv:proof_eq7},
\begin{align*}
&\mathbb{E}[\Phi_{\tau_{k+1}}(x_{k+1})-\Phi_{\tau_k}(x_{k})]\notag\\
&=\mathbb{E}[\Phi_{\tau_k}(x_{k+1})-\Phi_{\tau_k}(x_{k})]+\mathbb{E}[\Phi_{\tau_{k+1}}(x_{k+1})-\Phi_{\tau_k}(x_{k+1})]\notag\\
&\leq-\frac{\zeta_k}{2}\mathbb{E}[\|\nabla_x \Phi_{\tau_k}(x_{k})\|^2]+ \frac{2L_D^2\zeta_k}{w_k^2}\mathbb{E}[\|\pi_k-\pi_{\tau_k}^\star(x_k)\|^2] + \frac{2L_D^2\zeta_k}{w_k^2}\mathbb{E}[\|\pi_k^{\bias}-\pi_{w_k,\tau_k}^\star(x_k)\|^2]\notag\\
&\hspace{20pt}+\frac{256L_f^4L_V^2L_{V,2}^2 \zeta_k w_k^2}{C_L^4\underline{\sigma}^2\tau_k^4} + \frac{B_D^2 L_\Phi \zeta_k^2}{2\tau_k w_k^2} + \frac{16L_f|\Scal|\log|\Acal|}{(1-\gamma)C_L(k+1)}\notag\\
&\leq-\frac{\zeta_k}{4}\mathbb{E}[\|\nabla_x \Phi(x_{k})\|^2]+ \frac{2L_D^2\zeta_k}{w_k^2}\mathbb{E}[\|\pi_k-\pi_{\tau_k}^\star(x_k)\|^2] + \frac{2L_D^2\zeta_k}{w_k^2}\mathbb{E}[\|\pi_k^{\bias}-\pi_{w_k,\tau_k}^\star(x_k)\|^2]\notag\\
&\hspace{20pt}+\frac{L_\star^2 L_f^2 L_V^4L_{V,2}^2\zeta_k\tau_k^2}{2\underline{\sigma}^4}+\frac{256L_f^4L_V^2L_{V,2}^2 \zeta_k w_k^2}{C_L^4\underline{\sigma}^2\tau_k^4} + \frac{B_D^2 L_\Phi \zeta_k^2}{2\tau_k w_k^2} + \frac{16L_f|\Scal|\log|\Acal|}{(1-\gamma)C_L(k+1)},
\end{align*}
where the last inequality follows from the simple fact that $-\frac{a^2}{2}\leq-\frac{b^2}{4}+\frac{(a-b)^2}{2}$ for any scalar $a,b$.

\qed

\subsection{Proof of Proposition~\ref{prop:control_conv_fixedtau}}

The proof is almost identical to that of Proposition~\ref{prop:control_conv}. We include the proof here for completeness.

From Lemma~\ref{lem:Lipschitz_Phi}, we know that under a fixed regularization weight $\tau_0$, the objective $\Phi_{\tau_0}$ has $L_{\Phi,\tau_0}$ Lipschitz gradients, where we define 
\[L_{\Phi,\tau_0}\triangleq(1+\frac{2L_V}{C_L\tau_0})\Big(\frac{2L_fL_V}{C_L\tau_0}+\frac{2L_fL_VL_{V,2}}{\underline{\sigma}C_L\tau_0}+\frac{2L_fL_V^2L_{V,2}}{\underline{\sigma}^2C_L\tau_0}+\frac{2L_fL_V^2}{\underline{\sigma}C_L\tau}\Big).\]
This implies
\begin{align}
&\Phi_{\tau_0}(x_{k+1})-\Phi_{\tau_0}(x_{k})\notag\\
&\leq \langle\nabla_x \Phi_{\tau_0}(x_{k}), x_{k+1}-x_k\rangle + \frac{L_{\Phi,\tau_0}}{2}\|x_{k+1}-x_k\|^2\notag\\
&= -\zeta_k \langle\nabla_x \Phi_{\tau_0}(x_{k}),D_{w_k}(x_k,\pi_k,\pi_k^{\bias},s_k,a_k,\bar{s}_k,\bar{a}_k,\xi_k)\rangle+\frac{L_{\Phi,\tau_0} \zeta_k^2}{2}\|D_{w_k}(x_k,\pi_k,\pi_k^{\bias},s_k,a_k,\bar{s}_k,\bar{a}_k,\xi_k)\|^2\notag\\
&=-\zeta_k \|\nabla_x \Phi_{\tau_0}(x_{k})\|^2 + \zeta_k\langle\nabla_x \Phi_{\tau_0}(x_{k}),\nabla_x \Phi_{\tau_0}(x_{k})-D_{w_k}(x_k,\pi_k,\pi_k^{\bias},s_k,a_k,\bar{s}_k,\bar{a}_k,\xi_k)\rangle\notag\\
&\hspace{20pt}+\frac{L_{\Phi,\tau_0} \zeta_k^2}{2}\|D_{w_k}(x_k,\pi_k,\pi_k^{\bias},s_k,a_k,\bar{s}_k,\bar{a}_k,\xi_k)\|^2.
\end{align}

By the law of total expectation,
\begin{align}
&\mathbb{E}[\Phi_{\tau_0}(x_{k+1})-\Phi_{\tau_0}(x_{k})]\notag\\
&\leq -\zeta_k \mathbb{E}[\|\nabla_x \Phi_{\tau_0}(x_{k})\|^2] +\frac{L_{\Phi,\tau_0} \zeta_k^2}{2}\mathbb{E}[\|D_{w_k}(x_k,\pi_k,\pi_k^{\bias},s_k,a_k,\bar{s}_k,\bar{a}_k,\xi_k)\|^2]\notag\\
&\hspace{20pt}+ \zeta_k\mathbb{E}[\langle\nabla_x \Phi_{\tau_0}(x_{k}),\nabla_x \Phi_{\tau_0}(x_{k})-\mathbb{E}[D_{w_k}(x_k,\pi_k,\pi_k^{\bias},s_k,a_k,\bar{s}_k,\bar{a}_k,\xi_k)\mid\Fcal_{k-1}]\rangle]\notag\\
&=-\zeta_k \mathbb{E}[\|\nabla_x \Phi_{\tau_0}(x_{k})\|^2] +\frac{L_{\Phi,\tau_0} \zeta_k^2}{2}\mathbb{E}[\|D_{w_k}(x_k,\pi_k,\pi_k^{\bias},s_k,a_k,\bar{s}_k,\bar{a}_k,\xi_k)\|^2]\notag\\
&\hspace{20pt}+ \zeta_k\mathbb{E}[\langle\nabla_x \Phi_{\tau_0}(x_{k}),\nabla_x \Phi_{\tau_0}(x_{k})-\bar{D}_{w_k}(x_k,\pi_k,\pi_k^{\bias})\rangle]\notag\\
&\leq-\zeta_k \mathbb{E}[\|\nabla_x \Phi_{\tau_0}(x_{k})\|^2] +\frac{B_D^2 L_{\Phi,\tau_0} \zeta_k^2}{2w_k^2}+\frac{\zeta_k}{2}\mathbb{E}[\|\nabla_x \Phi_{\tau_0}(x_{k})\|^2]+\frac{\zeta_k}{2}\mathbb{E}[\|\nabla_x \Phi_{\tau_0}(x_{k})-\bar{D}_{w_k}(x_k,\pi_k,\pi_k^{\bias})\|^2]\notag\\
&=-\frac{\zeta_k}{2}\mathbb{E}[\|\nabla_x \Phi_{\tau_0}(x_{k})\|^2] +\frac{B_D^2 L_{\Phi,\tau_0} \zeta_k^2}{2w_k^2}\notag\\
&\hspace{20pt}+\frac{\zeta_k}{2}\mathbb{E}\left[\Big\|\Big(\nabla_x \Phi_{\tau_0}(x_{k})-\nabla_x \Phi_{w_k,\tau_0}(x_{k})\Big)+\Big(\bar{D}_{w_k}(x_k,\pi_{\tau_0}^\star(x_k),\pi_{w_k,\tau_0}^\star(x_k))-\bar{D}_{w_k}(x_k,\pi_k,\pi_k^{\bias})\Big)\Big\|^2\right]\notag\\
&\leq -\frac{\zeta_k}{2}\mathbb{E}[\|\nabla_x \Phi_{\tau_0}(x_{k})\|^2]+\zeta_k \mathbb{E}[\|\bar{D}_{w_k}(x_k,\pi_{\tau_0}^\star(x_k),\pi_{w_k,\tau_0}^\star(x_k))-\bar{D}_{w_k}(x_k,\pi_k,\pi_k^{\bias})\|^2]\notag\\
&\hspace{20pt}+\zeta_k \mathbb{E}[\|\nabla_x \Phi_{\tau_0}(x_{k})-\nabla_x \Phi_{w_k,\tau_0}(x_{k})\|^2]+\frac{B_D^2 L_{\Phi,\tau_0} \zeta_k^2}{2w_k^2},\label{prop:control_conv_fixedtau:proof_eq1}
\end{align}
where the second inequality applies Lemma~\ref{lem:bounded_DFG} and the last equation follows from \eqref{eq:grad_Phi_barD_equality}.

To bound the second term on the right hand side of \eqref{prop:control_conv_fixedtau:proof_eq1}, we apply Lemma~\ref{lem:Lipschitz_DFG}
\begin{align}
&\|\bar{D}_{w_k}(x_k,\pi_{\tau_0}^\star(x_k),\pi_{w_k,\tau_0}^\star(x_k))-\bar{D}_{w_k}(x_k,\pi_k,\pi_k^{\bias})\|^2\notag\\
&\leq \frac{L_D^2}{w_k^2}\Big(\|\pi_k-\pi_{\tau_0}^\star(x_k)\|+\|\pi_k^{\bias}-\pi_{w_k,\tau_0}^\star(x_k)\|\Big)^2\notag\\
&\leq \frac{2L_D^2}{w_k^2}\|\pi_k-\pi_{\tau_0}^\star(x_k)\|^2+\frac{2L_D^2}{w_k^2}\|\pi_k^{\bias}-\pi_{w_k,\tau_0}^\star(x_k)\|^2.\label{prop:control_conv_fixedtau:proof_eq2}
\end{align}

For the third term of \eqref{prop:control_conv_fixedtau:proof_eq1}, we have from Lemma~\ref{eq:grad_gap_sigma_tau}
\begin{align}
\|\nabla_x\Phi_{\tau_0}(x_k)-\nabla_x\Phi_{w_k,\tau_0}(x_k)\|^2
&\leq\Big(\frac{4L_f L_V w_k}{C_L \underline{\sigma} \tau_0}(L_f+\frac{2L_f L_{V,2}}{C_L\tau_0})\Big)^2=C_{2,\tau_0}w_k^2,\label{prop:control_conv_fixedtau:proof_eq3}
\end{align}
where we define $C_{2,\tau_0} = \Big(\frac{4L_f L_V}{C_L \underline{\sigma} \tau_0}(L_f+\frac{2L_f L_{V,2}}{C_L\tau_0})\Big)^2$.

Substituting \eqref{prop:control_conv_fixedtau:proof_eq2} and \eqref{prop:control_conv_fixedtau:proof_eq3} into \eqref{prop:control_conv_fixedtau:proof_eq1},
\begin{align*}
&\mathbb{E}[\Phi_{\tau_0}(x_{k+1})-\Phi_{\tau_0}(x_k)]\notag\\
&\leq -\frac{\zeta_k}{2}\mathbb{E}[\|\nabla_x \Phi_{\tau_0}(x_{k})\|^2]+ \frac{2L_D^2\zeta_k}{w_k^2}\mathbb{E}[\|\pi_k-\pi_{\tau_0}^\star(x_k)\|^2] + \frac{2L_D^2\zeta_k}{w_k^2}\mathbb{E}[\|\pi_k^{\bias}-\pi_{w_k,\tau_0}^\star(x_k)\|^2]\\
&\hspace{20pt}+C_{2,\tau_0}\zeta_k w_k^2 + \frac{B_D^2 L_{\Phi,\tau_0} \zeta_k^2 }{2w_k^2}.
\end{align*}

\qed

\subsection{Proof of Proposition~\ref{prop:policy_conv}}

The proof depends on an intermediate result that bounds an important cross term. We state it in the lemma below and defer its proof to Section~\ref{sec:proof:lem:policy_conv_cross_term}.

\begin{lem}\label{lem:policy_conv_cross_term}
Under the assumptions and step sizes of Proposition~\ref{prop:policy_conv}, we have for all $k\geq0$
\begin{align*}
&\mathbb{E}[-\langle\nabla_x J_{\tau_{k}}(x_{k},\pi_{\theta_{k}})-\nabla_x J_{\tau_{k}}(x_{k},\pi_{\theta_{\tau_k}^\star(x_k)}),x_{k+1}-x_k\rangle\rangle]\notag\\
&\leq \frac{ C_L^2\alpha_k\tau_k^2}{64}\mathbb{E}[\|\pi_k-\pi_{\tau_k}^\star(x_k)\|^2]\notag\\
&\hspace{20pt}+\frac{64 L_D^2 L_V^2 \zeta_k^2}{ C_L^2\alpha_k w_k^2 \tau_k^2} \mathbb{E}[\|\pi_k-\pi_{\tau_k}^\star(x_k)\|^2+\|\pi_k^{\bias}-\pi_{w_k,\tau_k}^\star(x_k)\|^2]+\frac{32L_V^2\zeta_k^2}{ C_L^2\alpha_k\tau_k^2}\mathbb{E}[\|\nabla_{x} \Phi_{w_k,\tau_k}(x_k)\|^2].
\end{align*}
\end{lem}

We now proceed to the proof of Proposition~\ref{prop:policy_conv}.
We consider the following decomposition and bound each term on the right hand side individually.
\begin{align}
&-J_{\tau_{k+1}}(x_{k+1},\pi_{\theta_{k+1}})+J_{\tau_{k}}(x_{k},\pi_{\theta_{k}})\notag\\
&=\Big(-J_{\tau_{k}}(x_{k+1},\pi_{\theta_{k+1}})+J_{\tau_k}(x_{k+1},\pi_{\theta_k})\Big) + \Big(-J_{\tau_{k}}(x_{k+1},\pi_{\theta_{k}})+J_{\tau_{k}}(x_{k},\pi_{\theta_{k}})\Big)\notag\\
&\hspace{20pt}+\Big(-J_{\tau_{k+1}}(x_{k+1},\pi_{\theta_{k+1}})+J_{\tau_{k}}(x_{k+1},\pi_{\theta_{k+1}})\Big).\label{prop:policy_conv:proof_eq0}
\end{align}

\noindent\textbf{Bound the First Term of \eqref{prop:policy_conv:proof_eq0}.}
As $J_{\tau}$ has $L_V$-Lipschitz gradients, 
\begin{align}
&-J_{\tau_{k}}(x_{k+1},\pi_{\theta_{k+1}})+J_{\tau_k}(x_{k+1},\pi_{\theta_k}) \notag\\
&\leq \langle -\nabla_\theta  J_{\tau_k}(x_{k+1},\pi_{\theta_k}),\theta_{k+1}-\theta_k\rangle+\frac{L_V}{2}\|\theta_{k+1} - \theta_k\|^2\notag\\
&=-\alpha_k\langle \nabla_\theta  J_{\tau_k}(x_{k+1},\pi_{\theta_k}), F_{0,\tau_k}(x_k,\theta_k,\hat{V}_{k},s_k,a_k,s_k',\xi_k) \rangle  + \frac{L_V\alpha_k^2}{2}\|F_{0,\tau_k}(x_k,\theta_k,\hat{V}_{k},s_k,a_k,s_k',\xi_k)\|^2\notag\\
&= -\alpha_k\langle \nabla_\theta  J_{\tau_k}(x_{k+1},\pi_{\theta_k}), F_{0,\tau_k}(x_k,\theta_k,\hat{V}_{k},s_k,a_k,s_k',\xi_k) - \bar{F}_{0,\tau_k}(x_k,\theta_k,\hat{V}_{k})\rangle\notag\\
&\hspace{20pt}-\alpha_k\langle \nabla_\theta  J_{\tau_k}(x_{k+1},\pi_{\theta_k}), \bar{F}_{0,\tau_k}(x_k,\theta_k,\hat{V}_{k})-\bar{F}_{0,\tau_k}(x_k,\theta_k,V_{\tau_k}^{x_k,\pi_{\theta_k}})\rangle\notag\\
&\hspace{20pt}-\alpha_k \langle\nabla_\theta  J_{\tau_k}(x_{k+1},\pi_{\theta_k}), \nabla_\theta  J_{\tau_k}(x_{k},\pi_{\theta_k}) \rangle+\frac{L_V\alpha_k^2}{2}\|F_{0,\tau_k}(x_k,\theta_k,\hat{V}_{k},s_k,a_k,s_k',\xi_k)\|^2,\label{prop:policy_conv:proof_eq1}
\end{align}
where the final equation follows from $\nabla_\theta  J_{\tau_k}(x_{k},\pi_{\theta_k})=\bar{F}_{0,\tau_k}(x_k,\theta_k,V_{\tau_k}^{x_k,\pi_{\theta_k}})$.

To bound the first term of \eqref{prop:policy_conv:proof_eq1},
\begin{align}
&-\alpha_k\mathbb{E}[\langle \nabla_\theta  J_{\tau_k}(x_{k+1},\pi_{\theta_k}), F_{0,\tau_k}(x_k,\theta_k,\hat{V}_{k},s_k,a_k,s_k',\xi_k) - \bar{F}_{0,\tau_k}(x_k,\theta_k,\hat{V}_{k})\rangle]\notag\\
&=-\alpha_k\mathbb{E}[\langle \nabla_\theta  J_{\tau_k}(x_k,\pi_{\theta_k}), \mathbb{E}[F_{0,\tau_k}(x_k,\theta_k,\hat{V}_{k},s_k,a_k,s_k',\xi_k)\mid\Fcal_{k-1}] - \bar{F}_{0,\tau_k}(x_k,\theta_k,\hat{V}_{k})\rangle]\notag\\
&\hspace{20pt}+\alpha_k\mathbb{E}[\langle \nabla_\theta  J_{\tau_k}(x_k,\pi_{\theta_k})-\nabla_\theta  J_{\tau_k}(x_{k+1},\pi_{\theta_k}), F_{0,\tau_k}(x_k,\theta_k,\hat{V}_{k},s_k,a_k,s_k',\xi_k) - \bar{F}_{0,\tau_k}(x_k,\theta_k,\hat{V}_{k})\rangle]\notag\\
&= \alpha_k\mathbb{E}[\langle \nabla_\theta  J_{\tau_k}(x_k,\pi_{\theta_k})-\nabla_\theta  J_{\tau_k}(x_{k+1},\pi_{\theta_k}), F_{0,\tau_k}(x_k,\theta_k,\hat{V}_{k},s_k,a_k,s_k',\xi_k) - \bar{F}_{0,\tau_k}(x_k,\theta_k,\hat{V}_{k})\rangle]\notag\\
&\leq \alpha_k\cdot L_V \mathbb{E}[\|x_{k+1}-x_k\|]\cdot 2B_F\notag\\
&\leq 2B_F L_V\alpha_k\cdot\frac{B_D\zeta_k}{w_k}\notag\\
&=\frac{2B_D B_F L_V\zeta_k\alpha_k}{w_k},\label{prop:policy_conv:proof_eq2}
\end{align}
where the second inequality follows from Lemma~\ref{lem:bounded_DFG}.

For the second term of \eqref{prop:policy_conv:proof_eq1}, 
\begin{align}
&-\alpha_k\langle \nabla_\theta  J_{\tau_k}(x_{k+1},\pi_{\theta_k}), \bar{F}_{0,\tau_k}(x_k,\theta_k,\hat{V}_{k})-\bar{F}_{0,\tau_k}(x_k,\theta_k,V_{\tau_k}^{x_k,\pi_{\theta_k}})\rangle\notag\\
&\leq \frac{\alpha_k}{8}\|\nabla_\theta  J_{\tau_k}(x_{k+1},\pi_{\theta_k})\|^2 + 2\alpha_k\|\bar{F}_{0,\tau_k}(x_k,\theta_k,\hat{V}_{k})-\bar{F}_{0,\tau_k}(x_k,\theta_k,V_{\tau_k}^{x_k,\pi_{\theta_k}})\|^2\notag\\
&\leq \frac{\alpha_k}{4}\|\nabla_\theta  J_{\tau_k}(x_{k},\pi_{\theta_k})\|^2 +\frac{\alpha_k}{4}\|\nabla_\theta  J_{\tau_k}(x_{k+1},\pi_{\theta_k})-\nabla_\theta  J_{\tau_k}(x_{k},\pi_{\theta_k})\|^2 \notag\\
&\hspace{20pt}+2\alpha_k\|\bar{F}_{0,\tau_k}(x_k,\theta_k,\hat{V}_{k})-\bar{F}_{0,\tau_k}(x_k,\theta_k,V_{\tau_k}^{x_k,\pi_{\theta_k}})\|^2\notag\\
&\leq\frac{\alpha_k}{4}\|\nabla_\theta  J_{\tau_k}(x_{k},\pi_{\theta_k})\|^2+\frac{L_V^2\alpha_k}{4}\Big(\|x_{k+1}-x_k\|^2\Big)+ 2L_F^2\alpha_k\|V_{\tau_k}^{x_k,\pi_{\theta_k}}-\hat{V}_k\|^2\notag\\
&\leq\frac{\alpha_k}{4}\|\nabla_\theta  J_{\tau_k}(x_{k},\pi_{\theta_k})\|^2+\frac{L_V^2\alpha_k}{4}\cdot\Big(\frac{B_D\zeta_k}{w_k}\Big)^2+ 2L_F^2\alpha_k \varepsilon_k^V\notag\\
&\leq\frac{\alpha_k}{4}\|\nabla_\theta  J_{\tau_k}(x_{k},\pi_{\theta_k})\|^2 + \frac{B_D^2 L_V^2\zeta_k^2\alpha_k}{4w_k^2} + 2L_F^2\alpha_k \varepsilon_k^V,\label{prop:policy_conv:proof_eq3}
\end{align}
where the third inequality is a result of the Lipschitz continuity of $J_{\tau_k}$ and $\bar{F}_{0,\tau_k}$.

For the third term of \eqref{prop:policy_conv:proof_eq1},
\begin{align}
&-\alpha_k \langle\nabla_\theta  J_{\tau_k}(x_{k+1},\pi_{\theta_k}), \nabla_\theta  J_{\tau_k}(x_{k},\pi_{\theta_k}) \rangle\notag\\
&= -\alpha_k \|\nabla_\theta  J_{\tau_k}(x_{k},\pi_{\theta_k})\|^2 + \alpha_k \langle\nabla_\theta  J_{\tau_k}(x_{k},\pi_{\theta_k}) -\nabla_\theta  J_{\tau_k}(x_{k+1},\pi_{\theta_k}), \nabla_\theta  J_{\tau_k}(x_{k},\pi_{\theta_k})\rangle\notag\\
&\leq -\frac{\alpha_k}{2} \|\nabla_\theta  J_{\tau_k}(x_{k},\pi_{\theta_k})\|^2 + \frac{\alpha_k}{2} \|\nabla_\theta  J_{\tau_k}(x_{k+1},\pi_{\theta_k})-\nabla_\theta  J_{\tau_k}(x_{k},\pi_{\theta_k})\|^2\notag\\
&\leq -\frac{\alpha_k}{2} \|\nabla_\theta  J_{\tau_k}(x_{k},\pi_{\theta_k})\|^2 + \frac{L_V^2\alpha_k}{2}\Big(\|x_{k+1}-x_k\|^2\Big)\notag\\
&\leq -\frac{\alpha_k}{2} \|\nabla_\theta  J_{\tau_k}(x_{k},\pi_{\theta_k})\|^2 + \frac{L_V^2\alpha_k}{2}\cdot\Big(\frac{B_D\zeta_k}{w_k}\Big)^2\notag\\
&\leq -\frac{\alpha_k}{2} \|\nabla_\theta  J_{\tau_k}(x_{k},\pi_{\theta_k})\|^2 + \frac{B_D^2 L_V^2\zeta_k^2\alpha_k}{2w_k^2},\label{prop:policy_conv:proof_eq4}
\end{align}
where the second inequality follows from the $L_V$-smoothness continuity of $J_\tau$ from \eqref{lem:Lipschitz_V:eq2} of Lemma~\ref{lem:Lipschitz_V}.

Substituting \eqref{prop:policy_conv:proof_eq2}-\eqref{prop:policy_conv:proof_eq4} into \eqref{prop:policy_conv:proof_eq1}, 
\begin{align}
&\mathbb{E}[-J_{\tau_{k+1}}(x_{k+1},\pi_{\theta_{k+1}})+J_{\tau_{k}}(x_{k},\pi_{\theta_{k}})] \notag\\
&\leq \frac{2B_D B_F L_V\zeta_k\alpha_k}{w_k}+\frac{\alpha_k}{4}\mathbb{E}[\|\nabla_\theta  J_{\tau_k}(x_{k},\pi_{\theta_k})\|^2] + \frac{B_D^2 L_V^2\zeta_k^2\alpha_k}{4w_k^2} + 2L_F^2\alpha_k \mathbb{E}[\varepsilon_k^V]\notag\\
&\hspace{20pt}-\frac{\alpha_k}{2} \mathbb{E}[\|\nabla_\theta  J_{\tau_k}(x_{k},\pi_{\theta_k})\|^2] + \frac{B_D^2 L_V^2\zeta_k^2\alpha_k}{2w_k^2}+\frac{L_V\alpha_k^2}{2}\mathbb{E}[\|F_{0,\tau_k}(x_k,\theta_k,\hat{V}_{k},s_k,a_k,s_k',\xi_k)\|^2]\notag\\
&\leq-\frac{\alpha_k}{4} \mathbb{E}[\|\nabla_\theta  J_{\tau_k}(x_{k},\pi_{\theta_k})\|^2] + 2L_F^2\alpha_k\mathbb{E}[\varepsilon_k^V] + \frac{3B_D^2 L_V^2\zeta_k^2\alpha_k}{4w_k^2} + \frac{2B_D B_F L_V\zeta_k\alpha_k}{w_k}+\frac{B_F^2 L_V\alpha_k^2}{2}\notag\\
&\leq-\frac{\alpha_k}{8} \mathbb{E}[\|\nabla_\theta  J_{\tau_k}(x_{k},\pi_{\theta_k})\|^2] + 2L_F^2\alpha_k\mathbb{E}[\varepsilon_k^V] + \frac{3B_D^2 L_V^2\zeta_k^2\alpha_k}{4w_k^2} + \frac{2B_D B_F L_V\zeta_k\alpha_k}{w_k}+\frac{B_F^2 L_V\alpha_k^2}{2}\notag\\
&\hspace{20pt}-\frac{C_L^2\alpha_k\tau_k^2}{32}\mathbb{E}[\|\pi_{\theta_k}-\pi_{\tau_k}^\star(x_k)\|^2],\label{prop:policy_conv:proof_eq5}
\end{align}
where the last inequality plugs in the relationship
\begin{align}
\|\nabla_\theta  J_{\tau_k}(x_{k},\pi_{\theta_k})\|\geq\frac{C_L\tau_k}{2}\|\pi_{\theta_k}-\pi_{\tau_k}^\star(x_k)\|.\label{prop:policy_conv:proof_eq5.1}
\end{align}
Note that $J_{\tau}(x,\pi)=\lim_{w\rightarrow0}w \Lcal_{w,\tau}(x,\pi)$, which implies that \eqref{prop:policy_conv:proof_eq5.1} follows from \eqref{eq:PL_QG_combined}.

\noindent\textbf{Bound the Second Term of \eqref{prop:policy_conv:proof_eq0}.}
We use $\theta_{\tau}^\star(x)$ to denote a softmax parameter that encodes the policy $\pi_{\tau}^\star(x)$.
Again, as $J_{\tau}$ has $L_V$-Lipschitz gradients, 
\begin{align*}
&-J_{\tau_{k}}(x_{k+1},\pi_{\theta_{k}})+J_{\tau_{k}}(x_{k},\pi_{\theta_{k}})\notag\\
&\leq -\langle \nabla_x J_{\tau_{k}}(x_{k},\pi_{\theta_{k}}),x_{k+1}-x_k\rangle+\frac{L_V}{2}\|x_{k+1}-x_k\|^2\notag\\
&=-\langle\nabla_x J_{\tau_{k}}(x_{k},\pi_{\theta_{k}})-\nabla_x J_{\tau_{k}}(x_{k},\pi_{\theta_{\tau_k}^\star(x_k)}),x_{k+1}-x_k\rangle+\frac{L_V}{2}\|x_{k+1}-x_k\|^2\notag\\
&\hspace{20pt}-\langle\nabla_x J_{\tau_{k}}(x_{k},\pi_{\theta_{\tau_k}^\star(x_k)}),x_{k+1}-x_k\rangle\notag\\
&=-\langle\nabla_x J_{\tau_{k}}(x_{k},\pi_{\theta_{k}})-\nabla_x J_{\tau_{k}}(x_{k},\pi_{\theta_{\tau_k}^\star(x_k)}),x_{k+1}-x_k\rangle+\frac{L_V}{2}\|x_{k+1}-x_k\|^2\notag\\
&\hspace{20pt}-\langle\nabla_x \ell_{\tau_{k}}(x_{k}),x_{k+1}-x_k\rangle\notag\\
&\leq-\langle\nabla_x J_{\tau_{k}}(x_{k},\pi_{\theta_{k}})-\nabla_x J_{\tau_{k}}(x_{k},\pi_{\theta_{\tau_k}^\star(x_k)}),x_{k+1}-x_k\rangle+\frac{L_V}{2}\|x_{k+1}-x_k\|^2\notag\\
&\hspace{20pt}+\Big(-\ell_{\tau_k}(x_{k+1}) + \ell_{\tau_k}(x_{k})\Big)+\frac{L_\Phi}{2\tau_k}\|x_{k+1}-x_k\|^2\notag\\
&\leq -\langle\nabla_x J_{\tau_{k}}(x_{k},\pi_{\theta_{k}})-\nabla_x J_{\tau_{k}}(x_{k},\pi_{\theta_{\tau_k}^\star(x_k)}),x_{k+1}-x_k\rangle\notag\\
&\hspace{20pt}+\Big(-J_{\tau_k}(x_{k+1},\pi_{\tau_k}^\star(x_{k+1})) + J_{\tau_k}(x_{k},\pi_{\tau_k}^\star(x_k))\Big)+\frac{L_\Phi}{\tau_k}\|x_{k+1}-x_k\|^2,
\end{align*}
where the last inequality is due to $L_V\leq L_\Phi$ and the step size condition $\tau_k\leq1$, and the second equation uses the relationship $\nabla_x J_{\tau_{k}}(x_{k},\pi_{\theta_{\tau_k}^\star(x_k)}) = \nabla_x \ell_{\tau_{k}}(x_{k})$, which is due to $\nabla_\theta  J_{\tau_{k}}(x_{k},\pi_{\theta_{\tau_k}^\star(x_k)})=0$ by the first-order optimality condition. The second inequality is due to the fact that $\ell_{\tau}$ is $\frac{L_\Phi}{\tau}$-smooth when $\tau\leq1$ (established in Lemma~\ref{lem:Lipschitz_Phi}) and that for an $L$-smooth function $f$, we have 
\[f(y)-f(x)\leq\langle\nabla f(x),y-x\rangle+\frac{L}{2}\|x-y\|^2.\]

Taking the expectation and plugging in the result from Lemma~\ref{lem:policy_conv_cross_term},
\begin{align}
&\mathbb{E}[-J_{\tau_{k}}(x_{k+1},\pi_{\theta_{k}})+J_{\tau_{k}}(x_{k},\pi_{\theta_{k}})]\notag\\
&\leq \frac{ C_L^2\alpha_k\tau_k^2}{64}\mathbb{E}[\|\pi_k-\pi_{\tau_k}^\star(x_k)\|^2]\notag\\
&\hspace{20pt}+\frac{64 L_D^2 L_V^2 \zeta_k^2}{ C_L^2\alpha_k w_k^2 \tau_k^2} \mathbb{E}[\|\pi_k-\pi_{\tau_k}^\star(x_k)\|^2+\|\pi_k^{\bias}-\pi_{w_k,\tau_k}^\star(x_k)\|^2]+\frac{32L_V^2\zeta_k^2}{ C_L^2\alpha_k\tau_k^2}\mathbb{E}[\|\nabla_{x} \Phi_{w_k,\tau_k}(x_k)\|^2]\notag\\
&\hspace{20pt}+\mathbb{E}[-J_{\tau_k}(x_{k+1},\pi_{\tau_k}^\star(x_{k+1})) + J_{\tau_k}(x_{k},\pi_{\tau_k}^\star(x_k))]+\frac{L_\Phi}{\tau_k}\mathbb{E}[\|x_{k+1}-x_k\|^2]\notag\\
&\leq \frac{ C_L^2\alpha_k\tau_k^2}{64}\mathbb{E}[\|\pi_k-\pi_{\tau_k}^\star(x_k)\|^2]+\frac{64 L_D^2 L_V^2 \zeta_k^2}{ C_L^2\alpha_k w_k^2 \tau_k^2} \mathbb{E}[\|\pi_k-\pi_{\tau_k}^\star(x_k)\|^2+\|\pi_k^{\bias}-\pi_{w_k,\tau_k}^\star(x_k)\|^2]\notag\\
&\hspace{20pt}+\mathbb{E}[-J_{\tau_k}(x_{k+1},\pi_{\tau_k}^\star(x_{k+1})) + J_{\tau_k}(x_{k},\pi_{\tau_k}^\star(x_k))]+\frac{32L_V^2 \zeta_k^2}{ C_L^2\alpha_k\tau_k^2}\mathbb{E}[\|\nabla_{x} \Phi_{w_k,\tau_k}(x_k)\|^2]+\frac{B_D^2 L_\Phi\zeta_k^2}{w_k^2\tau_k},\label{prop:policy_conv:proof_eq6}
\end{align}
where the last inequality follows from $\|x_{k+1}-x_k\|\leq\frac{B_D \zeta_k}{w_k}$.

\noindent\textbf{Bound the Third Term of \eqref{prop:policy_conv:proof_eq0}.}
\begin{align}
-J_{\tau_{k+1}}(x_{k+1},\pi_{\theta_{k+1}})+J_{\tau_{k}}(x_{k+1},\pi_{\theta_{k+1}})
&=\frac{\tau_{k}-\tau_{k+1}}{1-\gamma}\mathbb{E}_{s\sim d_\rho^{\pi_{k+1}}}[E(\pi_{k+1},s)]\notag\\
&\leq \frac{\log|\Acal|(\tau_k-\tau_{k+1})}{(1-\gamma)}\notag\\
&\leq \frac{8\log|\Acal|\tau_k}{3(1-\gamma)(k+1)},\label{prop:policy_conv:proof_eq7}
\end{align}
where the second inequality follows from Lemma~\ref{lem:tau_diff}.

Collecting the bounds in \eqref{prop:policy_conv:proof_eq5}-\eqref{prop:policy_conv:proof_eq7} and plugging them into \eqref{prop:policy_conv:proof_eq0}, we get
\begin{align*}
&\mathbb{E}[-J_{\tau_{k+1}}(x_{k+1},\pi_{\theta_{k+1}})+J_{\tau_{k}}(x_{k},\pi_{\theta_{k}})]\notag\\
&\leq-\frac{\alpha_k}{8} \mathbb{E}[\|\nabla_\theta  J_{\tau_k}(x_{k},\pi_{\theta_k})\|^2] + 2L_F^2\alpha_k\mathbb{E}[\varepsilon_k^V] + \frac{3B_D^2 L_V^2\zeta_k^2\alpha_k}{4w_k^2} + \frac{2B_D B_F L_V\zeta_k\alpha_k}{w_k}+\frac{B_F^2 L_V\alpha_k^2}{2}\notag\\
&\hspace{20pt}-\frac{C_L^2\alpha_k\tau_k^2}{32}\mathbb{E}[\|\pi_{\theta_k}-\pi_{\tau_k}^\star(x_k)\|^2]\notag\\
&\hspace{20pt}+\frac{ C_L^2\alpha_k\tau_k^2}{64}\mathbb{E}[\|\pi_k-\pi_{\tau_k}^\star(x_k)\|^2]+\frac{64 L_D^2 L_V^2 \zeta_k^2}{ C_L^2\alpha_k w_k^2 \tau_k^2} \mathbb{E}[\|\pi_k-\pi_{\tau_k}^\star(x_k)\|^2+\|\pi_k^{\bias}-\pi_{w_k,\tau_k}^\star(x_k)\|^2]\notag\\
&\hspace{20pt}+\mathbb{E}[-J_{\tau_k}(x_{k+1},\pi_{\tau_k}^\star(x_{k+1})) + J_{\tau_k}(x_{k},\pi_{\tau_k}^\star(x_k))]+\frac{32L_V^2 \zeta_k^2}{ C_L^2\alpha_k\tau_k^2}\mathbb{E}[\|\nabla_{x} \Phi_{w_k,\tau_k}(x_k)\|^2]+\frac{B_D^2 L_\Phi\zeta_k^2}{w_k^2\tau_k}\notag\\
&\hspace{20pt}+\frac{8\log|\Acal|\tau_k}{3(1-\gamma)(k+1)}\notag\\
&\leq-\frac{\alpha_k}{8} \mathbb{E}[\|\nabla_\theta  J_{\tau_k}(x_{k},\pi_{\theta_k})\|^2] + 2L_F^2\alpha_k\mathbb{E}[\varepsilon_k^V] +\frac{32L_V^2 \zeta_k^2}{ C_L^2\alpha_k\tau_k^2}\mathbb{E}[\|\nabla_{x} \Phi_{w_k,\tau_k}(x_k)\|^2]\notag\\
&\hspace{20pt}-\frac{ C_L^2\alpha_k\tau_k^2}{64}\mathbb{E}[\|\pi_k-\pi_{\tau_k}^\star(x_k)\|^2]+\frac{64 L_D^2 L_V^2 \zeta_k^2}{ C_L^2\alpha_k w_k^2 \tau_k^2} \mathbb{E}[\|\pi_k-\pi_{\tau_k}^\star(x_k)\|^2+\|\pi_k^{\bias}-\pi_{w_k,\tau_k}^\star(x_k)\|^2]\notag\\
&\hspace{20pt}+ \frac{2B_D^2 L_\Phi\zeta_k^2}{w_k^2\tau_k} + \frac{2B_D B_F L_V\zeta_k\alpha_k}{w_k} + \frac{B_F^2 L_V \alpha_k^2}{2}+\frac{8\log|\Acal|\tau_k}{3(1-\gamma)(k+1)}\notag\\
&\hspace{20pt}+\mathbb{E}[-J_{\tau_k}(x_{k+1},\pi_{\tau_k}^\star(x_{k+1})) + J_{\tau_k}(x_{k},\pi_{\tau_k}^\star(x_k))],
\end{align*}
where in the last inequality we have combined the terms $\frac{3B_D^2 L_V^2\zeta_k^2\alpha_k}{4w_k^2}$ and $\frac{B_D^2 L_\Phi\zeta_k^2}{w_k^2\tau_k}$ under the step size conditions $\tau_k\leq1$ and $\alpha_k\leq\frac{4L_\Phi}{3L_V^2}$.

Recall the definition of $\varepsilon_k^{\theta}$ in \eqref{eq:def_metrics}. We can re-arrange the terms in the inequality above and obtain
\begin{align*}
&\mathbb{E}[\varepsilon_{k+1}^{\theta}-\varepsilon_k^{\theta}]\notag\\
&=\mathbb{E}[-J_{\tau_{k+1}}(x_{k+1},\pi_{\theta_{k+1}})+J_{\tau_{k}}(x_{k},\pi_{\theta_{k}})] \notag\\
&\hspace{20pt}+ \mathbb{E}[J_{\tau_k}(x_{k+1},\pi_{\tau_k}^\star(x_{k+1})) - J_{\tau_k}(x_{k},\pi_{\tau_k}^\star(x_k))]\notag\\
&\hspace{20pt}+ \mathbb{E}[J_{\tau_{k+1}}(x_{k+1},\pi_{\tau_k}^\star(x_{k+1}))-J_{\tau_k}(x_{k+1},\pi_{\tau_k}^\star(x_{k+1}))]\notag\\
&\leq -\frac{\alpha_k}{8} \mathbb{E}[\|\nabla_\theta  J_{\tau_k}(x_{k},\pi_{\theta_k})\|^2] + 2L_F^2\alpha_k\mathbb{E}[\varepsilon_k^V] +\frac{32L_V^2 \zeta_k^2}{ C_L^2\alpha_k\tau_k^2}\mathbb{E}[\|\nabla_{x} \Phi_{w_k,\tau_k}(x_k)\|^2]\notag\\
&\hspace{20pt}-\frac{ C_L^2\alpha_k\tau_k^2}{64}\mathbb{E}[\|\pi_k-\pi_{\tau_k}^\star(x_k)\|^2]+\frac{64 L_D^2 L_V^2 \zeta_k^2}{ C_L^2\alpha_k w_k^2 \tau_k^2} \mathbb{E}[\|\pi_k-\pi_{\tau_k}^\star(x_k)\|^2+\|\pi_k^{\bias}-\pi_{w_k,\tau_k}^\star(x_k)\|^2]\notag\\
&\hspace{20pt}+ \frac{2B_D^2 L_\Phi\zeta_k^2}{w_k^2\tau_k} + \frac{2B_D B_F L_V\zeta_k\alpha_k}{w_k} + \frac{B_F^2 L_V \alpha_k^2}{2}+\frac{16\log|\Acal|\tau_k}{3(1-\gamma)(k+1)},
\end{align*}
where the bound on $J_{\tau_{k+1}}(x_{k+1},\pi_{\tau_k}^\star(x_{k+1}))-J_{\tau_k}(x_{k+1},\pi_{\tau_k}^\star(x_{k+1}))\leq\frac{8\log|\Acal|\tau_k}{3(1-\gamma)(k+1)}$ can be obtained in a manner similar to \eqref{prop:policy_conv:proof_eq7}.

\qed

\subsection{Proof of Proposition~\ref{prop:policybias_conv}}

The proof depends on an intermediate result that bounds an important cross term. We state it in the lemma below and defer its proof to Section~\ref{sec:proof:lem:policybias_conv_cross_term}.

\begin{lem}\label{lem:policybias_conv_cross_term}
Under the assumptions and step sizes of Proposition~\ref{prop:policybias_conv}, we have for all $k\geq0$
\begin{align*}
&\mathbb{E}[\langle\nabla_x \Lcal_{w_{k},\tau_{k}}(x_{k},\pi_{\theta_{k}^{\bias}})-\nabla_x \Lcal_{w_{k},\tau_{k}}(x_{k},\pi_{\theta_{w_k,\tau_k}^\star(x_k)}),x_{k+1}-x_k\rangle]\notag\\
&\leq \frac{ C_L^2\alpha_k\tau_k^2}{64}\mathbb{E}[\|\pi_k^{\bias}-\pi_{w_k,\tau_k}^\star(x_k)\|^2]\notag\\
&\hspace{20pt}+\frac{64 L_D^2 L_L^2 \zeta_k^2}{ C_L^2\alpha_k w_k^2 \tau_k^2} \mathbb{E}[\|\pi_k-\pi_{\tau_k}^\star(x_k)\|^2+\|\pi_k^{\bias}-\pi_{w_k,\tau_k}^\star(x_k)\|^2]+\frac{32L_L^2 \zeta_k^2}{ C_L^2\alpha_k\tau_k^2}\mathbb{E}[\|\nabla_{x} \Phi_{w_k,\tau_k}(x_k)\|^2].
\end{align*}
\end{lem}

We now proceed to the proof of the proposition.
We define the re-weighted functions
\[ \Lcal^{\reweight}_{w,\tau}(x,\pi)\triangleq w\Lcal_{w,\tau}(x,\pi) = w f(x,\pi)+(J_\tau(x,\pi_\tau^\star(x))-J_\tau(x,\pi)),\quad\Phi^{\reweight}_{w,\tau}(x)\triangleq w\Phi_{w,\tau}(x).\]

We consider the following decomposition and bound each term on the right hand side individually.
\begin{align}
&\Lcal^{\reweight}_{w_{k+1},\tau_{k+1}}(x_{k+1},\pi_{\theta_{k+1}^{\bias}})-\Lcal^{\reweight}_{w_{k},\tau_{k}}(x_{k},\pi_{\theta_{k}^{\bias}})\notag\\
&=\Big(\Lcal^{\reweight}_{w_{k},\tau_{k}}(x_{k+1},\pi_{\theta_{k+1}^{\bias}})-\Lcal^{\reweight}_{w_k,\tau_k}(x_{k+1},\pi_{\theta_k^{\bias}})\Big) + \Big(\Lcal^{\reweight}_{w_{k},\tau_{k}}(x_{k+1},\pi_{\theta_{k}^{\bias}})-\Lcal^{\reweight}_{w_{k},\tau_{k}}(x_{k},\pi_{\theta_{k}^{\bias}})\Big)\notag\\
&\hspace{20pt}+\Big(\Lcal^{\reweight}_{w_{k+1},\tau_{k+1}}(x_{k+1},\pi_{\theta_{k+1}^{\bias}})-\Lcal^{\reweight}_{w_{k},\tau_{k}}(x_{k+1},\pi_{\theta_{k+1}^{\bias}})\Big).\label{prop:policybias_conv:proof_eq0}
\end{align}

\noindent\textbf{Bound the First Term of \eqref{prop:policybias_conv:proof_eq0}.}
As $\Lcal^{\reweight}_{w,\tau}$ has $L_L$-Lipschitz gradients with respect to $\theta$ (shown in Lemma~\ref{lem:Lipschitz_Lcal}) under the condition $w,\tau\leq1$, 
\begin{align}
&\Lcal^{\reweight}_{w_{k},\tau_{k}}(x_{k+1},\pi_{\theta_{k+1}^{\bias}})-\Lcal^{\reweight}_{w_k,\tau_k}(x_{k+1},\pi_{\theta_k^{\bias}}) \notag\\
&\leq \langle \nabla_\theta  \Lcal^{\reweight}_{w_k,\tau_k}(x_{k+1},\pi_{\theta_k^{\bias}}),\theta_{k+1}^{\bias}-\theta_k^{\bias}\rangle+\frac{L_L}{2}\|\theta_{k+1}^{\bias} - \theta_k^{\bias}\|^2\notag\\
&=\alpha_k\langle \nabla_\theta  \Lcal^{\reweight}_{w_k,\tau_k}(x_{k+1},\pi_{\theta_k^{\bias}}), F_{w_k,\tau_k}(x_k,\theta_k^{\bias},\hat{V}_{k}^{\bias},\bar{s}_k,\bar{a}_k,\bar{s}_k',\xi_k) \rangle \notag\\
&\hspace{20pt} + \frac{L_L\alpha_k^2}{2}\|F_{w_k,\tau_k}(x_k,\theta_k^{\bias},\hat{V}_{k}^{\bias},\bar{s}_k,\bar{a}_k,\bar{s}_k',\xi_k)\|^2\notag\\
&= \alpha_k\langle \nabla_\theta  \Lcal^{\reweight}_{w_k,\tau_k}(x_{k+1},\pi_{\theta_k^{\bias}}), F_{w_k,\tau_k}(x_k,\theta_k^{\bias},\hat{V}_{k}^{\bias},\bar{s}_k,\bar{a}_k,\bar{s}_k',\xi_k) - \bar{F}_{w_k,\tau_k}(x_k,\theta_k^{\bias},\hat{V}_{k}^{\bias})\rangle\notag\\
&\hspace{20pt}+\alpha_k\langle \nabla_\theta  \Lcal^{\reweight}_{w_k,\tau_k}(x_{k+1},\pi_{\theta_k^{\bias}}), \bar{F}_{w_k,\tau_k}(x_k,\theta_k^{\bias},\hat{V}_{k}^{\bias})-\bar{F}_{w_k,\tau_k}(x_k,\theta_k^{\bias},V_{\tau_k}^{x_k,\pi_{\theta_k^{\bias}}})\rangle\notag\\
&\hspace{20pt}-\alpha_k \langle\nabla_\theta  \Lcal^{\reweight}_{w_k,\tau_k}(x_{k+1},\pi_{\theta_k^{\bias}}), \nabla_\theta  \Lcal^{\reweight}_{w_k,\tau_k}(x_{k},\pi_{\theta_k^{\bias}}) \rangle\notag\\
&\hspace{20pt}+\frac{L_L\alpha_k^2}{2}\|F_{w_k,\tau_k}(\theta_k,\omega_k,\hat\mu_k,\hat{V}_{f,k},s_k,a_k,b_k,s_k',\xi_k)\|^2,\label{prop:policybias_conv:proof_eq1}
\end{align}
where the final equation follows from 
\[\nabla_\theta  \Lcal^{\reweight}_{w_k,\tau_k}(x_{k},\pi_{\theta_k^{\bias}})=w_k\nabla_\theta  \Lcal_{w_k,\tau_k}(x_{k},\pi_{\theta_k^{\bias}})=-\bar{F}_{w_k,\tau_k}(x_k,\theta_k^{\bias},V_{\tau_k}^{x_k,\pi_{\theta_k^{\bias}}}).\]

To bound the first term of \eqref{prop:policybias_conv:proof_eq1},
\begin{align}
&\alpha_k\mathbb{E}[\langle \nabla_\theta  \Lcal^{\reweight}_{w_k,\tau_k}(x_{k+1},\pi_{\theta_k^{\bias}}), F_{w_k,\tau_k}(x_k,\theta_k^{\bias},\hat{V}_{k}^{\bias},\bar{s}_k,\bar{a}_k,\bar{s}_k',\xi_k) - \bar{F}_{w_k,\tau_k}(x_k,\theta_k^{\bias},\hat{V}_{k}^{\bias})\rangle]\notag\\
&=\alpha_k\mathbb{E}[\langle \nabla_\theta  \Lcal^{\reweight}_{w_k,\tau_k}(x_k,\pi_{\theta_k^{\bias}}), \mathbb{E}[F_{w_k,\tau_k}(x_k,\theta_k^{\bias},\hat{V}_{k}^{\bias},\bar{s}_k,\bar{a}_k,\bar{s}_k',\xi_k)\mid\Fcal_{k-1}] - \bar{F}_{w_k,\tau_k}(x_k,\theta_k^{\bias},\hat{V}_{k}^{\bias})\rangle]\notag\\
&\hspace{20pt}-\alpha_k\mathbb{E}[\langle \nabla_\theta  \Lcal^{\reweight}_{w_k,\tau_k}(x_k,\pi_{\theta_k^{\bias}})\hspace{-2pt}-\hspace{-2pt}\nabla_\theta  \Lcal^{\reweight}_{w_k,\tau_k}(x_{k+1},\pi_{\theta_k^{\bias}}), F_{w_k,\tau_k}(x_k,\theta_k^{\bias},\hat{V}_{k}^{\bias},\bar{s}_k,\bar{a}_k,\bar{s}_k',\xi_k) \hspace{-2pt}-\hspace{-2pt} \bar{F}_{w_k,\tau_k}(x_k,\theta_k^{\bias},\hat{V}_{k}^{\bias})\rangle]\notag\\
&= -\alpha_k\mathbb{E}[\langle \nabla_\theta  \Lcal^{\reweight}_{w_k,\tau_k}(x_k,\pi_{\theta_k^{\bias}})-\nabla_\theta  \Lcal^{\reweight}_{w_k,\tau_k}(x_{k+1},\pi_{\theta_k^{\bias}}), F_{w_k,\tau_k}(x_k,\theta_k^{\bias},\hat{V}_{k}^{\bias},\bar{s}_k,\bar{a}_k,\bar{s}_k',\xi_k) - \bar{F}_{w_k,\tau_k}(x_k,\theta_k^{\bias},\hat{V}_{k}^{\bias})\rangle]\notag\\
&\leq \alpha_k\cdot L_L \mathbb{E}[\|x_{k+1}-x_k\|] \cdot 2B_F\notag\\
&\leq 2B_F L_L\alpha_k\cdot\frac{B_D\zeta_k}{w_k}\notag\\
&=\frac{2B_D B_F L_L\zeta_k\alpha_k}{w_k},\label{prop:policybias_conv:proof_eq2}
\end{align}
where the first inequality is due to \eqref{lem:Lipschitz_Lcal:eq1} of Lemma~\ref{lem:Lipschitz_Lcal}, and the second inequality follows from Lemma~\ref{lem:bounded_DFG}.

For the second term of \eqref{prop:policybias_conv:proof_eq1}, 
\begin{align}
&\alpha_k\langle \nabla_\theta  \Lcal^{\reweight}_{w_k,\tau_k}(x_{k+1},\pi_{\theta_k^{\bias}}), \bar{F}_{w_k,\tau_k}(x_k,\theta_k^{\bias},\hat{V}_{k}^{\bias})-\bar{F}_{w_k,\tau_k}(x_k,\theta_k^{\bias},V_{\tau_k}^{x_k,\pi_{\theta_k^{\bias}}})\rangle\notag\\
&\leq \frac{\alpha_k}{8}\|\nabla_\theta  \Lcal^{\reweight}_{w_k,\tau_k}(x_{k+1},\pi_{\theta_k^{\bias}})\|^2 + 2\alpha_k\|\bar{F}_{w_k,\tau_k}(x_k,\theta_k^{\bias},\hat{V}_{k}^{\bias})-\bar{F}_{w_k,\tau_k}(x_k,\theta_k^{\bias},V_{\tau_k}^{x_k,\pi_{\theta_k^{\bias}}})\|^2\notag\\
&\leq \frac{\alpha_k}{4}\|\nabla_\theta  \Lcal^{\reweight}_{w_k,\tau_k}(x_{k},\pi_{\theta_k^{\bias}})\|^2 +\frac{\alpha_k}{4}\|\nabla_\theta  \Lcal^{\reweight}_{w_k,\tau_k}(x_{k},\pi_{\theta_k^{\bias}})-\nabla_\theta  \Lcal^{\reweight}_{w_k,\tau_k}(x_{k+1},\pi_{\theta_k^{\bias}})\|^2 \notag\\
&\hspace{20pt}+2\alpha_k\|\bar{F}_{w_k,\tau_k}(x_k,\theta_k^{\bias},\hat{V}_{k}^{\bias})-\bar{F}_{w_k,\tau_k}(x_k,\theta_k^{\bias},V_{\tau_k}^{x_k,\pi_{\theta_k^{\bias}}})\|^2\notag\\
&\leq\frac{\alpha_k}{4}\|\nabla_\theta  \Lcal^{\reweight}_{w_k,\tau_k}(x_{k},\pi_{\theta_k^{\bias}})\|^2+\frac{L_L^2\alpha_k}{4}\Big(\|x_{k+1}-x_k\|^2\Big)+ 2L_F^2\alpha_k\|V_{\tau_k}^{x_k,\pi_{\theta_k^{\bias}}}-\hat{V}_k^{\bias}\|^2\notag\\
&\leq\frac{\alpha_k}{4}\|\nabla_\theta  \Lcal^{\reweight}_{w_k,\tau_k}(x_{k},\pi_{\theta_k^{\bias}})\|^2+\frac{L_L^2\alpha_k}{4}\cdot\Big(\frac{B_D\zeta_k}{w_k}\Big)^2+ 2L_F^2\alpha_k\varepsilon_k^{V,\bias}\notag\\
&\leq\frac{\alpha_k}{4}\|\nabla_\theta  \Lcal^{\reweight}_{w_k,\tau_k}(x_{k},\pi_{\theta_k^{\bias}})\|^2 + \frac{B_D^2 L_L^2\zeta_k^2\alpha_k}{4w_k^2} + 2L_F^2\alpha_k\varepsilon_k^{V,\bias},\label{prop:policybias_conv:proof_eq3}
\end{align}
where the third inequality is again a result of the Lipschitz continuity of $\nabla_\theta\Lcal^{\reweight}_{w_k,\tau_k}$ from Lemma~\ref{lem:Lipschitz_V} and the Lipschitz continuity of $\bar{F}_{w_k,\tau_k}$.

For the third term of \eqref{prop:policybias_conv:proof_eq1},
\begin{align}
&-\alpha_k \langle\nabla_\theta  \Lcal^{\reweight}_{w_k,\tau_k}(x_{k+1},\pi_{\theta_k^{\bias}}), \nabla_\theta  \Lcal^{\reweight}_{w_k,\tau_k}(x_{k},\pi_{\theta_k^{\bias}}) \rangle\notag\\
&= -\alpha_k \|\nabla_\theta  \Lcal^{\reweight}_{w_k,\tau_k}(x_{k},\pi_{\theta_k^{\bias}})\|^2 + \alpha_k \langle\nabla_\theta  \Lcal^{\reweight}_{w_k,\tau_k}(x_{k},\pi_{\theta_k^{\bias}}) -\nabla_\theta  \Lcal^{\reweight}_{w_k,\tau_k}(x_{k+1},\pi_{\theta_k^{\bias}}), \nabla_\theta  \Lcal^{\reweight}_{w_k,\tau_k}(x_{k},\pi_{\theta_k^{\bias}})\rangle\notag\\
&\leq -\frac{\alpha_k}{2} \|\nabla_\theta  \Lcal^{\reweight}_{w_k,\tau_k}(x_{k},\pi_{\theta_k^{\bias}})\|^2 + \frac{\alpha_k}{2} \|\nabla_\theta  \Lcal^{\reweight}_{w_k,\tau_k}(x_{k+1},\pi_{\theta_k^{\bias}})-\nabla_\theta  \Lcal^{\reweight}_{w_k,\tau_k}(x_{k},\pi_{\theta_k^{\bias}})\|^2\notag\\
&\leq -\frac{\alpha_k}{2} \|\nabla_\theta  \Lcal^{\reweight}_{w_k,\tau_k}(x_{k},\pi_{\theta_k^{\bias}})\|^2 + \frac{L_L^2\alpha_k}{2}\Big(\|x_{k+1}-x_k\|^2\Big)\notag\\
&\leq -\frac{\alpha_k}{2} \|\nabla_\theta  \Lcal^{\reweight}_{w_k,\tau_k}(x_{k},\pi_{\theta_k^{\bias}})\|^2 + \frac{L_L^2\alpha_k}{2}\cdot\Big(\frac{B_D\zeta_k}{w_k}\Big)^2\notag\\
&\leq -\frac{\alpha_k}{2} \|\nabla_\theta  \Lcal^{\reweight}_{w_k,\tau_k}(x_{k},\pi_{\theta_k^{\bias}})\|^2 + \frac{B_D^2 L_L^2\zeta_k^2\alpha_k}{2w_k^2},\label{prop:policybias_conv:proof_eq4}
\end{align}
where the second inequality again follows from the $L_L$-smoothness of $\Lcal^{\reweight}_{w_k,\tau_k}$ shown in \eqref{lem:Lipschitz_Lcal:eq1} of Lemma~\ref{lem:Lipschitz_Lcal}.

Substituting \eqref{prop:policybias_conv:proof_eq2}-\eqref{prop:policybias_conv:proof_eq4} into \eqref{prop:policybias_conv:proof_eq1}, 
\begin{align}
&\mathbb{E}[\Lcal^{\reweight}_{w_k,\tau_k}(x_{k+1},\pi_{\theta_k^{\bias}})-\Lcal^{\reweight}_{w_{k},\tau_{k}}(x_{k+1},\pi_{\theta_{k+1}^{\bias}})] \notag\\
&\leq \frac{2B_D B_F L_L\zeta_k\alpha_k}{w_k}+\frac{\alpha_k}{4}\mathbb{E}[\|\nabla_\theta  \Lcal^{\reweight}_{w_k,\tau_k}(x_{k},\pi_{\theta_k^{\bias}})\|^2] + \frac{B_D^2 L_L^2\zeta_k^2\alpha_k}{4w_k^2} + 2L_F^2\alpha_k \mathbb{E}[\varepsilon_k^{V,\bias}]\notag\\
&\hspace{20pt}-\frac{\alpha_k}{2} \mathbb{E}[\|\nabla_\theta  \Lcal^{\reweight}_{w_k,\tau_k}(x_{k},\pi_{\theta_k^{\bias}})\|^2] + \frac{B_D^2 L_L^2\zeta_k^2\alpha_k}{2w_k^2}+\frac{L_L\alpha_k^2}{2}\mathbb{E}[\|F_{w_k,\tau_k}(\theta_k,\omega_k,\hat\mu_k,\hat{V}_{f,k},s_k,a_k,b_k,s_k',\xi_k)\|^2]\notag\\
&\leq-\frac{\alpha_k}{4} \mathbb{E}[\|\nabla_\theta  \Lcal^{\reweight}_{w_k,\tau_k}(x_{k},\pi_{\theta_k^{\bias}})\|^2] + 2L_F^2\alpha_k\mathbb{E}[\varepsilon_k^{V,\bias}] + \frac{3B_D^2 L_L^2\zeta_k^2\alpha_k}{4w_k^2} + \frac{2B_D B_F L_L\zeta_k\alpha_k}{w_k}+\frac{B_F^2 L_L\alpha_k^2}{2}\notag\\
&\leq-\frac{\alpha_k}{8} \mathbb{E}[\|\nabla_\theta  \Lcal^{\reweight}_{w_k,\tau_k}(x_{k},\pi_{\theta_k^{\bias}})\|^2] + 2L_F^2\alpha_k \mathbb{E}[\varepsilon_k^{V,\bias}] + \frac{3B_D^2 L_L^2\zeta_k^2\alpha_k}{4w_k^2} + \frac{2B_D B_F L_L\zeta_k\alpha_k}{w_k}+\frac{B_F^2 L_L\alpha_k^2}{2}\notag\\
&\hspace{20pt}-\frac{C_L^2\alpha_k\tau_k^2}{32}\mathbb{E}[\|\pi_{\theta_k^{\bias}}-\pi_{w_k,\tau_k}^\star(x_k)\|^2],\label{prop:policybias_conv:proof_eq5}
\end{align}
where the last inequality follows from \eqref{eq:PL_QG_combined}, which states that
\begin{align*}
\|\nabla_\theta \Lcal^{\reweight}_{w_k,\tau_k}(x_k,\pi_{\theta_k^{\bias}})\|\geq\frac{C_L\tau_k}{2}\|\pi_{\theta_k^{\bias}}-\pi_{w_k,\tau_k}^\star(x_k)\|.
\end{align*}

\noindent\textbf{Bound the Second Term of \eqref{prop:policybias_conv:proof_eq0}.}
We use $\theta_{w,\tau}^\star(x)$ to denote a softmax parameter that encodes the policy $\pi_{w,\tau}^\star(x)$.
We know from Lemma~\ref{lem:Lipschitz_Lcal} that $\Lcal^{\reweight}_{w,\tau}$ is $\frac{L_L}{\tau}$-smooth with respect to $x$ under $w,\tau\leq1$, 
\begin{align*}
&\Lcal^{\reweight}_{w_{k},\tau_{k}}(x_{k+1},\pi_{\theta_{k}^{\bias}})-\Lcal^{\reweight}_{w_{k},\tau_{k}}(x_{k},\pi_{\theta_{k}^{\bias}})\notag\\
&\leq \langle \nabla_x \Lcal^{\reweight}_{w_{k},\tau_{k}}(x_{k},\pi_{\theta_{k}^{\bias}}),x_{k+1}-x_k\rangle+\frac{L_L}{2\tau_k}\|x_{k+1}-x_k\|^2\notag\\
&=\langle\nabla_x \Lcal^{\reweight}_{w_{k},\tau_{k}}(x_{k},\pi_{\theta_{k}^{\bias}})-\nabla_x \Lcal^{\reweight}_{w_{k},\tau_{k}}(x_{k},\pi_{\theta_{w_k,\tau_k}^\star(x_k)}),x_{k+1}-x_k\rangle+\frac{L_L}{2\tau_k}\|x_{k+1}-x_k\|^2\notag\\
&\hspace{20pt}+\langle\nabla_x \Lcal^{\reweight}_{w_{k},\tau_{k}}(x_{k},\pi_{\theta_{w_k,\tau_k}^\star(x_k)}),x_{k+1}-x_k\rangle\notag\\
&=\langle\nabla_x \Lcal^{\reweight}_{w_{k},\tau_{k}}(x_{k},\pi_{\theta_{k}^{\bias}})-\nabla_x \Lcal^{\reweight}_{w_{k},\tau_{k}}(x_{k},\pi_{\theta_{w_k,\tau_k}^\star(x_k)}),x_{k+1}-x_k\rangle+\frac{L_L}{2\tau_k}\|x_{k+1}-x_k\|^2\notag\\
&\hspace{20pt}+\langle\nabla_x \Phi^{\reweight}_{w_{k},\tau_{k}}(x_{k}),x_{k+1}-x_k\rangle\notag\\
&\leq\langle\nabla_x \Lcal^{\reweight}_{w_{k},\tau_{k}}(x_{k},\pi_{\theta_{k}^{\bias}})-\nabla_x \Lcal^{\reweight}_{w_{k},\tau_{k}}(x_{k},\pi_{\theta_{w_k,\tau_k}^\star(x_k)}),x_{k+1}-x_k\rangle+\frac{L_L}{2\tau_k}\|x_{k+1}-x_k\|^2\notag\\
&\hspace{20pt}+\Big(\Phi^{\reweight}_{w_k,\tau_k}(x_{k+1},)-\Phi^{\reweight}_{w_k,\tau_k}(x_{k})\Big)+\frac{L_\Phi}{2\tau_k}\|x_{k+1}-x_k\|^2\notag\\
&\leq \langle\nabla_x \Lcal^{\reweight}_{w_{k},\tau_{k}}(x_{k},\pi_{\theta_{k}^{\bias}})-\nabla_x \Lcal^{\reweight}_{w_{k},\tau_{k}}(x_{k},\pi_{\theta_{w_k,\tau_k}^\star(x_k)}),x_{k+1}-x_k\rangle\notag\\
&\hspace{20pt}+\Big(\Lcal^{\reweight}_{w_k,\tau_k}(x_{k+1},\pi_{w_k,\tau_k}^\star(x_{k+1}))-\Lcal^{\reweight}_{w_k,\tau_k}(x_{k},\pi_{w_k,\tau_k}^\star(x_k))\Big)+\frac{L_\Phi}{\tau_k}\|x_{k+1}-x_k\|^2,
\end{align*}
where the last inequality is due to $L_L\leq L_\Phi$, and the second equation uses the relationship $\nabla_x \Lcal^{\reweight}_{w_{k},\tau_{k}}(x_{k},\pi_{\theta_{w_k,\tau_k}^\star(x_k)}) = \nabla_x \Phi^{\reweight}_{w_{k},\tau_{k}}(x_{k})$, which is due to $\nabla_\theta  \Lcal_{w_{k},\tau_{k}}(x_{k},\pi_{\theta_{w_k,\tau_k}^\star(x_k)})=0$ by the first-order optimality condition. The second inequality is due to the fact that $\Phi_{w,\tau}$ is $\frac{L_\Phi}{w\tau}$-smooth when $w,\tau\leq1$ (established in Lemma~\ref{lem:Lipschitz_Phi}) and that for an $L$-smooth function $f$, we have 
\[-f(y) +f(x)\leq\langle-\nabla f(x),y-x\rangle+\frac{L}{2}\|x-y\|^2.\]

Taking the expectation and plugging in the result from Lemma~\ref{lem:policybias_conv_cross_term},
\begin{align}
&\mathbb{E}[\Lcal^{\reweight}_{w_{k},\tau_{k}}(x_{k+1},\pi_{\theta_{k}^{\bias}})-\Lcal^{\reweight}_{w_{k},\tau_{k}}(x_{k},\pi_{\theta_{k}^{\bias}})]\notag\\
&\leq\frac{ C_L^2\alpha_k\tau_k^2}{64}\mathbb{E}[\|\pi_k^{\bias}-\pi_{w_k,\tau_k}^\star(x_k)\|^2]\notag\\
&\hspace{20pt}+\frac{64 L_D^2 L_L^2 \zeta_k^2}{ C_L^2\alpha_k w_k^2 \tau_k^2} \mathbb{E}[\|\pi_k-\pi_{\tau_k}^\star(x_k)\|^2+\|\pi_k^{\bias}-\pi_{w_k,\tau_k}^\star(x_k)\|^2]+\frac{32L_L^2 \zeta_k^2}{ C_L^2\alpha_k\tau_k^2}\mathbb{E}[\|\nabla_{x} \Phi_{w_k,\tau_k}(x_k)\|^2]\notag\\
&\hspace{20pt}+\mathbb{E}[\Lcal^{\reweight}_{w_k,\tau_k}(x_{k+1},\pi_{w_k,\tau_k}^\star(x_{k+1}))-\Lcal^{\reweight}_{w_k,\tau_k}(x_{k},\pi_{w_k,\tau_k}^\star(x_k))]+\frac{L_\Phi}{\tau_k}\mathbb{E}[\|x_{k+1}-x_k\|^2]\notag\\
&\leq \frac{ C_L^2\alpha_k\tau_k^2}{64}\mathbb{E}[\|\pi_k^{\bias}-\pi_{w_k,\tau_k}^\star(x_k)\|^2]+\frac{64 L_D^2 L_L^2 \zeta_k^2}{ C_L^2\alpha_k w_k^2\tau_k^2} \mathbb{E}[\|\pi_k-\pi_{\tau_k}^\star(x_k)\|^2+\|\pi_k^{\bias}-\pi_{w_k,\tau_k}^\star(x_k)\|^2]\notag\\
&\hspace{20pt}+\mathbb{E}[\Lcal^{\reweight}_{w_k,\tau_k}(x_{k+1},\pi_{w_k,\tau_k}^\star(x_{k+1}))-\Lcal^{\reweight}_{w_k,\tau_k}(x_{k},\pi_{w_k,\tau_k}^\star(x_k))]+\frac{32L_L^2 \zeta_k^2}{ C_L^2\alpha_k\tau_k^2}\mathbb{E}[\|\nabla_{x} \Phi_{w_k,\tau_k}(x_k)\|^2]+\frac{B_D^2 L_\Phi\zeta_k^2}{w_k^2\tau_k},\label{prop:policybias_conv:proof_eq6}
\end{align}
where the last inequality follows from $\|x_{k+1}-x_k\|\leq\frac{B_D \zeta_k}{w_k}$.

\noindent\textbf{Bound the Third Term of \eqref{prop:policybias_conv:proof_eq0}.}
By the definition of $\Lcal_{w,\tau}$ in \eqref{eq:def_L_w_tau},
\begin{align}
&\Lcal^{\reweight}_{w_{k+1},\tau_{k+1}}(x_{k+1},\pi_{\theta_{k+1}^{\bias}})-\Lcal^{\reweight}_{w_{k},\tau_{k}}(x_{k+1},\pi_{\theta_{k+1}^{\bias}})\notag\\
&=(w_{k+1}-w_{k}) f(x_{k+1},\pi_{\theta_{k+1}^{\bias}})\notag\\
&\hspace{20pt}+\Big(J_{\tau_{k+1}}(x_{k+1},\pi_{\tau_{k+1}}^\star(x_{k+1}))-J_{\tau_{k}}(x_{k+1},\pi_{\tau_k}^{\star}(x_{k+1}))\Big)-\Big(J_{\tau_{k+1}}(x_{k+1},\pi_{\theta_{k+1}^{\bias}})-J_{\tau_{k}}(x_{k+1},\pi_{\theta_{k+1}^{\bias}})\Big)\notag\\
&\leq (w_{k+1}-w_{k}) f(x_{k+1},\pi_{\theta_{k+1}^{\bias}}) + \frac{\log|\Acal|(\tau_{k}-\tau_{k+1})}{1-\gamma}+\frac{\tau_k-\tau_{k+1}}{1-\gamma}\mathbb{E}_{s\sim d_\rho^{\pi_{k+1}^{\bias}}}[ E(\pi_{k+1}^{\bias},s)]\notag\\
&\leq 0+\frac{\log|\Acal|(\tau_{k}-\tau_{k+1})}{1-\gamma}+\frac{\tau_k-\tau_{k+1}}{1-\gamma}\mathbb{E}_{s\sim d_\rho^{\pi_{k+1}^{\bias}}}[ E(\pi_{k+1}^{\bias},s)]\notag\\
&\leq \frac{16\log|\Acal|\tau_k}{3(1-\gamma)(k+1)},\label{prop:policybias_conv:proof_eq7}
\end{align}
where the first inequality follows from \citet{zeng2022regularized}[Lemma 3], the second inequality is due to the fact that $f$ is non-negative from Assumption~\ref{assump:f}, and the third inequality follows from Lemma~\ref{lem:tau_diff} and the relationship $E(\pi,s)\leq\log|\Acal|$ for any policy $\pi$.

Collecting the bounds in \eqref{prop:policybias_conv:proof_eq5}-\eqref{prop:policybias_conv:proof_eq7} and plugging them into \eqref{prop:policybias_conv:proof_eq0}, we get
\begin{align*}
&\mathbb{E}[\Lcal^{\reweight}_{w_{k+1},\tau_{k+1}}(x_{k+1},\pi_{\theta_{k+1}^{\bias}})-\Lcal^{\reweight}_{w_{k},\tau_{k}}(x_{k},\pi_{\theta_{k}^{\bias}})]\notag\\
&\leq-\frac{\alpha_k}{8} \mathbb{E}[\|\nabla_\theta  \Lcal^{\reweight}_{w_k,\tau_k}(x_{k},\pi_{\theta_k^{\bias}})\|^2] + 2L_F^2\alpha_k\mathbb{E}[\varepsilon_k^{V,\bias}] + \frac{3B_D^2 L_L^2\zeta_k^2\alpha_k}{4w_k^2} + \frac{2B_D B_F L_L\zeta_k\alpha_k}{w_k}+\frac{B_F^2 L_L\alpha_k^2}{2}\notag\\
&\hspace{20pt}-\frac{C_L^2\alpha_k\tau_k^2}{32}\mathbb{E}[\|\pi_{\theta_k^{\bias}}-\pi_{w_k,\tau_k}^\star(x_k)\|^2]\notag\\
&\hspace{20pt}+\frac{ C_L^2\alpha_k\tau_k^2}{64}\mathbb{E}[\|\pi_k^{\bias}-\pi_{w_k,\tau_k}^\star(x_k)\|^2]+\frac{64 L_D^2 L_L^2 \zeta_k^2}{ C_L^2\alpha_k w_k^2 \tau_k^2} \mathbb{E}[\|\pi_k-\pi_{\tau_k}^\star(x_k)\|^2+\|\pi_k^{\bias}-\pi_{w_k,\tau_k}^\star(x_k)\|^2]\notag\\
&\hspace{20pt}+\mathbb{E}[\Lcal^{\reweight}_{w_k,\tau_k}(x_{k+1},\pi_{w_k,\tau_k}^\star(x_{k+1}))-\Lcal^{\reweight}_{w_k,\tau_k}(x_{k},\pi_{w_k,\tau_k}^\star(x_k))]+\frac{32L_L^2 \zeta_k^2}{ C_L^2\alpha_k\tau_k^2}\mathbb{E}[\|\nabla_{x} \Phi_{w_k,\tau_k}(x_k)\|^2]+\frac{B_D^2 L_\Phi\zeta_k^2}{w_k^2\tau_k}\notag\\
&\hspace{20pt}+\frac{16\log|\Acal|\tau_k}{3(1-\gamma)(k+1)}\notag\\
&\leq-\frac{\alpha_k}{8} \mathbb{E}[\|\nabla_\theta  \Lcal^{\reweight}_{w_k,\tau_k}(x_{k},\pi_{\theta_k^{\bias}})\|^2] + 2L_F^2\alpha_k\mathbb{E}[\varepsilon_k^{V,\bias}] +\frac{32L_L^2 \zeta_k^2}{ C_L^2\alpha_k\tau_k^2}\mathbb{E}[\|\nabla_{x} \Phi_{w_k,\tau_k}(x_k)\|^2]\notag\\
&\hspace{20pt}-\frac{ C_L^2\alpha_k\tau_k^2}{64}\mathbb{E}[\|\pi_k^{\bias}-\pi_{w_k,\tau_k}^\star(x_k)\|^2]+\frac{64 L_D^2 L_L^2 \zeta_k^2}{ C_L^2\alpha_k w_k^2\tau_k^2} \mathbb{E}[\|\pi_k-\pi_{\tau_k}^\star(x_k)\|^2+\|\pi_k^{\bias}-\pi_{w_k,\tau_k}^\star(x_k)\|^2]\notag\\
&\hspace{20pt}+ \frac{2B_D^2 L_\Phi\zeta_k^2}{w_k^2\tau_k} + \frac{2B_D B_F L_L\zeta_k\alpha_k}{w_k}+\frac{B_F^2 L_L\alpha_k^2}{2}+\frac{16\log|\Acal|\tau_k}{3(1-\gamma)(k+1)}\notag\\
&\hspace{20pt}+\mathbb{E}[\Lcal^{\reweight}_{w_k,\tau_k}(x_{k+1},\pi_{w_k,\tau_k}^\star(x_{k+1}))-\Lcal^{\reweight}_{w_k,\tau_k}(x_{k},\pi_{w_k,\tau_k}^\star(x_k))],
\end{align*}
where in the last inequality we have combined the terms $\frac{3B_D^2 L_L^2\zeta_k^2\alpha_k}{4w_k^2}$ and $\frac{B_D^2 L_\Phi\zeta_k^2}{w_k^2\tau_k}$ under the step size conditions $\tau_k\leq1$ and $\alpha_k\leq\frac{4L_\Phi}{3L_L^2}$.

Recall the definition of $\varepsilon_k^{\theta,\bias}$ in \eqref{eq:def_metrics}. We can re-arrange the terms in the inequality above and obtain
\begin{align*}
&\mathbb{E}[\varepsilon_{k+1}^{\theta,\bias}-\varepsilon_k^{\theta,\bias}]\notag\\
&=\mathbb{E}[\Lcal^{\reweight}_{w_{k+1},\tau_{k+1}}(x_{k+1},\pi_{\theta_{k+1}^{\bias}})-\Lcal^{\reweight}_{w_{k},\tau_{k}}(x_{k},\pi_{\theta_{k}^{\bias}})] \notag\\
&\hspace{20pt}- \mathbb{E}[\Lcal^{\reweight}_{w_k,\tau_k}(x_{k+1},\pi_{w_k,\tau_k}^\star(x_{k+1}))-\Lcal^{\reweight}_{w_k,\tau_k}(x_{k},\pi_{w_k,\tau_k}^\star(x_k))]\notag\\
&\hspace{20pt}- \mathbb{E}[\Lcal^{\reweight}_{w_{k+1},\tau_{k+1}}(x_{k+1},\pi_{w_k,\tau_k}^\star(x_{k+1}))-\Lcal^{\reweight}_{w_k,\tau_k}(x_{k+1},\pi_{w_k,\tau_k}^\star(x_{k+1}))]\notag\\
&\leq -\frac{\alpha_k}{8} \mathbb{E}[\|\nabla_\theta  \Lcal^{\reweight}_{w_k,\tau_k}(x_{k},\pi_{\theta_k^{\bias}})\|^2] + 2L_F^2\alpha_k\mathbb{E}[\varepsilon_k^{V,\bias}] +\frac{32L_L^2 \zeta_k^2}{ C_L^2\alpha_k\tau_k^2}\mathbb{E}[\|\nabla_{x} \Phi_{w_k,\tau_k}(x_k)\|^2]\notag\\
&\hspace{20pt}-\frac{ C_L^2\alpha_k\tau_k^2}{64}\mathbb{E}[\|\pi_k^{\bias}-\pi_{w_k,\tau_k}^\star(x_k)\|^2]+\frac{64 L_D^2 L_L^2 \zeta_k^2}{ C_L^2\alpha_k w_k^2\tau_k^2} \mathbb{E}[\|\pi_k-\pi_{\tau_k}^\star(x_k)\|^2+\|\pi_k^{\bias}-\pi_{w_k,\tau_k}^\star(x_k)\|^2]\notag\\
&\hspace{20pt}+ \frac{2B_D^2 L_\Phi\zeta_k^2}{w_k^2\tau_k} + \frac{2B_D B_F L_L\zeta_k\alpha_k}{w_k}+\frac{B_F^2 L_L\alpha_k^2}{2}+\frac{32\log|\Acal|\tau_k}{3(1-\gamma)(k+1)},
\end{align*}
where the bound on $\Lcal^{\reweight}_{w_{k+1},\tau_{k+1}}(x_{k+1},\pi_{w_k,\tau_k}^\star(x_{k+1}))-\Lcal^{\reweight}_{w_k,\tau_k}(x_{k+1},\pi_{w_k,\tau_k}^\star(x_{k+1}))\leq\frac{16\log|\Acal|\tau_k}{3(1-\gamma)(k+1)}$ can be obtained in a manner similar to \eqref{prop:policybias_conv:proof_eq7}.

\qed

\subsection{Proof of Proposition~\ref{prop:value_conv}}

We first establish the convergence of $\varepsilon_k^V$. To this end, we introduce the following technical lemma, which bounds an important cross term.

\begin{lem}\label{lem:V_cross_term}
Under the assumptions and step sizes of Proposition~\ref{prop:value_conv}, we have for all $k\geq0$
\begin{align*}
&\mathbb{E}[\langle\hat{V}_{k} - V_{\tau_{k}}^{x_{k},\pi_{\theta_{k}}}+\beta_k \bar{G}_{\tau_k}(x_k,\theta_k,\hat{V}_{k}), V_{\tau_{k}}^{x_{k},\pi_{\theta_{k}}}-V_{\tau_{k+1}}^{x_{k+1},\pi_{\theta_{k+1}}}\rangle]\notag\\
&\leq \frac{(1-\gamma)\beta_k}{2}\mathbb{E}[\|\hat{V}_{k} - V_{\tau_{k}}^{x_{k},\pi_{\theta_{k}}}+\beta_k \bar{G}_{\tau_k}(x_k,\theta_k,\hat{V}_{k})\|^2]+\frac{6L_V^2\zeta_k^2}{(1-\gamma)\beta_k}\mathbb{E}[\|\nabla_{x} \Phi_{w_k,\tau_k}(x_k)\|^2]\notag\\
&\hspace{20pt} + \frac{12L_V^2 L_D^2\zeta_k^2}{(1-\gamma)\beta_k}\mathbb{E}[\|\pi_k-\pi_{\tau_k}^\star(x_k)\|^2] + \frac{12L_V^2 L_D^2\zeta_k^2}{(1-\gamma)\beta_k}\mathbb{E}[\|\pi_k^{\bias}-\pi_{w_k,\tau_k}^\star(x_k)\|^2]\notag\\
&\hspace{20pt}+\frac{6L_V^2\alpha_k^2}{(1-\gamma)\beta_k}\mathbb{E}[\|\nabla_\theta J_{\tau_k}(x_k,\pi_{\theta_k})\|^2]+ \frac{6L_V^2 L_F^2 \alpha_k^2}{(1-\gamma)\beta_k}\mathbb{E}[\varepsilon_k^V]+\frac{6B_F^2 L_V\tau_0\alpha_k^2}{\alpha_0}+\frac{32L_V^2\tau_k^2}{3(1-\gamma)\beta_k(k+1)^2}.
\end{align*}
\end{lem}

We defer the proof of the lemma to Appendix~\ref{proof:V_cross_term}.

By the update rule in \eqref{alg:analysis:critic}, we have
\begin{align}
\varepsilon_{k+1}^V &= \|\hat{V}_{k+1}-V_{\tau_{k+1}}^{x_{k+1},\pi_{\theta_{k+1}}}\|^2\notag\\
&=\|\Pi_{B_V}\Big( \hat{V}_{k} + \beta_k G_{\tau_k}(x_k,\theta_k,\hat{V}_{k},s_k,a_k,s_k')\Big)-V_{\tau_{k+1}}^{x_{k+1},\pi_{\theta_{k+1}}}\|^2\notag\\
&\leq \Big\| \hat{V}_{k} + \beta_k G_{\tau_k}(x_k,\theta_k,\hat{V}_{k},s_k,a_k,s_k')-V_{\tau_{k+1}}^{x_{k+1},\pi_{\theta_{k+1}}}\Big\|^2\notag\\
&= \Big\|\hat{V}_{k} - V_{\tau_{k}}^{x_{k},\pi_{\theta_{k}}} + \beta_k \bar{G}_{\tau_k}(x_k,\theta_k,\hat{V}_{k}) + \beta_k\Big(G_f(x_k,\theta_k,\hat{V}_{k},s_k,a_k,s_k') -\bar{G}_{\tau_k}(x_k,\theta_k,\hat{V}_{k})\Big)\notag\\
&\hspace{20pt}+V_{\tau_{k}}^{x_{k},\pi_{\theta_{k}}}-V_{\tau_{k+1}}^{x_{k+1},\pi_{\theta_{k+1}}}\Big\|^2\notag\\
&\leq \Big\|\hat{V}_{k} - V_{\tau_{k}}^{x_{k},\pi_{\theta_{k}}}+\beta_k \bar{G}_{\tau_k}(x_k,\theta_k,\hat{V}_{k})\Big\|^2\notag\\
&\hspace{20pt}+2\beta_k^2\|G_{\tau_k}(x_k,\theta_k,\hat{V}_{k},s_k,a_k,s_k') -\bar{G}_{\tau_k}(x_k,\theta_k,\hat{V}_{k})\|^2\notag\\
&\hspace{20pt}+2\Big\|V_{\tau_{k}}^{x_{k},\pi_{\theta_{k}}}-V_{\tau_{k+1}}^{x_{k+1},\pi_{\theta_{k+1}}}\Big\|^2\notag\\
&\hspace{20pt}+\beta_k\langle\hat{V}_{k} - V_{\tau_{k}}^{x_{k},\pi_{\theta_{k}}}+\beta_k \bar{G}_{\tau_k}(x_k,\theta_k,\hat{V}_{k}), G_{\tau_k}(x_k,\theta_k,\hat{V}_{k},s_k,a_k,s_k') -\bar{G}_{\tau_k}(x_k,\theta_k,\hat{V}_{k})\rangle\notag\\
&\hspace{20pt}+\langle\hat{V}_{k} - V_{\tau_{k}}^{x_{k},\pi_{\theta_{k}}}+\beta_k \bar{G}_{\tau_k}(x_k,\theta_k,\hat{V}_{k}), V_{\tau_{k}}^{x_{k},\pi_{\theta_{k}}}-V_{\tau_{k+1}}^{x_{k+1},\pi_{\theta_{k+1}}}\rangle,\label{prop:value_conv:proof_eq1}
\end{align}
where the first inequality follows from the fact that the projection to a convex set is non-expansive.

To bound the first term of \eqref{prop:value_conv:proof_eq1},
\begin{align}
&\Big\|\hat{V}_{k} - V_{\tau_{k}}^{x_{k},\pi_{\theta_{k}}}+\beta_k \bar{G}_{\tau_k}(x_k,\theta_k,\hat{V}_{k})\Big\|^2\notag\\
&= \|\hat{V}_{k} - V_{\tau_{k}}^{x_{k},\pi_{\theta_{k}}}\|^2+\beta_k^2\|\bar{G}_{\tau_k}(x_k,\theta_k,\hat{V}_{k})\|^2+2\beta_k\langle \hat{V}_{k} - V_{\tau_{k}}^{x_{k},\pi_{\theta_{k}}}, \bar{G}_{\tau_k}(x_k,\theta_k,\hat{V}_{k})\rangle\notag\\
&=\|\hat{V}_{k} - V_{\tau_{k}}^{x_{k},\pi_{\theta_{k}}}\|^2+\beta_k^2\|\bar{G}_{\tau_k}(x_k,\theta_k,\hat{V}_{k})-\bar{G}_{\tau_k}(x_k,\theta_k,V_{\tau_k}^{x_k,\pi_{\theta_k}})\|^2\notag\\
&\hspace{20pt}+2\beta_k\langle \hat{V}_{k} - V_{\tau_{k}}^{x_{k},\pi_{\theta_{k}}}, \bar{G}_{\tau_k}(x_k,\theta_k,\hat{V}_{k})-\bar{G}_{\tau_k}(x_k,\theta_k,V_{\tau_k}^{x_k,\pi_{\theta_k}})\rangle\notag\\
&=\|\hat{V}_{k} - V_{\tau_{k}}^{x_{k},\pi_{\theta_{k}}}\|^2+\beta_k^2\|\bar{G}_{\tau_k}(x_k,\theta_k,\hat{V}_{k})-\bar{G}_{\tau_k}(x_k,\theta_k,V_{\tau_k}^{x_k,\pi_{\theta_k}})\|^2\notag\\
&\hspace{20pt}+2\beta_k\Big(\hat{V}_{k} - V_{\tau_{k}}^{x_{k},\pi_{\theta_{k}}}\Big)^{\top}(\gamma P^{\pi_{\theta_k}}-I)\Big( \hat{V}_{k} - V_{\tau_{k}}^{x_{k},\pi_{\theta_{k}}}\Big)\notag\\
&\leq \|\hat{V}_{k} - V_{\tau_{k}}^{x_{k},\pi_{\theta_{k}}}\|^2+L_G^2\beta_k^2\|\hat{V}_{k} - V_{\tau_k}^{x_k,\pi_{\theta_k}}\|^2+2(\gamma-1)\beta_k\|\hat{V}_{k} - V_{\tau_{k}}^{x_{k},\pi_{\theta_{k}}}\|^2\notag\\
&\leq(1-(1-\gamma)\beta_k)\varepsilon_k^V,\label{prop:value_conv:proof_eq2}
\end{align}
where the second equation uses the relationship $\bar{G}_{\tau}(x,\theta,V_{\tau}^{x,\pi_{\theta}})=0$ for any $\tau,x,\theta$, and the last inequality follows from the step size condition $\beta_k\leq\frac{1-\gamma}{L_G^2}$.

For the third term of \eqref{prop:value_conv:proof_eq1},
\begin{align*}
&2\|V_{\tau_{k}}^{x_{k},\pi_{\theta_{k}}}-V_{\tau_{k+1}}^{x_{k+1},\pi_{\theta_{k+1}}}\|^2\notag\\
&\leq4\|V_{\tau_{k}}^{x_{k+1},\pi_{\theta_{k+1}}}-V_{\tau_{k+1}}^{x_{k+1},\pi_{\theta_{k+1}}}\|^2+4\|V_{\tau_{k}}^{x_{k},\pi_{\theta_{k}}}-V_{\tau_{k}}^{x_{k+1},\pi_{\theta_{k+1}}}\|^2\notag\\
&\leq4|\Scal|\|V_{\tau_{k}}^{x_{k+1},\pi_{\theta_{k+1}}}-V_{\tau_{k+1}}^{x_{k+1},\pi_{\theta_{k+1}}}\|_{\infty}^2+8L_V^2\Big(\|x_k-x_{k+1}\|^2+\|\pi_{\theta_{k+1}}-\pi_{\theta_{k}}\|^2\Big)\notag\\
&\leq4|\Scal|\cdot\Big(\frac{\tau_k-\tau_{k+1}}{1-\gamma}\log|\Acal|\Big)^2+8L_V^2\Big(\frac{B_D^2\zeta_k^2}{w_k^2}+ B_F^2\alpha_k^2\Big),
\end{align*}
where the second inequality follows from the Lipschitz continuity of the value function established in Lemma~\ref{lem:Lipschitz_V}, and the third inequality applies \citet{zeng2022regularized}[Lemma 3].
Plugging in the bound on $\tau_k-\tau_{k+1}$ from Lemma~\ref{lem:tau_diff}, we get
\begin{align}
2\|V_{\tau_{k}}^{x_{k},\pi_{\theta_{k}}}-V_{\tau_{k+1}}^{x_{k+1},\pi_{\theta_{k+1}}}\|^2&\leq4|\Scal|\cdot\Big(\frac{8\tau_k}{3(1-\gamma)(k+1)}\log|\Acal|\Big)^2+8L_V^2\Big(\frac{B_D^2\zeta_k^2}{w_k^2}+ B_F^2\alpha_k^2\Big)\notag\\
&\leq\frac{256|\Scal|\log^2|\Acal|\tau_k^2}{3(1-\gamma)^2 (k+1)^2}+16B_F^2 L_V^2\alpha_k^2,\label{prop:value_conv:proof_eq3}
\end{align}
where we use the step size condition $\zeta_k\leq\frac{B_F\alpha_k w_k}{B_D}$.

For the fourth term of \eqref{prop:value_conv:proof_eq1}, we have in expectation
\begin{align}
&\mathbb{E}[\langle\hat{V}_{k} - V_{\tau_{k}}^{x_{k},\pi_{\theta_{k}}}+\beta_k \bar{G}_{\tau_k}(x_k,\theta_k,\hat{V}_{k}), G_{\tau_k}(x_k,\theta_k,\hat{V}_{k},s_k,a_k,s_k') -\bar{G}_{\tau_k}(x_k,\theta_k,\hat{V}_{k})\rangle]\notag\\
&=\mathbb{E}[\langle\hat{V}_{k} - V_{\tau_{k}}^{x_{k},\pi_{\theta_{k}}}+\beta_k \bar{G}_{\tau_k}(x_k,\theta_k,\hat{V}_{k}), \mathbb{E}[G_{\tau_k}(x_k,\theta_k,\hat{V}_{k},s_k,a_k,s_k') -\bar{G}_{\tau_k}(x_k,\theta_k,\hat{V}_{k})\mid\Fcal_{k-1}]\rangle]\notag\\
&=0.\label{prop:value_conv:proof_eq4}
\end{align}

Collecting the bounds from \eqref{prop:value_conv:proof_eq2}-\eqref{prop:value_conv:proof_eq4} and Lemma~\ref{lem:V_cross_term}, 
\begin{align*}
&\mathbb{E}[\varepsilon_{k+1}^V] \notag\\
&= (1-(1-\gamma)\beta_k)\mathbb{E}[\varepsilon_k^V]+2\beta_k^2\mathbb{E}[\|G_{\tau_k}(x_k,\theta_k,\hat{V}_{k},s_k,a_k,s_k') -\bar{G}_{\tau_k}(x_k,\theta_k,\hat{V}_{k})\|^2]\notag\\
&\hspace{20pt}+\frac{256|\Scal|\log^2|\Acal|\tau_k^2}{3(1-\gamma)^2 (k+1)^2} + 16B_F^2 L_V^2\alpha_k^2\notag\\
&\hspace{20pt}+\frac{(1-\gamma)\beta_k}{2}\mathbb{E}[\|\hat{V}_{k} - V_{\tau_{k}}^{x_{k},\pi_{\theta_{k}}}+\beta_k \bar{G}_{\tau_k}(x_k,\theta_k,\hat{V}_{k})\|^2]+\frac{6L_V^2\zeta_k^2}{(1-\gamma)\beta_k}\mathbb{E}[\|\nabla_{x} \Phi_{w_k,\tau_k}(x_k)\|^2]\notag\\
&\hspace{20pt} + \frac{12L_V^2 L_D^2\zeta_k^2}{(1-\gamma)\beta_k}\mathbb{E}[\|\pi_k-\pi_{\tau_k}^\star(x_k)\|^2] + \frac{12L_V^2 L_D^2\zeta_k^2}{(1-\gamma)\beta_k}\mathbb{E}[\|\pi_k^{\bias}-\pi_{w_k,\tau_k}^\star(x_k)\|^2]\notag\\
&\hspace{20pt}+\frac{6L_V^2\alpha_k^2}{(1-\gamma)\beta_k}\mathbb{E}[\|\nabla_\theta J_{\tau_k}(x_k,\pi_{\theta_k})\|^2]+ \frac{6L_V^2 L_F^2 \alpha_k^2}{(1-\gamma)\beta_k}\mathbb{E}[\varepsilon_k^V]+\frac{6B_F^2 L_V\tau_0\alpha_k^2}{\alpha_0}+\frac{32L_V^2\tau_k^2}{3(1-\gamma)\beta_k(k+1)^2}\notag\\
&\leq \Big(1-(1-\gamma)\beta_k\Big)\mathbb{E}[\varepsilon_k^V]+\Big(1-(1-\gamma)\beta_k\Big)\frac{(1-\gamma)\beta_k}{2}\mathbb{E}[\varepsilon_k^V]+\frac{6L_V^2 L_F^2 \alpha_k^2}{(1-\gamma)\beta_k}\mathbb{E}[\varepsilon_k^V]+8B_G\beta_k^2\notag\\
&\hspace{20pt}+\frac{256|\Scal|\log^2|\Acal|\tau_k^2}{3(1-\gamma)^2 (k+1)^2}+16B_F^2 L_V^2\alpha_k^2+\frac{6L_V^2\zeta_k^2}{(1-\gamma)\beta_k}\mathbb{E}[\|\nabla_{x} \Phi_{w_k,\tau_k}(x_k)\|^2]\notag\\
&\hspace{20pt} + \frac{12L_V^2 L_D^2\zeta_k^2}{(1-\gamma)\beta_k}\mathbb{E}[\|\pi_k-\pi_{\tau_k}^\star(x_k)\|^2] + \frac{12L_V^2 L_D^2\zeta_k^2}{(1-\gamma)\beta_k}\mathbb{E}[\|\pi_k^{\bias}-\pi_{w_k,\tau_k}^\star(x_k)\|^2]\notag\\
&\hspace{20pt}+\frac{6L_V^2\alpha_k^2}{(1-\gamma)\beta_k}\mathbb{E}[\|\nabla_\theta J_{\tau_k}(x_k,\pi_{\theta_k})\|^2]+\frac{6B_F^2 L_V\tau_0\alpha_k^2}{\alpha_0}+\frac{32L_V^2\tau_k^2}{3(1-\gamma)\beta_k(k+1)^2}\notag\\
&\leq \Big(1-\frac{(1-\gamma)\beta_k}{4}\Big)\mathbb{E}[\varepsilon_k^V] + \frac{12L_V^2 L_D^2\zeta_k^2}{(1-\gamma)\beta_k}\mathbb{E}[\|\pi_k-\pi_{\tau_k}^\star(x_k)\|^2] + \frac{12L_V^2 L_D^2\zeta_k^2}{(1-\gamma)\beta_k}\mathbb{E}[\|\pi_k^{\bias}-\pi_{w_k,\tau_k}^\star(x_k)\|^2]\notag\\
&\hspace{20pt}+\frac{6L_V^2\alpha_k^2}{(1-\gamma)\beta_k}\mathbb{E}[\|\nabla_\theta J_{\tau_k}(x_k,\pi_{\theta_k})\|^2]+\frac{6L_V^2\zeta_k^2}{(1-\gamma)\beta_k}\mathbb{E}[\|\nabla_{x} \Phi_{w_k,\tau_k}(x_k)\|^2]\notag\\
&\hspace{20pt}+\frac{22B_F^2 L_V\tau_0\alpha_k^2}{\alpha_0}+\frac{64L_V^2\tau_k^2}{3(1-\gamma)\beta_k(k+1)^2} + 8B_G\beta_k^2,
\end{align*}
where the second inequality simplifies and combines terms based on the step size conditions $\frac{\alpha_k}{\beta_k}\leq\frac{1-\gamma}{2\sqrt{6}L_V L_F}$, $\beta_k\leq\frac{(1-\gamma)L_V^2}{8|\Scal|\log^2|\Acal|}$, and $\frac{\alpha_0}{\tau_0}\leq\frac{1}{L_V}$.

The bound on $\mathbb{E}[\varepsilon_{k+1}^{V,\bias}]$ can be derived using an identical argument.

\qed

\section{Proof of Supporting Results}\label{sec:proof_lemma}

\subsection{Proof of Lemma~\ref{lem:pi_star_unique}}\label{sec:pi_star_unique}

\textbf{Uniqueness of $\pi^\star(x)$.}
We first take the approach of proof by contradiction to show that $\pi^\star(x)$ is unique for any $x$. 

Given $x$, suppose that there exist two distinct optimal solutions of \eqref{eq:def_pi_star}, which we denote by $\pi_1,\pi_2$. From the definition of $\Pi^\star(x)$, we know that $\pi_1,\pi_2$ satisfy
\begin{align}
J(x,\pi_1)\geq J(x,\pi),\quad J(x,\pi_2)\geq J(x,\pi),\quad\forall \pi.\label{lem:pi_star_unique:proof_eq1}
\end{align}

We construct another policy $\pi'$ as follows, inspired by the proof of Theorem 1 in \citet{zeng2023connected}. For all state $s$ and action $a$, $\pi'(a\mid s)$ is expressed as
\begin{align}
\pi'(a\mid s)=\frac{d_\rho^{\pi_1}(s) \pi_1(a \mid s)+d_\rho^{\pi_2}(s) \pi_2(a \mid s)}{d_\rho^{\pi_1}(s)+d_\rho^{\pi_2}(s)}.\label{lem:pi_star_unique:proof_eq2}
\end{align}
Note that Assumption~\ref{assump:exploration} guarantees $d_\rho^{\pi}(s)\geq(1-\gamma)\rho_{\min}>0$ for all $s$, ensuring $\pi'$ is well-defined.

Our first step is to show that $\pi'$ is also an optimal policy, i.e., $\pi'\in\Pi^\star(x)$. To see this, we define a modified transition kernel $\Pcal_\gamma$ such that $\Pcal_\gamma(s'\mid s,a)=(1-\gamma)\rho(s')+\gamma \Pcal(s'\mid s,a)$ for any $\pi$, and $P_\gamma^{\pi}$ such that $P_\gamma^{\pi}(s'\mid s)=(1-\gamma)\rho(s')+\gamma P^{\pi}(s'\mid s)$ for any $\pi$. 

We also define a vector $B\in\mathbb{R}^{|\Scal|}$
\begin{align*}
B=\Pcal_\gamma^{\pi'}\cdot\left(\frac{1}{2}d_\rho^{\pi_1}+\frac{1}{2}d_\rho^{\pi_2}\right).
\end{align*}

We can express each entry of $B$ in the following way
\begin{align*}
    B(s') &= \sum_{s,a}\Pcal_\gamma(s'\mid s,a)\pi'(a\mid s)\left(\frac{1}{2}d_\rho^{\pi_1}(s)+\frac{1}{2}d_\rho^{\pi_2}(s)\right)\notag\\
    &=\sum_{s,a}\Pcal_\gamma(s'\mid s,a)\frac{d_\rho^{\pi_1}(s)\pi_1(a\mid s)+d_\rho^{\pi_2}(s)\pi_2(a\mid s)}{d_\rho^{\pi_1}(s)+d_\rho^{\pi_2}(s)}\left(\frac{1}{2} d_\rho^{\pi_1}(s)+\frac{1}{2}d_\rho^{\pi_2}(s)\right)\notag\\
    &=\frac{1}{2}\sum_{s,a}\Pcal_\gamma(s'\mid s,a) d_\rho^{\pi_1}(s)\pi_1(a\mid s)+\frac{1}{2}\sum_{s,a}\Pcal(s'\mid s,a) d_\rho^{\pi_2}(s)\pi_2(a\mid s)\notag\\
    &=\frac{1}{2}\sum_{s,a}P_\gamma^{\pi_1}(s'\mid s) d_\rho^{\pi_1}(s)+\frac{1}{2}\sum_{s,a}P_\gamma^{\pi_2}(s'\mid s) d_\rho^{\pi_2}(s)\notag\\
    &=\frac{1}{2}d_\rho^{\pi_1}(s')+\frac{1}{2}d_\rho^{\pi_2}(s'),
\end{align*}
which leads to
\begin{align}
P_\gamma^{\pi'}\cdot\left(\frac{1}{2}d_\rho^{\pi_1}+\frac{1}{2}d_\rho^{\pi_2}\right)=\frac{1}{2}d_\rho^{\pi_1}+\frac{1}{2}d_\rho^{\pi_2}.\label{lem:pi_star_unique:proof_eq2.5}
\end{align}

The Markov chain induced by $P_\gamma^{\pi}$ is always ergodic, assumed in Assumption~\ref{assump:exploration}. Under ergodicity, it is known that $d_\rho^\pi$ (properly normalized) is the unique eigenvector of $P_\gamma^\pi$ associated with eigenvalue 1.
Therefore, \eqref{lem:pi_star_unique:proof_eq2.5} implies that $\frac{1}{2}d_\rho^{\pi_1}+\frac{1}{2}d_\rho^{\pi_2}$ is the discounted visitation distribution induced by policy $\pi'$, i.e.,
\begin{align}
d_\rho^{\pi'}=\frac{1}{2}d_\rho^{\pi_1}+\frac{1}{2}d_\rho^{\pi_2}.\label{lem:pi_star_unique:proof_eq3}
\end{align}

We use $\hat{d}_\rho^{\pi}$ to denote the extend discounted visitation distribution over state and action such that
\[\hat{d}_\rho^{\pi}(s,a)=d_\rho^{\pi}(s)\pi(a\mid s).\]
We have from \eqref{lem:pi_star_unique:proof_eq3}
\begin{align*}
    \hat{d}_\rho^{\pi'}(s,a) &= d_\rho^{\pi'}(s)\pi'(a\mid s)\\
    &=\left(\frac{1}{2}d_\rho^{\pi_1}(s)+\frac{1}{2}d_\rho^{\pi_2}(s)\right)\frac{d_\rho^{\pi_1}(s)\pi_1(a\mid s)+d_\rho^{\pi_2}(s)\pi_2(a\mid s)}{d_\rho^{\pi_1}(s)+d_\rho^{\pi_2}\mu_{\pi_2}(s)}\\
    &=\frac{1}{2} d_\rho^{\pi_1}(s)\pi_1(a\mid s)+\frac{1}{2}d_\rho^{\pi_1}(s)\pi_2(a\mid s)\\
    &=\frac{1}{2} \hat{d}_\rho^{\pi_1}(s,a) + \frac{1}{2} \hat{d}_\rho^{\pi_2}(s,a).
\end{align*}

Note that the cumulative return $J$ is linear in the space of extended discounted visitation distribution. We have
\begin{align*}
J(x,\pi')=\langle r_x, \hat{d}_\rho^{\pi'}\rangle=\frac{1}{2}\langle r_x, \frac{1}{2} \hat{d}_\rho^{\pi_1}(s,a) + \frac{1}{2} \hat{d}_\rho^{\pi_2}(s,a)\rangle=\frac{1}{2}J(x,\pi_1)+\frac{1}{2}J(x,\pi_2).
\end{align*}

In view of \eqref{lem:pi_star_unique:proof_eq1}, this implies that $\pi'$ is an optimal policy, i.e., $\pi'\in\Pi^\star(x)$.

Since $\pi'$ is in the constraint set for the optimization problem in \eqref{eq:def_pi_star}, we can create a contradiction that $\pi_1,\pi_2$ are the two distinct maximizers of \eqref{eq:def_pi_star} if $\pi'$ has a larger weighted entropy. 
The entropy function $E(\dot,s)$ is strictly concave for all state $s$ for policies in the interior of the simplex (note that $\pi_1,\pi_2$ must in the interior of the simplex, as they cannot be the optimal solution of \eqref{eq:def_pi_star} otherwise). Recall from \eqref{lem:pi_star_unique:proof_eq2} that $\pi'(\cdot\mid s)$ is a convex combination of $\pi_1(\cdot\mid s),\pi_2(\cdot\mid s)$, and by the property of strictly concave functions,
\begin{align*}
E(\pi',s)>\frac{d_\rho^{\pi_1}(s)E(\pi_1, s)+d_\rho^{\pi_2}(s)E(\pi_2, s)}{d_\rho^{\pi_1}(s)+d_\rho^{\pi_2}(s)}.
\end{align*}

We denote $\Ecal_\pi=\mathbb{E}_{s\sim d_\rho^\pi}[E(\pi,s)]$. 
\begin{align*}
\Ecal_{\pi'}&=\langle d_\rho^{\pi'}, E(\pi',\cdot)\rangle\notag\\
&=\sum_{s}\big(\frac{1}{2}d_\rho^{\pi_1}(s)+\frac{1}{2}d_\rho^{\pi_2}(s)\Big)E(\pi',s)\notag\\
&>\sum_{s}\big(\frac{1}{2}d_\rho^{\pi_1}(s)+\frac{1}{2}d_\rho^{\pi_2}(s)\Big)\frac{d_\rho^{\pi_1}(s)E(\pi_1, s)+d_\rho^{\pi_2}(s)E(\pi_2, s)}{d_\rho^{\pi_1}(s)+d_\rho^{\pi_2}(s)}\notag\\
&=\frac{1}{2}\langle d_\rho^{\pi_1},E(\pi_1,s)\rangle+\frac{1}{2}\langle d_\rho^{\pi_2},E(\pi_2,s)\rangle\notag\\
&=\frac{1}{2}\Ecal_{\pi_1}+\frac{1}{2}\Ecal_{\pi_2}.
\end{align*}
This contradicts the condition that $\pi_1,\pi_2$ maximize the weighted entropy within the constraint set $\Pi^\star(x)$ and concludes our proof on the uniqueness of $\pi^\star(x)$.

\textbf{Limit Point of $\{\pi_{\tau}^\star(x)\}_\tau$.} 
We then show the limit point of $\{\pi_{\tau}^\star(x)\}_\tau$ is $\pi^\star(x)$ as $\tau\rightarrow 0$.
As $x$ is fixed here, we simply denote $\pi_\tau=\pi_\tau^\star(x)$. We define for simplicity $\bar{E}(\pi)\triangleq\frac{1}{1-\gamma}\mathbb{E}_{s\sim d_\rho^{\pi}}[E(\pi,s)]$.
By the Bolzano–Weierstrass theorem, as the sequence $\{\pi_{\tau}\}$ is bounded, it has a convergent subsequence. Let $\tau_n\rightarrow0$ and $\pi_{\tau_n}\rightarrow\bar{\pi}$. We first need to show $\bar{\pi}\in\Pi^\star(x)$. By the definition of $\pi_\tau^\star(x)$,
\begin{align}
J(x,\pi_\tau)+\tau\bar{E}(\pi_\tau) \geq J(x,\pi^\star(x))+\tau\bar{E}(\pi^\star(x)),\label{sec:limit_pi_tau_star:proof_eq1}
\end{align}
leading to 
\begin{align*}
J(x,\pi_\tau) \geq J(x,\pi^\star(x))+\tau\Big(\bar{E}(\pi^\star(x))-\bar{E}(\pi_\tau)\Big).
\end{align*}

As $J$ is continuous, we take $n\rightarrow\infty$
\[\lim\sup J(x,\pi_\tau)\geq J(x,\pi^\star(x)),\]
implying $J(x,\bar{\pi})$. This means $\bar{\pi}\in\Pi^\star$.

Then, to show $\bar{\pi}$ is the maximum entropy solution, we re-arrange the terms in \eqref{sec:limit_pi_tau_star:proof_eq1}
\[\bar{E}(\pi_\tau)-\bar{E}(\pi^\star(x))\geq\frac{J(x,\pi^\star(x))-J(x,\pi_\tau)}{\tau}\geq0,\]
where the second inequality is due to the definition of $\pi^\star(x)$. 

Taking the limit,
\[\lim\sup\bar{E}(\pi_\tau)\geq\bar{E}(\pi^\star(x)).\]

As the limit point $\bar{\pi}$ is in $\Pi^\star$ and we have $\pi^\star(x)=\argmax_{\pi\in\Pi^\star}\bar{E}(\pi)$, then it holds that $\lim\sup\bar{E}(\pi_\tau)=\bar{E}(\pi^\star(x))=\bar{E}(\bar{\pi})$. This allows us to conclude that $\pi_\tau^\star(x)\rightarrow\pi^\star(x)$ as $\tau\rightarrow0$.

\qed

\subsection{Proof of Lemma~\ref{lem:tau_diff}}
We apply \citet{zeng2022regularized}[Lemma 7]
\begin{align*}
\tau_k-\tau_{k+1}&=\frac{\tau_0}{(k+1)^{c_\tau}}-\frac{\tau_0}{(k+2)^{c_\tau}}\leq \frac{8\tau_0}{3(k+1)^{c_\tau+1}}=\frac{8\tau_k}{3(k+1)}.
\end{align*}

\qed

\subsection{Proof of Lemma~\ref{lem:Lipschitz_V}}

We derive the inequalities on the value function $V_{\tau}^{x,\pi_{\theta}}$ and note that the ones on the cumulative return $J_{\tau}(x,\pi_{\theta})$ immediately follows as the cumulative return is simply an weighted average of the value function
\[J_{\tau}(x,\pi_{\theta})=\mathbb{E}_{s\sim\rho}[V_{\tau}^{x,\pi_{\theta}}(s)].\]

Fixing $x$, we know that $V_{\tau}^{x,\pi}$ is the standard policy optimization objective (in a standard MDP) as a function of policy $\pi$. It is well-known that the following inequalities hold (for example, see Lemma B.5 of \citet{zeng2021decentralized} and Lemma 5 of \citet{zeng2022regularized})
\begin{gather}
\|V^{x,\pi_{\theta}}-V^{x,\pi_{\theta'}}\|\leq \frac{2}{(1-\gamma)^2}\|\theta-\theta'\|,\label{lem:Lipschitz_V:proof_eq1}\\
\|\nabla_\theta V^{x,\pi_{\theta}}-\nabla_\theta V^{x,\pi_{\theta'}}\|\leq \frac{8}{(1-\gamma)^3}\|\theta-\theta'\|.
\label{lem:Lipschitz_V:proof_eq2}
\end{gather}

We define the shorthand notation
\[\Hcal(\theta,s)\triangleq\mathbb{E}_{a_k\sim\pi(\cdot\mid s_k),s_{k+1}\sim\Pcal(\cdot\mid s_k,a_k)}\left[\sum_{k=0}^{\infty}-\gamma^k \log\pi_{\theta}(a_k\mid s_k) \mid s_0=s\right].\]
This implies $V_\tau^{x,\pi_{\theta}}(s)=V^{x,\pi_{\theta}}(s)+\tau\Hcal(\theta,s)$. We also define the aggregate notation 
\[\Hcal(\theta)=[\Hcal(\theta,s_1),\Hcal(\theta,s_2),\cdots]\in\mathbb{R}^{|\Scal|}.\]
Adapting Lemma 6 of \citet{zeng2022regularized}, we have for all $s\in\Scal$
\begin{gather}
\|\Hcal(\theta,s)-\Hcal(\theta',s)\|\leq\frac{4+8\log|\Acal|}{(1-\gamma)^3}\|\theta-\theta'\|,\label{lem:Lipschitz_V:proof_eq3.0}\\
\|\nabla_\theta\Hcal(\theta,s)-\nabla_\theta\Hcal(\theta',s)\|\leq\frac{4+8\log|\Acal|}{(1-\gamma)^3}\|\theta-\theta'\|.\label{lem:Lipschitz_V:proof_eq4.0}
\end{gather}
We obviously have the following inequalities from \eqref{lem:Lipschitz_V:proof_eq3.0} and \eqref{lem:Lipschitz_V:proof_eq4.0}
\begin{gather}
\|\Hcal(\theta)-\Hcal(\theta')\|\leq\frac{(4+8\log|\Acal|)\sqrt{|\Scal|}}{(1-\gamma)^3}\|\theta-\theta'\|,\label{lem:Lipschitz_V:proof_eq3}\\
\|\nabla_\theta\Hcal(\theta)-\nabla_\theta\Hcal(\theta')\|\leq\frac{(4+8\log|\Acal|)\sqrt{|\Scal|}}{(1-\gamma)^3}\|\theta-\theta'\|.\label{lem:Lipschitz_V:proof_eq4}
\end{gather}

Note that \eqref{lem:Lipschitz_V:proof_eq4.0} also implies
\[\|\nabla_{\theta,\theta}\mathbb{E}_{s\sim d_\rho^{\pi_{\theta}}}[E(\pi_{\theta},s)]\|\leq \frac{4+8\log|\Acal|}{(1-\gamma)^3}\]
hence leading to \eqref{lem:Lipschitz_V:eq5}.

In addition, we have from \eqref{eq:def_V_tau}
\begin{align}
|V_\tau^{x,\pi_{\theta}}-V_\tau^{x',\pi_{\theta}}|&=|\mathbb{E}_{s'\sim d_s^{\pi},\,a'\sim\pi(\cdot\mid s')}[r_x(s,a)-r_{x'}(s,a)]|\notag\\
&\leq \mathbb{E}_{s'\sim d_s^{\pi},\,a'\sim\pi(\cdot\mid s')}[|r_x(s,a)-r_{x'}(s,a)|]\notag\\
&\leq \mathbb{E}_{s'\sim d_s^{\pi},\,a'\sim\pi(\cdot\mid s')}[L_r\|x-x'\|]\notag\\
&\leq L_r\|x-x'\|,\label{lem:Lipschitz_V:proof_eq5}
\end{align}
where the first inequality is a result of Jensen's inequality and the second inequality is due to Assumption~\ref{assump:reward}.

We can express $\nabla_\theta V^{x,\pi_{\theta}}$ as follows \citep{agarwal2021theory}
\begin{align*}
\nabla_{\theta_{s',a'}}V^{x,\pi_{\theta}}(s)=\frac{1}{1-\gamma}d_s^{\pi_{\theta}}(s')\pi_{\theta}(a'\mid s')A^{x,\pi_{\theta}}(s',a').
\end{align*}
This implies
\begin{align*}
|\nabla_{\theta_{s',a'}}V^{x,\pi_{\theta}}(s)-\nabla_{\theta_{s',a'}}V^{x',\pi_{\theta}}(s)|&=\frac{1}{1-\gamma}d_s^{\pi_{\theta}}(s')\pi_{\theta}(a'\mid s')|A^{x,\pi_{\theta}}(s',a')-A^{x',\pi_{\theta}}(s',a')|\notag\\
&\leq \frac{1}{1-\gamma}|d_s^{\pi_{\theta}}(s')||\pi_{\theta}(a'\mid s')|\cdot 2L_r\|x-x'\|\notag\\
&\leq\frac{2}{1-\gamma}\|x-x'\|,
\end{align*}
where the first inequality is due to the fact that the advantage function is $2L_r$-Lipschitz with respect to $x$, since it is the difference of the value function and Q function, which are themselves $L_r$-Lipschitz with respect to $x$ as can be seen from \eqref{lem:Lipschitz_V:proof_eq5}.
Then, as the entropy regularizer is not a function of $x$, we have
\begin{align}
\|\nabla_\theta V_\tau^{x,\pi_{\theta}}-\nabla_\theta V_\tau^{x,\pi_{\theta'}}\|&=\|\nabla_\theta V^{x,\pi_{\theta}}-\nabla_\theta V^{x,\pi_{\theta'}}\|\notag\\
&\leq \frac{2L_r|\Scal||\Acal|}{1-\gamma}\|x-x'\|.\label{lem:Lipschitz_V:proof_eq6}
\end{align}

Combining \eqref{lem:Lipschitz_V:proof_eq1}, \eqref{lem:Lipschitz_V:proof_eq3}, and \eqref{lem:Lipschitz_V:proof_eq5}, we have for all $\tau\leq 1$
\begin{align*}
\|V_\tau^{x,\pi_{\theta}}-V_\tau^{x',\pi_{\theta'}}\|&\leq\|V_\tau^{x,\pi_{\theta}}-V_\tau^{x',\pi_{\theta}}\|+\|V_\tau^{x',\pi_{\theta}}-V_\tau^{x',\pi_{\theta'}}\|\notag\\
&\leq \|V_\tau^{x,\pi_{\theta}}-V_\tau^{x',\pi_{\theta}}\|+\tau\|\Hcal(\theta)-\Hcal(\theta')\|+\|V^{x',\pi_{\theta}}-V^{x',\pi_{\theta'}}\|\notag\\
&\leq L_r\|x-x'\|+\frac{(4+8\log|\Acal|)\sqrt{|\Scal|}}{(1-\gamma)^3}\|\theta-\theta'\|+\frac{2}{(1-\gamma)^2}\|\theta-\theta'\|\notag\\
&\leq L_r\|x-x'\|+\frac{(6+8\log|\Acal|)\sqrt{|\Scal|}}{(1-\gamma)^3}\|\theta-\theta'\|.
\end{align*}
This shows the Lipschitz continuity of the value function.

To show smoothness with respect to $\theta$, we combine \eqref{lem:Lipschitz_V:proof_eq2}, \eqref{lem:Lipschitz_V:proof_eq4}, and \eqref{lem:Lipschitz_V:proof_eq6}
\begin{align*}
\|\nabla_\theta V_\tau^{x,\pi_{\theta}}-\nabla_\theta V_\tau^{x',\pi_{\theta'}}\|&\leq\|\nabla_\theta V_\tau^{x,\pi_{\theta}}-\nabla_\theta V_\tau^{x',\pi_{\theta}}\|+\|\nabla_\theta V_\tau^{x',\pi_{\theta}}-\nabla_\theta V_\tau^{x',\pi_{\theta'}}\|\notag\\
&\leq \|\nabla_\theta V_\tau^{x,\pi_{\theta}}-\nabla_\theta V_\tau^{x',\pi_{\theta}}\|+\tau\|\nabla_\theta \Hcal(\theta)-\nabla_\theta \Hcal(\theta')\|+\|\nabla_\theta V^{x',\pi_{\theta}}-\nabla_\theta V^{x',\pi_{\theta'}}\|\notag\\
&\leq \frac{2L_r|\Scal||\Acal|}{1-\gamma}\|x-x'\|+\frac{(4+8\log|\Acal|)\sqrt{|\Scal|}}{(1-\gamma)^3}\|\theta-\theta'\|+\frac{8}{(1-\gamma)^3}\|\theta-\theta'\|\notag\\
&\leq \frac{2L_r|\Scal||\Acal|}{1-\gamma}\|x-x'\|+\frac{(12+8\log|\Acal|)\sqrt{|\Scal|}}{(1-\gamma)^3}\|\theta-\theta'\|.
\end{align*}

To show \eqref{lem:Lipschitz_V:eq2.5},
\begin{align*}
&\|\nabla_x J_\tau(x,\pi)-\nabla_x J_\tau(x',\pi')\| \notag\\
&\leq \|\nabla_x J_\tau(x,\pi)-\nabla_x J_\tau(x,\pi')\| + \|\nabla_x J_\tau(x,\pi')-\nabla_x J_\tau(x',\pi')\|\notag\\
&\leq \|\sum_{s,a}(d_\rho^{\pi}(s,a)-d_\rho^{\pi'}(s,a))\nabla_x r_x(s,a)\| + \mathbb{E}_{s\sim d_\rho^{\pi'},\,a\sim\pi'(\cdot\mid s)}[\|\nabla_x r_x(s,a)-\nabla_x r_{x'}(s,a)\|]\notag\\
&\leq L_r\|d_\rho^{\pi}-d_\rho^{\pi'}\|+\mathbb{E}_{s\sim d_\rho^{\pi'},\,a\sim\pi'(\cdot\mid s)}[L_r\|x-x'\|]\notag\\
&=\left\|(1-\gamma)\Big((I-\gamma P^{\pi})^{-1}-(I-\gamma P^{\pi'})^{-1}\Big)\rho\right\|+L_r\|x-x'\|\notag\\
&\leq (1-\gamma)\cdot\frac{\gamma}{(1-\gamma)^2}\|\pi-\pi'\|\|\rho\|+L_r\|x-x'\|\notag\\
&\leq L_V(\|\pi-\pi'\|+\|x-x'\|),
\end{align*}
where the fourth inequality is due to the fact that $(I-\gamma P^\pi)$ is $\gamma/(1-\gamma)^2$ Lipschitz in $\pi$.

Finally, we show $\nabla_{x,\theta}^2 J_\tau(x,\pi_\theta)$ and $\nabla_{\theta,\theta}^2 J_\tau(x,\pi_\theta)$ are Lipschitz. Since $\nabla_{\theta,\theta}\mathbb{E}_{s\sim d_\rho^{\pi_{\theta}}}[E(\pi_{\theta},s)]$ does not depend on $x$ and can be shown to have Lipschitz Hessians by extending the argument in \citet{mei2020global}[Lemma 14] (we skip showing the exact constant here), the problem reduces to showing that $\nabla_{x,\theta}^2 J(x,\pi_\theta)$ and $\nabla_{\theta,\theta}^2 J(x,\pi_\theta)$ are Lipschitz.

From \eqref{eq:grad_J_eq1}, we have
\begin{gather}
\nabla_{x,\theta}
^2 J(x,\pi_\theta) = \frac{1}{1-\gamma} \mathbb{E}_{s\sim d_{\rho}^{\pi_\theta},a\sim\pi_\theta(\cdot\mid s),s'\sim\Pcal(\cdot\mid s,a)}\left[\Big(\nabla_x r_x(s, a)+\gamma \nabla_x V_\tau^{x,\pi_\theta}(s')\Big) \nabla_\theta \log \pi_\theta(a \mid s)\right].\label{lem:Lipschitz_V:proof_eq7}
\end{gather}

We define $\operatorname{traj}=\{s_0,a_0,s_1,a_1,\cdots\}$ and use $p(\pi,\operatorname{traj})$ to denote the probability that the trajectory $\operatorname{traj}$ is generated under the policy $\pi$, i.e.,
\[p(\pi,\operatorname{traj})=\rho(s_0)\prod_{k=0}^\infty \pi(a_k\mid s_k)\Pcal(s_{k+1}\mid s_k,a_k).\]
Adapting the result from \citet{shen2019hessian}, we have 
\begin{gather}
\nabla_{\theta,\theta}
^2 J(x,\pi_\theta) = \mathbb{E}_{\operatorname{traj}\sim p(\pi_\theta,\cdot)}\Big[\nabla_\theta\phi(x,\pi,\operatorname{traj})\nabla_\theta p(\pi_\theta,\operatorname{traj})^\top+\nabla_{\theta,\theta}^2\phi(x,\pi,\operatorname{traj})\Big],\label{lem:Lipschitz_V:proof_eq8}
\end{gather}
where we define $\phi(x,\pi,\operatorname{traj})=\sum_{t=0}^\infty(\sum_{k=t}^{\infty}\gamma^k (r_x(s_k,a_k))^k)\log\pi(a_t\mid s_t)$.

As the right hand side expressions in \eqref{lem:Lipschitz_V:proof_eq7} and \eqref{lem:Lipschitz_V:proof_eq8} are the composition of Lipschitz functions, we know that $\nabla_{x,\theta}
^2 J(x,\pi_\theta)$ and $\nabla_{x,\theta}
^2 J(\theta,\pi_\theta)$ are Lipschitz.

\qed

\subsection{Proof of Lemma~\ref{lem:Lipschitz_discountedvisitation}}\label{sec:proof_lem:Lipschitz_discountedvisitation}

We can write the discounted visitation distribution as follows
\begin{align*}
d_\rho^\pi = (1-\gamma)(I-\gamma P^\pi)^{-1}\rho,
\end{align*}
which implies
\begin{align*}
\|d_\rho^\pi-d_\rho^{\pi'}\|&\leq(1-\gamma)\|(I-\gamma P^\pi)^{-1}\rho-(I-\gamma P^{\pi'})^{-1}\|\|\rho\|\notag\\
&= (1-\gamma)\|\rho\| \Big\|(I-\gamma P^{\pi'})^{-1}\Big((I-\gamma P^{\pi'})-(I-\gamma P^\pi)\Big)(I-\gamma P^\pi)^{-1}\Big\|\notag\\
&\leq (1-\gamma)\cdot 1\cdot\|(I-\gamma P^\pi)^{-1}\|\|(I-\gamma P^{\pi'})^{-1}\|\cdot\gamma\|P^\pi-P^{\pi'}\|\notag\\
&\leq \frac{\gamma}{1-\gamma}\|\pi-\pi'\|,
\end{align*}
where the third inequality follows from the standard result $\|P^\pi-P^{\pi'}\|\leq\sqrt{|\Acal|}\|\pi-\pi'\|$.

\qed

\subsection{Proof of Lemma~\ref{lem:bounded_DFG}}

By the definition of $D_w$ in \eqref{eq:def_D},
\begin{align*}
\|D_w(x,\pi,\pi^{\bias}, s, a, \bar{s},\bar{a},\xi)\| &= \left\|\widetilde\nabla_x f(x,\pi^{\bias},\xi)+\frac{1}{w}\Big(\nabla_x r_x(s,a) + \nabla_x r_x(\bar{s},\bar{a})\Big)\right\|\notag\\
&\leq \|\widetilde\nabla_x f(x,\pi^{\bias},\xi)\| + \frac{1}{w}\|\nabla_x r_x(s,a) + \nabla_x r_x(\bar{s},\bar{a})\|\notag\\
&\leq L_f + \frac{1}{w}(L_r+L_r)\notag\\
&\leq\frac{3L_r}{w},
\end{align*}
where the second inequality follows from Assumption~\ref{assump:f} and the condition $\|\nabla_x r_x(s,a)\|\leq L_r,\,\forall x,s,a$, which follows from Assumption~\ref{assump:reward}, and the last inequality is due to $w\leq\frac{L_r}{L_f}$.

To bound $F$, note that $\nabla_{\theta}\log\pi_\theta(a\mid s)$ has the following closed-form expression entry-wise
\begin{align}
\frac{\partial \log \pi_\theta(a \mid s)}{\partial \theta_{s', a'}}=\1\left[s=s^{\prime}\right]\left(\1\left[a=a'\right]-\pi_\theta\left(a' \mid s\right)\right).\label{lem:bounded_DFG:proof_eq1}
\end{align}
This implies
\begin{align}
\|\nabla_{\theta}\log\pi_\theta(a\mid s)\|_2\leq\|\nabla_{\theta}\log\pi_\theta(a\mid s)\|_1\leq 1+1=2.\label{lem:bounded_DFG:proof_eq2}
\end{align}

By the definition of $F_{w,\tau}$ in \eqref{eq:def_F},
\begin{align*}
\|F_{w,\tau}(x,\theta,V,s,a,s',\xi)\|&\leq\|\nabla_{\theta}\log\pi_\theta(a\mid s)\| \Big|r_x(s,a)-\tau \log\pi_{\theta}(a\mid s) +\gamma V(s')-V(s)\Big|+w\|\widetilde\nabla_\theta f(x,\pi_\theta,\xi)\|\notag\\
&\leq2(1+\tau|\log\underline{\pi}|+\gamma B_V+B_V)+w L_f\notag\\
&\leq2(1+\gamma)B_V+2|\log\underline{\pi}|+2+L_r,
\end{align*}
where the second inequality applies \eqref{lem:bounded_DFG:proof_eq2}, and the last inequality follows from $w\leq\frac{L_r}{L_f},\tau\leq1$.

Finally, by the definition of $G$ in \eqref{eq:def_G},
\begin{align*}
\|G_{\tau}(x,\theta,V,s,a,s')\| &\leq \|e_{s}\|\Big|r_x(s,a)-\tau \log\pi_{\theta}(a\mid s)+\gamma V(s')-V(s)\Big|\notag\\
&\leq 1\cdot(1+\tau|\log\underline{\pi}|+\gamma B_V+B_V)\notag\\
&\leq(1+\gamma)B_V+|\log\underline{\pi}|+1.
\end{align*}

\qed

\subsection{Proof of Lemma~\ref{lem:Lipschitz_DFG}}

By the definition of $\bar{D}_w$ in \eqref{eq:def_bar_D},
\begin{align}
&\|\bar{D}_w(x_1,\pi_1,\pi_1^{\bias})-\bar{D}_w(x_2,\pi_2,\pi_2^{\bias})\|\notag\\
&=\Big\|\mathbb{E}_{s \sim d_\rho^{\pi_1}, a \sim \pi_1(\cdot \mid s), \bar{s} \sim d_\rho^{\pi_1^{\bias}}, \bar{a} \sim \pi_1^{\bias}(\cdot \mid \bar{s})),\xi\sim\mu}[D_w(x_1,\pi_1,\pi_1^{\bias}, s, a, \bar{s},\bar{a},\xi)]\notag\\
&\hspace{20pt}-\mathbb{E}_{s \sim d_\rho^{\pi_2}, a \sim \pi_2(\cdot \mid s), \bar{s} \sim d_\rho^{\pi_2^{\bias}}, \bar{a} \sim \pi_2^{\bias}(\cdot \mid \bar{s})),\xi\sim\mu}[D_w(x_2,\pi_2,\pi_2^{\bias}, s, a, \bar{s},\bar{a},\xi)]\Big\|\notag\\
&=\Big\|\hspace{-5pt}\sum_{s,a,\bar{s},\bar{a},\xi}\hspace{-5pt}\Big(d_\rho^{\pi_1}(s)\pi_1(a \mid s) d_\rho^{\pi_1^{\bias}}(\bar{s}) \pi_1^{\bias}(\bar{a} \mid \bar{s})-d_\rho^{\pi_2}(s)\pi_2(a \mid s) d_\rho^{\pi_2^{\bias}}(\bar{s}) \pi_2^{\bias}(\bar{a} \mid \bar{s})\Big) \mu(\xi) D_w(x_2,\pi_2,\pi_2^{\bias}, s, a, \bar{s},\bar{a},\xi)\notag\\
&\hspace{20pt}+\mathbb{E}_{s \sim d_\rho^{\pi_1}, a \sim \pi_1(\cdot \mid s), \bar{s} \sim d_\rho^{\pi_1^{\bias}}, \bar{a} \sim \pi_1^{\bias}(\cdot \mid \bar{s})),\xi\sim\mu}[D_w(x_1,\pi_1,\pi_1^{\bias}, s, a, \bar{s},\bar{a},\xi)-D_w(x_2,\pi_2,\pi_2^{\bias}, s, a, \bar{s},\bar{a},\xi)]\Big\|\notag\\
&\leq\Big\|\mathbb{E}_{s \sim d_\rho^{\pi_1}, a \sim \pi_1(\cdot \mid s), \bar{s} \sim d_\rho^{\pi_1^{\bias}}, \bar{a} \sim \pi_1^{\bias}(\cdot \mid \bar{s})),\xi\sim\mu}[D_w(x_1,\pi_1,\pi_1^{\bias}, s, a, \bar{s},\bar{a},\xi)-D_w(x_2,\pi_2,\pi_2^{\bias}, s, a, \bar{s},\bar{a},\xi)]\Big\|\notag\\
&\hspace{20pt}+\frac{B_D}{w}\Big|\sum_{s,a,\bar{s},\bar{a}}\Big(d_\rho^{\pi_1}(s)\pi_1(a \mid s) d_\rho^{\pi_1^{\bias}}(\bar{s}) \pi_1^{\bias}(\bar{a} \mid \bar{s})-d_\rho^{\pi_2}(s)\pi_2(a \mid s) d_\rho^{\pi_2^{\bias}}(\bar{s}) \pi_2^{\bias}(\bar{a} \mid \bar{s})\Big)\Big|\notag\\
&\leq\Big\|\mathbb{E}_{s \sim d_\rho^{\pi_1}, a \sim \pi_1(\cdot \mid s), \bar{s} \sim d_\rho^{\pi_1^{\bias}}, \bar{a} \sim \pi_1^{\bias}(\cdot \mid \bar{s})),\xi\sim\mu}[D_w(x_1,\pi_1,\pi_1^{\bias}, s, a, \bar{s},\bar{a},\xi)-D_w(x_2,\pi_2,\pi_2^{\bias}, s, a, \bar{s},\bar{a},\xi)]\Big\|\notag\\
&\hspace{20pt}+\frac{B_D}{w}\Big|\sum_{s,a,\bar{s},\bar{a}}\Big(d_\rho^{\pi_1}(s)\pi_1(a \mid s) -d_\rho^{\pi_2}(s)\pi_2(a \mid s)\Big) d_\rho^{\pi_1^{\bias}}(\bar{s}) \pi_1^{\bias}(\bar{a} \mid \bar{s})\Big|\notag\\
&\hspace{20pt}+\frac{B_D}{w}\Big|\sum_{s,a,\bar{s},\bar{a}}\Big(d_\rho^{\pi_1^{\bias}}(\bar{s}) \pi_1^{\bias}(\bar{a} \mid \bar{s})-d_\rho^{\pi_2^{\bias}}(\bar{s}) \pi_2^{\bias}(\bar{a} \mid \bar{s})\Big) d_\rho^{\pi_2}(s)\pi_2(a \mid s) \Big|\notag\\
&=\Big\|\mathbb{E}_{s \sim d_\rho^{\pi_1}, a \sim \pi_1(\cdot \mid s), \bar{s} \sim d_\rho^{\pi_1^{\bias}}, \bar{a} \sim \pi_1^{\bias}(\cdot \mid \bar{s})),\xi\sim\mu}[D_w(x_1,\pi_1,\pi_1^{\bias}, s, a, \bar{s},\bar{a},\xi)-D_w(x_2,\pi_2,\pi_2^{\bias}, s, a, \bar{s},\bar{a},\xi)]\Big\|\notag\\
&\hspace{20pt}+\frac{B_D}{w}\Big(\big|\sum_{s,a}(d_\rho^{\pi_1}(s)\pi_1(a \mid s) -d_\rho^{\pi_2}(s)\pi_2(a \mid s))\big| + \big|\sum_{\bar{s},\bar{a}}(d_\rho^{\pi_1^{\bias}}(\bar{s}) \pi_1^{\bias}(\bar{a} \mid \bar{s})-d_\rho^{\pi_2^{\bias}}(\bar{s}) \pi_2^{\bias}(\bar{a} \mid \bar{s}))\big|\Big).\label{lem:Lipschitz_DFG:proof_eq1}
\end{align}

To bound the first term of \eqref{lem:Lipschitz_DFG:proof_eq1}, note that by Jensen's inequality it suffices to bound the norm of the term within the expectation
\begin{align}
&\|D_w(x_1,\pi_1,\pi_1^{\bias}, s, a, \bar{s},\bar{a},\xi)-D_w(x_2,\pi_2,\pi_2^{\bias}, s, a, \bar{s},\bar{a},\xi)\|\notag\\
&=\left\|\widetilde\nabla_{x}f(x_1,\pi_1^{\bias},\xi)+\frac{1}{w}\Big(\nabla_x r_{x_1}(s,a)-\nabla_x r_{x_1}(\bar{s},\bar{a})\Big)-\widetilde\nabla_{x}f(x_2,\pi_2^{\bias},\xi)-\frac{1}{w}\Big(\nabla_x r_{x_2}(s,a)-\nabla_x r_{x_2}(\bar{s},\bar{a})\Big)\right\|\notag\\
&\leq \|\widetilde\nabla_{x}f(x_1,\pi_1^{\bias},\xi)-\widetilde\nabla_{x}f(x_2,\pi_2^{\bias},\xi)\| + \frac{1}{w}\|\nabla_x r_{x_1}(s,a)-\nabla_x r_{x_2}(s,a)\| + \frac{1}{w}\|\nabla_x r_{x_1}(\bar{s},\bar{a})-\nabla_x r_{x_2}(\bar{s},\bar{a})\|\notag\\
&\leq L_f\Big(\|x_1-x_2\|+\|\pi_1^{\bias}-\pi_2^{\bias}\|\Big) + \frac{2L_r}{w}\|x_1-x_2\|.\label{lem:Lipschitz_DFG:proof_eq2}
\end{align}

To bound the second term of \eqref{lem:Lipschitz_DFG:proof_eq1}, 
\begin{align}
&\big|\sum_{s,a}(d_\rho^{\pi_1}(s)\pi_1(a \mid s) -d_\rho^{\pi_2}(s)\pi_2(a \mid s))\big| + \big|\sum_{\bar{s},\bar{a}}(d_\rho^{\pi_1^{\bias}}(\bar{s}) \pi_1^{\bias}(\bar{a} \mid \bar{s})-d_\rho^{\pi_2^{\bias}}(\bar{s}) \pi_2^{\bias}(\bar{a} \mid \bar{s}))\big|\notag\\
&\leq \big|\sum_{s,a}(d_\rho^{\pi_1}(s) -d_\rho^{\pi_2}(s))\pi_1(a \mid s)\big| + \big|\sum_{s,a}(\pi_1(a \mid s) -\pi_2(a \mid s))d_\rho^{\pi_2}(s)\big|\notag\\
&\hspace{20pt} + \big|\sum_{\bar{s},\bar{a}}(d_\rho^{\pi_1^{\bias}}(\bar{s}) -d_\rho^{\pi_2^{\bias}}(\bar{s}))\pi_1^{\bias}(\bar{a} \mid \bar{s})\big| + \big|\sum_{\bar{s},\bar{a}}(\pi_1^{\bias}(\bar{a} \mid \bar{s}) -\pi_2^{\bias}(\bar{a} \mid \bar{s}))d_\rho^{\pi_2^{\bias}}(\bar{s})\big|\notag\\
&\leq \|d_\rho^{\pi_1}-d_\rho^{\pi_2}\|+\|\pi_1-\pi_2\|+\|d_\rho^{\pi_1^{\bias}}-d_\rho^{\pi_2^{\bias}}\|+\|\pi_1^{\bias}-\pi_2^{\bias}\|\notag\\
&\leq \frac{1}{1-\gamma}\|\pi_1-\pi_2\|+\frac{1}{1-\gamma}\|\pi_1^{\bias}-\pi_2^{\bias}\|,\label{lem:Lipschitz_DFG:proof_eq3}
\end{align}
where the final inequality follows from Lemma~\ref{lem:Lipschitz_discountedvisitation}.

Substituting \eqref{lem:Lipschitz_DFG:proof_eq2} and \eqref{lem:Lipschitz_DFG:proof_eq3} into \eqref{lem:Lipschitz_DFG:proof_eq1}, we have
\begin{align}
&\|\bar{D}_w(x_1,\pi_1,\pi_1^{\bias})-\bar{D}_w(x_2,\pi_2,\pi_2^{\bias})\|\notag\\
&\leq L_f\Big(\|x_1-x_2\|+\|\pi_1^{\bias}-\pi_2^{\bias}\|\Big) + \frac{2L_r}{w}\|x_1-x_2\| + \frac{B_D}{(1-\gamma)w}\|\pi_1-\pi_2\|+\frac{B_D}{(1-\gamma)w}\|\pi_1^{\bias}-\pi_2^{\bias}\|\notag\\
&= (L_f+\frac{2L_r}{w})\|x_1-x_2\|+(L_f+\frac{B_D}{(1-\gamma)w})\|\pi_1^{\bias}-\pi_2^{\bias}\|+\frac{B_D}{(1-\gamma)w}\|\pi_1-\pi_2\|\notag\\
&\leq \frac{3L_r}{w}\|x_1-x_2\|+\frac{2B_D}{(1-\gamma)w}\|\pi_1^{\bias}-\pi_2^{\bias}\|+\frac{B_D}{(1-\gamma)w}\|\pi_1-\pi_2\|,
\end{align}
where the second inequality simplifies terms under the condition $w\leq\min\{\frac{L_r}{L_f},\frac{B_D}{(1-\gamma)L_f}\}$.

Similarly, by the definition of $\bar{F}_{w,\tau}$ in \eqref{eq:def_bar_F},
\begin{align}
&\|\bar{F}_{w,\tau}(x_1,\theta_1,V_1) - \bar{F}_{w,\tau}(x_2,\theta_2,V_2)\|\notag\\
&=\Big\|\mathbb{E}_{s \sim d_\rho^{\pi_{\theta_1}}, a \sim \pi_{\theta_1}(\cdot \mid s), s' \sim \Pcal(\cdot\mid s,a),\xi\sim\mu}[F_{w,\tau}(x_1,\theta_1,V_1, s, a, s',\xi)]\notag\\
&\hspace{20pt}-\mathbb{E}_{s \sim d_\rho^{\pi_{\theta_2}}, a \sim \pi_{\theta_2}(\cdot \mid s), s' \sim \Pcal(\cdot\mid s,a),\xi\sim\mu}[F_{w,\tau}(x_2,\theta_2,V_2, s, a, s',\xi)]\Big\|\notag\\
&=\Big\|\hspace{-5pt}\sum_{s,a,s',\xi}\Big(d_\rho^{\pi_{\theta_1}}(s)\pi_{\theta_1}(a \mid s) \Pcal(s'\mid s,a)-d_\rho^{\pi_{\theta_2}}(s)\pi_{\theta_2}(a \mid s) \Pcal(s'\mid s,a)\Big) \mu(\xi) F_{w,\tau}(x_2,\theta_2,V_2, s, a, s',\xi)\notag\\
&\hspace{20pt}+\mathbb{E}_{s \sim d_\rho^{\pi_{\theta_1}}, a \sim \pi_{\theta_1}(\cdot \mid s), s' \sim \Pcal(\cdot\mid s,a),\xi\sim\mu}[F_{w,\tau}(x_1,\theta_1,V_1, s, a, s',\xi)-F_{w,\tau}(x_2,\theta_2,V_2, s, a, s',\xi)]\Big\|\notag\\
&\leq\Big\|\mathbb{E}_{s \sim d_\rho^{\pi_{\theta_1}}, a \sim \pi_{\theta_1}(\cdot \mid s), s' \sim \Pcal(\cdot\mid s,a),\xi\sim\mu}[F_{w,\tau}(x_1,\theta_1,V_1, s, a, s',\xi)-F_{w,\tau}(x_2,\theta_2,V_2, s, a, s',\xi)]\Big\|\notag\\
&\hspace{20pt}+B_F\Big|\sum_{s,a,s'}\Big(d_\rho^{\pi_{\theta_1}}(s)\pi_{\theta_1}(a \mid s) \Pcal(s'\mid s,a)-d_\rho^{\pi_{\theta_2}}(s)\pi_{\theta_2}(a \mid s) \Pcal(s'\mid s,a)\Big)\Big|\notag\\
&=\Big\|\mathbb{E}_{s \sim d_\rho^{\pi_{\theta_1}}, a \sim \pi_{\theta_1}(\cdot \mid s), s' \sim \Pcal(\cdot\mid s,a),\xi\sim\mu}[F_{w,\tau}(x_1,\theta_1,V_1, s, a, s',\xi)-F_{w,\tau}(x_2,\theta_2,V_2, s, a, s',\xi)]\Big\|\notag\\
&\hspace{20pt}+B_F\Big|\sum_{s,a}\Big(d_\rho^{\pi_{\theta_1}}(s)\pi_{\theta_1}(a \mid s) -d_\rho^{\pi_{\theta_2}}(s)\pi_{\theta_2}(a \mid s)\Big)\Big|.\label{lem:Lipschitz_DFG:proof_eq4}
\end{align}

To bound the first term of \eqref{lem:Lipschitz_DFG:proof_eq1}, note that by Jensen's inequality it suffices to bound the norm of the term within the expectation
\begin{align}
&\|F_{w,\tau}(x_1,\theta_1,V_1, s, a, s',\xi) - F_{w,\tau}(x_2,\theta_2,V_2, s, a, s',\xi)\|\notag\\
&=\Big\|\big(r_{x_1}(s,a)-\tau \log\pi_{\theta_1}(a\mid s)+\gamma V_1(s')-V_1(s)\big)\nabla_{\theta}\log\pi_{\theta_1}(a\mid s)-w\widetilde\nabla_\theta f(x_1,\pi_{\theta_1},\xi)\notag\\
&\hspace{20pt}-\big(r_{x_2}(s,a)-\tau \log\pi_{\theta_2}(a\mid s)+\gamma V_2(s')-V_2(s)\big)\nabla_{\theta}\log\pi_{\theta_2}(a\mid s) + w\widetilde\nabla_\theta f(x_2,\pi_{\theta_2},\xi)\Big\|\notag\\
&\leq w\|\widetilde\nabla_{\theta}f(x_1,\pi_{\theta_1},\xi)-\widetilde\nabla_{\theta}f(x_2,\pi_{\theta_2},\xi)\| \notag\\
&\hspace{20pt}+ \left\|\nabla_\theta\log\pi_{\theta_1}(a\mid s)-\nabla_\theta\log\pi_{\theta_2}(a\mid s)\right\| |r_{x_1}(s,a)-\tau \log\pi_{\theta_1}(a\mid s)+\gamma V_1(s')-V_1(s)|\notag\\
&\hspace{20pt}+ \left\|\nabla_\theta\log\pi_{\theta_2}(a\mid s)\right\| |r_{x_1}(s,a)-r_{x_2}(s,a)-\tau \log\pi_{\theta_1}(a\mid s)+\tau\log\pi_{\theta_2}(a\mid s)+\gamma V_1(s')-\gamma V_2(s')-V_1(s)+V_2(s)|\notag\\
&\leq L_f w(\|x_1-x_2\|+\|\pi_{\theta_1}-\pi_{\theta_2}\|)\notag\\
&\hspace{20pt}+|r_{x_1}(s,a)-\tau\log\pi_{\theta_1}(a\mid s)+\gamma V_1(s')-V_1(s)|\|\theta_1-\theta_2\|\notag\\
&\hspace{20pt}+\left\|\nabla_\theta\log\pi_{\theta_2}(a\mid s)\right\| \Big(L_r\|x_1-x_2\|+\frac{\tau}{\underline{\pi}}\|\theta_1-\theta_2\|+(1+\gamma)\|V_1-V_2\|\Big)\notag\\
&\leq L_r(\|x_1-x_2\|+\|\theta_1-\theta_2\|)+(1+|\log\underline{\pi}|+\frac{\gamma}{1-\gamma}+\frac{1}{1-\gamma})\|\theta_1-\theta_2\|\notag\\
&\hspace{20pt}+2\Big(L_r\|x_1-x_2\|+\frac{1}{\underline{\pi}}\|\theta_1-\theta_2\|+(1+\gamma)\|V_1-V_2\|\Big)\notag\\
&\leq3L_r\|x_1-x_2\|+(L_r+\frac{2}{\underline{\pi}}+|\log\underline{\pi}|+\frac{2}{1-\gamma}+1)\|\theta_1-\theta_2\|+4\|V_1-V_2\|,\label{lem:Lipschitz_DFG:proof_eq5}
\end{align}
where the second inequality is due to the $1$-Lipschitz continuity of $\nabla_\theta\log\pi_{\theta}$ and the $\log|\Acal|$-Lipschitz continuity of the entropy function with respect to the softmax parameter, and the third inequality is due to $w\leq\frac{L_r}{L_f},\tau\leq1$ and the fact that $\|\nabla_{\theta}\log\pi_\theta(a\mid s)\|_2\leq2$ for all $\theta$.

To bound the second term of \eqref{lem:Lipschitz_DFG:proof_eq4}, 
\begin{align}
\big|\sum_{s,a}(d_\rho^{\pi_{\theta_1}}(s)\pi_{\theta_1}(a \mid s) - d_\rho^{\pi_{\theta_2}}(s)\pi_{\theta_2}(a \mid s))\big|
&\leq \big|\sum_{s,a}(d_\rho^{\pi_{\theta_1}}(s) -d_\rho^{\pi_{\theta_2}}(s))\pi_{\theta_1}(a \mid s)\big| \notag\\
&\hspace{20pt}+ \big|\sum_{s,a}(\pi_{\theta_1}(a \mid s) - \pi_{\theta_2}(a \mid s))d_\rho^{\pi_{\theta_2}}(s)\big|\notag\\
&\leq \|d_\rho^{\pi_{\theta_1}}-d_\rho^{\pi_{\theta_2}}\|+\|\pi_{\theta_1}-\pi_{\theta_2}\|\notag\\
&\leq \frac{1}{1-\gamma}\|\pi_{\theta_1}-\pi_{\theta_2}\|\notag\\
&\leq \frac{1}{1-\gamma}\|\theta_1-\theta_2\|,\label{lem:Lipschitz_DFG:proof_eq6}
\end{align}
where the third inequality follows from Lemma~\ref{lem:Lipschitz_discountedvisitation}.

Substituting \eqref{lem:Lipschitz_DFG:proof_eq5} and \eqref{lem:Lipschitz_DFG:proof_eq6} into \eqref{lem:Lipschitz_DFG:proof_eq3}, we have
\begin{align*}
&\|\bar{F}_{w,\tau}(x_1,\theta_1,V_1) - \bar{F}_{w,\tau}(x_2,\theta_2,V_2)\|\notag\\
&\leq 3L_r\|x_1-x_2\|+(L_r+\frac{2}{\underline{\pi}}+|\log\underline{\pi}|+\frac{2}{1-\gamma}+1)\|\theta_1-\theta_2\|+4\|V_1-V_2\|+\frac{B_F}{1-\gamma}\|\theta_1-\theta_2\|\notag\\
&= 3L_r\|x_1-x_2\|+(L_r+\frac{2}{\underline{\pi}}+|\log\underline{\pi}|+\frac{2+B_F}{1-\gamma}+1)\|\theta_1-\theta_2\|+4\|V_1-V_2\|.
\end{align*}

By the definition of $\bar{G}_\tau$ in \eqref{eq:def_bar_G},
\begin{align}
&\|\bar{G}_{\tau}(x_1,\theta_1,V_1) - \bar{G}_{\tau}(x_2,\theta_2,V_2)\|\notag\\
&=\Big\|\mathbb{E}_{s \sim d_\rho^{\pi_{\theta_1}}, a \sim \pi_{\theta_1}(\cdot \mid s), s' \sim \Pcal(\cdot\mid s,a)}[G_{\tau}(x_1,\theta_1,V_1, s, a, s')]\notag\\
&\hspace{20pt}-\mathbb{E}_{s \sim d_\rho^{\pi_{\theta_2}}, a \sim \pi_{\theta_2}(\cdot \mid s), s' \sim \Pcal(\cdot\mid s,a)}[G_{\tau}(x_2,\theta_2,V_2, s, a, s')]\Big\|\notag\\
&=\Big\|\hspace{-5pt}\sum_{s,a,s'}\Big(d_\rho^{\pi_{\theta_1}}(s)\pi_{\theta_1}(a \mid s) \Pcal(s'\mid s,a)-d_\rho^{\pi_{\theta_2}}(s)\pi_{\theta_2}(a \mid s) \Pcal(s'\mid s,a)\Big) G_{\tau}(x_2,\theta_2,V_2, s, a, s',\xi)\notag\\
&\hspace{20pt}+\mathbb{E}_{s \sim d_\rho^{\pi_{\theta_1}}, a \sim \pi_{\theta_1}(\cdot \mid s), s' \sim \Pcal(\cdot\mid s,a)}[G_{\tau}(x_1,\theta_1,V_1, s, a, s')-G_{\tau}(x_2,\theta_2,V_2, s, a, s')]\Big\|\notag\\
&\leq\Big\|\mathbb{E}_{s \sim d_\rho^{\pi_{\theta_1}}, a \sim \pi_{\theta_1}(\cdot \mid s), s' \sim \Pcal(\cdot\mid s,a)}[G_{\tau}(x_1,\theta_1,V_1, s, a, s')-G_{\tau}(x_2,\theta_2,V_2, s, a, s')]\Big\|\notag\\
&\hspace{20pt}+B_G\Big|\sum_{s,a,s'}\Big(d_\rho^{\pi_{\theta_1}}(s)\pi_{\theta_1}(a \mid s) \Pcal(s'\mid s,a)-d_\rho^{\pi_{\theta_2}}(s)\pi_{\theta_2}(a \mid s) \Pcal(s'\mid s,a)\Big)\Big|\notag\\
&=\Big\|\mathbb{E}_{s \sim d_\rho^{\pi_{\theta_1}}, a \sim \pi_{\theta_1}(\cdot \mid s), s' \sim \Pcal(\cdot\mid s,a),\xi\sim\mu}[F_{w,\tau}(x_1,\theta_1,V_1, s, a, s',\xi)-F_{w,\tau}(x_2,\theta_2,V_2, s, a, s',\xi)]\Big\|\notag\\
&\hspace{20pt}+B_G\Big|\sum_{s,a}\Big(d_\rho^{\pi_{\theta_1}}(s)\pi_{\theta_1}(a \mid s) -d_\rho^{\pi_{\theta_2}}(s)\pi_{\theta_2}(a \mid s)\Big)\Big|\notag\\
&\leq \Big\|\mathbb{E}_{s \sim d_\rho^{\pi_{\theta_1}}, a \sim \pi_{\theta_1}(\cdot \mid s), s' \sim \Pcal(\cdot\mid s,a),\xi\sim\mu}[F_{w,\tau}(x_1,\theta_1,V_1, s, a, s',\xi)-F_{w,\tau}(x_2,\theta_2,V_2, s, a, s',\xi)]\Big\|\notag\\
&\hspace{20pt}+\frac{B_G}{1-\gamma}\|\theta_1-\theta_2\|,\label{lem:Lipschitz_DFG:proof_eq7}
\end{align}
where the last inequality follows from \eqref{lem:Lipschitz_DFG:proof_eq6}.

The first of \eqref{lem:Lipschitz_DFG:proof_eq7} can be bounded as follows. Again, by Jensen's inequality it suffices to bound the norm of the term within the expectation
\begin{align}
&\|G_{\tau}(x_1,\theta_1,V_1, s, a, s') - G_{\tau}(x_2,\theta_2,V_2, s, a, s')\|\notag\\
&=\Big\|\big(r_{x_1}(s,a)-\tau \log\pi_{\theta_1}(a\mid s)+\gamma V_1(s')-V_1(s)\big)e_s-\big(r_{x_2}(s,a)-\tau \log\pi_{\theta_2}(a\mid s)+\gamma V_2(s')-V_2(s)\big)e_s\Big\|\notag\\
&\leq L_r\|x_1-x_2\|+\frac{\tau}{\underline{\pi}}\|\theta_1-\theta_2\|+(1+\gamma)\|V_1-V_2\|,\label{lem:Lipschitz_DFG:proof_eq8}
\end{align}
where in the last inequality we again use the $\log|\Acal|$-Lipschitz continuity of the entropy function with respect to the softmax parameter.

Combining \eqref{lem:Lipschitz_DFG:proof_eq7} and \eqref{lem:Lipschitz_DFG:proof_eq8}, we have under $\tau$
\begin{align*}
\|\bar{G}_{\tau}(x_1,\theta_1,V_1) - \bar{G}_{\tau}(x_2,\theta_2,V_2)\|&\leq L_r\|x_1-x_2\|+\frac{\tau}{\underline{\pi}}\|\theta_1-\theta_2\|+(1+\gamma)\|V_1-V_2\|\notag\\
&\hspace{20pt}+\frac{B_G}{1-\gamma}\|\theta_1-\theta_2\|\notag\\
&\leq L_r\|x_1-x_2\|+(\frac{B_G}{1-\gamma}+\frac{1}{\underline{\pi}})\|\theta_1-\theta_2\|+2\|V_1-V_2\|.
\end{align*}

\qed

\subsection{Proof of Lemma~\ref{lem:Lipschitz_BR}}

The proof proceeds in a manner similar to \citet{kwon2023fully}[Lemma 3.2], with strong convexity replaced by the PL condition. 

First, we consider a fixed $\tau$. Recall the definition of $\pi_{w,\tau}^\star$ in Section~\ref{sec:penalty_reformulation}. 
Let $\theta_{w_1,\tau}^\star(x_1)$ denote a softmax parameter that encodes $\pi_{w_1,\tau}^\star(x_1)$.
The optimality condition of $\theta_{w_1,\tau}^\star(x_1)$ indicates
\begin{align}
\nabla_\theta \Lcal_{w_1,\tau}(x_1,\pi_{\theta_{w_1,\tau}^\star(x_1)}) = \nabla_\theta f(x_1,\pi_{\theta_{w_1,\tau}^\star(x_1)})-\frac{1}{w_1}\nabla_\pi J_{\tau}(x_1,\pi_{\theta_{w_1,\tau}^\star(x_1)})=0,\label{lem:Lipschitz_BR:proof_eq1}
\end{align}
which obviously implies
\begin{align}
\|\nabla_\theta J_{\tau}(x_1,\pi_{\theta_{w_1,\tau}^\star(x_1)})\|=w_1\|\nabla_\theta f(x_1,\pi_{\theta_{w_1,\tau}^\star(x_1)})\|\leq L_f w_1.\label{lem:Lipschitz_BR:proof_eq2}
\end{align}

Applying the relationship in \eqref{lem:Lipschitz_BR:proof_eq1}, we have
\begin{align*}
&\nabla_{\theta} \Lcal_{w_2,\tau}(x_2,\pi_{\theta_{w_1,\tau}^\star(x_1)})\notag\\
&=\nabla_{\theta}f(x_2,\pi_{\theta_{w_1,\tau}^\star(x_1)})-\frac{1}{w_2}\nabla_{\pi}J_\tau(x_2,\pi_{\theta_{w_1,\tau}^\star(x_1)})\notag\\
&=\Big(\nabla_{\theta}f(x_2,\pi_{\theta_{w_1,\tau}^\star(x_1)})-\nabla_{\theta}f(x_1,\pi_{\theta_{w_1,\tau}^\star(x_1)})\Big)-\frac{1}{w_2}\Big(\nabla_{\theta} J_\tau(x_2,\pi_{\theta_{w_1,\tau}^\star(x_1)})-\nabla_{\theta}J_\tau(x_1,\pi_{\theta_{w_1,\tau}^\star(x_1)})\Big)\notag\\
&\hspace{20pt}+\nabla_{\theta}f(x_1,\pi_{\theta_{w_1,\tau}^\star(x_1)})-\frac{1}{w_2}\nabla_{\pi}J_\tau(x_1,\pi_{\theta_{w_1,\tau}^\star(x_1)})\notag\\
&=\Big(\nabla_{\theta}f(x_2,\pi_{\theta_{w_1,\tau}^\star(x_1)})-\nabla_{\theta} f(x_1,\pi_{\theta_{w_1,\tau}^\star(x_1)})\Big) -\frac{1}{w_2}\Big(\nabla_{\theta} J_\tau(x_2,\pi_{\theta_{w_1,\tau}^\star(x_1)})-\nabla_{\theta} J_\tau(x_1,\pi_{\theta_{w_1,\tau}^\star(x_1)})\Big)\notag\\
&\hspace{20pt}-(\frac{1}{w_2}-\frac{1}{w_1})\nabla_{\theta} J_\tau(x_1,\pi_{\theta_{w_1,\tau}^\star(x_1)}).
\end{align*}

Taking the norm,
\begin{align}
&\|\nabla_{\theta} \Lcal_{w_2,\tau}(x_2,\pi_{\theta_{w_1,\tau}^\star(x_1)})\|\notag\\
&\leq \|\nabla_{\theta}f(x_2,\pi_{w_1,\tau}^\star(x_1))-\nabla_{\pi}f(x_1,\pi_{w_1,\tau}^\star(x_1))\|+\frac{1}{w_2}\|\nabla_{\pi}J_\tau(x_2,\pi_{w_1,\tau}^\star(x_1))-\nabla_{\theta}J_\tau(x_1,\pi_{\theta_{w_1,\tau}^\star(x_1)})\|\notag\\
&\hspace{20pt}+|\frac{1}{w_2}-\frac{1}{w_1}|\|\nabla_{\theta} J_\tau(x_1,\pi_{\theta_{w_1,\tau}^\star(x_1)})\|\notag\\
&\leq L_f\|x_1-x_2\|+\frac{L_V}{w_2}\|x_1-x_2\| + |\frac{1}{w_2}-\frac{1}{w_1}|\|\nabla_{\theta} J_\tau(x_1,\pi_{\theta_{w_1,\tau}^\star(x_1)})\|\notag\\
&\leq (L_f+\frac{L_V}{w_2}) \|x_1-x_2\|+ \frac{L_f|w_1-w_2|}{w_2},\label{lem:Lipschitz_BR:proof_eq3}
\end{align}
where the second inequality follows from the Lipschitz continuous gradients of $f$ and $J_\tau$, and the third inequality plugs in \eqref{lem:Lipschitz_BR:proof_eq2}.

Due to \eqref{eq:PL_QG_combined}, we have 
\begin{align}
\|\nabla_\theta \Lcal_{w_2,\tau}(x,\pi_{\theta_{w_1,\tau}^\star(x_1)})\|\geq\frac{C_L\tau}{2w_2}\|\pi_{w_1,\tau}^\star(x_1)-\pi_{w_2,\tau}^\star(x_2)\|.\label{lem:Lipschitz_BR:proof_eq4}
\end{align}

Combining \eqref{lem:Lipschitz_BR:proof_eq3} and \eqref{lem:Lipschitz_BR:proof_eq4}, 
\begin{align*}
\frac{C_L\tau}{2w_2}\|\pi_{w_1,\tau}^\star(x_1)-\pi_{w_2,\tau}^\star(x_2)\|\leq (L_f+\frac{L_V}{w_2}) \|x_1-x_2\|+ \frac{L_f|w_1-w_2|}{w_2}.
\end{align*}
For any $w_2>0$, this simplifies to
\begin{align}
\|\pi_{w_1,\tau}^\star(x_1)-\pi_{w_2,\tau}^\star(x_2)\|\leq (\frac{2L_f w_2}{C_L\tau} + \frac{2L_V}{C_L\tau})\|x_1-x_2\|+\frac{2L_f|w_1-w_2|}{C_L\tau}.\label{lem:Lipschitz_BR:proof_eq5}
\end{align}

Recognizing $\pi_{\tau}^\star(x)=\lim_{w\rightarrow 0^+}\pi_{w,\tau}^\star(x)$, we have from \eqref{lem:Lipschitz_BR:proof_eq5}
\begin{align*}
\|\pi_{\tau}^\star(x)-\pi_{w,\tau}^\star(x)\|\leq\frac{2L_f w}{C_L\tau}.
\end{align*}

Now, we fix $w,x$ and show the bound on $\|\pi_{w,\tau_1}^\star(x)-\pi_{w,\tau_2}^\star(x)\|$. We use $\theta_{w,\tau_1}(x)^\star$ to denote a softmax parameter for $\pi_{w,\tau_1}(x)^\star$. The optimality condition of $\theta_{w,\tau_1}(x)^\star$ indicates
\begin{align*}
\nabla_\theta \Lcal_{w,\tau_1}(x,\pi_{\theta_{w,\tau_1}^\star(x)}) = \nabla_\theta f(x,\pi_{\theta_{w,\tau_1}^\star(x)})-\frac{1}{w}\nabla_\theta J_{\tau_1}(x,\pi_{\theta_{w,\tau_1}^\star(x)})=0.
\end{align*}

Applying the equation, we get
\begin{align}
\nabla_\theta \Lcal_{w,\tau_2}(x,\pi_{\theta_{w,\tau_1}^\star(x)})&=\nabla_{\theta}f(x,\pi_{\theta_{w,\tau_1}^\star(x)})-\frac{1}{w}\nabla_{\theta}J_{\tau_2}(x,\pi_{\theta_{w,\tau_1}^\star(x)})\notag\\
&=\frac{1}{w}\Big(\nabla_{\theta} J_{\tau_1}(x,\pi_{\theta_{w,\tau_1}^\star(x)})-\nabla_{\theta} J_{\tau_2}(x,\pi_{\theta_{w,\tau_1}^\star(x)})\Big).\label{lem:Lipschitz_BR:proof_eq6}
\end{align}

The regularized RL objective has a closed-form expression (see \citet{mei2020global}[Lemma 10])
\begin{align*}
\frac{\partial J_\tau(x,\pi_{\theta})}{\partial \theta(s,a)}=\frac{d_{\rho}^{\pi_{\theta}}(s)}{1-\gamma}\cdot\pi_\theta(a\mid s)\cdot A_\tau^{x,\pi_{\theta}}(s,a),
\end{align*}
which in combination with \eqref{lem:Lipschitz_BR:proof_eq6} implies
\begin{align}
\|\nabla_\theta \Lcal_{w,\tau_2}(x,\pi_{\theta_{w,\tau_1}^\star(x)})\|&\leq\|\nabla_\theta \Lcal_{w,\tau_2}(x,\pi_{\theta_{w,\tau_1}^\star(x)})\|_1\notag\\
&\leq\frac{1}{(1-\gamma)w}\sum_{s,a}\pi_{\theta_{w,\tau_1}^\star(x)}(a\mid s)\Big|A_{\tau_1}^{x,\pi_{\theta_{w,\tau_1}^\star(x)}}(s,a)-A_{\tau_2}^{x,\pi_{\theta_{w,\tau_1}^\star(x)}}(s,a)\Big|.\label{lem:Lipschitz_BR:proof_eq7}
\end{align}

Due to \citet{zeng2022regularized}[Lemma 3], we have for any $s$
\begin{align*}
|V_{\tau_1}^{x,\pi}(s)-V_{\tau_2}^{x,\pi}(s)|\leq|\tau_1-\tau_2|\log|\Acal|.
\end{align*}
The definitions of the Q function and advantage function in \eqref{eq:def_Q_A} imply
\begin{align*}
&|Q_{\tau_1}^{x,\pi}(s,a)-Q_{\tau_2}^{x,\pi}(s,a)|\leq\gamma|V_{\tau_1}^{x,\pi}(s)-V_{\tau_2}^{x,\pi}(s)|\leq\gamma|\tau_1-\tau_2|\log|\Acal|,
\end{align*}
and
\begin{align}
&\sum_a\pi(a\mid s)|A_{\tau_1}^{x,\pi}(s,a)-A_{\tau_2}^{x,\pi}(s,a)|\notag\\
&\leq\sum_a\pi(a\mid s)|Q_{\tau_1}^{x,\pi}(s,a)-Q_{\tau_2}^{x,\pi}(s,a)|+|V_{\tau_1}^{x,\pi}(s)-V_{\tau_2}^{x,\pi}(s)|+|\tau_1-\tau_2|E(\pi,s)\notag\\
&\leq 3|\tau_1-\tau_2|\log|\Acal|.\label{lem:Lipschitz_BR:proof_eq8}
\end{align}

Plugging \eqref{lem:Lipschitz_BR:proof_eq8} into \eqref{lem:Lipschitz_BR:proof_eq7},
\begin{align*}
\|\nabla_\theta \Lcal_{w,\tau_2}(x,\pi_{\theta_{w,\tau_1}^\star(x)})\| &\leq \frac{3|\tau_1-\tau_2||\Scal|\log|\Acal|}{(1-\gamma)w}.
\end{align*}
Again, due to \eqref{eq:PL_QG_combined},
\begin{align*}
\frac{C_L\tau_2}{2w}\|\pi_{w,\tau_1}^\star(x)-\pi_{w,\tau_2}^\star(x)\|\leq\|\nabla_\pi \Lcal_{w,\tau_2}(x,\pi_{w,\tau_1}^\star(x))\|\leq \frac{3|\tau_1-\tau_2||\Scal|\log|\Acal|}{(1-\gamma)w}.
\end{align*}
This leads to
\begin{align}
\|\pi_{w,\tau_1}^\star(x)-\pi_{w,\tau_2}^\star(x)\|\leq\frac{6|\tau_1-\tau_2||\Scal|\log|\Acal|}{(1-\gamma)C_L\tau_2}.\label{lem:Lipschitz_BR:proof_eq9}
\end{align}

Putting \eqref{lem:Lipschitz_BR:proof_eq5} and \eqref{lem:Lipschitz_BR:proof_eq9} together,
\begin{align*}
\|\pi_{w_1,\tau_1}^\star(x_1)-\pi_{w_2,\tau_2}^\star(x_2)\|&\leq\|\pi_{w_1,\tau_1}^\star(x_1)-\pi_{w_2,\tau_1}^\star(x_2)\|+\|\pi_{w_2,\tau_2}^\star(x_2)-\pi_{w_2,\tau_1}^\star(x_2)\|\notag\\
&\leq (\frac{2L_f w_2}{C_L\tau_1} + \frac{2L_V}{C_L\tau_1})\|x_1-x_2\|+\frac{2L_f|w_1-w_2|}{C_L\tau_1}+\frac{6|\tau_1-\tau_2||\Scal|\log|\Acal|}{(1-\gamma)C_L\tau_1}.
\end{align*}

\qed

\subsection{Proof of Lemma~\ref{lem:distance_pistar}}
We know from Lemma~\ref{lem:pi_star_unique} that $\pi^\star(x)$, defined in \eqref{eq:def_pi_star}, is the limit point of $\pi_\tau^\star(x)$ as $\tau\rightarrow 0$ 
Let $\theta_\tau^\star(x)$ denote a softmax parameter for $\pi_\tau^\star(x)$.
By the first-order optimality condition, we have
\[\nabla_\theta J_\tau(x,\pi_{\theta_\tau^\star(x)})=0.\]
We further differentiate with respect to $\tau$. Due to the differentiation chain rule, 
\begin{align*}
\frac{d}{d\tau}\nabla_\theta J_\tau(x,\pi_{\theta_\tau^\star(x)})=\nabla_{\tau,\theta}^2 J_\tau(x,\pi_{\theta_\tau^\star(x)})+\nabla_{\theta,\theta}^2 J_\tau(x,\pi_{\theta_\tau^\star(x)})\cdot\frac{d\theta_\tau^\star(x)}{d\tau}=0.
\end{align*}

As $\|\nabla_{\theta,\theta}^2 J_\tau(x,\pi_{\theta_\tau^\star(x)})\|$ is lower bounded by $\underline{\sigma}$ due to Assumption~\ref{assump:invertible}, we have for any $\tau\geq0$
\begin{align}
\left\|\frac{d\theta_\tau^\star(x)}{d\tau}\right\|&=\left\|-\Big(\|\nabla_{\theta,\theta}^2 J_\tau(x,\pi_{\theta_\tau^\star(x)})\Big)^{-1}\nabla_{\tau,\theta}^2 J_\tau(x,\pi_{\theta_\tau^\star(x)})\right\|\notag\\
&\leq \left\|\Big(\|\nabla_{\theta,\theta}^2 J_\tau(x,\pi_{\theta_\tau^\star(x)})\Big)^{-1}\right\|\|\nabla_{\tau,\theta}^2 J_\tau(x,\pi_{\theta_\tau^\star(x)})\|\notag\\
&\leq\frac{1}{\underline{\sigma}}\|\nabla_{\tau,\theta}^2 J_\tau(x,\pi_{\theta_\tau^\star(x)})\|.\label{lem:distance_pistar:proof_eq1}
\end{align}

It is clear from \eqref{eq:def_V_tau} and \eqref{eq:def_J_regularized}
\begin{align*}
\nabla_{\tau} J_\tau(x,\pi_\theta)=\frac{1}{1-\gamma}\mathbb{E}_{s\sim d_\rho^{\pi_\theta},\,a\sim\pi_\theta(\cdot\mid s)}[E(\pi_\theta,s)].
\end{align*}

Therefore,
\begin{align}
\nabla_{\tau,\theta}^2 J_\tau(x,\pi_\theta)=\frac{1}{1-\gamma}\nabla_\theta\mathbb{E}_{s\sim d_\rho^{\pi_\theta},\,a\sim\pi_\theta(\cdot\mid s)}[E(\pi_\theta,s)].\label{lem:distance_pistar:proof_eq2}
\end{align}

\citet{zeng2022regularized}[Lemma 6] shows that $\mathbb{E}_{s\sim d_\rho^{\pi_\theta},\,a\sim\pi_\theta(\cdot\mid s)}[E(\pi_\theta,s)]$ is Lipschitz with constant $\frac{4+8\log|\Acal|}{(1-\gamma)^3}$, which is equivalent to
\begin{align}
\|\nabla_\theta\mathbb{E}_{s\sim d_\rho^{\pi_\theta},\,a\sim\pi_\theta(\cdot\mid s)}[E(\pi_\theta,s)]\|\leq\frac{4+8\log|\Acal|}{(1-\gamma)^3},\quad\forall\theta.\label{lem:distance_pistar:proof_eq3}
\end{align}

Combining \eqref{lem:distance_pistar:proof_eq1}-\eqref{lem:distance_pistar:proof_eq3}, we have
\begin{align*}
\left\|\frac{d\theta_\tau^\star(x)}{d\tau}\right\|=\frac{1}{\underline{\sigma}}\cdot\frac{1}{1-\gamma}\cdot\frac{4+8\log|\Acal|}{(1-\gamma)^3}=\frac{4+8\log|\Acal|}{\underline{\sigma}(1-\gamma)^4}=L_\star.
\end{align*}

This implies $\theta_\tau^\star(x)$ is $L_\star$-Lipschitz with respect to $\tau$ for $\tau\geq0$.

\qed

\subsection{Proof of Lemma~\ref{lem:Lipschitz_Lcal}}

It can be seen from \eqref{eq:def_Phi_w_tau} that the gradients of the Lagrangian function have the following closed-form expressions
\begin{align}
\nabla_x\Lcal_{w,\tau}(x,\pi)&=\nabla_x f(x,\pi)+\frac{1}{w}\Big(\nabla_x J_\tau(x,\pi_\tau^\star(x))+\nabla_x \pi_\tau^\star(x) \nabla_\pi J_\tau(x,\pi_\tau^\star(x))-\nabla_x J_\tau(x,\pi)\Big)\notag\\
&=\nabla_x f(x,\pi)+\frac{1}{w}\Big(\nabla_x J_\tau(x,\pi_\tau^\star(x))-\nabla_x J_\tau(x,\pi)\Big),\label{eq:grad_Lagrangian_x}\\
\nabla_\pi\Lcal_{w,\tau}(x,\pi)&=\nabla_\pi f(x,\pi)-\frac{1}{w}\nabla_\pi J_\tau(x,\pi),\label{eq:grad_Lagrangian_pi}
\end{align}
where the equation \eqref{eq:grad_Lagrangian_x} is due to the optimality condition of $J_\tau$ at $\pi_\tau^\star(x)$.

According to \eqref{eq:grad_Lagrangian_x}, we have
\begin{align*}
&\|\nabla_x\Lcal_{w,\tau}(x,\pi)-\nabla_x\Lcal_{w,\tau}(x',\pi')\|\notag\\
&=\|\nabla_x f(x,\pi)+\frac{1}{w}\Big(\nabla_x J_\tau(x,\pi_\tau^\star(x))-\nabla_x J_\tau(x,\pi)\Big)-\nabla_x f(x',\pi')-\frac{1}{w}\Big(\nabla_x J_\tau(x',\pi_\tau^\star(x'))-\nabla_x J_\tau(x',\pi')\Big)\|\notag\\
&\leq \|\nabla_x f(x,\pi)-\nabla_x f(x',\pi')\|+\frac{1}{w}\|\nabla_x J_\tau(x,\pi_\tau^\star(x))-\nabla_x J_\tau(x',\pi_\tau^\star(x'))\|\notag\\
&\hspace{20pt}+\frac{1}{w}\|\nabla_x J_\tau(x,\pi)-\nabla_x J_\tau(x',\pi')\|\notag\\
&\leq L_f(\|x-x'\|+\|\pi-\pi'\|) + \frac{L_V}{w}(\|x-x'\|+\|\pi_\tau^\star(x)-\pi_\tau^\star(x')\|) + \frac{L_V}{w}(\|x-x'\| + \|\pi-\pi'\|).
\end{align*}

Recognizing $\pi_{\tau}^\star(x)=\lim_{w\rightarrow 0^+}\pi_{w,\tau}^\star(x)$, we have from Lemma~\ref{lem:Lipschitz_BR}
\begin{align*}
\|\pi_\tau^\star(x)-\pi_\tau^\star(x')\|\leq\frac{2L_V}{C_L\tau}\|x-x'\|.
\end{align*}

Combining the two inequalities above and imposing the condition $w,\tau\leq1$, we get
\begin{align*}
\|\nabla_x\Lcal_{w,\tau}(x,\pi)-\nabla_x\Lcal_{w,\tau}(x',\pi')\|&\leq (L_V+L_f+\frac{L_V(C_L+2L_V)}{C_L})\frac{1}{w\tau}\|x-x'\|+\frac{L_f+L_V}{w}\|\pi-\pi'\|\notag\\
&\leq\frac{L_L}{w\tau}\|x-x'\|+\frac{L_L}{w}\|\theta-\theta'\|.
\end{align*}

According to \eqref{eq:grad_Lagrangian_pi}, we have
\begin{align*}
&\|\nabla_\theta\Lcal_{w,\tau}(x,\pi_\theta)-\nabla_\theta\Lcal_{w,\tau}(x',\pi_{\theta'})\|\notag\\
&=\|\nabla_\theta f(x,\pi_\theta)-\frac{1}{w}\nabla_\theta J_\tau(x,\pi_\theta)-\nabla_\theta f(x',\pi_{\theta'})-\frac{1}{w}\nabla_\theta J_\tau(x',\pi_{\theta'})\|\notag\\
&\leq\|\nabla_\theta f(x,\pi_\theta)-\nabla_\theta f(x',\pi_{\theta'})\|+\frac{1}{w}\|\nabla_\theta J_\tau(x,\pi_\theta)-\nabla_\theta J_\tau(x',\pi_{\theta'})\|\notag\\
&\leq L_f(\|x-x'\|+\|\pi_\theta-\pi_{\theta'}\|)+\frac{L_V}{w}(\|x-x'\|+\|\pi_\theta-\pi_{\theta'}\|)\notag\\
&\leq \frac{L_L}{w}\|x-x'\|+\frac{L_L}{w}\|\theta-\theta'\|.
\end{align*}

\qed

\subsection{Proof of Lemma~\ref{lem:Lipschitz_Phi}}
Let $\theta_\tau^\star(x)$ denote a softmax parameter of $\pi_\tau^\star(x)$.
By the definition of $\ell_\tau$, we have
\begin{align*}
\nabla\ell_\tau(x)=\nabla_x J_\tau(x,\pi_{\theta_\tau^\star(x)}),
\end{align*}
due to $\nabla_\theta J_\tau(x,\pi_{\theta_\tau^\star(x)})=0$.

As $J_\tau$ is $L_V$-smooth (from Lemma~\ref{lem:Lipschitz_V}) this implies
\begin{align*}
\|\nabla \ell_\tau(x_1)-\nabla \ell_\tau(x_2)\| &= \|\nabla_x J_\tau(x_1,\pi_{\theta_\tau^\star(x_1)})-\nabla_x J_\tau(x_2,\pi_{\theta_\tau^\star(x_2)})\|\notag\\
&\leq L_V\|x_1-x_2\|+L_V\|\pi_\tau^\star(x_1)-\pi_\tau^\star(x_2)\|\notag\\
&\leq L_V\|x_1-x_2\|+L_V\cdot\frac{2L_V}{C_L\tau}\|x_1-x_2\|\notag\\
&\leq \Big(L_V+\frac{2L_V^2}{C_L\tau}\Big)\|x_1-x_2\|,
\end{align*}
where the second inequality follows from Lemma~\ref{lem:Lipschitz_BR} by recognizing that $\pi_{\tau}^\star(x)=\lim_{w\rightarrow0}\pi_{w,\tau}^\star(x)$.

We next show the smoothness of $\Phi_{w,\tau}$. From \eqref{eq:grad_Phi_sigma_tau} it can be seen
\begin{align*}
&\|\nabla_x \Phi_{w,\tau}(x_1)-\nabla_x \Phi_{w,\tau}(x_2)\| \notag\\
&\leq \Big\|\nabla_x f(x_1,\pi_\tau^\star(x_1))-\nabla_x f(x_2,\pi_\tau^\star(x_2))\Big\| + \frac{1}{w}\Big\|\nabla_x J_{\tau}(x_1,\pi_\tau^\star(x_1))-\nabla_x J_{\tau}(x_2,\pi_\tau^\star(x_2))\Big\|\notag\\
&\hspace{20pt}+\frac{1}{w}\Big\|\nabla_x J_{\tau}(x_1,\pi_{w,\tau}^\star(x_1))-\nabla_x J_{\tau}(x_2,\pi_{w,\tau}^\star(x_2))\Big\|\notag\\
&\leq L_f\|x_1-x_2\|+\Big(L_f+\frac{L_V}{w}\Big)\|\pi_\tau^\star(x_1)-\pi_\tau^\star(x_2)\|+\frac{L_V}{w}\|\pi_{w,\tau}^\star(x_1)-\pi_{w,\tau}^\star(x_2)\|\notag\\
&\leq L_f\|x_1-x_2\|+\Big(L_f+\frac{L_V}{w}\Big)\cdot\frac{2L_V}{C_L\tau}\|x_1-x_2\|+\frac{L_V}{w}\cdot\Big(\frac{2L_f w_2}{C_L\tau}+\frac{2L_V}{C_L\tau}\Big)\|x_1-x_2\|\notag\\
&\leq \Big(L_f+\frac{4L_f L_V}{C_L\tau}+\frac{4L_V^2}{C_L w\tau}\Big)\|x_1-x_2\|.
\end{align*}

Finally, we show the smoothness of $\Phi_\tau$. 
Let $\theta_\tau^\star(x)$ denote one of the softmax parameters that encodes $\pi_\tau^\star(x)$. We can express the hyper-gradient of $\Phi_\tau$ as follows
\begin{align*}
\nabla_x\Phi_\tau(x)=\nabla_x f(x,\pi_{\theta_\tau^\star(x)})-\nabla_{x,\theta}^2 J_\tau(x,\pi_{\theta_\tau^\star(x)})\nabla_{\theta,\theta}^2 J_\tau(x,\pi_{\theta_\tau^\star(x)})^{-1}\nabla_{\theta}f(x,\pi_{\theta_\tau^\star(x)}).
\end{align*}

This implies
\begin{align}
&\|\nabla_x\Phi_\tau(x_1)-\nabla_x\Phi_\tau(x_2)\|\notag\\
&\leq\underbrace{\|\nabla_x f(x_1,\pi_{\theta_\tau^\star(x_1)})-\nabla_x f(x_2,\pi_{\theta_\tau^\star(x_2)})\|}_{T_1}\notag\\
&\hspace{-20pt} + \underbrace{\|\nabla_{x,\theta}^2 J_\tau(x_1,\pi_{\theta_\tau^\star(x_1)})\nabla_{\theta,\theta}^2 J_\tau(x_1,\pi_{\theta_\tau^\star(x_1)})^{-1}\nabla_{\theta}f(x_1,\pi_{\theta_\tau^\star(x_1)})-\nabla_{x,\theta}^2 J_\tau(x_2,\pi_{\theta_\tau^\star(x_2)})\nabla_{\theta,\theta}^2 J_\tau(x_1,\pi_{\theta_\tau^\star(x_1)})^{-1}\nabla_{\theta}f(x_1,\pi_{\theta_\tau^\star(x_1)})\|}_{T_2}\notag\\
&\hspace{-20pt} + \underbrace{\|\nabla_{x,\theta}^2 J_\tau(x_2,\pi_{\theta_\tau^\star(x_2)})\nabla_{\theta,\theta}^2 J_\tau(x_1,\pi_{\theta_\tau^\star(x_1)})^{-1}\nabla_{\theta}f(x_1,\pi_{\theta_\tau^\star(x_1)})-\nabla_{x,\theta}^2 J_\tau(x_2,\pi_{\theta_\tau^\star(x_2)})\nabla_{\theta,\theta}^2 J_\tau(x_2,\pi_{\theta_\tau^\star(x_2)})^{-1}\nabla_{\theta}f(x_1,\pi_{\theta_\tau^\star(x_1)})\|}_{T_3}\notag\\
&\hspace{-20pt} + \underbrace{\|\nabla_{x,\theta}^2 J_\tau(x_2,\pi_{\theta_\tau^\star(x_2)})\nabla_{\theta,\theta}^2 J_\tau(x_2,\pi_{\theta_\tau^\star(x_2)})^{-1}\nabla_{\theta}f(x_1,\pi_{\theta_\tau^\star(x_1)})-\nabla_{x,\theta}^2 J_\tau(x_2,\pi_{\theta_\tau^\star(x_2)})\nabla_{\theta,\theta}^2 J_\tau(x_2,\pi_{\theta_\tau^\star(x_2)})^{-1}\nabla_{\theta}f(x_2,\pi_{\theta_\tau^\star(x_2)})\|}_{T_4}.\label{lem:Lipschitz_Phi:proof_eq1}
\end{align}

We treat each term of \eqref{lem:Lipschitz_Phi:proof_eq1} individually. First, we bound $T_1$ using the smoothness property of $f$
\begin{align}
T_1 &\leq L_f\Big(\|x_1-x_2\|+\|\pi_\tau^\star(x_1)-\pi_\tau^\star(x_2)\|\Big)\notag\\
&\leq (L_f+\frac{2L_f L_V}{C_L\tau})\|x_1-x_2\|,\label{lem:Lipschitz_Phi:proof_eq2}
\end{align}
where to derive the second inequality, we plug in the result from Lemma~\ref{lem:Lipschitz_BR} to get $\|\pi_\tau^\star(x_1)-\pi_\tau^\star(x_2)\|\leq\frac{2L_V}{C_L\tau}\|x_1-x_2\|$ (note that $\pi_\tau^\star(x)=\lim_{w\rightarrow0}\pi_{w,\tau}^\star(x)$).

As we have $\nabla_\theta f(x,\pi_\theta)\leq L_f$ from Assumption~\ref{assump:f} and $\|\nabla_{\theta,\theta}^2 J_\tau(x,\pi_{\theta_\tau^\star(x)})^{-1}\|\leq\frac{1}{\underline{\sigma}}$ due to Assumption~\ref{assump:invertible},
\begin{align}
T_2 &\leq \|\nabla_{x,\theta}^2 J_\tau(x_1,\pi_{\theta_\tau^\star(x_1)})-\nabla_{x,\theta}^2 J_\tau(x_2,\pi_{\theta_\tau^\star(x_2)})\|\|\nabla_{\theta,\theta}^2 J_\tau(x_1,\pi_{\theta_\tau^\star(x_1)})^{-1}\nabla_{\theta}f(x_1,\pi_{\theta_\tau^\star(x_1)})\|\notag\\
&\leq \frac{L_f}{\underline{\sigma}}\cdot L_{V,2}\Big(\|x_1-x_2\|+\|\pi_\tau^\star(x_1)-\pi_\tau^\star(x_2)\|\Big)\notag\\
&\leq(\frac{L_f L_{V,2}}{\underline{\sigma}}+\frac{2L_fL_VL_{V,2}}{\underline{\sigma}C_L\tau})\|x_1-x_2\|,\label{lem:Lipschitz_Phi:proof_eq3}
\end{align}
where second inequality is due to Lemma~\ref{lem:Lipschitz_V}, and the last inequality follows from an argument similar to the one in \eqref{lem:Lipschitz_Phi:proof_eq2}.

Similarly, for $T_3$
\begin{align}
T_3 &\leq \|\nabla_{x,\theta}^2 J_\tau(x_2,\pi_{\theta_\tau^\star(x_2)})\| \|\nabla_{\theta,\theta}^2 J_\tau(x_1,\pi_{\theta_\tau^\star(x_1)})^{-1}-\nabla_{\theta,\theta}^2 J_\tau(x_2,\pi_{\theta_\tau^\star(x_2)})^{-1}\| \|\nabla_{\theta}f(x_1,\pi_{\theta_\tau^\star(x_1)})\|\notag\\
&\leq L_f L_V \|\nabla_{\theta,\theta}^2 J_\tau(x_1,\pi_{\theta_\tau^\star(x_1)})^{-1}\|\|\nabla_{\theta,\theta}^2 J_\tau(x_2,\pi_{\theta_\tau^\star(x_2)})-\nabla_{\theta,\theta}^2 J_\tau(x_1,\pi_{\theta_\tau^\star(x_1)})^{-1}\|\|\nabla_{\theta,\theta}^2 J_\tau(x_2,\pi_{\theta_\tau^\star(x_2)})^{-1}\|\notag\\
&\leq \frac{L_f L_V}{\underline{\sigma}^2}\cdot L_{V,2}\Big(\|x_1-x_2\|+\|\pi_\tau^\star(x_1)-\pi_\tau^\star(x_2)\|\Big)\notag\\
&\leq (\frac{L_f L_V L_{V,2}}{\underline{\sigma}^2}+\frac{2L_f L_V^2 L_{V,2}}{\underline{\sigma}^2C_L\tau})\|x_1-x_2\|.\label{lem:Lipschitz_Phi:proof_eq4}
\end{align}

For the final term, we have
\begin{align}
T_4 &\leq \|\nabla_{x,\theta}^2 J_\tau(x_2,\pi_{\theta_\tau^\star(x_2)})\nabla_{\theta,\theta}^2 J_\tau(x_2,\pi_{\theta_\tau^\star(x_2)})^{-1}\|\|\nabla_{\theta}f(x_1,\pi_{\theta_\tau^\star(x_1)})-\nabla_{\theta}f(x_2,\pi_{\theta_\tau^\star(x_2)})\|\notag\\
&\leq \frac{L_V}{\underline{\sigma}}\|\nabla_{\theta}f(x_1,\pi_{\theta_\tau^\star(x_1)})-\nabla_{\theta}f(x_2,\pi_{\theta_\tau^\star(x_2)})\|\notag\\
&\leq \frac{L_V}{\underline{\sigma}}\cdot L_f\Big(\|x_1-x_2\|+\|\pi_\tau^\star(x_1)-\pi_\tau^\star(x_2)\|\Big)\notag\\
&\leq (\frac{L_f L_V}{\underline{\sigma}}+\frac{2L_fL_V^2}{\underline{\sigma}C_L\tau})\|x_1-x_2\|.\label{lem:Lipschitz_Phi:proof_eq5}
\end{align}

We combine \eqref{lem:Lipschitz_Phi:proof_eq2}-\eqref{lem:Lipschitz_Phi:proof_eq5}
\begin{align*}
\|\nabla_x\Phi_\tau(x_1)-\nabla_x\Phi_\tau(x_2)\| &\leq (L_f+\frac{2L_f L_V}{C_L\tau})\|x_1-x_2\|+(\frac{L_f L_{V,2}}{\underline{\sigma}}+\frac{2L_fL_VL_{V,2}}{\underline{\sigma}C_L\tau})\|x_1-x_2\|\notag\\
&\hspace{20pt}+(\frac{L_f L_V L_{V,2}}{\underline{\sigma}^2}+\frac{2L_f L_V^2 L_{V,2}}{\underline{\sigma}^2C_L\tau})\|x_1-x_2\|+(\frac{L_f L_V}{\underline{\sigma}}+\frac{2L_fL_V^2}{\underline{\sigma}C_L\tau})\|x_1-x_2\|\notag\\
&\leq (1+\frac{2L_V}{C_L\tau})\Big(\frac{2L_fL_V}{C_L\tau}+\frac{2L_fL_VL_{V,2}}{\underline{\sigma}C_L\tau}+\frac{2L_fL_V^2L_{V,2}}{\underline{\sigma}^2C_L\tau}+\frac{2L_fL_V^2}{\underline{\sigma}C_L\tau}\Big)\|x_1-x_2\|.
\end{align*}

Imposing the step size condition $\tau\leq\frac{2L_V}{C_L}$, we get
\begin{align*}
\|\nabla_x\Phi_\tau(x_1)-\nabla_x\Phi_\tau(x_2)\| 
&\leq \Big(\frac{4L_fL_V}{C_L\tau}+\frac{4L_fL_VL_{V,2}}{\underline{\sigma}C_L\tau}+\frac{4L_fL_V^2L_{V,2}}{\underline{\sigma}^2C_L\tau}+\frac{4L_fL_V^2}{\underline{\sigma}C_L\tau}\Big)\|x_1-x_2\|\notag\\
&\leq \frac{L_\Phi}{\tau}\|x_1-x_2\|.
\end{align*}

\qed

\subsection{Proof of Lemma~\ref{eq:grad_gap_sigma_tau}}

Let $\theta_\tau^\star(x)$ denote a parameter representing $\pi_\tau^\star(x)$ through the softmax function.
Define for $\tau>0$
\[\nabla\Phi_{\tau}(x)=\nabla_x f(x,\pi_{\tau}^\star(x))-\nabla_{x,\theta}^2 J_\tau(x,\pi_{\theta_{\tau}^\star(x)})\nabla_{\theta,\theta}^2 J_{\tau}(x,\pi_{\theta_{\tau}^\star(x)})^{-1}\nabla_{\theta}f(x,\pi_{\theta_{\tau}^\star(x)}).\]
We consider the following decomposition
\begin{align}
\|\nabla_x\Phi(x)-\nabla_x\Phi_{w,\tau}(x)\| &\leq \|\nabla_x\Phi_\tau(x)-\nabla_x\Phi_{w,\tau}(x)\| + \|\nabla_x\Phi(x)-\nabla_x\Phi_{\tau}(x)\|.\label{eq:grad_gap_sigma_tau:proof_eq1}
\end{align}
We first bound the first term of \eqref{eq:grad_gap_sigma_tau:proof_eq1}.

To derive the bound on $\|\nabla_x\Phi_\tau(x)-\nabla_x\Phi_{w,\tau}(x)\|$, we take an argument similar to \citet{kwon2023fully}[Lemma A.2], which we adapt to the case of a non-convex lower level objective. Note that $\lambda$ in \citet{kwon2023fully} plays the same role as our $1/w$. \citet{kwon2023fully}[Lemma A.2] is still valid without lower level convexity, with the lower bound on $\|\nabla_{\theta,\theta}^2 J_\tau(x,\pi_{\theta_{\tau}^\star(x)})\|$ changed from the strong convexity coefficient to $\underline{\sigma}$. This allows us to write
\begin{align*}
&\left\|\nabla_x \Phi_\tau(x)-\nabla_x \Lcal_{w,\tau}(x, \pi_\theta)+\nabla_{x,\theta}^2 J_\tau(x,\pi_{\theta_{\tau}^\star(x)})\nabla_{\theta,\theta}^2 J_\tau(x,\pi_{\theta_{\tau}^\star(x)})^{-1}\nabla_{\theta} \Lcal_{w,\tau}(x, \pi_\theta)\right\|\notag\\
&\leq \frac{2L_V}{\underline{\sigma}}\|\pi_\theta-\pi_{\tau}^\star(x)\|\Big(L_f+\frac{L_{V,2}}{w}\|\pi_\theta-\pi_{\tau}^\star(x)\|\Big).
\end{align*}

Recognizing $\nabla_{\theta} \Lcal_{w,\tau}(x, \pi_{\theta_{w,\tau}^\star(x)})=0$, we have
\begin{align}
\|\nabla_x \Phi_\tau(x)-\nabla_x \Phi_{w,\tau}(x)\|
&=\|\nabla_x \Phi_\tau(x)-\nabla_x \Lcal_{w,\tau}(x, \pi_{w,\tau}^\star(x))\|\notag\\
&\leq \frac{2L_V}{\underline{\sigma}}\|\pi_{w,\tau}^\star(x)-\pi_{\tau}^\star(x)\|\Big(L_f+\frac{L_{V,2}}{w}\|\pi_{w,\tau}^\star(x)-\pi_{\tau}^\star(x)\|\Big)\notag\\
&\leq \frac{2L_V}{\underline{\sigma}}\cdot\frac{2L_f w}{C_L \tau}\Big(L_f+\frac{L_{V,2}}{w}\cdot\frac{2L_f w}{C_L \tau}\Big)\notag\\
&=\frac{4L_f L_V w}{C_L \underline{\sigma} \tau}(L_f+\frac{2L_f L_{V,2}}{C_L\tau}),\label{eq:grad_gap_sigma_tau:proof_eq2}
\end{align}
where the second inequality follows from Lemma~\ref{lem:Lipschitz_BR}.

Next, we bound $\|\nabla_x\Phi_{\tau}(x)-\nabla_x\Phi(x)\|$.
\begin{align}
\nabla_x\Phi_{\tau}(x)-\nabla_x\Phi(x) &= \underbrace{\nabla_x f(x,\pi_{\theta_{\tau}^\star(x)})-\nabla_x f(x,\pi_{\theta^\star(x)})}_{T_1}\notag\\
&\hspace{-100pt}+\underbrace{\nabla_{x,\theta}^2 J(x,\pi_{\theta^\star(x)})\nabla_{\theta,\theta}^2 J(x,\pi_{\theta^\star(x)})^{-1}\nabla_{\theta}f(x,\pi_{\theta^\star(x)})-\nabla_{x,\theta}^2 J(x,\pi_{\theta_{\tau}^\star(x)})\nabla_{\theta,\theta}^2 J(x,\pi_{\theta_{\tau}^\star(x)})^{-1}\nabla_{\theta}f(x,\pi_{\theta_{\tau}^\star(x)})}_{T_2}\notag\\
&\hspace{-100pt}+\underbrace{\nabla_{x,\theta}^2 J(x,\pi_{\theta_{\tau}^\star(x)})\nabla_{\theta,\theta}^2 J(x,\pi_{\theta_{\tau}^\star(x)})^{-1}\nabla_{\theta}f(x,\pi_{\theta_{\tau}^\star(x)})-\nabla_{x,\theta}^2 J_\tau(x,\pi_{\theta_{\tau}^\star(x)})\nabla_{\theta,\theta}^2 J_{\tau}(x,\pi_{\theta_{\tau}^\star(x)})^{-1}\nabla_{\theta}f(x,\pi_{\theta_{\tau}^\star(x)})}_{T_3}.\label{eq:grad_gap_sigma_tau:proof_eq3}
\end{align}

To treat $T_1$, we have from the Lipschitz continuity of $f$
\begin{align*}
\|T_1\|\leq L_f\|\pi_{\tau}^\star(x)-\pi^\star(x)\|.
\end{align*}

For $T_2$,
\begin{align*}
\|T_2\|&\leq \|\nabla_{x,\theta}^2 J(x,\pi_{\theta^\star(x)})-\nabla_{x,\theta}^2 J(x,\pi_{\theta_{\tau}^\star(x)})\|\|\nabla_{\theta,\theta}^2 J(x,\pi_{\theta^\star(x)})^{-1}\|\|\nabla_{\theta} f(x,\pi_{\theta^\star(x)})\|\notag\\
&\hspace{20pt}+\|\nabla_{x,\theta}^2 J(x,\pi_{\theta_{\tau}^\star(x)})\| \|\nabla_{\theta,\theta}^2 J(x,\pi_{\theta^\star(x)})^{-1}-\nabla_{\theta,\theta}^2 J(x,\pi_{\theta_{\tau}^\star(x)})^{-1}\| \|\nabla_{\theta}f(x,\pi_{\theta^\star(x)})\|\notag\\
&\hspace{20pt}+\|\nabla_{x,\theta}^2 J(x,\pi_{\theta_{\tau}^\star(x)})\|\|\nabla_{\theta,\theta}^2 J(x,\pi_{\theta_{\tau}^\star(x)})^{-1}\|\|\nabla_{\theta}f(x,\pi_{\theta^\star(x)})-\nabla_{\theta}f(x,\pi_{\theta_{\tau}^\star(x)})\|\notag\\
&\leq L_{V,2}\|\pi_{\tau}^\star(x)-\pi^\star(x)\|\cdot\frac{1}{\underline{\sigma}}\cdot L_f\notag\\
&\hspace{20pt}+L_V\cdot\|\nabla_{\theta,\theta}^2 J(x,\pi_{\theta_{\tau}^\star(x)})^{-1}\|\|\nabla_{\theta,\theta}^2 J(x,\pi_{\theta_{\tau}^\star(x)})-\nabla_{\theta,\theta}^2 J(x,\pi_{\theta^\star(x)})\|\|\nabla_{\theta,\theta}^2 J(x,\pi_{\theta^\star(x)})^{-1}\|\notag\\
&\hspace{20pt}+L_V\cdot\frac{1}{\underline{\sigma}}\cdot L_f\|\pi_{\tau}^\star(x)-\pi^\star(x)\|\notag\\
&\leq \frac{L_f(L_V+L_{V,2})}{\underline{\sigma}}\|\pi_{\tau}^\star(x)-\pi^\star(x)\|+L_V\cdot\frac{1}{\underline{\sigma}}\cdot L_{V,2}\|\pi_{\tau}^\star(x)-\pi^\star(x)\|\cdot\frac{1}{\underline{\sigma}}\notag\\
&\leq \frac{L_f L_V+L_f L_{V,2}+L_V L_{V,2}}{\underline{\sigma}^2}\|\pi_{\tau}^\star(x)-\pi^\star(x)\|.
\end{align*}

We then bound $T_3$. Note that
\begin{align*}
J_\tau(x,\pi)-J(x,\pi)=\frac{\tau}{1-\gamma}\mathbb{E}_{s\sim d_\rho^{\pi}}[E(\pi,s)],
\end{align*}
which is independent of $x$. This implies $\nabla_{x,\theta}^2 J(x,\pi_{\theta})=\nabla_{x,\theta}^2 J_\tau(x,\pi_{\theta})$. Using this relationship, we have
\begin{align*}
\|T_3\|&=\left\|\nabla_{x,\theta}^2 J(x,\pi_{\theta})\Big(\nabla_{\theta,\theta}^2 J(x,\pi_{\theta_{\tau}^\star(x)})^{-1}\nabla_{\theta}f(x,\pi_{\theta_{\tau}^\star(x)})-\nabla_{\theta,\theta}^2 J_{\tau}(x,\pi_{\theta_{\tau}^\star(x)})^{-1}\nabla_{\theta}f(x,\pi_{\theta_{\tau}^\star(x)})\Big)\right\|\notag\\
&\leq L_{V,2}L_f \|\nabla_{\theta,\theta}^2 J_\tau(x,\pi_{\theta_{\tau}^\star(x)})^{-1}\| \|\nabla_{\theta,\theta}^2 J_\tau(x,\pi_{\theta_{\tau}^\star(x)})-\nabla_{\theta,\theta}^2 J(x,\pi_{\theta_{\tau}^\star(x)})\|\|\nabla_{\theta,\theta}^2 J(x,\pi_{\theta_{\tau}^\star(x)})^{-1}\| \notag\\
&\leq L_f L_{V,2}\cdot\frac{1}{\underline{\sigma}}\cdot\frac{\tau}{1-\gamma}\|\nabla_{\theta,\theta}^2 \mathbb{E}_{s\sim d_\rho^{\pi_{\tau}^\star(x)}}[E(\pi_{\tau}^\star(x),s)]\|\cdot\frac{1}{\underline{\sigma}}\notag\\
&\leq \frac{L_f L_{V,2}(4+8\log|\Acal|)\tau}{(1-\gamma)^4\underline{\sigma}^2},
\end{align*}
where the third inequality follows from the fact that $\mathbb{E}_{s\sim d_\rho^{\pi}}[E(\pi,s)]$ is $\frac{4+8\log|\Acal|}{(1-\gamma)^3}$-Lipschitz (see, for example, Lemma 6 of \citet{zeng2022regularized}).

Collecting the bounds on $T_1$-$T_3$ and substituting them into \eqref{eq:grad_gap_sigma_tau:proof_eq3},
\begin{align}
\|\nabla_x \Phi_\tau(x)-\nabla_x \Phi_{w,\tau}(x)\|
&\leq \|T_1\|+\|T_2\|+\|T_3\|\notag\\
&\leq \frac{L_f (L_V+1)+L_f L_{V,2}+L_V L_{V,2}}{\underline{\sigma}^2}\|\pi_{\tau}^\star(x)-\pi^\star(x)\|+\frac{L_f L_{V,2}(4+8\log|\Acal|)\tau}{(1-\gamma)^4\underline{\sigma}^2}\notag\\
&\leq \frac{L_\star L_f (L_V+1)+L_\star L_f L_{V,2}+L_\star L_V L_{V,2}}{\underline{\sigma}^2}\tau+\frac{L_f L_{V,2}(4+8\log|\Acal|)\tau}{(1-\gamma)^4\underline{\sigma}^2}\notag\\
&\leq \frac{L_\star L_f (L_V+1)+L_\star L_f L_{V,2}+L_\star L_V L_{V,2}+L_f L_{V,2}(4+8\log|\Acal|)}{(1-\gamma)^4\underline{\sigma}^2}\tau,\label{eq:grad_gap_sigma_tau:proof_eq4}
\end{align}
where the third inequality follows from Lemma~\ref{lem:distance_pistar}.

Substituting \eqref{eq:grad_gap_sigma_tau:proof_eq2} and \eqref{eq:grad_gap_sigma_tau:proof_eq4} into \eqref{eq:grad_gap_sigma_tau:proof_eq1}, 
\begin{align*}
\|\nabla_x\Phi(x)-\nabla_x\Phi_{w,\tau}(x)\| &\leq \|\nabla_x\Phi_\tau(x)-\nabla_x\Phi_{w,\tau}(x)\| + \|\nabla_x\Phi(x)-\nabla_x\Phi_{\tau}(x)\|\notag\\
&\leq\frac{4L_f L_V w}{C_L \underline{\sigma} \tau}(L_f+\frac{2L_f L_{V,2}}{C_L\tau})\notag\\
&\hspace{20pt}+\frac{L_\star L_f (L_V+1)+L_\star L_f L_{V,2}+L_\star L_V L_{V,2}+L_f L_{V,2}(4+8\log|\Acal|)}{(1-\gamma)^4\underline{\sigma}^2}\tau.
\end{align*}

\qed

\subsection{Proof of Lemma~\ref{lem:Phi_tau}}

Let $\theta^\star(x)$ and $\theta_\tau^\star(x)$ denote one of the softmax parameters that encodes $\pi^\star(x)$ and $\pi_\tau^\star(x)$. Recall the gradient expression $\nabla_x\Phi_\tau(x)$ from \eqref{eq:hypergrad_Phi_tau}. 
We can similarly write
\begin{align}
\nabla_x\Phi(x)
&=\nabla_x f(x,\pi^\star(x))-\nabla_{x,\pi}^2 J(x,\pi^\star(x))\nabla_{\pi,\pi}^2 J(x,\pi^\star(x))^{-1}\nabla_{\pi}f(x,\pi^\star(x)).\label{eq:hypergrad_Phi}
\end{align}

Combining \eqref{eq:hypergrad_Phi_tau} and \eqref{eq:hypergrad_Phi},
\begin{align}
&\|\nabla_x\Phi_\tau(x)-\nabla_x\Phi(x)\|\notag\\
&\leq\underbrace{\|\nabla_x f(x,\pi_{\theta_\tau^\star(x)})-\nabla_x f(x,\pi_{\theta^\star(x)})\|}_{T_1}\notag\\
&\hspace{20pt} + \underbrace{\|\nabla_{x,\theta}^2 J_\tau(x,\pi_{\theta_\tau^\star(x)})\nabla_{\theta,\theta}^2 J_\tau(x,\pi_{\theta_\tau^\star(x)})^{-1}\nabla_{\theta}f(x,\pi_{\theta_\tau^\star(x)})\hspace{-2pt}-\hspace{-2pt}\nabla_{x,\theta}^2 J(x,\pi_{\theta^\star(x)})\nabla_{\theta,\theta}^2 J_\tau(x,\pi_{\theta_\tau^\star(x)})^{-1}\nabla_{\theta}f(x,\pi_{\theta_\tau^\star(x)})\|}_{T_2}\notag\\
&\hspace{20pt} + \underbrace{\|\nabla_{x,\theta}^2 J(x,\pi_{\theta^\star(x)})\nabla_{\theta,\theta}^2 J_\tau(x,\pi_{\theta_\tau^\star(x)})^{-1}\nabla_{\theta}f(x,\pi_{\theta_\tau^\star(x)})-\nabla_{x,\theta}^2 J(x,\pi_{\theta^\star(x)})\nabla_{\theta,\theta}^2 J(x,\pi_{\theta^\star(x)})^{-1}\nabla_{\theta}f(x,\pi_{\theta_\tau^\star(x)})\|}_{T_3}\notag\\
&\hspace{20pt} + \underbrace{\|\nabla_{x,\theta}^2 J(x,\pi_{\theta^\star(x)})\nabla_{\theta,\theta}^2 J(x,\pi_{\theta^\star(x)})^{-1}\nabla_{\theta}f(x,\pi_{\theta_\tau^\star(x)})-\nabla_{x,\theta}^2 J(x,\pi_{\theta^\star(x)})\nabla_{\theta,\theta}^2 J(x,\pi_{\theta^\star(x)})^{-1}\nabla_{\theta}f(x,\pi_{\theta^\star(x)})\|}_{T_4}.\label{lem:Phi_tau:proof_eq1}
\end{align}

We treat each term of \eqref{lem:Phi_tau:proof_eq1} individually. First, we bound $T_1$ using the smoothness property of $f$
\begin{align}
T_1 &\leq L_f\|\pi_\tau^\star(x)-\pi^\star(x)\|\leq  L_\star L_f \tau,\label{lem:Phi_tau:proof_eq2}
\end{align}
where the second inequality follows from Lemma~\ref{lem:distance_pistar}.

We have $\|\nabla_\theta f(x,\pi_\theta)\|\leq L_f$ from Assumption~\ref{assump:f} and $\|\nabla_{\theta,\theta}^2 J_\tau(x,\pi_{\theta_\tau^\star(x)})^{-1}\|\leq\frac{1}{\underline{\sigma}}$ due to Assumption~\ref{assump:invertible}. This allows us to bound $T_2$ as follows
\begin{align}
T_2 &\leq \|\nabla_{x,\theta}^2 J_\tau(x,\pi_{\theta_\tau^\star(x)})-\nabla_{x,\theta}^2 J(x,\pi_{\theta^\star(x)})\|\|\nabla_{\theta,\theta}^2 J_\tau(x,\pi_{\theta_\tau^\star(x)})^{-1}\nabla_{\theta}f(x,\pi_{\theta_\tau^\star(x)})\|\notag\\
&\leq \frac{L_f}{\underline{\sigma}}\Big(\|\nabla_{x,\theta}^2 J_\tau(x,\pi_{\theta_\tau^\star(x)})-\nabla_{x,\theta}^2 J(x,\pi_{\theta_\tau^\star(x)})\|+\|\nabla_{x,\theta}^2 J(x,\pi_{\theta_\tau^\star(x)})-\nabla_{x,\theta}^2 J(x,\pi_{\theta^\star(x)})\|\Big)\notag\\
&=\frac{L_f}{\underline{\sigma}}\|\nabla_{x,\theta}^2 J(x,\pi_{\theta_\tau^\star(x)})-\nabla_{x,\theta}^2 J(x,\pi_{\theta^\star(x)})\|\notag\\
&\leq \frac{L_f}{\underline{\sigma}}\cdot L_{V,2}\|\pi_{\theta_\tau^\star(x)}-\pi_{\theta^\star(x)}\|\notag\\
&= \frac{L_f}{\underline{\sigma}}\cdot L_{V,2}\|\pi_\tau^\star(x)-\pi^\star(x)\|\notag\\
&\leq\frac{L_\star L_f L_{V,2} \tau}{\underline{\sigma}},\label{lem:Phi_tau:proof_eq3}
\end{align}
where the first equation is due to the fact that $J_\tau(x,\pi)-J(x,\pi)$ is independent of $x$, so the derivative with respect to $x$ is zero. The last inequality plugs in \eqref{lem:Phi_tau:proof_eq2}.

Similarly, for $T_3$
\begin{align}
T_3 &\leq \|\nabla_{x,\theta}^2 J(x,\pi_{\theta^\star(x)})\| \|\nabla_{\theta,\theta}^2 J_\tau(x,\pi_{\theta_\tau^\star(x)})^{-1}-\nabla_{\theta,\theta}^2 J(x,\pi_{\theta^\star(x)})^{-1}\| \|\nabla_{\theta}f(x,\pi_{\theta_\tau^\star(x)})\|\notag\\
&\leq L_f L_V \|\nabla_{\theta,\theta}^2 J_\tau(x,\pi_{\theta_\tau^\star(x)})^{-1}\|\|\nabla_{\theta,\theta}^2 J_\tau(x,\pi_{\theta_\tau^\star(x)})-\nabla_{\theta,\theta}^2 J(x,\pi_{\theta^\star(x)})\|\|\nabla_{\theta,\theta}^2 J(x,\pi_{\theta_\tau^\star(x)})^{-1}\|\notag\\
&\leq \frac{L_f L_V}{\underline{\sigma}^2}\Big(\|\nabla_{\theta,\theta}^2 J_\tau(x,\pi_{\theta_\tau^\star(x)})-\nabla_{\theta,\theta}^2 J_\tau(x,\pi_{\theta^\star(x)})\|+\|\nabla_{\theta,\theta}^2 J_\tau(x,\pi_{\theta^\star(x)})-\nabla_{\theta,\theta}^2 J(x,\pi_{\theta^\star(x)})\|\Big)\notag\\
&\leq \frac{L_f L_V}{\underline{\sigma}^2}\Big(L_{V,2}\|\pi_\tau^\star(x)-\pi^\star(x)\|+L_V\tau\Big)\notag\\
&\leq \frac{L_\star L_f L_V L_{V,2}\tau}{\underline{\sigma}^2}+\frac{L_f L_V^2 \tau}{\underline{\sigma}^2},\label{lem:Phi_tau:proof_eq4}
\end{align}
where the third inequality again follows from Assumption~\ref{assump:invertible}, and the fourth inequality is due to \eqref{lem:Lipschitz_V:eq5} of Lemma~\ref{lem:Lipschitz_V} (note that $J_\tau(x,\pi)-J(x,\pi)=\tau\mathbb{E}_{s\sim d_\rho^\pi}[E(\pi,s)]$). The last inequality again plugs in \eqref{lem:Phi_tau:proof_eq2}.

For the final term, we have
\begin{align}
T_4 &\leq \|\nabla_{x,\theta}^2 J(x,\pi_{\theta^\star(x)})\nabla_{\theta,\theta}^2 J(x,\pi_{\theta^\star(x)})^{-1}\|\|\nabla_{\theta}f(x,\pi_{\theta_\tau^\star(x)})-\nabla_{\theta}f(x,\pi_{\theta^\star(x)})\|\notag\\
&\leq \frac{L_V}{\underline{\sigma}}\|\nabla_{\theta}f(x,\pi_{\theta_\tau^\star(x)})-\nabla_{\theta}f(x,\pi_{\theta^\star(x)})\|\notag\\
&\leq \frac{L_V}{\underline{\sigma}}\|\nabla_{\pi}f(x,\pi_\tau^\star(x))-\nabla_{\pi}f(x,\pi^\star(x))\|\notag\\
&\leq \frac{L_V}{\underline{\sigma}}\|\pi_\tau^\star(x)-\pi^\star(x)\|\notag\\
&\leq \frac{L_\star L_V\tau}{\underline{\sigma}},\label{lem:Phi_tau:proof_eq5}
\end{align}
where the third inequality is due to the $1$-Lipschitz continuity of softmax function, and the fourth inequality is due to Assumption~\ref{assump:f}.

Combining \eqref{lem:Phi_tau:proof_eq2}-\eqref{lem:Phi_tau:proof_eq5} leads to
\begin{align*}
\|\nabla_x\Phi_\tau(x)-\nabla_x\Phi(x)\|&\leq L_\star L_f\tau+\frac{L_\star L_f L_{V,2} \tau}{\underline{\sigma}}+\frac{L_\star L_f L_V L_{V,2}\tau}{\underline{\sigma}^2}+\frac{L_\star L_V^2 \tau}{\underline{\sigma}}+\frac{L_\star  L_V\tau}{\underline{\sigma}}\notag\\
&\leq \frac{L_\star L_f L_V^2 L_{V,2}\tau}{\underline{\sigma}^2}.
\end{align*}

\qed

\subsection{Proof of Lemma~\ref{lem:policy_conv_cross_term}}\label{sec:proof:lem:policy_conv_cross_term}

As $\nabla_x J_{\tau_{k}}(x_{k},\pi_{\theta_{k}})-\nabla_x J_{\tau_{k}}(x_{k},\pi_{\theta_{\tau_k}^\star(x_k)})$ does not depend on the randomness at iteration $k$, we have
\begin{align*}
&\mathbb{E}[-\langle\nabla_x J_{\tau_{k}}(x_{k},\pi_{\theta_{k}})-\nabla_x J_{\tau_{k}}(x_{k},\pi_{\theta_{\tau_k}^\star(x_k)}),x_{k+1}-x_k\rangle\rangle]\notag\\
&=\zeta_k\mathbb{E}[\langle\nabla_x J_{\tau_{k}}(x_{k},\pi_{\theta_{k}})-\nabla_x J_{\tau_{k}}(x_{k},\pi_{\theta_{\tau_k}^\star(x_k)}),\mathbb{E}[D_{w_k}(x_k,\pi_k,\pi_k^{\bias},s_k,a_k,\bar{s}_k,\bar{a}_k,\xi_k)\mid\Fcal_{k-1}]\rangle]\notag\\
&=\zeta_k\mathbb{E}[\langle\nabla_x J_{\tau_{k}}(x_{k},\pi_{\theta_{k}})-\nabla_x J_{\tau_{k}}(x_{k},\pi_{\theta_{\tau_k}^\star(x_k)}),\bar{D}_{w_k}(x_k,\pi_k,\pi_k^{\bias})\rangle]\notag\\
&=\zeta_k\mathbb{E}[\langle\nabla_x J_{\tau_{k}}(x_{k},\pi_{\theta_{k}})-\nabla_x J_{\tau_{k}}(x_{k},\pi_{\theta_{\tau_k}^\star(x_k)}),\nabla_x\Phi_{w_k,\tau_k}(x_k)\rangle]\notag\\
&\hspace{20pt}-\zeta_k\mathbb{E}[\langle\nabla_x J_{\tau_{k}}(x_{k},\pi_{\theta_{k}})-\nabla_x J_{\tau_{k}}(x_{k},\pi_{\theta_{\tau_k}^\star(x_k)}),\bar{D}_{w_k}(x_k,\pi_{\tau_k}^\star(x_k),\pi_{w_k,\tau_k}^\star(x_k))-\bar{D}_{w_k}(x_k,\pi_k,\pi_k^{\bias})\rangle],
\end{align*}
where the third equation is from \eqref{eq:grad_Phi_barD_equality}.

By Young's inequality,
\begin{align*}
&\mathbb{E}[-\langle\nabla_x J_{\tau_{k}}(x_{k},\pi_{\theta_{k}})-\nabla_x J_{\tau_{k}}(x_{k},\pi_{\theta_{\tau_k}^\star(x_k)}),x_{k+1}-x_k\rangle\rangle]\notag\\
&\leq \frac{C_L^2\alpha_k\tau_k^2}{128L_V^2}\mathbb{E}[\|\nabla_x J_{\tau_{k}}(x_{k},\pi_{\theta_{k}})-\nabla_x J_{\tau_{k}}(x_{k},\pi_{\theta_{\tau_k}^\star(x_k)})\|^2]+\frac{32L_V^2 \zeta_k^2}{ C_L^2\alpha_k\tau_k^2}\mathbb{E}[\|\nabla_{x} \Phi_{w_k,\tau_k}(x_k)\|^2]\notag\\
&\hspace{20pt}+\frac{C_L^2\alpha_k\tau_k^2}{128L_V^2}\mathbb{E}[\|\nabla_x J_{\tau_{k}}(x_{k},\pi_{\theta_{k}})-\nabla_x J_{\tau_{k}}(x_{k},\pi_{\theta_{\tau_k}^\star(x_k)})\|^2]\notag\\
&\hspace{20pt}+\frac{32L_V^2 \zeta_k^2}{C_L^2\alpha_k\tau_k^2}\mathbb{E}[\|\bar{D}_{w_k}(x_k,\pi_{\tau_k}^\star(x_k),\pi_{w_k,\tau_k}^\star(x_k))-\bar{D}_{w_k}(x_k,\pi_k,\pi_k^{\bias})\|^2]\notag\\
&\leq \frac{C_L^2\alpha_k\tau_k^2}{64L_V^2}\mathbb{E}[\|\nabla_x J_{\tau_{k}}(x_{k},\pi_{\theta_{k}})-\nabla_x J_{\tau_{k}}(x_{k},\pi_{\theta_{\tau_k}^\star(x_k)})\|^2]+\frac{32L_V^2 \zeta_k^2}{ C_L^2\alpha_k\tau_k^2}\mathbb{E}[\|\nabla_{x} \Phi_{w_k,\tau_k}(x_k)\|^2]\notag\\
&\hspace{20pt}+\frac{64 L_D^2 L_V^2 \zeta_k^2}{ C_L^2\alpha_k w_k^2 \tau_k^2} \mathbb{E}[\|\pi_k-\pi_{\tau_k}^\star(x_k)\|^2+\|\pi_k^{\bias}-\pi_{w_k,\tau_k}^\star(x_k)\|^2]\notag\\
&\leq \frac{ C_L^2\alpha_k\tau_k^2}{64}\mathbb{E}[\|\pi_k-\pi_{\tau_k}^\star(x_k)\|^2]\notag\\
&\hspace{20pt}+\frac{64 L_D^2 L_V^2 \zeta_k^2}{ C_L^2\alpha_k w_k^2 \tau_k^2} \mathbb{E}[\|\pi_k-\pi_{\tau_k}^\star(x_k)\|^2+\|\pi_k^{\bias}-\pi_{w_k,\tau_k}^\star(x_k)\|^2]+\frac{32L_V^2\zeta_k^2}{ C_L^2\alpha_k\tau_k^2}\mathbb{E}[\|\nabla_{x} \Phi_{w_k,\tau_k}(x_k)\|^2],
\end{align*}
where the second inequality employs the Lipschitz continuity of $\bar{D}_{w_k}$ established in Lemma~\ref{lem:Lipschitz_DFG}.

\qed

\subsection{Proof of Lemma~\ref{lem:policybias_conv_cross_term}}\label{sec:proof:lem:policybias_conv_cross_term}

As $\nabla_x \Lcal^{\reweight}_{w_{k},\tau_{k}}(x_{k},\pi_{\theta_{k}^{\bias}})-\nabla_x \Lcal^{\reweight}_{w_{k},\tau_{k}}(x_{k},\pi_{\theta_{w_k,\tau_k}^\star(x_k)})$ does not depend on the randomness at iteration $k$, we have
\begin{align*}
&\mathbb{E}[\langle\nabla_x \Lcal^{\reweight}_{w_{k},\tau_{k}}(x_{k},\pi_{\theta_{k}^{\bias}})-\nabla_x \Lcal^{\reweight}_{w_{k},\tau_{k}}(x_{k},\pi_{\theta_{w_k,\tau_k}^\star(x_k)}),x_{k+1}-x_k\rangle]\notag\\
&=-\zeta_k\mathbb{E}[\langle\nabla_x \Lcal^{\reweight}_{w_{k},\tau_{k}}(x_{k},\pi_{\theta_{k}^{\bias}})-\nabla_x \Lcal^{\reweight}_{w_{k},\tau_{k}}(x_{k},\pi_{\theta_{w_k,\tau_k}^\star(x_k)}),\mathbb{E}[D_{w_k}(x_k,\pi_k,\pi_k^{\bias},s_k,a_k,\bar{s}_k,\bar{a}_k,\xi_k)\mid\Fcal_{k-1}]\rangle]\notag\\
&=-\zeta_k\mathbb{E}[\langle\nabla_x \Lcal^{\reweight}_{w_{k},\tau_{k}}(x_{k},\pi_{\theta_{k}^{\bias}})-\nabla_x \Lcal^{\reweight}_{w_{k},\tau_{k}}(x_{k},\pi_{\theta_{w_k,\tau_k}^\star(x_k)}),\bar{D}_{w_k}(x_k,\pi_k,\pi_k^{\bias})\rangle]\notag\\
&=-\zeta_k\mathbb{E}[\langle\nabla_x \Lcal^{\reweight}_{w_{k},\tau_{k}}(x_{k},\pi_{\theta_{k}^{\bias}})-\nabla_x \Lcal^{\reweight}_{w_{k},\tau_{k}}(x_{k},\pi_{\theta_{w_k,\tau_k}^\star(x_k)}),\nabla_x\Phi_{w_k,\tau_k}(x_k)\rangle]\notag\\
&\hspace{20pt}+\zeta_k\mathbb{E}[\langle\nabla_x \Lcal^{\reweight}_{w_{k},\tau_{k}}(x_{k},\pi_{\theta_{k}^{\bias}})-\nabla_x \Lcal^{\reweight}_{w_{k},\tau_{k}}(x_{k},\pi_{\theta_{w_k,\tau_k}^\star(x_k)}),\notag\\
&\hspace{100pt}\bar{D}_{w_k}(x_k,\pi_{\tau_k}^\star(x_k),\pi_{w_k,\tau_k}^\star(x_k))-\bar{D}_{w_k}(x_k,\pi_k,\pi_k^{\bias})\rangle],
\end{align*}
where the third equation is from \eqref{eq:grad_Phi_barD_equality}.

By Young's inequality,
\begin{align*}
&\mathbb{E}[\langle\nabla_x \Lcal^{\reweight}_{w_{k},\tau_{k}}(x_{k},\pi_{\theta_{k}^{\bias}})-\nabla_x \Lcal^{\reweight}_{w_{k},\tau_{k}}(x_{k},\pi_{\theta_{w_k,\tau_k}^\star(x_k)}),x_{k+1}-x_k\rangle]\notag\\
&\leq \frac{C_L^2\alpha_k\tau_k^2}{128L_L^2}\mathbb{E}[\|\nabla_x \Lcal^{\reweight}_{w_{k},\tau_{k}}(x_{k},\pi_{\theta_{k}^{\bias}})-\nabla_x \Lcal^{\reweight}_{w_{k},\tau_{k}}(x_{k},\pi_{\theta_{w_k,\tau_k}^\star(x_k)})\|^2]+\frac{32L_L^2 \zeta_k^2}{ C_L^2\alpha_k\tau_k^2}\mathbb{E}[\|\nabla_{x} \Phi_{w_k,\tau_k}(x_k)\|^2]\notag\\
&\hspace{20pt}+\frac{ C_L^2\alpha_k\tau_k^2}{128L_L^2}\mathbb{E}[\|\nabla_x \Lcal^{\reweight}_{w_{k},\tau_{k}}(x_{k},\pi_{\theta_{k}^{\bias}})-\nabla_x \Lcal^{\reweight}_{w_{k},\tau_{k}}(x_{k},\pi_{\theta_{w_k,\tau_k}^\star(x_k)})\|^2]\notag\\
&\hspace{20pt}+\frac{32L_L^2\zeta_k^2}{ C_L^2\alpha_k\tau_k^2}\mathbb{E}[\|\bar{D}_{w_k}(x_k,\pi_{\tau_k}^\star(x_k),\pi_{w_k,\tau_k}^\star(x_k))-\bar{D}_{w_k}(x_k,\pi_k,\pi_k^{\bias})\|^2]\notag\\
&\leq \frac{ C_L^2\alpha_k\tau_k^2}{64L_L^2}\mathbb{E}[\|\nabla_x \Lcal^{\reweight}_{w_{k},\tau_{k}}(x_{k},\pi_{\theta_{k}^{\bias}})-\nabla_x \Lcal^{\reweight}_{w_{k},\tau_{k}}(x_{k},\pi_{\theta_{w_k,\tau_k}^\star(x_k)})\|^2]+\frac{32L_L^2 \zeta_k^2}{ C_L^2\alpha_k\tau_k^2}\mathbb{E}[\|\nabla_{x} \Phi_{w_k,\tau_k}(x_k)\|^2]\notag\\
&\hspace{20pt}+\frac{64 L_D^2 L_L^2 \zeta_k^2}{ C_L^2\alpha_k w_k^2\tau_k^2} \mathbb{E}[\|\pi_k-\pi_{\tau_k}^\star(x_k)\|^2+\|\pi_k^{\bias}-\pi_{w_k,\tau_k}^\star(x_k)\|^2]\notag\\
&\leq \frac{ C_L^2\alpha_k\tau_k^2}{64}\mathbb{E}[\|\pi_k^{\bias}-\pi_{w_k,\tau_k}^\star(x_k)\|^2]\notag\\
&\hspace{20pt}+\frac{64 L_D^2 L_L^2 \zeta_k^2}{ C_L^2\alpha_k w_k^2 \tau_k^2} \mathbb{E}[\|\pi_k-\pi_{\tau_k}^\star(x_k)\|^2+\|\pi_k^{\bias}-\pi_{w_k,\tau_k}^\star(x_k)\|^2]+\frac{32L_L^2 \zeta_k^2}{ C_L^2\alpha_k\tau_k^2}\mathbb{E}[\|\nabla_{x} \Phi_{w_k,\tau_k}(x_k)\|^2],
\end{align*}
where the second inequality employs the Lipschitz continuity of $\bar{D}_{w_k}$ established in Lemma~\ref{lem:Lipschitz_DFG}.

\qed

\subsection{Proof of Lemma~\ref{lem:V_cross_term}}\label{proof:V_cross_term}

Within the proof of this lemma, we employ the shorthand notation $z_k = [x_k,\theta_k,\tau_k]$, $\ell(z_k)=V_{\tau_k}^{x_k,\pi_{\theta_k}}$, and 
\[y_k=\hat{V}_{k} - V_{\tau_{k}}^{x_{k},\pi_{\theta_{k}}}+\beta_k \bar{G}_{\tau_k}(x_k,\theta_k,\hat{V}_{k}).\]

As $V_\tau^{x,\pi_\theta}$ is smooth in $x,\theta,\tau$, we have from the mean-value theorem that there exists $z^m_{k+1}=m z_k+(1-m) z_{k+1}$ for some scalar $m\in[0,1]$ such that 
\begin{align}
&\ell(z_k)-\ell(z_{k+1})\notag\\
&= \nabla_{z}\ell(z^m_{k+1})^{\top}\Big(z_k-z_{k+1}\Big)\notag\\
&= \Big(\nabla_{x}V_{\tau_{k+1}^m}^{x_{k+1}^m,\pi_{\theta^m_{k+1}}}\Big)^{\top}\Big(x_k-x_{k+1}\Big) + \Big(\nabla_{\theta}V_{\tau_{k+1}^m}^{x_{k+1}^m,\pi_{\theta^m_{k+1}}}\Big)^{\top}\Big(\theta_k-\theta_{k+1}\Big)\notag\\
&\hspace{20pt} + \Big(\nabla_{\tau}V_{\tau_{k+1}^m}^{x_{k+1}^m,\pi_{\theta^m_{k+1}}}\Big)^{\top}\Big(\tau_k-\tau_{k+1}\Big)\notag\\
&=\zeta_k\Big(\nabla_{x}V_{\tau_{k+1}^m}^{x_{k+1}^m,\pi_{\theta^m_{k+1}}}\Big)^{\top}\bar{D}_{w_k}(x_k,\pi_{\theta_k},\pi_{\theta_k^{\bias}})\notag\\
&\hspace{20pt}+\zeta_k\Big(\nabla_{x}V_{\tau_{k+1}^m}^{x_{k+1}^m,\pi_{\theta^m_{k+1}}}\Big)^{\top}\Big(D_{w_k}(x_k,\pi_{\theta_k},\pi_{\theta_k^{\bias}},s_k,a_k,\bar{s}_k,\bar{a}_k,\xi_k)-\bar{D}_{w_k}(x_k,\pi_{\theta_k},\pi_{\theta_k^{\bias}})\Big)\notag\\
&\hspace{20pt}+\alpha_k\Big(\nabla_{\theta}V_{\tau_{k+1}^m}^{x_{k+1}^m,\pi_{\theta^m_{k+1}}}\Big)^{\top}\bar{F}_{0,\tau_k}(x_k,\theta_k,\hat{V}_{k})\notag\\
&\hspace{20pt}+\alpha_k\Big(\nabla_{\theta}V_{\tau_{k+1}^m}^{x_{k+1}^m,\pi_{\theta^m_{k+1}}}\Big)^{\top}\Big(F_{0,\tau_k}(x_k,\theta_k,\hat{V}_{k},s_k,a_k,s_k',\xi_k)-\bar{F}_{0,\tau_k}(x_k,\theta_k,\hat{V}_{k})\Big)\notag\\
&\hspace{20pt}+\Big(\nabla_{\tau}V_{\tau_{k+1}^m}^{x_{k+1}^m,\pi_{\theta^m_{k+1}}}\Big)^{\top}\Big(\tau_k-\tau_{k+1}\Big),\label{lem:V_cross_term:proof_eq1}
\end{align}
where we denote $x^m_{k+1}=m x_k+(1-m)x_{k+1}, \theta^m_{k+1}=m\theta_k+(1-m)\theta_{k+1},  \tau^m_{k+1}=m\tau_k+(1-m)\tau_{k+1}$.

Plugging \eqref{lem:V_cross_term:proof_eq1} into the cross term of interest, we have
\begin{align}
&\langle\hat{V}_{k} - V_{\tau_{k}}^{x_{k},\pi_{\theta_{k}}}+\beta_k \bar{G}_{\tau_k}(x_k,\theta_k,\hat{V}_{k}), V_{\tau_{k}}^{x_{k},\pi_{\theta_{k}}}-V_{\tau_{k+1}}^{x_{k+1},\pi_{\theta_{k+1}}}\rangle\notag\\
&=\langle y_k,\ell(z_k)-\ell(x_{k+1})\rangle\notag\\
&=\alpha_k \langle y_k, \Big(\nabla_{\theta}V_{\tau_{k+1}^m}^{x_{k+1}^m,\pi_{\theta^m_{k+1}}}\Big)^{\top}\bar{F}_{0,\tau_k}(x_k,\theta_k,\hat{V}_{k})\rangle\notag\\
&\hspace{20pt}+\alpha_k \langle y_k, \Big(\nabla_{\theta}V_{\tau_{k+1}^m}^{x_{k+1}^m,\pi_{\theta^m_{k+1}}}\Big)^{\top}\Big(F_{0,\tau_k}(x_k,\theta_k,\hat{V}_{k},s_k,a_k,s_k',\xi_k)-\bar{F}_{0,\tau_k}(x_k,\theta_k,\hat{V}_{k})\Big)\rangle\notag\\
&\hspace{20pt}+\zeta_k \langle y_k, \Big(\nabla_{x}V_{\tau_{k+1}^m}^{x_{k+1}^m,\pi_{\theta^m_{k+1}}}\Big)^{\top}\bar{D}_{w_k}(x_k,\pi_{\theta_k},\pi_{\theta_k^{\bias}})\rangle\notag\\
&\hspace{20pt}+\zeta_k \langle y_k, \Big(\nabla_{x}V_{\tau_{k+1}^m}^{x_{k+1}^m,\pi_{\theta^m_{k+1}}}\Big)^{\top}\Big(D_{w_k}(x_k,\pi_{\theta_k},\pi_{\theta_k^{\bias}},s_k,a_k,\bar{s}_k,\bar{a},\xi_k)-\bar{D}_{w_k}(x_k,\pi_{\theta_k},\pi_{\theta_k^{\bias}})\Big)\rangle\notag\\
&\hspace{20pt}+\langle y_k, \Big(\nabla_{\tau}V_{\tau_{k+1}^m}^{x_{k+1}^m,\pi_{\theta^m_{k+1}}}\Big)^{\top}\Big(\tau_k-\tau_{k+1}\Big)\rangle.\label{lem:V_cross_term:proof_eq2}
\end{align}

We bound each term of \eqref{lem:V_cross_term:proof_eq2} individually. First, by Young's inequality
\begin{align}
&\alpha_k \langle y_k, \Big(\nabla_{\theta}V_{\tau_{k+1}^m}^{x_{k+1}^m,\pi_{\theta^m_{k+1}}}\Big)^{\top}\bar{F}_{0,\tau_k}(x_k,\theta_k,\hat{V}_{k})\rangle\notag\\
&\leq L_V\alpha_k\|y_k\|\|\bar{F}_{0,\tau_k}(x_k,\theta_k,\hat{V}_{k})\|\notag\\
&\leq \frac{(1-\gamma)\beta_k}{12}\|y_k\|^2+\frac{3L_V^2\alpha_k^2}{(1-\gamma)\beta_k}\|\bar{F}_{0,\tau_k}(x_k,\theta_k,\hat{V}_{k})\|^2\notag\\
&\leq \frac{(1-\gamma)\beta_k}{12}\|y_k\|^2+\frac{6L_V^2\alpha_k^2}{(1-\gamma)\beta_k}\|\bar{F}_{0,\tau_k}(x_k,\theta_k,V_{\tau_k}^{x_k,\pi_{\theta_k}})\|^2\notag\\
&\hspace{20pt} +\frac{6L_V^2\alpha_k^2}{(1-\gamma)\beta_k}\|\bar{F}_{0,\tau_k}(x_k,\theta_k,\hat{V}_{k})-\bar{F}_{0,\tau_k}(x_k,\theta_k,V_{\tau_k}^{x_k,\pi_{\theta_k}})\|^2\notag\\
&= \frac{(1-\gamma)\beta_k}{12}\|y_k\|^2+\frac{6L_V^2\alpha_k^2}{(1-\gamma)\beta_k}\|\nabla_{\theta}J_{\tau_k}(x_k,\pi_{\theta_k})\|^2 \notag\\
&\hspace{20pt} +\frac{6L_V^2\alpha_k^2}{(1-\gamma)\beta_k}\|\bar{F}_{0,\tau_k}(x_k,\theta_k,\hat{V}_{k})-\bar{F}_{0,\tau_k}(x_k,\theta_k,V_{\tau_k}^{x_k,\pi_{\theta_k}})\|^2\notag\\
&\leq \frac{(1-\gamma)\beta_k}{12}\|y_k\|^2+\frac{6L_V^2\alpha_k^2}{(1-\gamma)\beta_k}\|\nabla_{\theta}J_{\tau_k}(x_k,\pi_{\theta_k})\|^2+ \frac{6L_V^2 L_F^2 \alpha_k^2}{(1-\gamma)\beta_k}\varepsilon_k^V,\label{lem:V_cross_term:proof_eq3}
\end{align}
where the equation follows from the fact that $\bar{F}_{0,\tau}(x,\theta,V_\tau^{x,\pi_\theta})=\nabla_\theta J_\tau(x,\pi_\theta)$ for any $x,\theta,\tau$ (see \eqref{eq:grad_J_barF_equality}), and the final inequality is due to the Lipschitz continuity of $\bar{F}_{0,\tau_k}$.

For the second term of \eqref{lem:V_cross_term:proof_eq2}, we take the expectation
\begin{align}
&\alpha_k\mathbb{E}\left[\langle y_k, \Big(\nabla_{\theta}V_{\tau_{k+1}^m}^{x_{k+1}^m,\pi_{\theta^m_{k+1}}}\Big)^{\top}\Big(F_{0,\tau_k}(x_k,\theta_k,\hat{V}_{k},s_k,a_k,s_k',\xi_k)-\bar{F}_{0,\tau_k}(x_k,\theta_k,\hat{V}_{k})\Big)\rangle\right]\notag\\
&=\alpha_k\mathbb{E}[\langle y_k, \Big(\nabla_{\theta}V_{\tau_{k+1}^m}^{x_{k+1}^m,\pi_{\theta^m_{k+1}}}-\nabla_\theta V_{\tau_{k}}^{x_{k},\pi_{\theta_{k}}}\Big)^{\top}\Big(F_{0,\tau_k}(x_k,\theta_k,\hat{V}_{k},s_k,a_k,s_k',\xi_k)-\bar{F}_{0,\tau_k}(x_k,\theta_k,\hat{V}_{k})\Big)]\notag\\
&\hspace{20pt}+\alpha_k\mathbb{E}[\langle y_k, \Big(\nabla_\theta V_{\tau_{k}}^{x_{k},\pi_{\theta_{k}}}\Big)^{\top}\Big(F_{0,\tau_k}(x_k,\theta_k,\hat{V}_{k},s_k,a_k,s_k',\xi_k)-\bar{F}_{0,\tau_k}(x_k,\theta_k,\hat{V}_{k})\Big)]\notag\\
&=\alpha_k\mathbb{E}[\langle y_k, \Big(\nabla_{\theta}V_{\tau_{k+1}^m}^{x_{k+1}^m,\pi_{\theta^m_{k+1}}}-\nabla_\theta V_{\tau_{k}}^{x_{k},\pi_{\theta_{k}}}\Big)^{\top}\Big(F_{0,\tau_k}(x_k,\theta_k,\hat{V}_{k},s_k,a_k,s_k',\xi_k)-\bar{F}_{0,\tau_k}(x_k,\theta_k,\hat{V}_{k})\Big)]\notag\\
&\leq 2B_F\alpha_k\mathbb{E}[\|y_k\|\|\nabla_{\theta}V_{\tau_{k+1}^m}^{x_{k+1}^m,\pi_{\theta^m_{k+1}}}-\nabla_\theta V_{\tau_{k}}^{x_{k},\pi_{\theta_{k}}}\|]\notag\\
&\leq 2B_F L_V\alpha_k\mathbb{E}[\|y_k\|\Big(\|\pi_{\theta^m_{k+1}}-\pi_{\theta_k}\|+\|x^m_{k+1}-x_k\|+|\tau_{k+1}^m-\tau_k|\Big)]\notag\\
&\leq 2B_F L_V\alpha_k\mathbb{E}[\|y_k\|\Big(\|\pi_{\theta_{k+1}}-\pi_{\theta_k}\|+\|x_{k+1}-x_k\|+|\tau_{k+1}-\tau_k|\Big)]\notag\\
&\leq 2B_F L_V\alpha_k\mathbb{E}[\|y_k\|]\Big(\alpha_k B_F+\frac{\zeta_k B_D}{w_k}+\frac{8\tau_k}{3(k+1)}\Big)\notag\\
&\leq 2B_F L_V\alpha_k\mathbb{E}[\|y_k\|] \cdot \frac{3\tau_0 B_F\alpha_k}{\alpha_0}\notag\\
&\leq \frac{3B_F^2 L_V\tau_0\alpha_k^2}{\alpha_0}\mathbb{E}[\|y_k\|^2]+\frac{3B_F^2 L_V\tau_0\alpha_k^2}{\alpha_0},\label{lem:V_cross_term:proof_eq4}
\end{align}
where the fifth inequality follows from the step size conditions $\zeta_k\leq\frac{B_F \alpha_k w_k}{B_D}$ and $\alpha_0\leq\min\{\tau_0,\frac{3B_F}{8}\}$, and the second equation follows from
\begin{align*}
&\mathbb{E}[\langle y_k, \Big(\nabla_\theta V_{\tau_{k}}^{x_{k},\pi_{\theta_{k}}}\Big)^{\top}\Big(F_{0,\tau_k}(x_k,\theta_k,\hat{V}_{k},s_k,a_k,s_k',\xi_k)-\bar{F}_{0,\tau_k}(x_k,\theta_k,\hat{V}_{k})\Big)]\notag\\
&=\mathbb{E}[\langle y_k, \Big(V_{\tau_{k}}^{x_{k},\pi_{\theta_{k}}}\Big)^{\top}\mathbb{E}\Big[F(\theta_k,\omega_k,\hat\mu_k,\hat{V}_{f,k},s_k,a_k,b_k,s_k')-\bar{F}(\theta_k,\omega_k,\hat\mu_k,\hat{V}_{f,k})\mid\Fcal_{k-1}\Big]]\notag\\
&=0.
\end{align*}

The third term of \eqref{lem:V_cross_term:proof_eq2} can be bounded similar to the first term,
\begin{align}
\zeta_k \langle y_k, \big(\nabla_{x}V_{\tau_{k+1}^m}^{x_{k+1}^m,\pi_{\theta^m_{k+1}}}\big)^{\top}\hspace{-2pt}\bar{D}_{w_k}(x_k,\pi_{\theta_k},\pi_{\theta_k^{\bias}})\rangle&\leq L_V\zeta_k\|y_k\|\|\bar{D}_{w_k}(x_k,\pi_{\theta_k},\pi_{\theta_k^{\bias}})\|\notag\\
&\leq\frac{(1-\gamma)\beta_k}{12}\|y_k\|^2+\frac{3L_V^2\zeta_k^2}{(1-\gamma)\beta_k}\|\bar{D}_{w_k}(x_k,\pi_{\theta_k},\pi_{\theta_k^{\bias}})\|^2\notag\\
&\leq \frac{(1-\gamma)\beta_k}{12}\|y_k\|^2+\frac{6L_V^2\zeta_k^2}{(1-\gamma)\beta_k}\|\bar{D}_{w_k}(x_k,\pi_{\tau_k}^\star(x_k),\pi_{w_k,\tau_k}^\star(x_k))\|^2\notag\\
&\hspace{20pt} +\frac{6L_V^2\zeta_k^2}{(1-\gamma)\beta_k}\|\bar{D}_{w_k}(x_k,\pi_{\theta_k},\pi_{\theta_k^{\bias}})-\bar{D}_{w_k}(x_k,\pi_{\tau_k}^\star(x_k),\pi_{w_k,\tau_k}^\star(x_k))\|^2\notag\\
&= \frac{(1-\gamma)\beta_k}{12}\|y_k\|^2+\frac{6L_V^2\zeta_k^2}{(1-\gamma)\beta_k}\|\nabla_{x} \Phi_{w_k,\tau_k}(x_k)\|^2 \notag\\
&\hspace{20pt}+\frac{6L_V^2\zeta_k^2}{(1-\gamma)\beta_k}\|\bar{D}_{w_k}(x_k,\pi_{\theta_k},\pi_{\theta_k^{\bias}})-\bar{D}_{w_k}(x_k,\pi_{\tau_k}^\star(x_k),\pi_{w_k,\tau_k}^\star(x_k))\|^2\notag\\
&\leq \frac{(1-\gamma)\beta_k}{12}\|y_k\|^2+\frac{6L_V^2\zeta_k^2}{(1-\gamma)\beta_k}\|\nabla_{x} \Phi_{w_k,\tau_k}(x_k)\|^2 \notag\\
&\hspace{20pt} + \frac{12L_V^2 L_D^2\zeta_k^2}{(1-\gamma)\beta_k}\|\pi_k-\pi_{\tau_k}^\star(x_k)\|^2 + \frac{12L_V^2 L_D^2\zeta_k^2}{(1-\gamma)\beta_k}\|\pi_k^{\bias}-\pi_{w_k,\tau_k}^\star(x_k)\|^2,\label{lem:V_cross_term:proof_eq5}
\end{align}
where the equation is due to the condition in \eqref{eq:grad_Phi_barD_equality}.

For the fourth term of \eqref{lem:V_cross_term:proof_eq2}, we again take the expectation and use the technique in \eqref{lem:V_cross_term:proof_eq4}
\begin{align}
&\zeta_k \mathbb{E}[\langle y_k, \Big(\nabla_{x}V_{\tau_{k+1}^m}^{x_{k+1}^m,\pi_{\theta^m_{k+1}}}\Big)^{\top}\Big(D_{w_k}(x_k,\pi_{\theta_k},\pi_{\theta_k^{\bias}},s_k,a_k,\bar{s}_k,\bar{a},\xi_k)-\bar{D}_{w_k}(x_k,\pi_{\theta_k},\pi_{\theta_k^{\bias}})\Big)\rangle]\notag\\
&=\zeta_k\mathbb{E}[\langle y_k, \Big(\nabla_{x}V_{\tau_{k+1}^m}^{x_{k+1}^m,\pi_{\theta^m_{k+1}}}-\nabla_{x}V_{\tau_{k}}^{x_{k},\pi_{\theta_{k}}}\Big)^{\top}\Big(D_{w_k}(x_k,\pi_{\theta_k},\pi_{\theta_k^{\bias}},s_k,a_k,\bar{s}_k,\bar{a},\xi_k)-\bar{D}_{w_k}(x_k,\pi_{\theta_k},\pi_{\theta_k^{\bias}})\Big)]\notag\\
&\hspace{20pt}+\zeta_k\mathbb{E}[\langle y_k, \Big(\nabla_{x}V_{\tau_{k}}^{x_{k},\pi_{\theta_{k}}}\Big)^{\top}\Big(D_{w_k}(x_k,\pi_{\theta_k},\pi_{\theta_k^{\bias}},s_k,a_k,\bar{s}_k,\bar{a},\xi_k)-\bar{D}_{w_k}(x_k,\pi_{\theta_k},\pi_{\theta_k^{\bias}})\Big)]\notag\\
&=\zeta_k\mathbb{E}[\langle y_k, \Big(\nabla_{x}V_{\tau_{k+1}^m}^{x_{k+1}^m,\pi_{\theta^m_{k+1}}}-\nabla_{x}V_{\tau_{k}}^{x_{k},\pi_{\theta_{k}}}\Big)^{\top}\Big(D_{w_k}(x_k,\pi_{\theta_k},\pi_{\theta_k^{\bias}},s_k,a_k,\bar{s}_k,\bar{a},\xi_k)-\bar{D}_{w_k}(x_k,\pi_{\theta_k},\pi_{\theta_k^{\bias}})\Big)]\notag\\
&\leq 2B_D\zeta_k\mathbb{E}[\|y_k\|\|\nabla_{x}V_{\tau_{k+1}^m}^{x_{k+1}^m,\pi_{\theta^m_{k+1}}}-\nabla_{x}V_{\tau_{k}}^{x_{k},\pi_{\theta_{k}}}\|]\notag\\
&\leq 2B_D L_V\zeta_k\mathbb{E}[\|y_k\|\Big(\|\pi_{\theta^m_{k+1}}-\pi_{\theta_k}\|+\|x^m_{k+1}-x_k\|+|\tau_{k+1}^m-\tau_k|\Big)]\notag\\
&\leq 2B_D L_V\zeta_k\mathbb{E}[\|y_k\|\Big(\|\pi_{\theta_{k+1}}-\pi_{\theta_k}\|+\|x_{k+1}-x_k\|+|\tau_{k+1}-\tau_k|\Big)]\notag\\
&\leq 2B_D L_V\zeta_k\mathbb{E}[\|y_k\|]\Big(\alpha_k B_F+\frac{\zeta_k B_D}{w_k}+\frac{8\tau_k}{3(k+1)}\Big)\notag\\
&\leq 2B_D L_V\zeta_k\mathbb{E}[\|y_k\|] \cdot \frac{3\tau_0 B_F\alpha_k}{\alpha_0}\notag\\
&\leq \frac{3B_F^2 L_V\tau_0\alpha_k^2}{\alpha_0}\mathbb{E}[\|y_k\|^2]+\frac{3B_F^2 L_V\tau_0\alpha_k^2}{\alpha_0},\label{lem:V_cross_term:proof_eq6}
\end{align}
where the last inequality follows from the step size condition $\frac{\zeta_k}{\alpha_k}\leq\frac{B_F}{B_D}$.

For the fifth term of \eqref{lem:V_cross_term:proof_eq2}, we apply Lemma~\ref{lem:tau_diff}
\begin{align}
\langle y_k, \Big(\nabla_{\tau}V_{\tau_{k+1}^m}^{x_{k+1}^m,\pi_{\theta^m_{k+1}}}\Big)^{\top}\Big(\tau_k-\tau_{k+1}\Big)\rangle&\leq L_V\|y_k\||\tau_k-\tau_{k+1}|\notag\\
&\leq L_V\|y_k\|\cdot\frac{8\tau_k}{3(k+1)}\notag\\
&\leq \frac{(1-\gamma)\beta_k}{6}\|y_k\|^2+\frac{32L_V^2\tau_k^2}{3(1-\gamma)\beta_k(k+1)^2}.\label{lem:V_cross_term:proof_eq7}
\end{align}

Collecting the results in \eqref{lem:V_cross_term:proof_eq3}-\eqref{lem:V_cross_term:proof_eq7} and substituting into \eqref{lem:V_cross_term:proof_eq2},
\begin{align*}
&\mathbb{E}[\langle\hat{V}_{k} - V_{\tau_{k}}^{x_{k},\pi_{\theta_{k}}}+\beta_k \bar{G}_{\tau_k}(x_k,\theta_k,\hat{V}_{k}), V_{\tau_{k}}^{x_{k},\pi_{\theta_{k}}}-V_{\tau_{k+1}}^{x_{k+1},\pi_{\theta_{k+1}}}\rangle]\notag\\
&\leq \frac{(1-\gamma)\beta_k}{12}\mathbb{E}[\|y_k\|^2]+\frac{6L_V^2\alpha_k^2}{(1-\gamma)\beta_k}\mathbb{E}[\|\nabla_{\theta}J_{\tau_k}(x_k,\pi_{\theta_k})\|^2]+ \frac{6L_V^2 L_F^2 \alpha_k^2}{(1-\gamma)\beta_k}\mathbb{E}[\varepsilon_k^V]\notag\\
&\hspace{20pt}+\frac{3B_F^2 L_V\tau_0\alpha_k^2}{\alpha_0}\mathbb{E}[\|y_k\|^2]+\frac{3B_F^2 L_V\tau_0\alpha_k^2}{\alpha_0}\notag\\
&\hspace{20pt}+\frac{(1-\gamma)\beta_k}{12}\mathbb{E}[\|y_k\|^2]+\frac{6L_V^2\zeta_k^2}{(1-\gamma)\beta_k}\mathbb{E}[\|\nabla_{x} \Phi_{w_k,\tau_k}(x_k)\|^2] \notag\\
&\hspace{20pt} + \frac{12L_V^2 L_D^2\zeta_k^2}{(1-\gamma)\beta_k}\mathbb{E}[\|\pi_k-\pi_{\tau_k}^\star(x_k)\|^2] + \frac{12L_V^2 L_D^2\zeta_k^2}{(1-\gamma)\beta_k}\mathbb{E}[\|\pi_k^{\bias}-\pi_{w_k,\tau_k}^\star(x_k)\|^2]\notag\\
&\hspace{20pt}+\frac{3B_F^2 L_V\tau_0\alpha_k^2}{\alpha_0}\mathbb{E}[\|y_k\|^2]+\frac{3B_F^2 L_V\tau_0\alpha_k^2}{\alpha_0}\notag\\
&\hspace{20pt}+\frac{(1-\gamma)\beta_k}{6}\mathbb{E}[\|y_k\|^2]+\frac{32L_V^2\tau_k^2}{3(1-\gamma)\beta_k(k+1)^2}\notag\\
&\leq \frac{(1-\gamma)\beta_k}{3}\mathbb{E}[\|y_k\|^2]+\frac{6B_F^2 L_V\tau_0\alpha_k^2}{\alpha_0}\mathbb{E}[\|y_k\|^2]+\frac{6L_V^2\zeta_k^2}{(1-\gamma)\beta_k}\mathbb{E}[\|\nabla_{x} \Phi_{w_k,\tau_k}(x_k)\|^2]\notag\\
&\hspace{20pt}+\frac{6L_V^2\alpha_k^2}{(1-\gamma)\beta_k}\mathbb{E}[\|\nabla_{\theta}J_{\tau_k}(x_k,\pi_{\theta_k})\|^2]+ \frac{6L_V^2 L_F^2 \alpha_k^2}{(1-\gamma)\beta_k}\mathbb{E}[\varepsilon_k^V]\notag\\
&\hspace{20pt} + \frac{12L_V^2 L_D^2\zeta_k^2}{(1-\gamma)\beta_k}\mathbb{E}[\|\pi_k-\pi_{\tau_k}^\star(x_k)\|^2] + \frac{12L_V^2 L_D^2\zeta_k^2}{(1-\gamma)\beta_k}\mathbb{E}[\|\pi_k^{\bias}-\pi_{w_k,\tau_k}^\star(x_k)\|^2]\notag\\
&\hspace{20pt}+\frac{6B_F^2 L_V\tau_0\alpha_k^2}{\alpha_0}+\frac{32L_V^2\tau_k^2}{3(1-\gamma)\beta_k(k+1)^2}\notag\\
&\leq \frac{(1-\gamma)\beta_k}{2}\mathbb{E}[\|y_k\|^2]+\frac{6L_V^2\zeta_k^2}{(1-\gamma)\beta_k}\mathbb{E}[\|\nabla_{x} \Phi_{w_k,\tau_k}(x_k)\|^2]+\frac{6L_V^2\alpha_k^2}{(1-\gamma)\beta_k}\mathbb{E}[\|\nabla_{\theta}J_{\tau_k}(x_k,\pi_{\theta_k})\|^2]+ \frac{6L_V^2 L_F^2 \alpha_k^2}{(1-\gamma)\beta_k}\mathbb{E}[\varepsilon_k^V]\notag\\
&\hspace{20pt} + \frac{12L_V^2 L_D^2\zeta_k^2}{(1-\gamma)\beta_k}\mathbb{E}[\|\pi_k-\pi_{\tau_k}^\star(x_k)\|^2] + \frac{12L_V^2 L_D^2\zeta_k^2}{(1-\gamma)\beta_k}\mathbb{E}[\|\pi_k^{\bias}-\pi_{w_k,\tau_k}^\star(x_k)\|^2]\notag\\
&\hspace{20pt}+\frac{6B_F^2 L_V\tau_0\alpha_k^2}{\alpha_0}+\frac{32L_V^2\tau_k^2}{3(1-\gamma)\beta_k(k+1)^2},
\end{align*}
where the terms are combined in the last inequality under the step size condition $\alpha_k\leq\beta_k$ and $\frac{\alpha_k}{\beta_k}\leq\frac{1-\gamma}{36B_F^2L_V\tau_0}$.

\qed

\section{Simulation Details -- GridWorld Goal Placement}\label{sec:simulation_details}

\begin{wrapfigure}{r}{0.35\textwidth}
\vspace{-45pt}
  \begin{center}
  \includegraphics[width=0.35\textwidth]{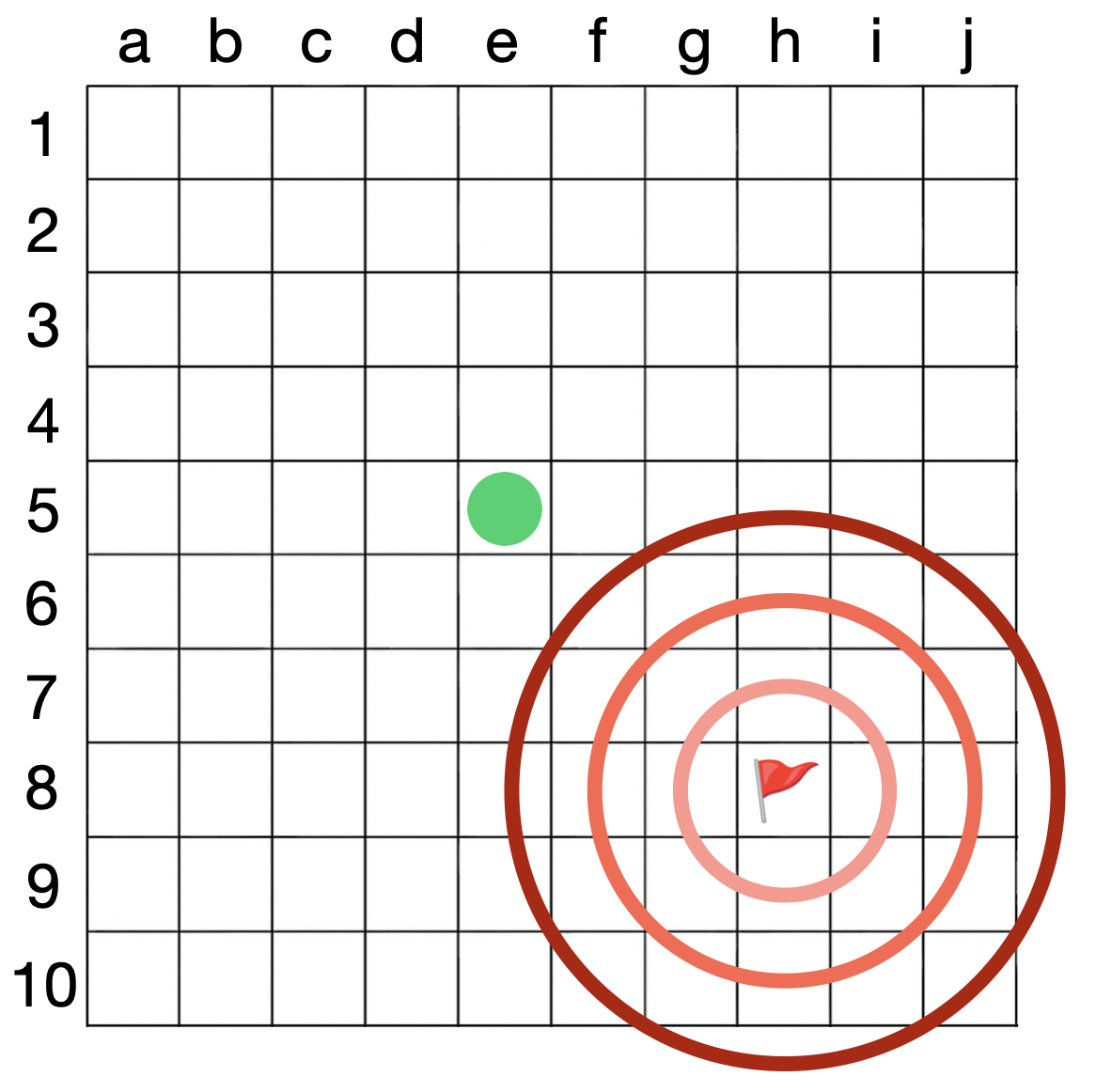}
  \end{center}
  \vspace{-5pt}
  \caption{GridWorld Illustration. The red flag is the goal in the lower-level MDP set by the upper-level decision variable. A state further away from the goal incurs a negative reward with higher magnitude. The green circle indicates the center of the grid, which defines a component of the upper-level objective.}
  \label{fig:grid}
\vspace{-20pt}
\end{wrapfigure}


The lower-level MDP is defined on a 10×10 grid, where each state corresponds to a position on the grid. At every state, the agent can choose from four possible actions: $\Acal = \{\textbf{UP}, \textbf{DOWN}, \textbf{LEFT}, \textbf{RIGHT}\}$. Each action moves the agent to the adjacent cell in the corresponding direction. If the current position lies on the boundary and the action would move the agent outside the grid, the state remains unchanged. The upper-level decision variable $x$ sets a goal state for the lower-level problem, shown by the flag in Figure~\ref{fig:grid}. Let $x=(\operatorname{coor}_1(\text{goal}),\operatorname{coor}_2(\text{goal}))$. The reward of state $s=(\operatorname{coor}_1(s),\operatorname{coor}_2(s))$ is
\begin{align*}
r(s) &= -\Big(\operatorname{coor}_1(s)-\operatorname{coor}_1(\text{goal})\Big)^2\notag\\
&\hspace{20pt}-\Big(\operatorname{coor}_2(s)-\operatorname{coor}_2(\text{goal})\Big)^2.
\end{align*}

We choose the upper-level objective $f$ such that $f(x,\pi)$ when $x$ is close to the center of the grid, indicated by green circle in Figure~\ref{fig:grid} and that $\pi$ has bias towards $\text{DOWN}$ and $\text{RIGHT}$ actions. Specifically, with the coordinate of the center cell denoted by $(\operatorname{coor}_1(\text{center}),\operatorname{coor}_2(\text{center}))$, we consider
\begin{align*}
f(x,\pi)&=\Big(\operatorname{coor}_1(\text{goal})-\operatorname{coor}_1(\text{center})\Big)^2\notag\\
&\hspace{20pt}+\Big(\operatorname{coor}_2(\text{goal})-\operatorname{coor}_2(\text{center})\Big)^2\notag\\
&\hspace{20pt}-\lambda\sum_{s}\Big(\pi(\text{DOWN}\mid s)+\pi(\text{RIGHT}\mid s)\Big),
\end{align*}
where $\lambda$ is a non-negative parameter.

We set $\lambda$ sufficiently large, under which the optimal solution to the bi-level problem is to set the goal on the bottom right corner. This is indeed the learned solution from Algorithm~\ref{alg:main}.


\end{document}